\pdfoutput=1
\documentclass[12pt, a4paper, reqno, twoside, makeidx]{amsart}
\usepackage[margin=2.8cm, marginpar=2cm]{geometry}
\usepackage{comment}
\usepackage{verbatim}
\usepackage{amsmath}
\usepackage{amsfonts}
\usepackage{amssymb}
\usepackage{graphicx}
\usepackage{color}
\usepackage{empheq}
\usepackage[all]{xy}
\usepackage{tikz-cd}
\usepackage{tikz}
\usetikzlibrary{calc}
\usepackage{amssymb}
\usepackage{amsmath, amsthm, amssymb, amsfonts, amscd}
\usepackage[english]{babel}
\usepackage[all]{xy}
\usepackage{url}
\usepackage{hyperref}
\usepackage{mathtools}
\usepackage{pinlabel}

\usepackage{tikzducks}
\usepackage{cite}
\usepackage{float}
\usepackage[latin1]{inputenc}
\usepackage{amssymb, amsmath, amsthm, amscd, mathtools}
\usepackage{graphicx,subfigure}
\usepackage{epstopdf}
\usepackage{tikz}
\usepackage{tikz-qtree}
\usepackage[font=small]{caption}
\usepackage{enumitem}
\usepackage[textsize=scriptsize]{todonotes}
\usetikzlibrary{shapes,arrows,calc}

\DeclarePairedDelimiter\abs{\lvert}{\rvert}

\theoremstyle{remark}

\theoremstyle{plain}
\newtheorem{thm}{Theorem}[section]

\newtheorem{defi}[thm]{Definition}

\theoremstyle{definition}

\title{Contagion Dynamics for Manifold Learning}
\author{Barbara I. Mahler}
\address{Mathematical Institute, University of Oxford, United Kingdom}
\email{mahler@maths.ox.ac.uk}
\begin{document}
\maketitle

\begin{abstract}
Contagion maps exploit activation times in threshold contagions to assign vectors in high-dimensional Euclidean space to the nodes of a network. 
A point cloud that is the image of a contagion map reflects both the structure underlying the network and the spreading behaviour of the contagion on it. Intuitively, such a point cloud exhibits features of the network's underlying structure if the contagion spreads along that structure, an observation which suggests contagion maps as a viable manifold-learning technique. We test contagion maps as a manifold-learning tool on a number of different real-world and synthetic data sets, and we compare their performance to that of Isomap, one of the most well-known manifold-learning algorithms. We find that, under certain conditions, contagion maps are able to reliably detect underlying manifold structure in noisy data, while Isomap fails due to noise-induced error. This consolidates contagion maps as a technique for manifold learning. 
\end{abstract}
\section{Introduction}
Manifold-learning techniques aim to identify low-dimensional manifold structure in high-dimensional data \cite{lee2007nonlinear}. High-dimensional point-cloud data may represent a large number of features on a collection of objects. Some of these features may be redundant or irrelevant, thus giving the data lower-dimensional intrinsic structure. Alternatively, high-dimensional point-cloud data with low-dimensional intrinsic structure may arise as a sample of points from a low-dimensional manifold that is embedded in a high-dimensional space. 

Consider, for instance, data points that lie on a plane in three-dimensional space. Principal component analysis (PCA), a classical dimensionality-reduction technique \cite{Sorzano2014}, can find the directions along which the data has maximum variance as well as the relative importance of these directions. In the case of the plane embedded in three-dimensional space, PCA returns three vectors: two of positive weight spanning the plane and one vector of zero weight that is orthogonal to the plane. PCA can thus identify the plane underlying the ostensibly three-dimensional data. More generally, consider data points that are concentrated around a low-dimensional manifold (reflecting the underlying information) that is embedded in a high-dimensional space. PCA is a \emph{linear} dimensionality-reduction method: If the manifold is nonlinear, PCA is unable to detect the low-dimensional space underlying the data set. This is where manifold-learning techniques (as a type of \emph{nonlinear} dimensionality reduction) can be effective. The purpose of manifold learning is to uncover low-dimensional manifold structure of a data set in a high-dimensional feature space, even if the structure of the data is curved. 

A common procedure for nonlinear dimensionality reduction is the following:
\begin{enumerate}[label=(\arabic*)]
\item Create a network on the data points, such as by defining an edge between any two nodes within some distance $\epsilon$ (producing the \emph{$\epsilon$-neighbourhood graph}) or by connecting each point to its $k$ closest neighbours (producing the \emph{$k$-nearest-neighbour graph}).

\item Define some notion of pairwise distance between data points based on this network (e.g.~shortest-path length for Isomap \cite{Tenenbaum2000}). The aim is to approximate the actual geodesic distance on the underlying manifold.

\item Map points to some space based on the pairwise distances. Possible ways of doing this include the following: (a) use a multidimensional scaling algorithm, which finds an embedding that preserves pairwise distances as well as possible; or (b) take distances to be coordinates in a space of dimension equal to the number of data points (the approach that contagion maps take as a manifold-learning technique \cite{Taylor2015}).
\end{enumerate}

\begin{figure}[H]
\includegraphics[scale=.3]{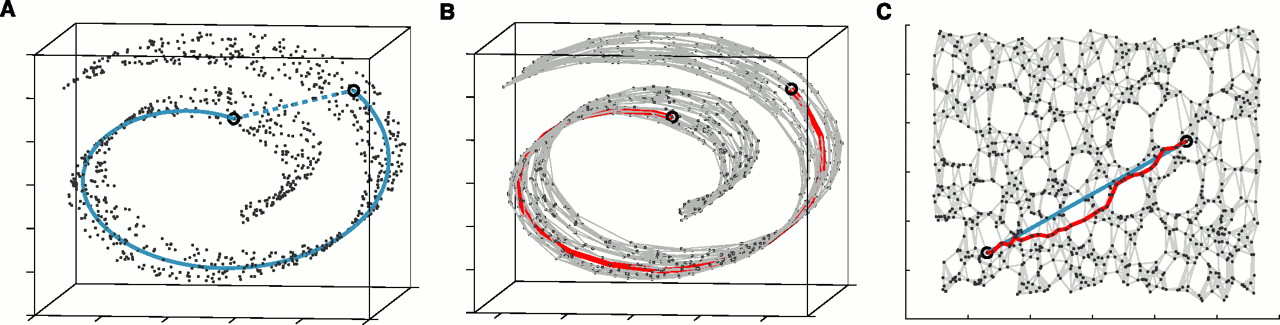}
\caption{(A) Points sampled from a Swiss roll. The distance between two points according to the ambient space can differ significantly from the geodesic distance between the same two points along the underlying two-dimensional manifold. (B) The Isomap algorithm first builds a 
neighbourhood graph on the point cloud and then finds the length of a shortest path between any pair of nodes in this network. (C) The length of a shortest path between two nodes in the network approximates the geodesic distance between the corresponding two points under favourable conditions \cite{Bernstein2000}. Isomap performs classical multidimensional scaling based on the set of shortest-path distances. (Figure taken from \cite{Tenenbaum2000}.)}\label{Swiss}
\end{figure}

Many well established manifold-learning techniques perform poorly when faced with noisy data. Isomap, for instance, can be very sensitive to noise. Consider, for example, a noisy point sample on the Swiss roll (see Figure~\ref{Swiss}). Noise can lead to two points on adjacent sheets lying close together. The $k$-nearest-neighbour graph might then have an edge that connects the corresponding nodes (i.e.~a `short-circuit error'), although the points lie far apart in the intrinsic geometry. Consequently, Isomap falsely considers the two points to be close, and thus fails as a manifold-learning technique in this case. 

Contagion maps \cite{Taylor2015} can circumvent Isomap's `short-circuit error' issue by exploiting the `social reinforcement' phenomenon that characterizes threshold contagions. 
When the threshold of a contagion is small enough to allow spreading via a single edge, the associated contagion map can be viewed as a variant of Isomap and is similarly sensitive to the type of noise described above. For larger thresholds, however, a single errant edge in the $k$-nearest-neighbour graph cannot carry a contagion, and, as a result, the contagion map does not view the two points as close and performs well as a manifold-learning technique. 

We use persistent homology, a method from topological data analysis \cite{Edelsbrunner2008}, as well as more established statistical techniques to perform manifold learning based on contagion dynamics, and we compare this approach to Isomap-type algorithms. 
One of the most common applications of persistent homology is the task of recovering a manifold from a random, potentially noisy sample of points from it \cite{carlsson2009topology}. This application illuminates the natural overlap of topological data analysis with manifold learning. Both are designed to find shape regardless of exact geometry (including measures of curvature and length), and both aim to be robust to noise. Traditionally, they differ in their respective approaches and, as a result, in their ability to identify different types of structural features. Persistent homology can, for example, identify a sphere (by its topological features in dimensions $1$ and $2$), but not a Swiss roll (as it has no non-trivial topological features). A manifold-learning algorithm like Isomap, on the other hand, can `unroll' the Swiss roll (under favourable conditions), but cannot see that a sphere is a two-dimensional manifold: Isomap detects a sphere's lowest embedding dimension $3$, but cannot see its intrinsic $2$-dimensional structure. In this paper we combine the two approaches, by processing our data via a manifold-learning-type procedure first and then computing persistent homology (along with some other measures) based on this processed data.

\section{Algorithms}\label{algorithms}
The fundamental hypothesis that our algorithms are built on is that our data come as samples from some underlying submanifold of $\mathbb{R}^n$, which we want to infer. To this end, we perform variants of a procedure whose basic steps are as follows. 

First, if a data set is given in the form of a point cloud, we start by constructing a \emph{neighbourhood graph} based on this point cloud whose nodes are the points and whose edges connect nearby pairs of nodes. We construct such a graph in two different ways: (1) by connecting each point to its $k$ closest neighbours, producing the \emph{$k$-nearest-neighbour graph}; and (2) by connecting any two points that are within $\epsilon$ from one another, producing the \emph{$\epsilon$-neighbourhood graph}. In both types of neighbourhood graph, one can either weight the edges by the corresponding pairwise distances, or treat the edges as unweighted\footnote{Note that, as the `weights' in this graph are really distances, a small weight of an edge indicates a strong connection between its incident nodes.}. 

Second, we calculate a notion of distance between points that is aimed to be an estimate for the actual geodesic distance on the underlying manifold. 
The idea is that only the pairwise Euclidean distances between neighbouring pairs of points approximate the geodesic distances sufficiently, and that estimates for geodesic distances between non-neighbouring pairs of points can be inferred from the distances between neighbouring points by `tracing' through the neighbourhood graph. 

The precise pairwise distances between adjacent points in a neighbourhood graph provide information that is relevant to estimating the geodesic distance between non-neighbouring points. Unweighted neighbourhood graphs forget this information and thus tend to lead to less accurate approximations to the geodesic distances. The loss of information from taking an unweighted graph can be greater for $k$-nearest-neighbour graphs than for $\epsilon$-neighbourhood graphs, because, in the latter case, the pairwise distances are within a range that is capped by $\epsilon$.  

\subsection{Contagion maps}\label{cont_maps}
First, given data in point-cloud form, our algorithm starts by building a neighbourhood graph $(V,E)$ on the point cloud. It does so by associating a set $V$ of nodes to the data points (where $i$ is the node associated to point $p_i$) and defining the edge set $E$ to build either (1) a $k$-nearest-neighbour graph, or (2) an $\epsilon$-neighbourhood graph:  
\begin{enumerate}[label=(\arabic*)]
\item Given some $k\in \mathbb{N}$, $(i,j)\in E$ if and only if $p_i$ is in the set of the $k$ nearest neighbours of $p_j$, or vice versa. 
\item  Given some $\epsilon \in \mathbb{R}_{>0}$, $(i,j)\in E$ if and only if $d_2(p_i,p_j) \leq \epsilon$. 
\end{enumerate}
If we are given data in the form of a network, we use this given (weighted or unweighted) network instead.

We denote the graph by $(V,E)$, the number of nodes (i.e.~the number of points in the case of point-cloud data) by $|V|=N$, and the graph's binary adjacency matrix by $A$.
We then consider a threshold contagion on this network. We denote the state of node $i \in V$ at time $t$ by $\eta_i(t)$, which takes the value $1$ if it is \emph{active} and the value $0$ if it is \emph{inactive}. Given a set of seed nodes consisting of a node $j \in V$ together with its immediate neighbours
\[
S^{(j)}=\{j\} \cup \{k \ | \ A_{jk} \neq 0\}
\]
that are active at time $t=0$, and a threshold $T$, we update node states synchronously in discrete time steps according to the following rule. If $\eta_i(t)=1$, then $\eta_i(t+1)=1$. If $\eta_i(t)=0$, then 
\begin{equation*}
	\eta_i(t+1)=1 \, \text{ if and only if } \, f_i > T \,,\text{\hspace{5mm} where \hspace{3mm}} f_i = \frac{1}{d}\sum_{k \in V} A_{ik}\eta_k(t) \, .
\end{equation*}
Given a network $(V,E)$ and a threshold $T$, a contagion seed yields a deterministic process, which we call a \emph{realization} of the contagion model with $T$ on $(V,E)$. The activation time of node $i$ in the realization seeded around node $j$ is the smallest $t$ such that $\eta_i(t)=1$, and we denote it by $x_j^{(i)}$. If node $i$ is never activated in the realization that is seeded around node $j \in V$, we set $x_j^{(i)}=2N$ (i.e.~larger than any actual activation time). 

One can now work directly with this set of activation times, that is, treat the activation times as estimates to the geodesic distance between points on the underlying manifold and use them to examine geometry, topology, and dimensionality of the data. Alternatively, one can work with points whose coordinate vectors are given by the columns of the dissimilarity matrix $D_{\rm cont}=\left(x_i^{(j)}\right)_{i,j \in V}$ that holds the activation times (or of a symmetrization of this dissimilarity matrix given by $D_{\rm cont}+\left(D_{\rm cont}\right)^{\rm T}$). 
Using terminology from \cite{Taylor2015}, producing such a point cloud is equivalent to mapping the nodes via a \emph{contagion map}. 

\begin{defi}
The \emph{regular contagion map} associated to $(V,E)$ and $T$ is the function from the set $V$ of nodes to $\mathbb{R}^{N}$ that is defined by
\begin{equation*} 
	i \mapsto x^{(i)}=[x_1^{(i)}, x_2^{(i)}, \dots , x_{N}^{(i)} ]^T\,.
\end{equation*}
The \emph{symmetric contagion map} associated to $(V,E)$ and $T$ is the function from the set $V$ of nodes to $\mathbb{R}^{N}$ that is defined by
\begin{equation*} 
i \mapsto [x_1^{(i)} + x_i^{(1)}, \dots , x_{N}^{(i)} + x_i^{(N)}]^T\,.
\end{equation*} 

\end{defi}
In this paper, we work with symmetric contagion maps exclusively. 

We examine dimensionality, topology, and geometry as follows. 
We perform classical multidimensional scaling (MDS) \cite{Torgerson1958} based either on the activation times directly, or on the Euclidean distances between points in the image of the contagion map. 
MDS aims to embed a point cloud in a given low-dimensional vector space in a way that preserves given distances between pairs of points as well as possible. It does so by minimizing a cost function called `strain': 
Given a matrix $D=\left(d_{ij}\right)_{i,j \in I}$ of pairwise distances, or `dissimilarities' (not necessarily satisfying the defining properties of a metric), and some Euclidean target space $Y$, MDS finds coordinate vectors $\left\{y_i \in Y \right\}_{i \in I}$ that minimize the cost function $$E=\|\tau(D)-\tau(D_Y)\|_{L^2} \ ,$$ where $D_Y=\left(\|y_i - y_j\|_2 \right)_{i,j \in I}$, $\|M\|_{L^2}=\sqrt{\sum_{i,j}M^2_{ij}}$, and $\tau(M)=-HSH/2$ with $S_{ij}=M^2_{ij}$ and $H_{ij}=\delta_{ij} - 1/\abs{I}$ \cite{Mardia1979}.

MDS is a linear dimensionality-reduction technique when applied to Euclidean distances. By applying it to sensible approximations to the geodesic distances between data points, we hope to recover potentially curved structure. In other words, we hope to use MDS to achieve non-linear dimensionality reduction. 

Given an embedding via MDS to Euclidean space of dimension $p$, we calculate its \emph{residual variances} \cite{Cox2010} 
\begin{equation*}
	R_p = 1 - \left(\rho^{(p)}\right)^2 \, ,
\end{equation*}
where $\rho^{(p)}$ is the Pearson correlation coefficient \cite{pearson1895notes} between the given pairwise distances $\left\{d_{ij} \right\}_{i,j \in I}$ and the corresponding pairwise Euclidean distances $\left\{\|y_i - y_j\|_2\right\}_{i,j \in I}$ between points in the embedding. 

We determine the \emph{approximate embedding dimension} $P$ of the data (according to the given pairwise distances) by finding the smallest dimension such that the residual variance of the embedding via MDS to that dimension is less than $5\%$, that is, $$P=\min \{p \ | \ R_p < 0.05 \} \ . \footnote{In practice, we put a cap of $100$ on $P$, so if the approximate embedding dimension is $100$ or larger, we record it as $100$. } $$

In addition to these dimensionality considerations via MDS, we analyse our data topologically by computing the persistent homology of the Vietoris--Rips filtration \cite{Ghrist2008} based either on the activation times directly, or on the Euclidean distances between points in the image of the contagion map. When a given base-geometry is given, we also examine our data geometrically through a Pearson correlation coefficient between that base-geometry and the given dissimilarity measure.  
Note that such a known base-geometry to compare our processed data to is not usually given in manifold-learning applications. The measure is, however, useful when testing the algorithm on benchmark data. 

When analysing the point cloud given by the columns of $D_{\rm cont}$ (in other words, the contagion map), we are essentially applying our methods to a dissimilarity matrix that encodes the pairwise distances between the column vectors of $D_{\rm cont}$. We denote the operator that maps $D_{\rm cont}$ to that matrix as $$p_{\rm dist} \ : \ \mathcal{M}_{m,n}\left(\mathbb{R} \right) \longrightarrow \mathcal{M}_{n,n}\left(\mathbb{R} \right) \, ,$$ with $\left(p_{\rm dist}(D)\right)_{ij}=d_2(D_{*i},D_{*j})$. 

\subsection{Isomap}\label{isomap}
The Isomap algorithm \cite{Tenenbaum2000} is essentially a combination of a shortest-path algorithm with MDS. In a sense, Isomap works as a `nonlinear version' of MDS which accommodates for potential curvature of data by incorporating a shortest-paths algorithm to estimate geodesic distances between data points.

The original Isomap algorithm proceeds as follows. 
First, given point-cloud data, Isomap starts by building a neighbourhood graph $(V,E)$ on the point cloud, as described for contagion maps in Section~\ref{cont_maps} above. Next, Isomap calculates the shortest-path lengths between pairs of nodes in this network using some shortest-path algorithm. We use the Floyd\textendash Warshall algorithm \cite{Floyd1962,Warshall1962} in the following work. The set of shortest-path lengths can be recorded in a \emph{dissimilarity matrix} $D_{\rm iso}=\left(d_{\rm G}(i,j) \right)_{i,j\in V}$ \footnote{The length of a shortest path between nodes $i$ and $j$ is denoted by $d_{\rm G}(i,j)$. Note that the function $d_{\rm G} \ : \ V \times V \rightarrow [0,\infty)$ is a metric on the set of nodes or (equivalently) on the set of data points.}, to which Isomap finally applies MDS to map the data points to a low-dimensional space.  

As for contagion maps, if data is given in the form of a network, we will work with this given network instead of a neighbourhood graph. 
Moreover, in addition to the original Isomap algorithm, which simply projects points via MDS based on the set of shortest-path lengths (i.e.~the entries in $D_{\rm iso}$), we calculate the residual variances of these projections, and we also examine this set topologically (via the persistent homology of the Vietoris--Rips filtration based on these shortest-path lengths) and geometrically (via a Pearson correlation \cite{pearson1895notes} with some given base-geometry), when possible. Furthermore, we analyse the point cloud given by the columns (or, equivalently, rows) of $D_{\rm iso}$, that is, we analyse the entries in $p_{\rm dist}(D_{\rm iso}) $. 

Note that a contagion map with threshold $T=0$ is approximately equivalent to a version of Isomap that uses an unweighted neighbourhood graph.

\subsection{Workflow}

\begin{figure}[H]
\begin{tikzpicture}
\node [draw, rectangle, align=center,anchor=west,xshift=1.0cm] (data) at (0,0) {point-cloud\\data};
\node [draw, rectangle, align=center,anchor=west,xshift=1.0cm] (graph) at (data.east) {neighbourhood\\graph};
\node [draw, rectangle, align=center,anchor=west,xshift=1.0cm] (dist) at (graph.east) {distance\\estimates};
\node [draw, rectangle, align=center,anchor=west,xshift=1.0cm, yshift=+1.0cm] (pdist) at (dist.north) {processed\\distance estimates};
\node [draw, rectangle, align=center,anchor=west,xshift=1.0cm, xshift=+2.0cm] (inference) at (dist.east) {structural\\inference };
\draw [->] (data) -- node{} (graph); 
\draw [->] (graph) -- node{} (dist);
\draw [->] (dist) -- node{} (inference); 
\draw [->] (pdist) -- node{} (inference);     
\draw [->] (dist) -- node{} (pdist); 

\end{tikzpicture}
\caption{Schematic representation of the workflow.}
\label{workflowline}
\end{figure}
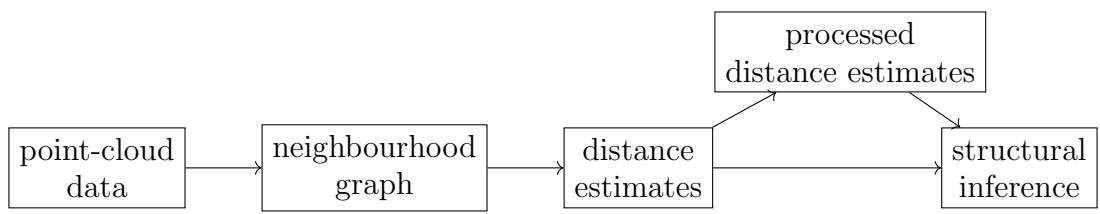

Our workflow is composed of multiple stages, at each of which one can choose from a number of different options. This leads to exponentially many possible procedures that one can use to analyse a given data set. First, given point-cloud data, one needs to choose the type of neighbourhood graph to build on this data set, as well as the defining parameter $k$ or $\epsilon$. Next, one needs to pick a way of estimating geodesic distances based on this graph. We choose either shortest-path distance or activation times in a threshold contagion; that is, we follow either the Isomap algorithm or that of contagion maps. In the case of contagion maps, one also needs to choose a threshold parameter $T$. Given the set of estimates for the geodesic distances, i.e.~the dissimilarity matrix that encodes the shortest-path ($D_{\rm iso}$) distances or activation times ($D_{\rm cont}$), one can apply further methods either to these estimates directly, or to the pairwise distances between the points whose coordinate vectors are the columns (or rows, by symmetry) of this dissimilarity matrix. For either of these choices one can finally apply methods to determine dimensionality, topology, and geometry. Figure~\ref{workflowline} shows a schematic representation of our workflow. We apply different subsets of the full analysis to the different data sets that we consider. 

\section{Results}\label{manifoldresults}
We apply Isomap, as well as contagion maps with several different thresholds, to three different data sets. First, we consider point samples from a Swiss roll, a classical benchmark data set for dimensionality reduction and one of the first data sets to which Isomap was applied \cite{Tenenbaum2000}. We then analyse a couple of representatives of a class of torus-based networks, which was studied in \cite{Mahler2020}. The toroidal structure underlying these networks allows us to examine the topological aspect of our methods by looking to recover the torus's nontrivial topological features. Finally, we consider a data set that represents the conformation space of the cyclo-octane molecule \cite{Martin2010}. This data set is known to have non-trivial topological features and is an example of a naturally occurring data set.

\subsection{Noisy samples from a Swiss roll}
We first examine point samples from a Swiss roll that are obtained by taking regularly spaced points on the Swiss roll surface and then adding various levels of noise to these points. This way of generating data makes it possible to have direct control over the data, so one can explore how different algorithms react to slight variations of the data (in terms of e.g.~density, noise level, or uniformity). 
We use this data set primarily to explore the effects of the parameter $k$ or $\epsilon$ when building a neighbourhood graph. 

\begin{figure}[H]
     \leftline{\hskip 0.4cm (a) \hskip 2.83cm (b) \hskip 2.83cm (c) \hskip 2.83cm (d)} 
     \begin{minipage}{0.22\textwidth}
        \centering
\hspace{4mm} \includegraphics[scale=.25]{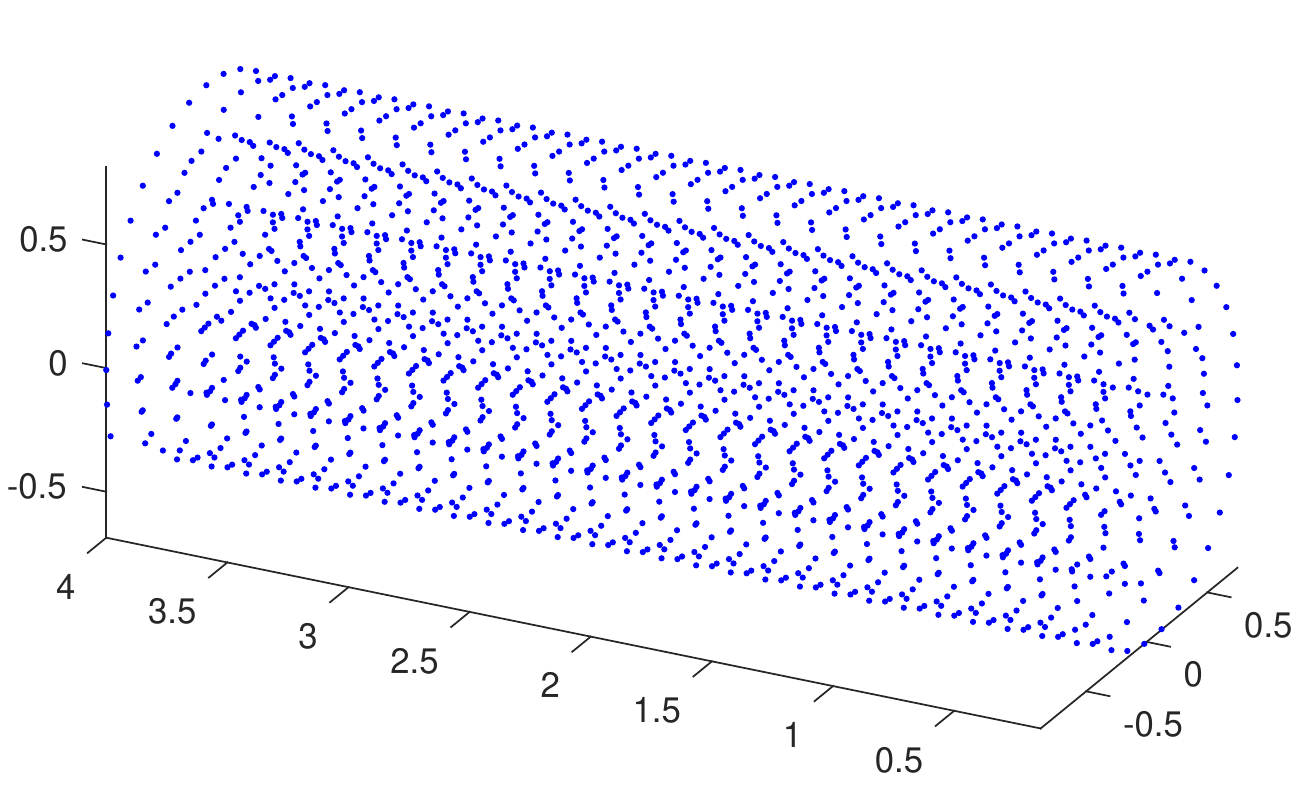}
\end{minipage}
\begin{minipage}{0.22\textwidth}
        \centering
\includegraphics[scale=.25]{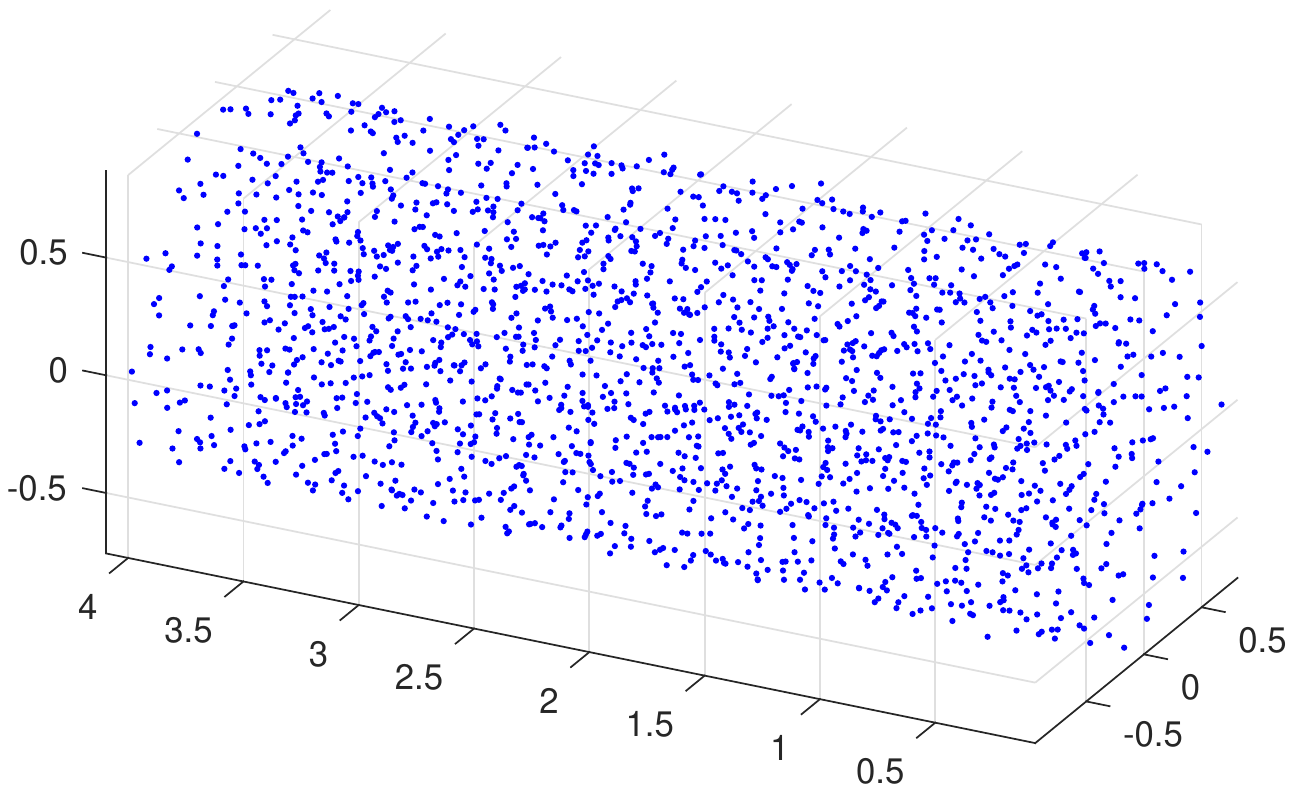}
\end{minipage}
\begin{minipage}{0.22\textwidth}
        \centering
\includegraphics[scale=.25]{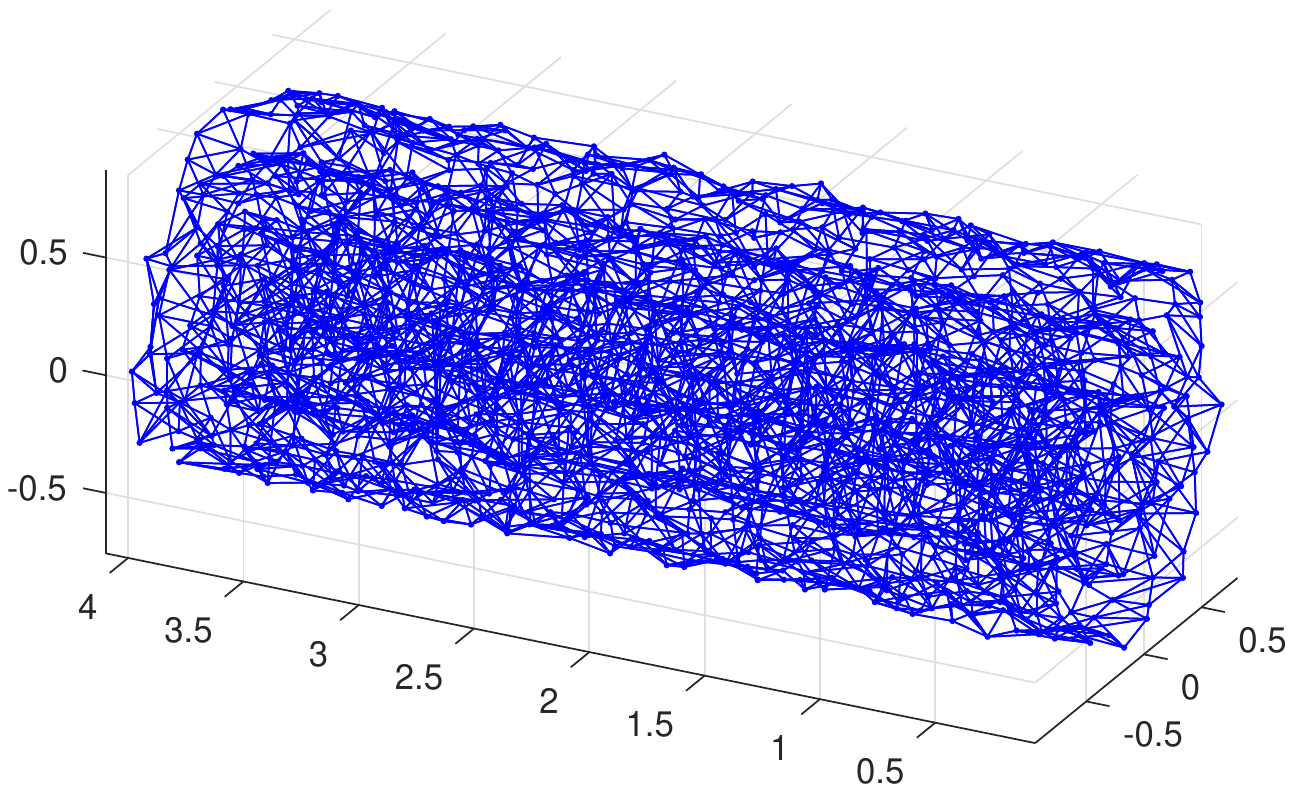}
\end{minipage}
\begin{minipage}{0.22\textwidth}
        \centering
\includegraphics[scale=.25]{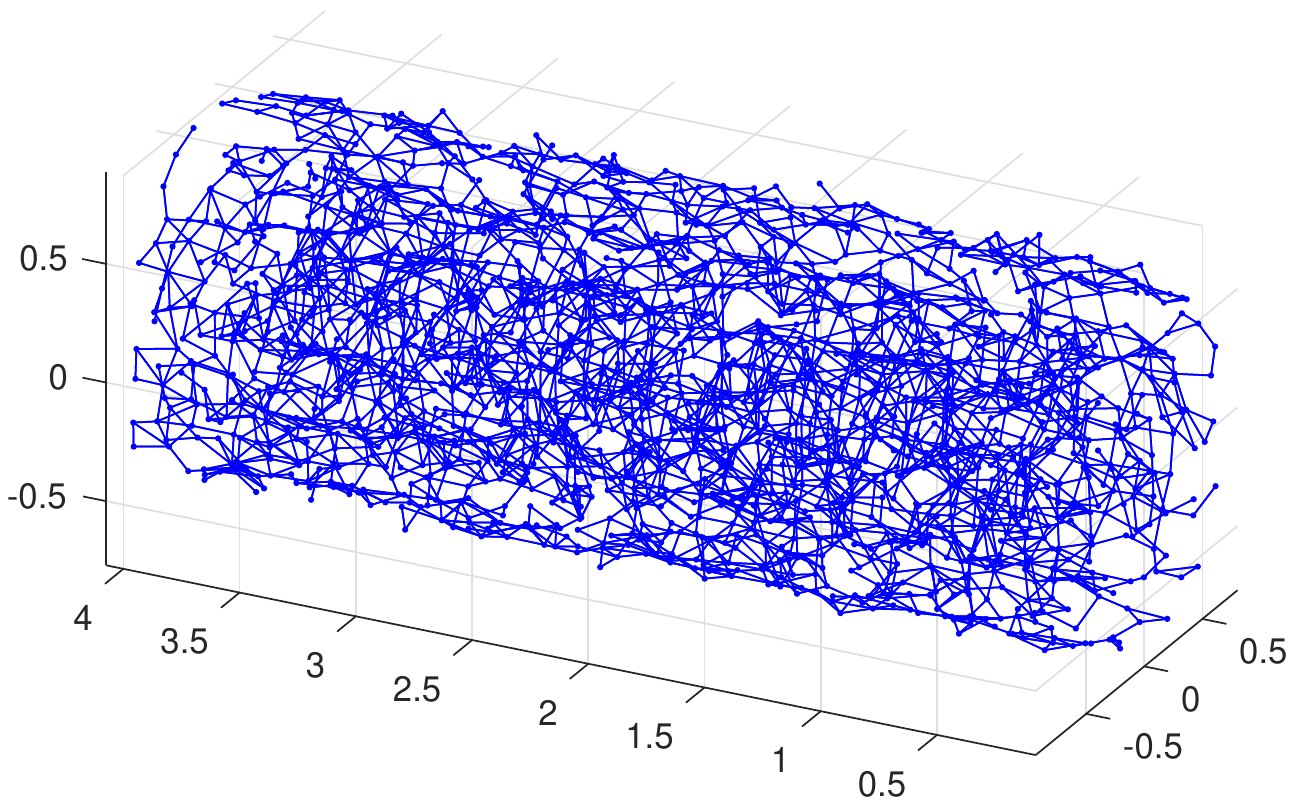}
\end{minipage}

\begin{minipage}{0.22\textwidth}
        \centering
\hspace{4mm} \includegraphics[scale=.25]{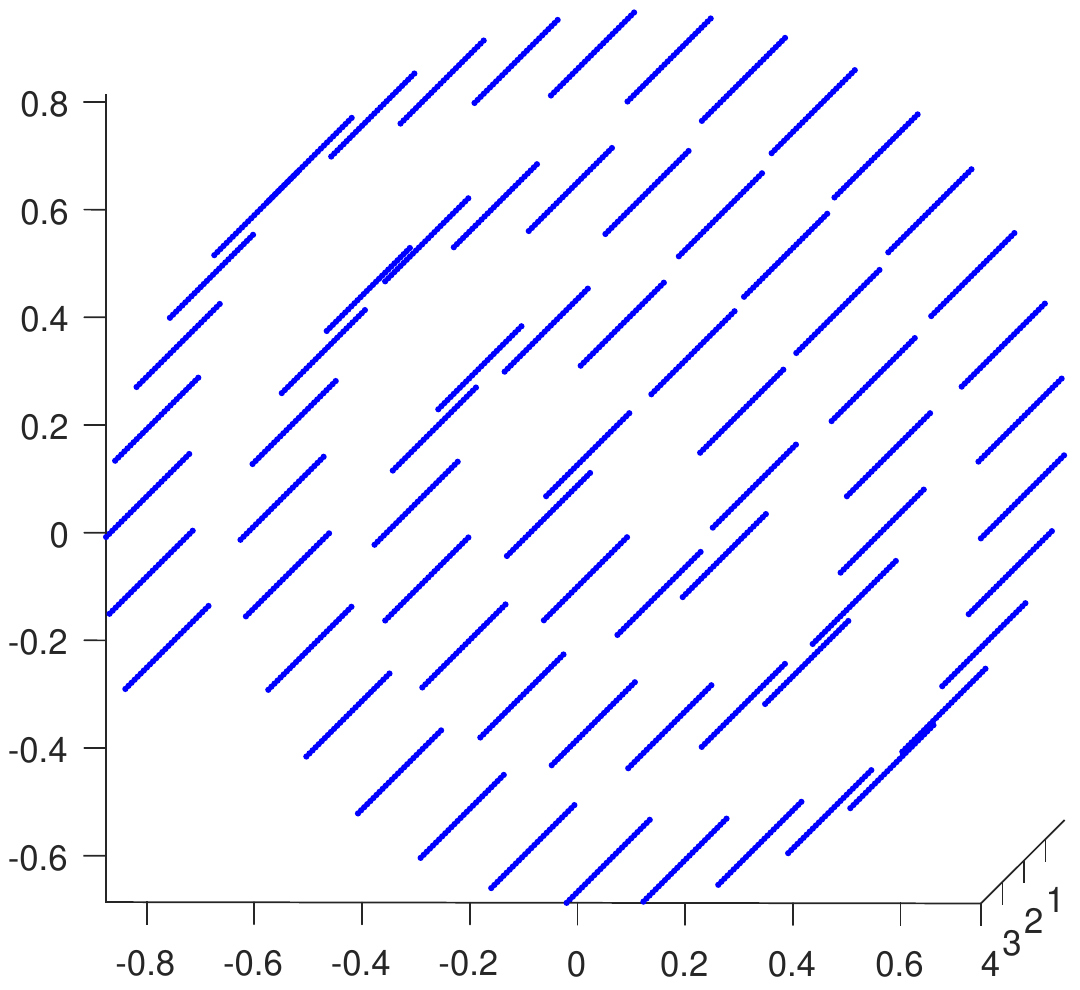}
\end{minipage}
\begin{minipage}{0.22\textwidth}
        \centering
\includegraphics[scale=.25]{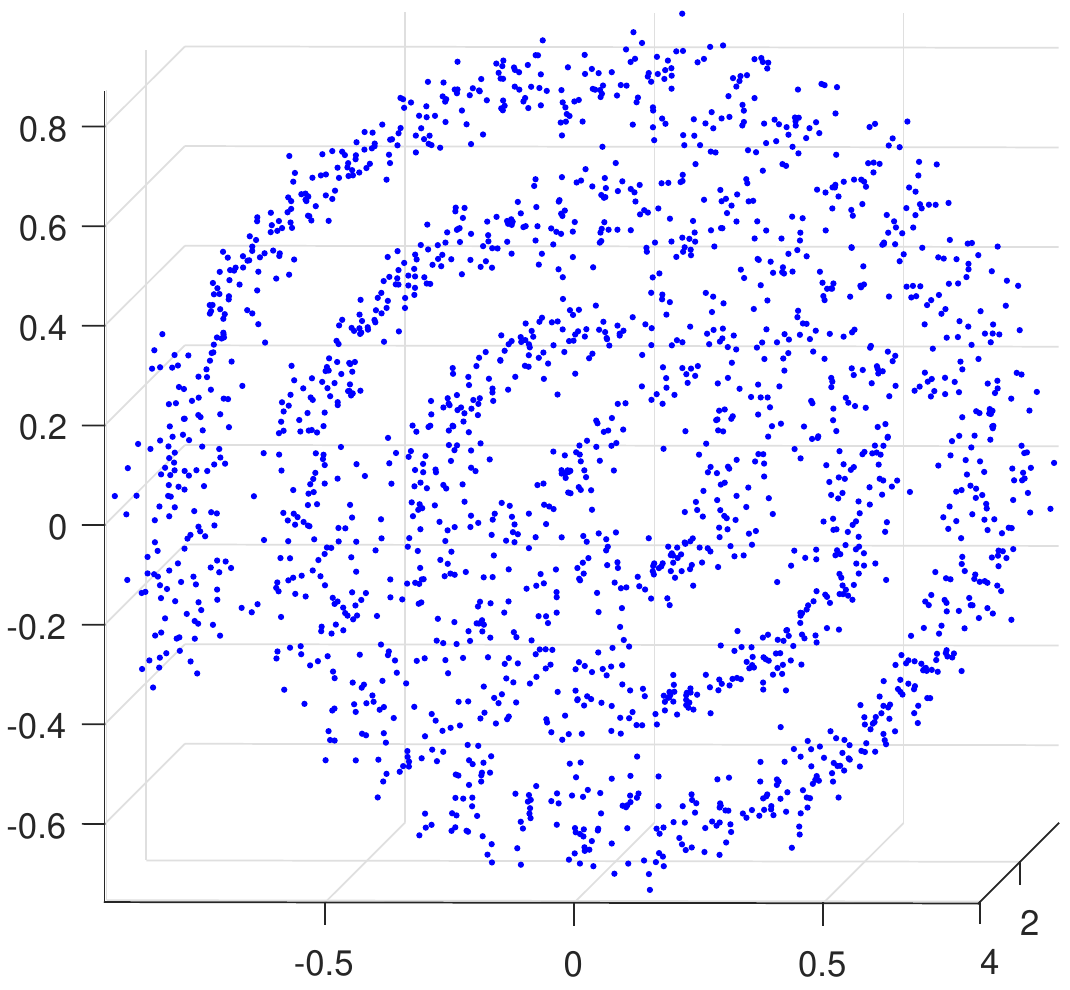}
\end{minipage}
\begin{minipage}{0.22\textwidth}
        \centering
\includegraphics[scale=.25]{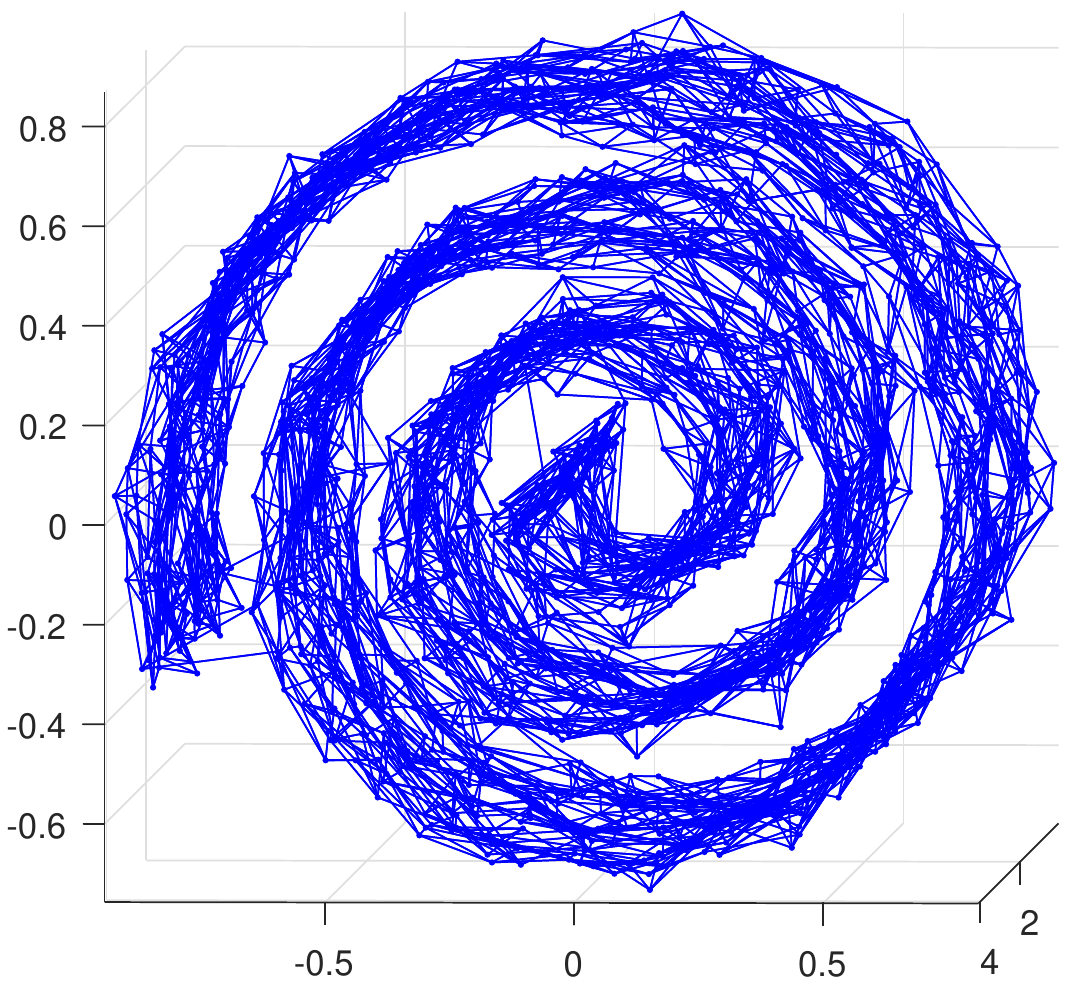}
\end{minipage}
\begin{minipage}{0.22\textwidth}
        \centering
\includegraphics[scale=.25]{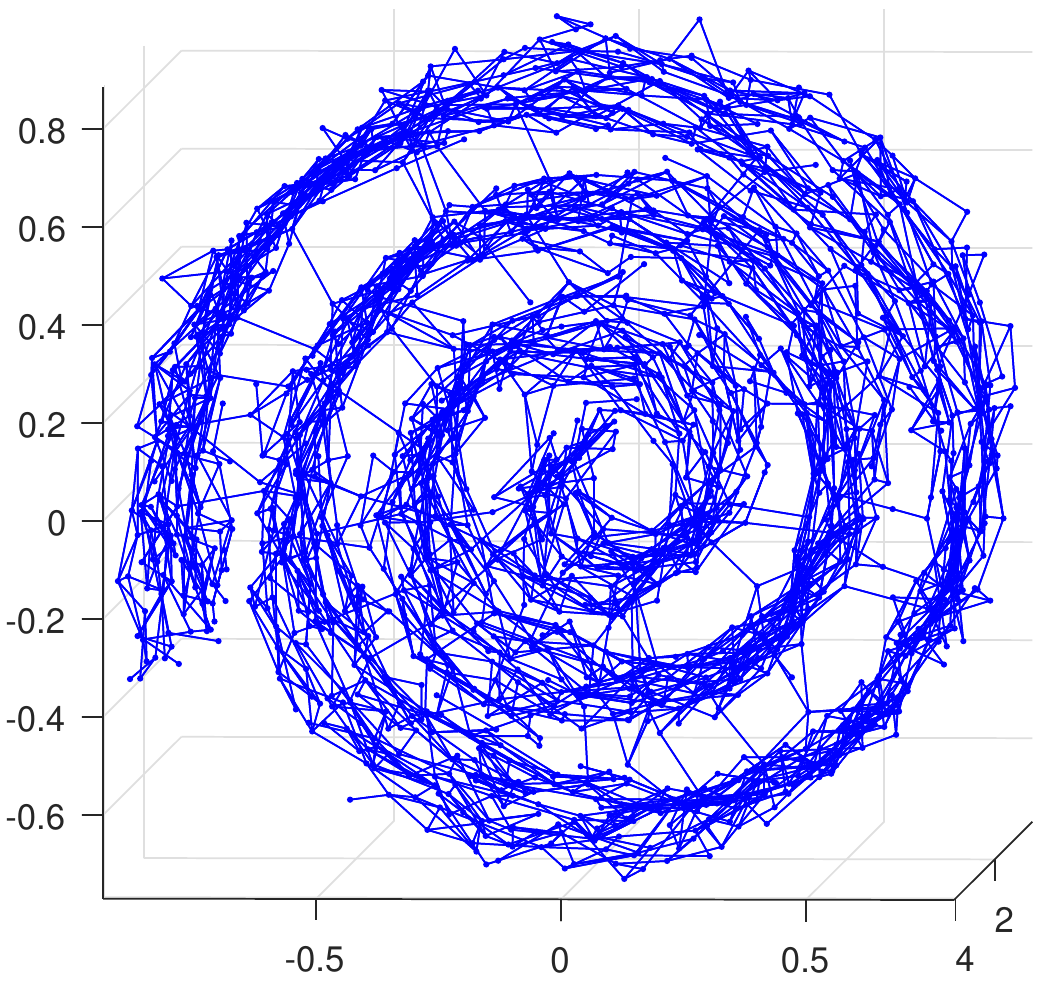}
\end{minipage}

\caption{(a) Regularly spaced points on a Swiss roll at density $50$ per unit square. (b) Gaussian noise is added at a signal-to-noise ratio of $30$. (c) The $5$-nearest-neighbour graph on the noisy point cloud. (d) The $0.18$-neighbourhood graph on the noisy point cloud. (We show two views of each plot.)}\label{Barbara_swiss_roll}
\end{figure}

We start by taking points on a Swiss roll at a density of approximately $50$ per unit square, regularly spaced with respect to the intrinsic geodesic distance on the Swiss roll (see Figure~\ref{Barbara_swiss_roll}(a)). 
We then add \emph{Gaussian noise} with a specified signal-to-noise ratio ($S/N$) to these regularly spaced points (see Figure~\ref{Barbara_swiss_roll}(b)). 
That is, for each point $p=[p_1,p_2,p_3]$ in this regularly space point sample, we add independent, identically distributed noise drawn from a zero-mean normal distribution to each of its coordinates to obtain a perturbation $p_{\rm noisy}$ of the point:
\[
p_{\rm noisy}=[p_1+n_1,p_2+n_2,p_3+n_3] \ ,
\]
where $n_i \sim \mathcal{N}\left(0,\sigma^2  \right)$ with $\sigma^2=10^{\frac{-S/N}{10}}$.

We test Isomap and contagion maps on this noisy point cloud to see how well each of them sees the underlying $2$-dimensional space. Each algorithm starts by building a neighbourhood graph on the points (see Figure~\ref{Barbara_swiss_roll}(c,d)).

We have already touched on Isomap's sensitivity to `short-circuit errors' in the introduction. However, a careful choice of $\epsilon$ (or $k$) when constructing the neighbourhood graph can mitigate such errors to some extent \cite{balasubramanian2002isomap}. The goal is to find an $\epsilon$ (or $k$) that is small enough to avoid short-circuit edges, but not so small that the resulting graph `corrupts' the underlying space. One may choose to simply vary $\epsilon$ over a range and implement Isomap on all of the neighbourhood graphs that thus arise. This approach is in the same vein as considering the full range of thresholds for the contagion map algorithm, and it works whenever there exists a range of $\epsilon$ (or $k$) for which the resulting neighbourhood graphs correctly represent the underlying topology. However, for some data sets --- particularly those that are sparse and incorporate a high level of noise --- it is impossible to find a value for $\epsilon$ (or $k$) that strikes a balance between covering the underlying topology and not making `short-circuit errors'. In other words, the range of $\epsilon$ (or $k$) that `corrupt' the underlying topology and the range of $\epsilon$ (or $k$) that make `short-circuit errors' overlap, leaving no values of $\epsilon$ (or $k$) that yield neighbourhood graphs that correctly represent the underlying topology. For such data sets, Isomap is inadequate as a manifold-learning tool, but contagion maps may be effective. 

\begin{figure}[H]
\centering

    \leftline{\hskip 0.00cm (a)} 
    \begin{minipage}{0.33\textwidth}
        \centering
\includegraphics[width=.7\textwidth]{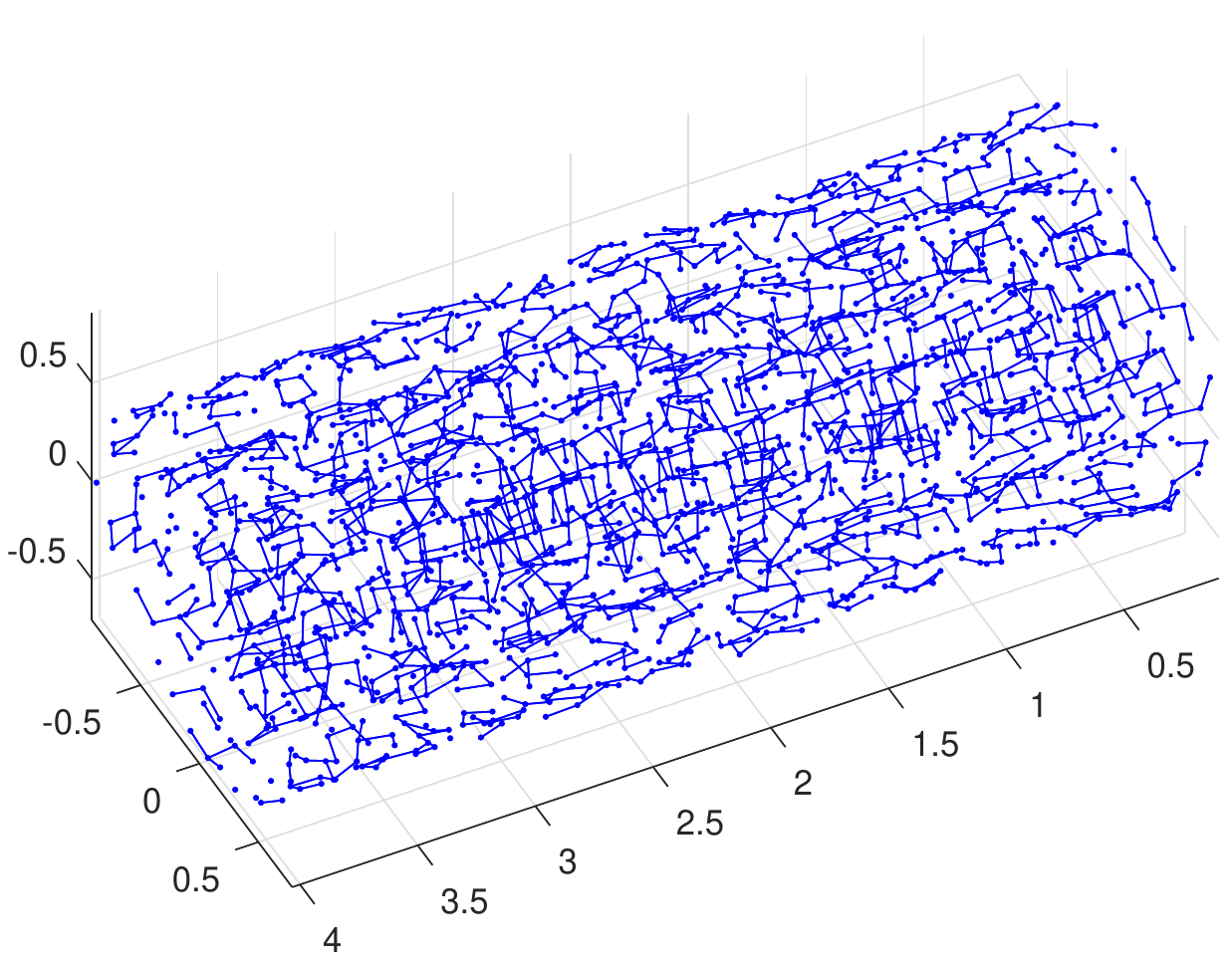}
    \end{minipage}\hfill
    \begin{minipage}{0.33\textwidth}
        \centering
\includegraphics[width=.7\textwidth]{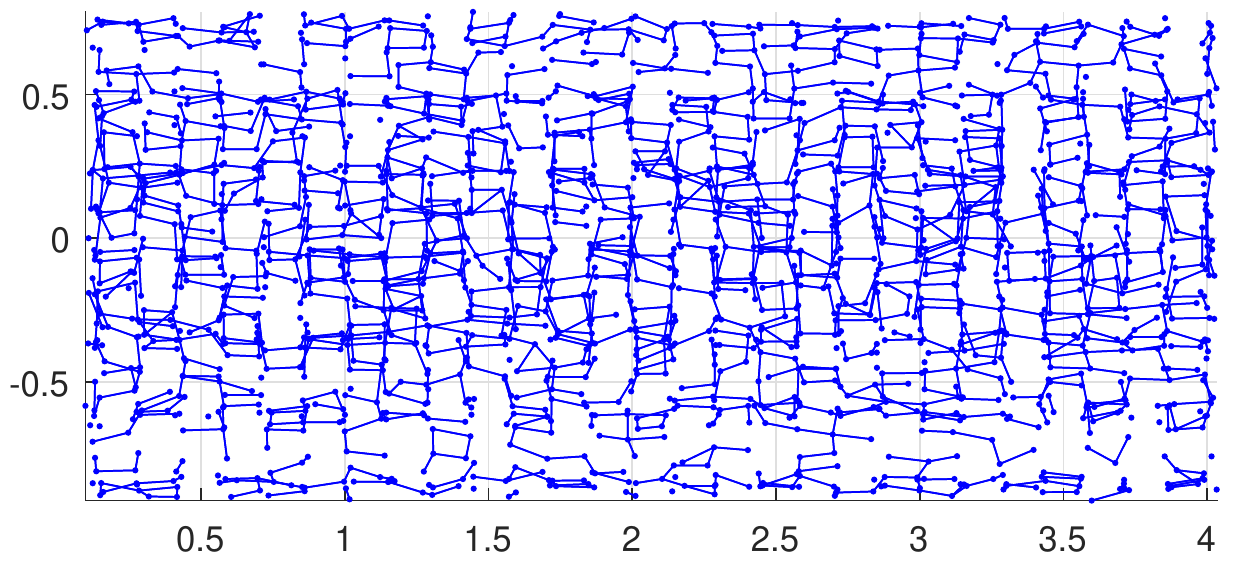}
    \end{minipage}\hfill
\begin{minipage}{0.33\textwidth}
        \centering
\includegraphics[width=.6\textwidth]{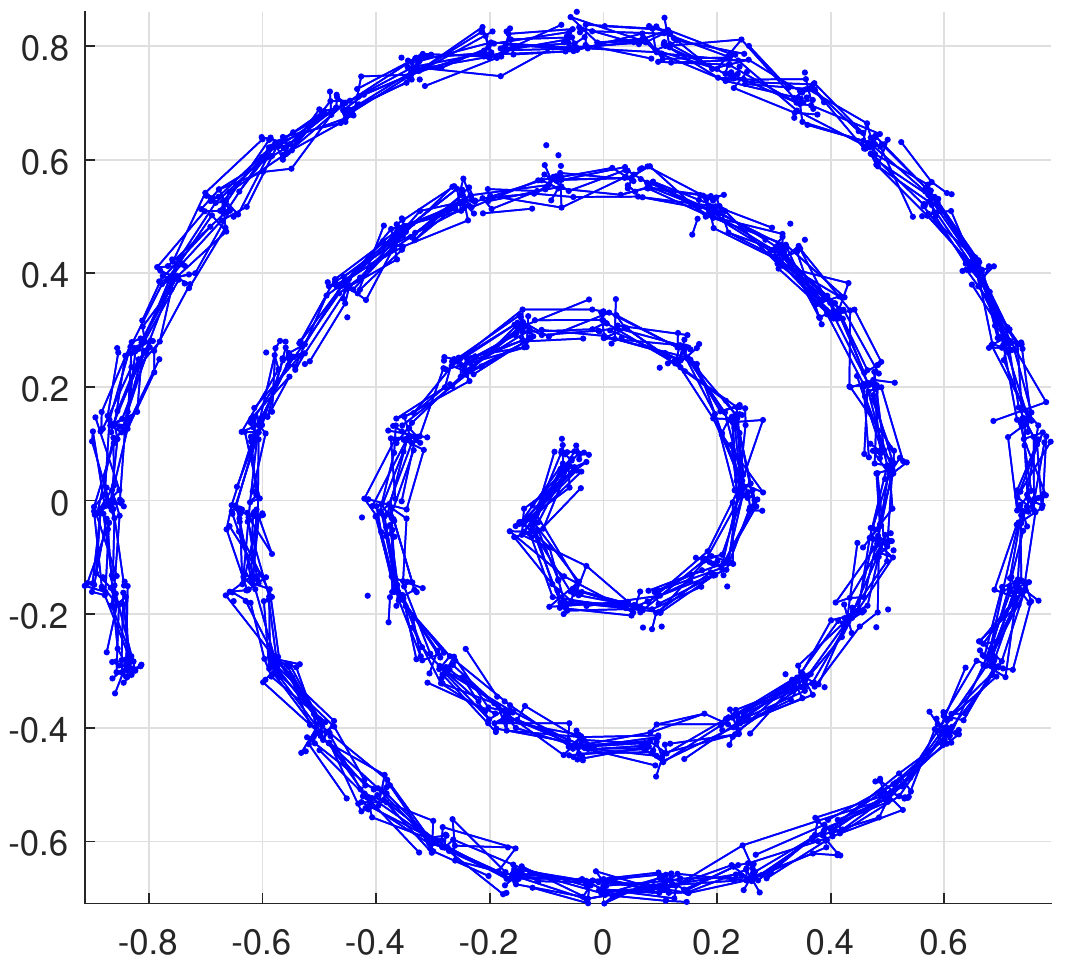}
    \end{minipage}\hfill
        
    \leftline{\hskip 0.00cm (b)} 
    \begin{minipage}{0.33\textwidth}
        \centering
\includegraphics[width=.7\textwidth]{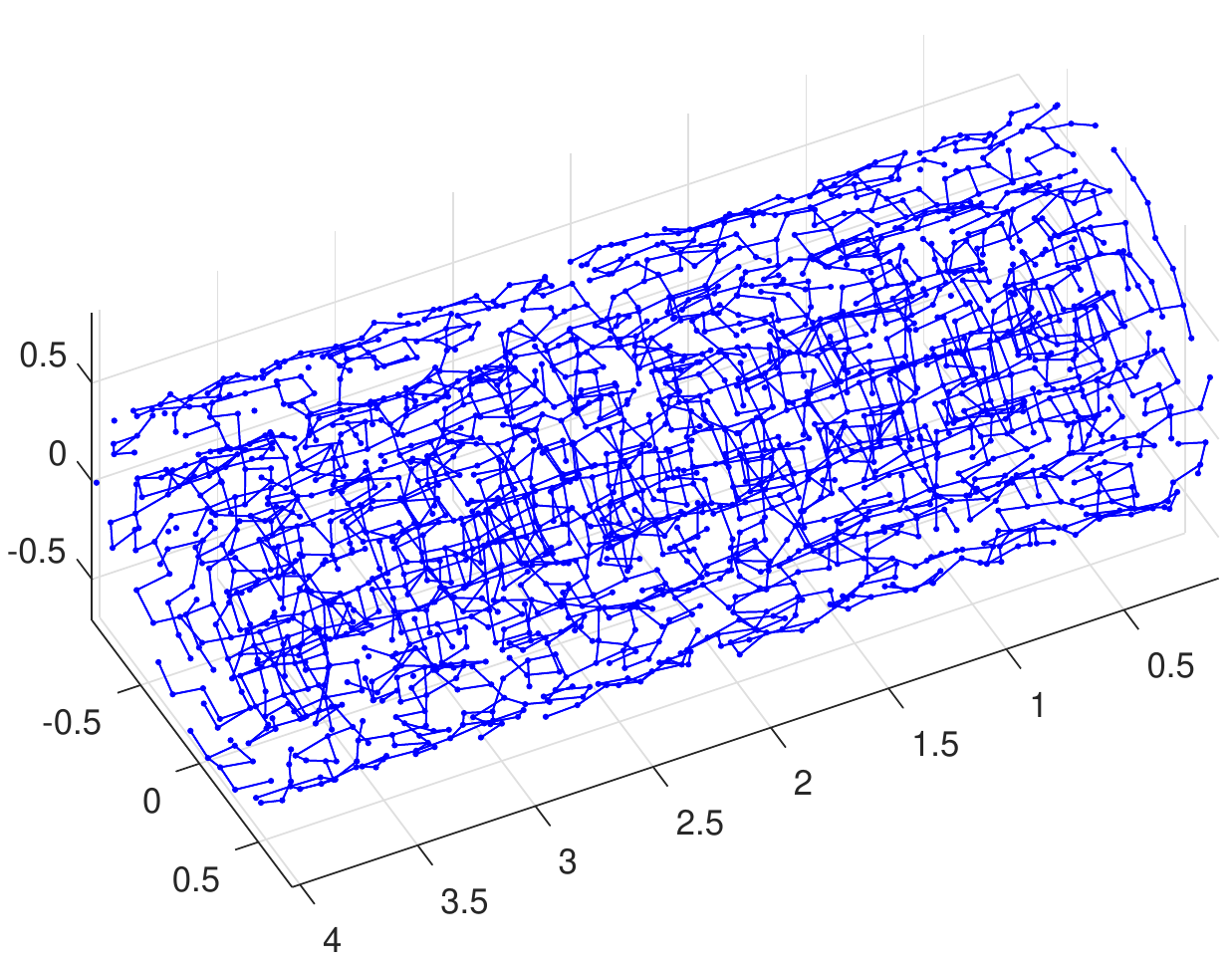}
    \end{minipage}\hfill
    \begin{minipage}{0.33\textwidth}
        \centering
\includegraphics[width=.7\textwidth]{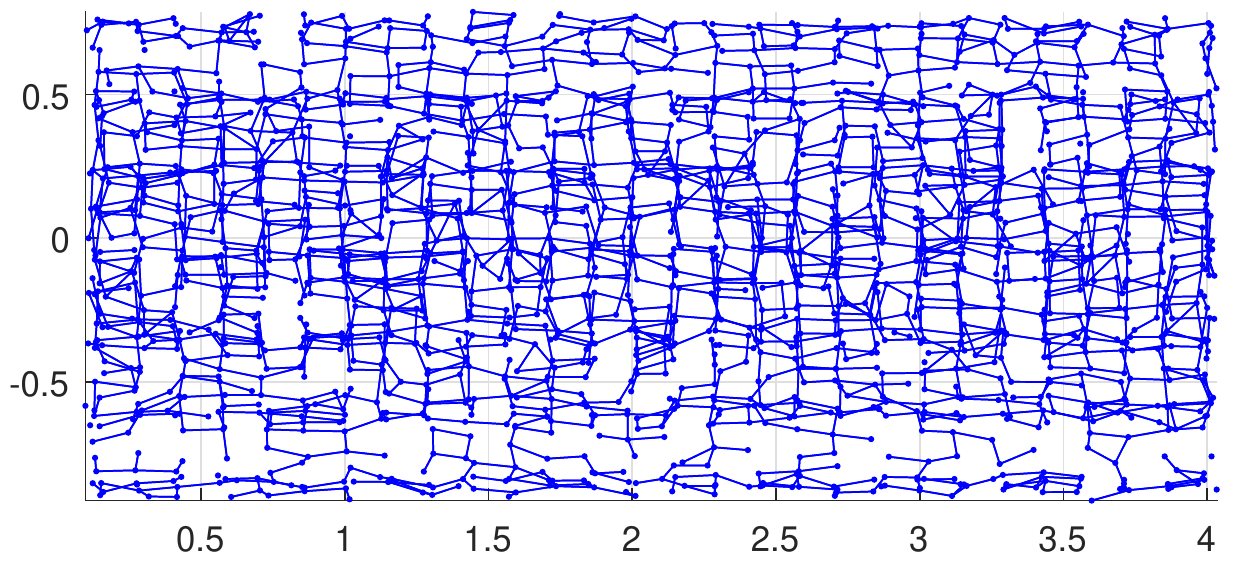}
    \end{minipage}\hfill
    \begin{minipage}{0.33\textwidth}
        \centering
\includegraphics[width=.6\textwidth]{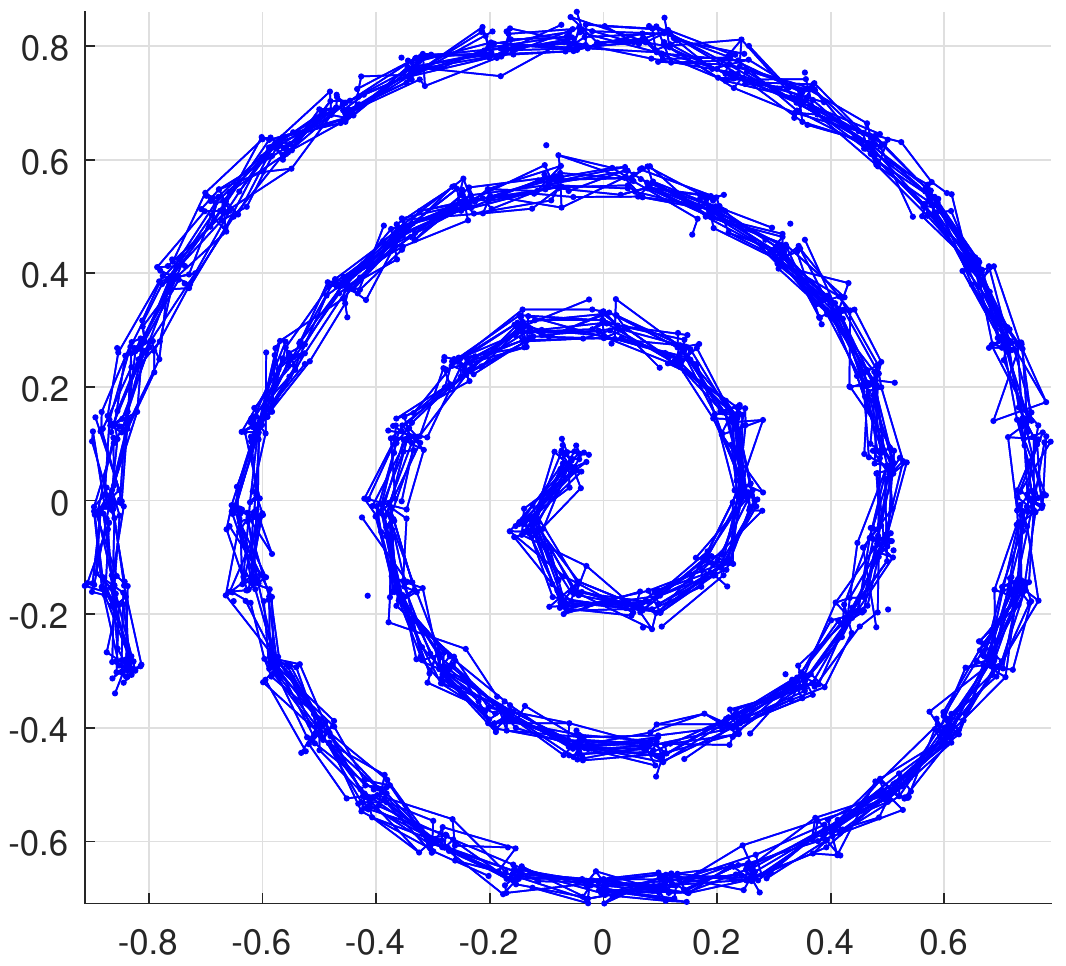}
    \end{minipage}\hfill
    
    \leftline{\hskip 0.00cm (c)} 
    \begin{minipage}{0.33\textwidth}
        \centering
\includegraphics[width=.7\textwidth]{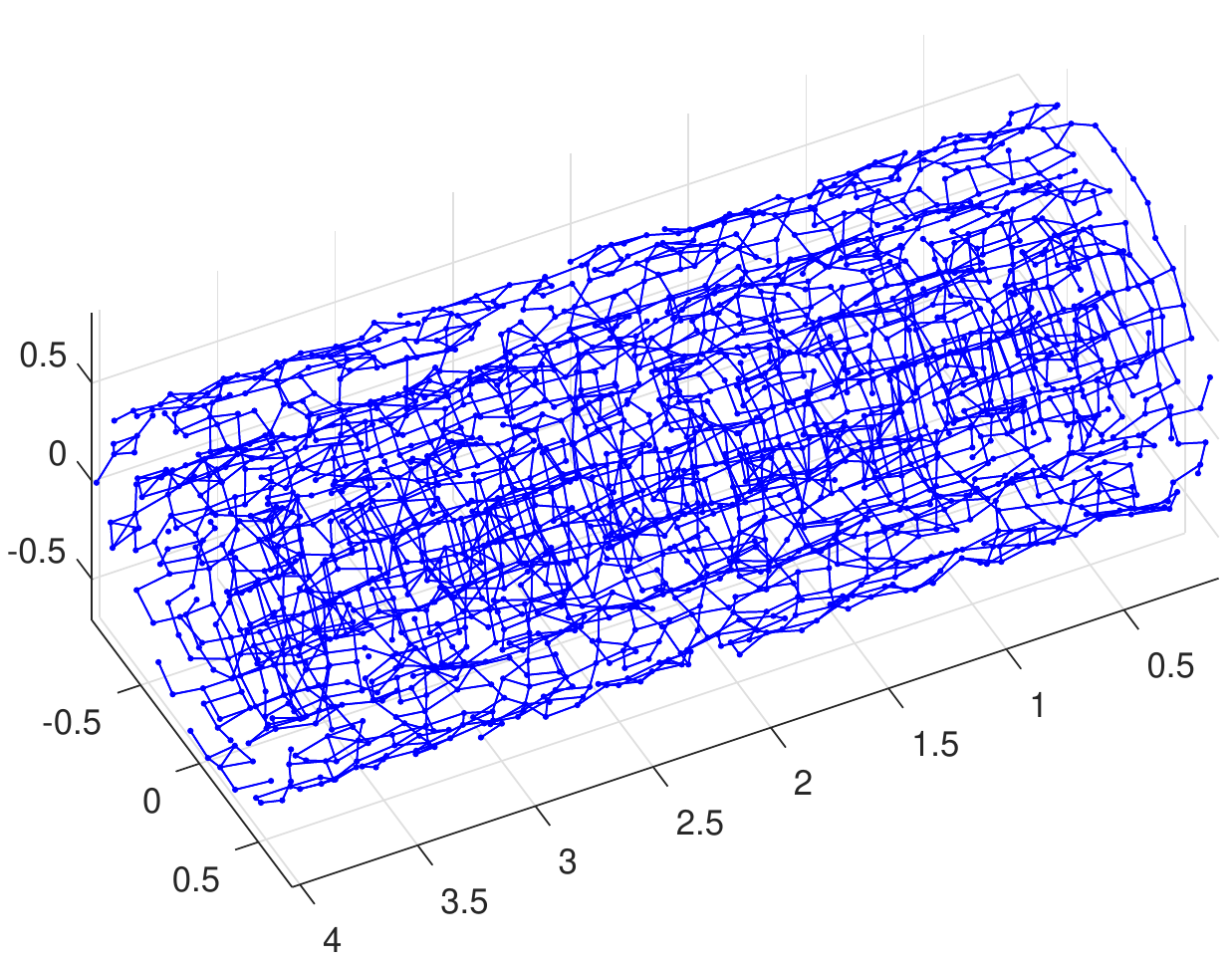}
    \end{minipage}\hfill
     \begin{minipage}{0.33\textwidth}
        \centering
\includegraphics[width=.7\textwidth]{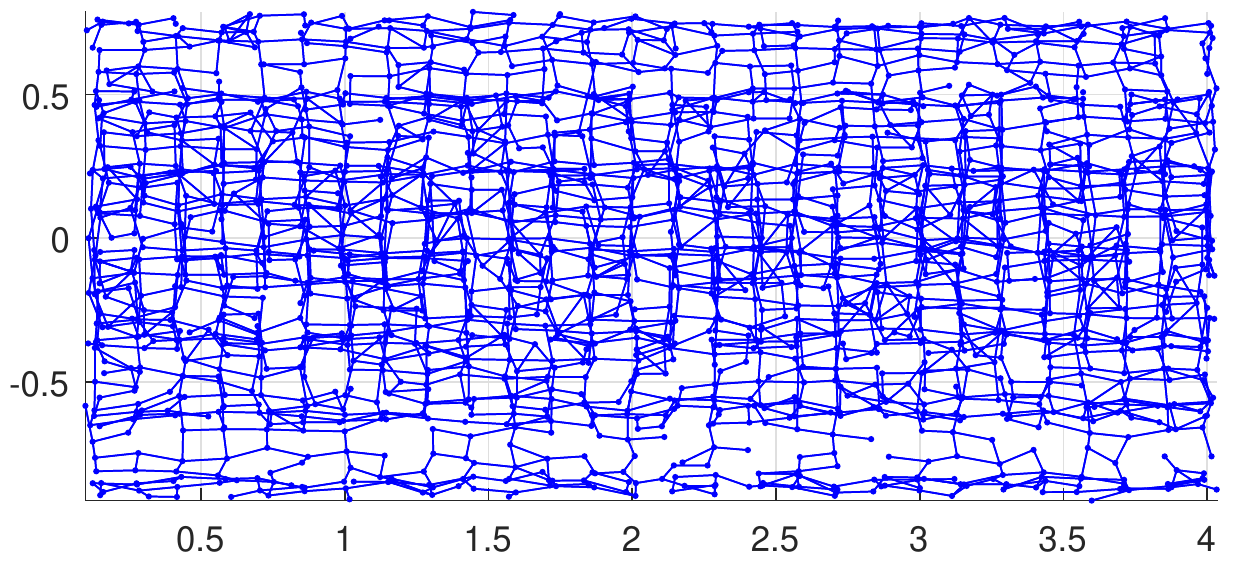}
    \end{minipage}\hfill
     \begin{minipage}{0.33\textwidth}
        \centering
\includegraphics[width=.6\textwidth]{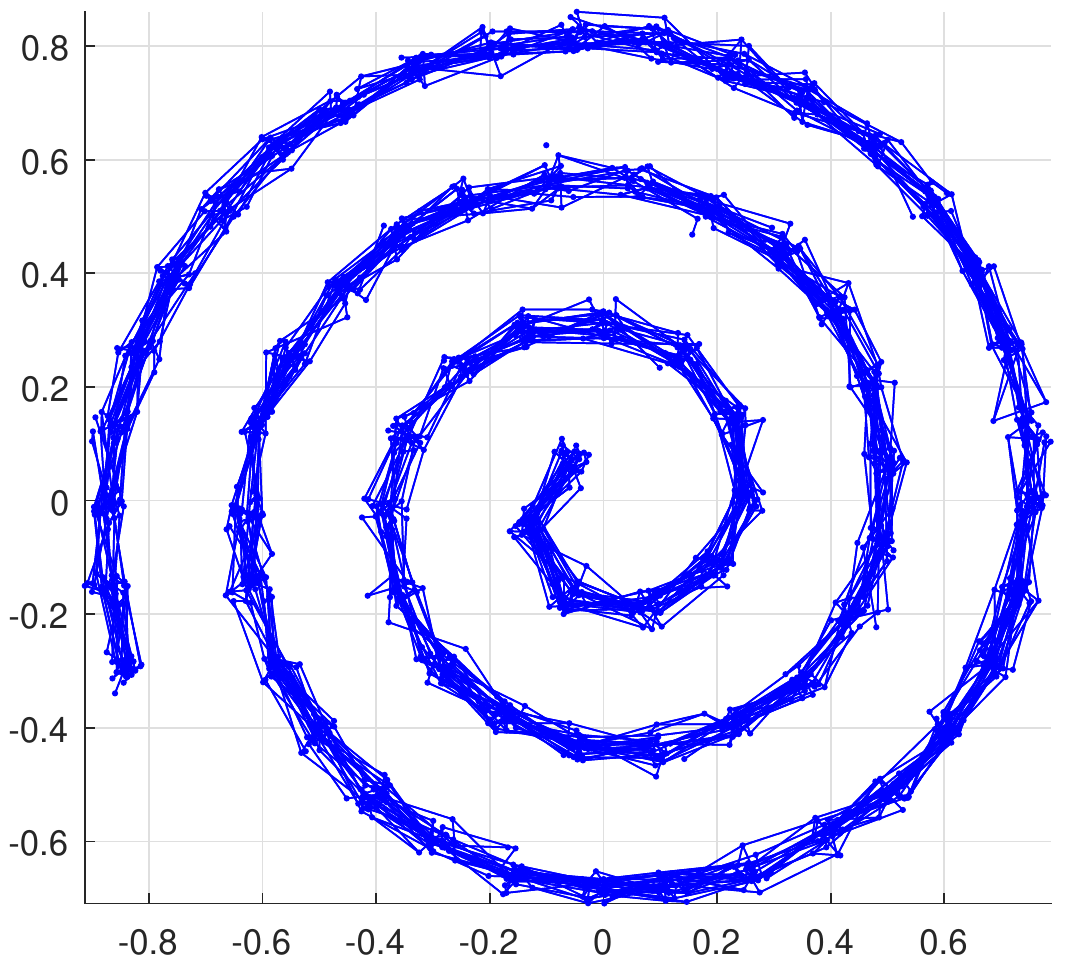}
    \end{minipage}\hfill
    
    \leftline{\hskip 0.00cm (d)} 
    \begin{minipage}{0.33\textwidth}
        \centering
\includegraphics[width=.7\textwidth]{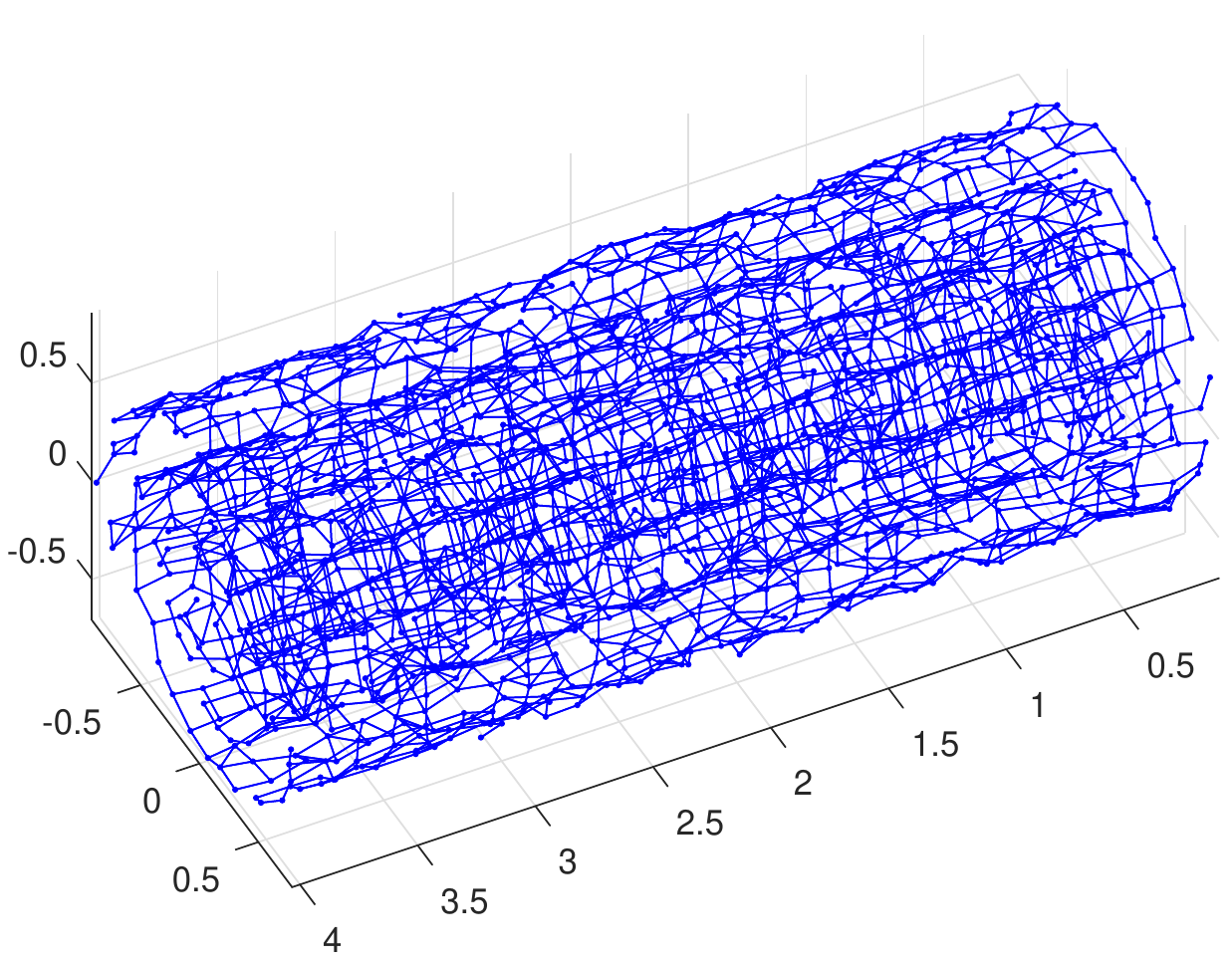}
    \end{minipage}\hfill
    \begin{minipage}{0.33\textwidth}
        \centering
\includegraphics[width=.7\textwidth]{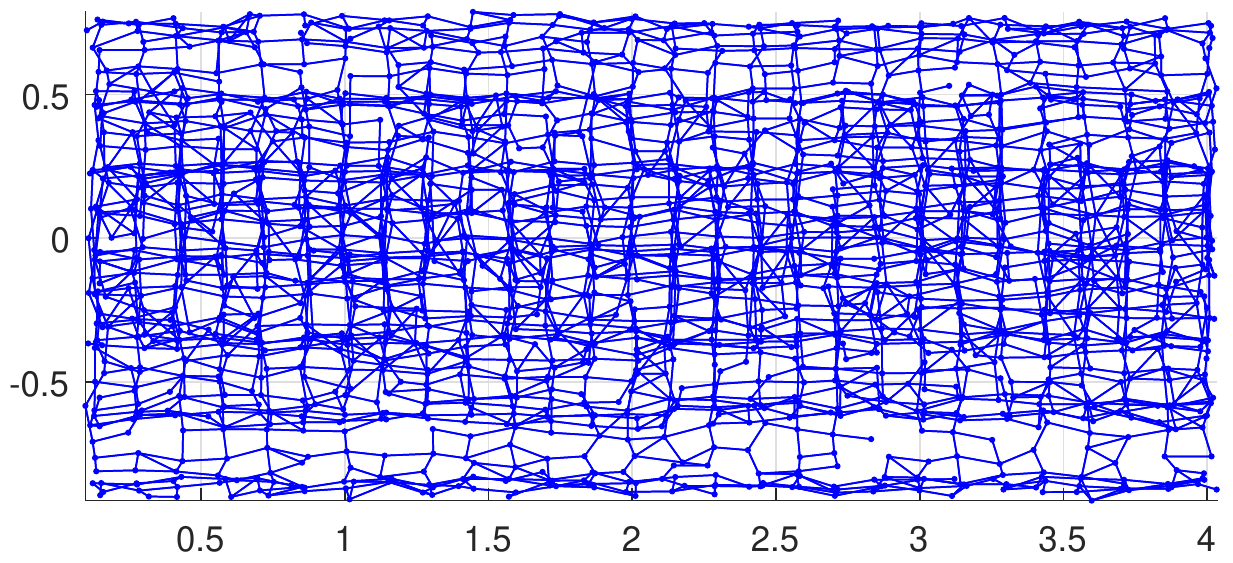}
    \end{minipage}\hfill
    \begin{minipage}{0.33\textwidth}
        \centering
\includegraphics[width=.6\textwidth]{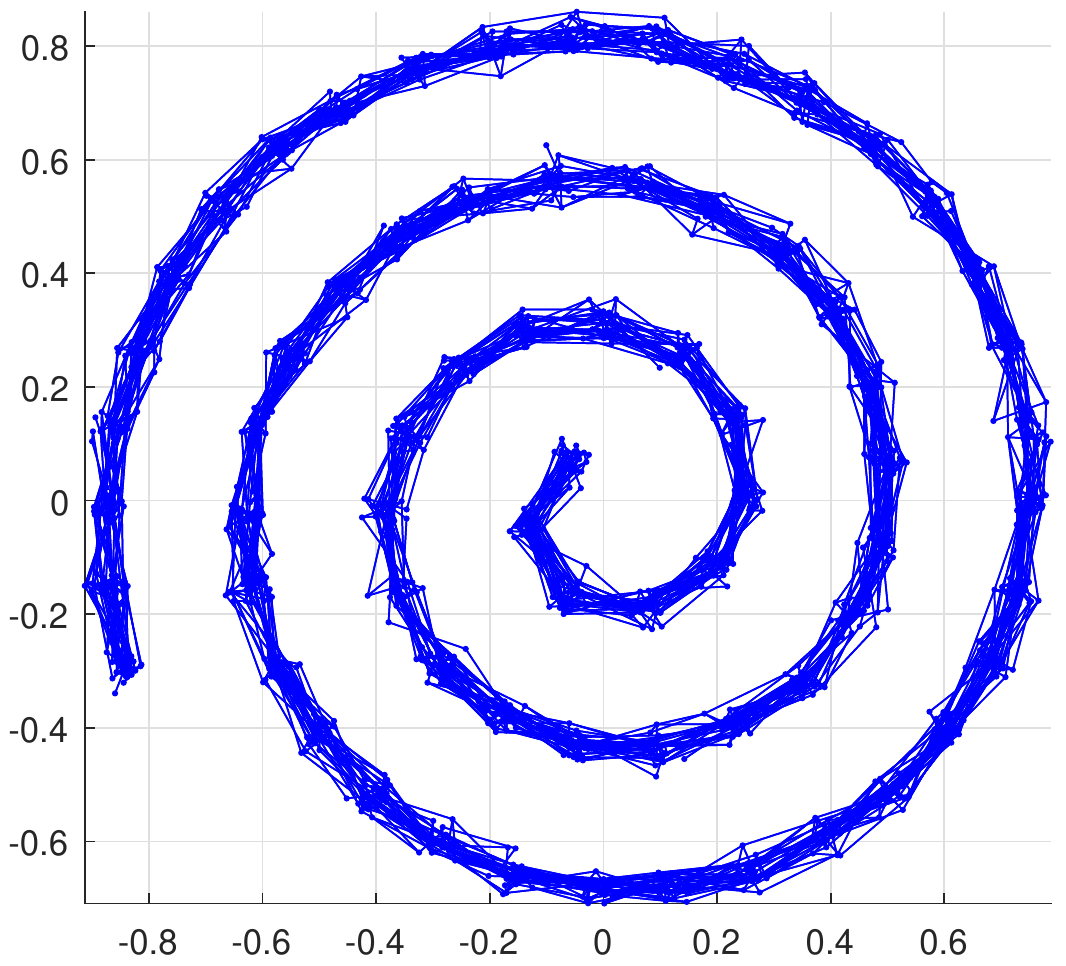}
    \end{minipage}\hfill
    
    \leftline{\hskip 0.00cm (e)} 
    \begin{minipage}{0.33\textwidth}
        \centering
\includegraphics[width=.7\textwidth]{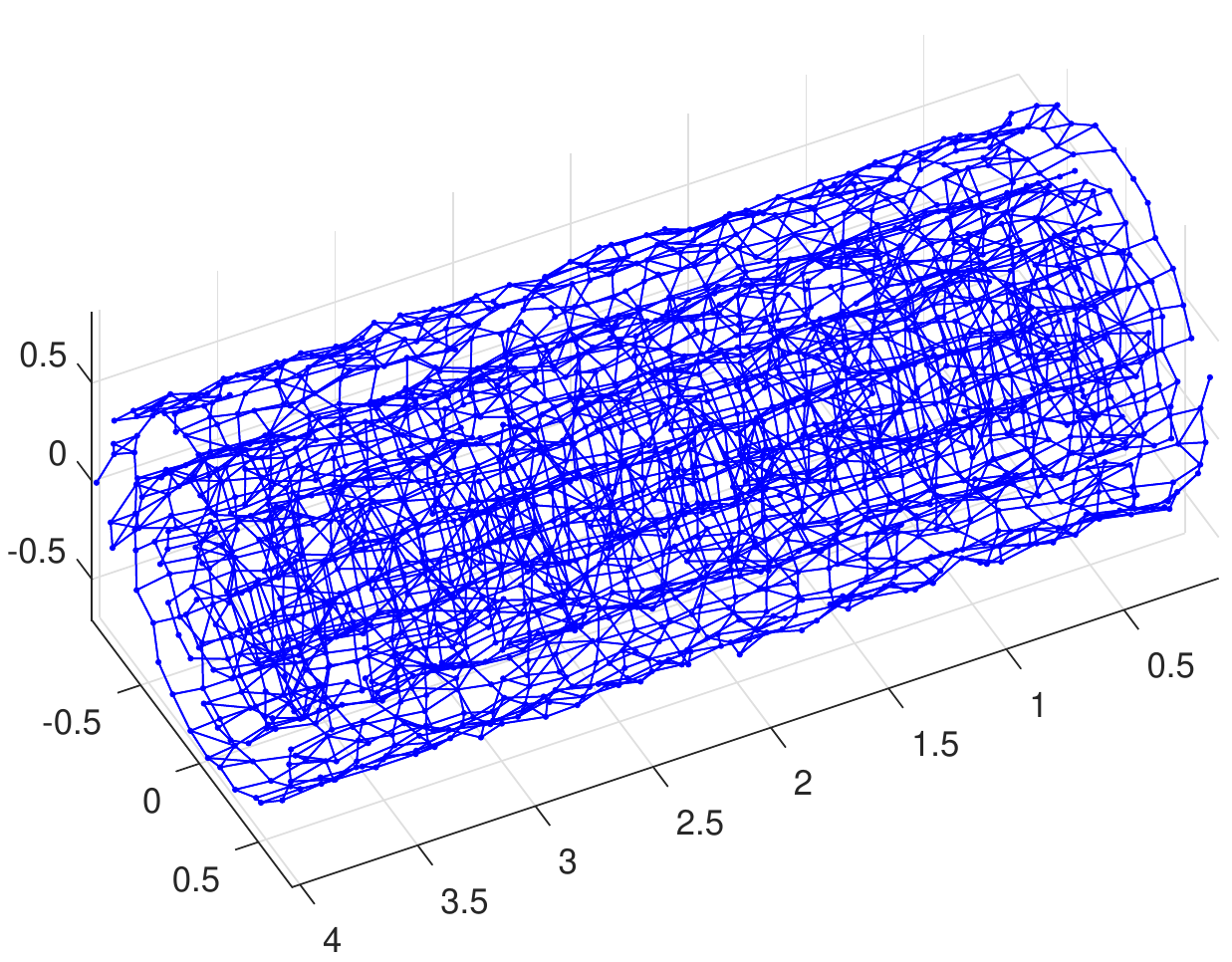}
    \end{minipage}\hfill
     \begin{minipage}{0.33\textwidth}
        \centering
\includegraphics[width=.7\textwidth]{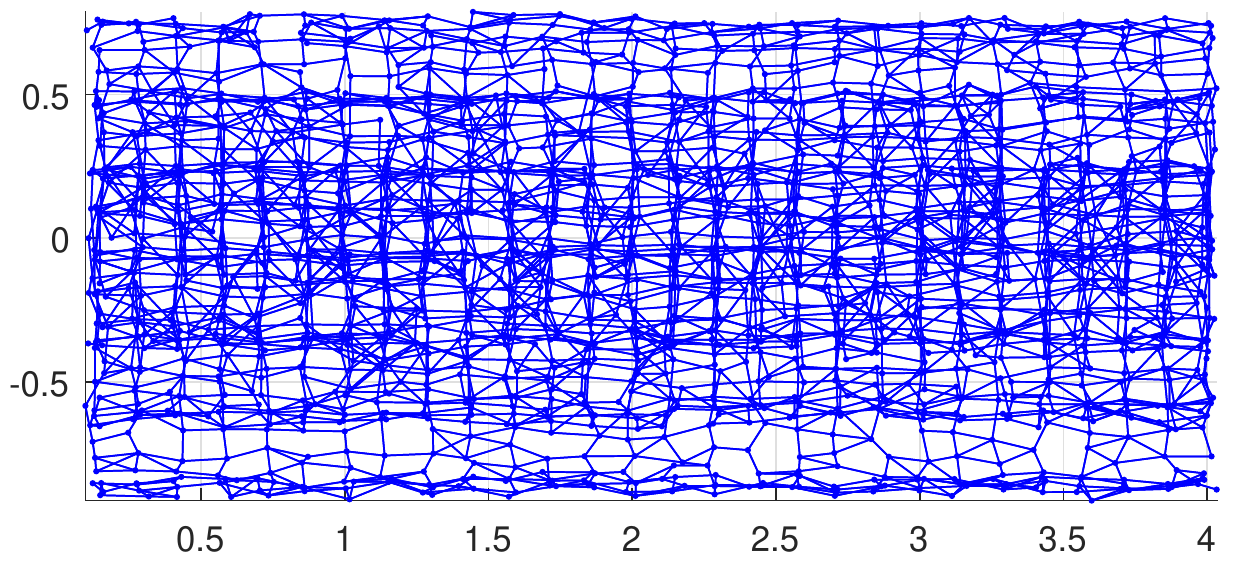}
    \end{minipage}\hfill
     \begin{minipage}{0.33\textwidth}
        \centering
\includegraphics[width=.6\textwidth]{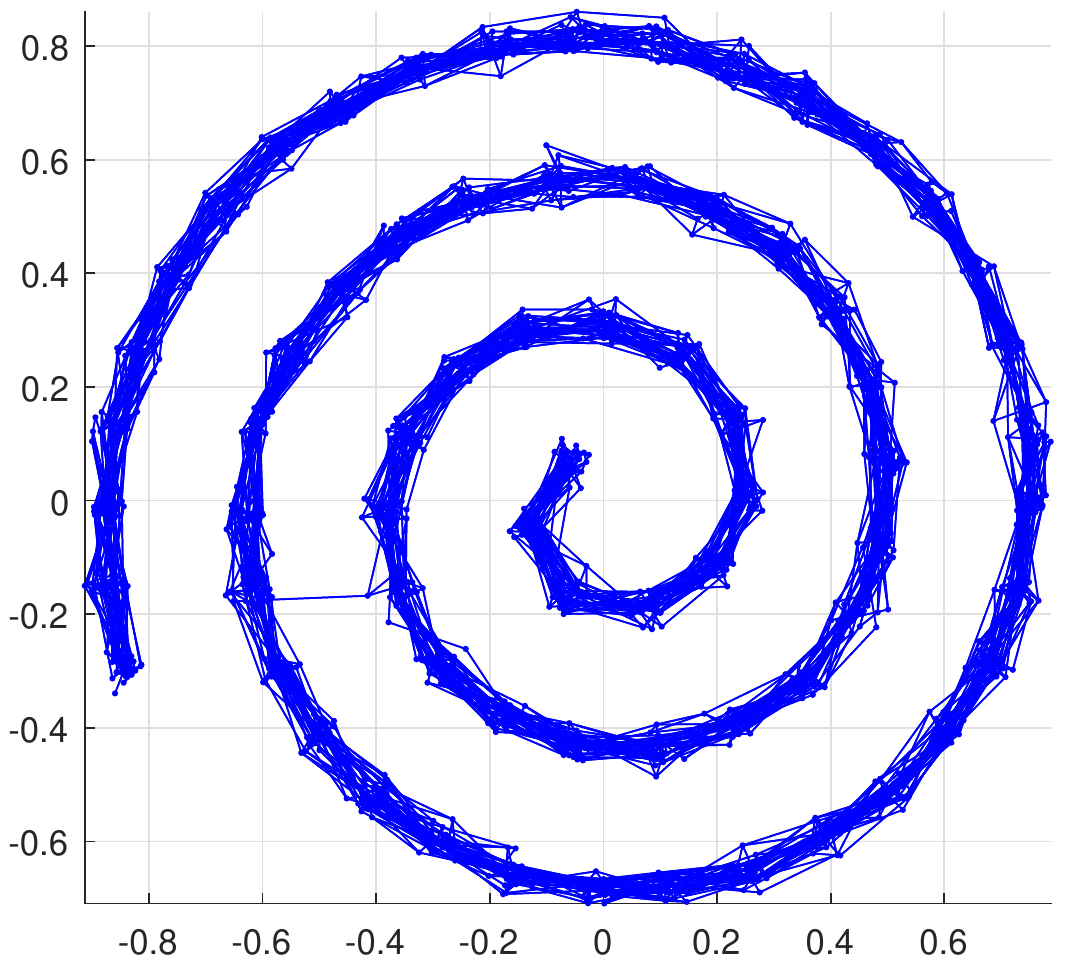}
    \end{minipage}\hfill
\caption{The $\epsilon$-neighbourhood graphs on a noisy point sample from a Swiss roll with $S/N=35$ for (a) $\epsilon=0.14$, (b) $\epsilon=0.15$, (c) $\epsilon=0.16$, (d) $\epsilon=0.17$, and (e) $\epsilon=0.18$. (We show three views of each plot.) If $\epsilon$ is too small (e.g.~$\epsilon=0.14$), the $\epsilon$-neighbourhood graph does not adequately `cover' the underlying Swiss roll. If $\epsilon$ is too large (e.g.~$\epsilon=0.18$), the $\epsilon$-neighbourhood graph includes inter-sheet edges, and thus does not represent the underlying Swiss roll. However, there exists a range of $\epsilon$ for which the $\epsilon$-neighbourhood graph covers the underlying surface and does not include inter-sheet edges, thus providing an authentic representation of the underlying Swiss roll. In other words, the pairwise distances between points whose corresponding nodes are adjacent in the $\epsilon$-neighbourhood graph for such $\epsilon$ approximate the actual geodesic sufficiently, and approximate pairwise geodesic distances between other point pairs can be inferred through the Isomap algorithm. This is an example of a data set for which Isomap is suitable as a manifold-learning technique with a careful choice of $\epsilon$.}
\label{epsilon_success}
\end{figure}

\begin{figure}[H]
\centering
    
    \leftline{\hskip 0.00cm (a)} 
    \begin{minipage}{0.33\textwidth}
        \centering
\includegraphics[width=.7\textwidth]{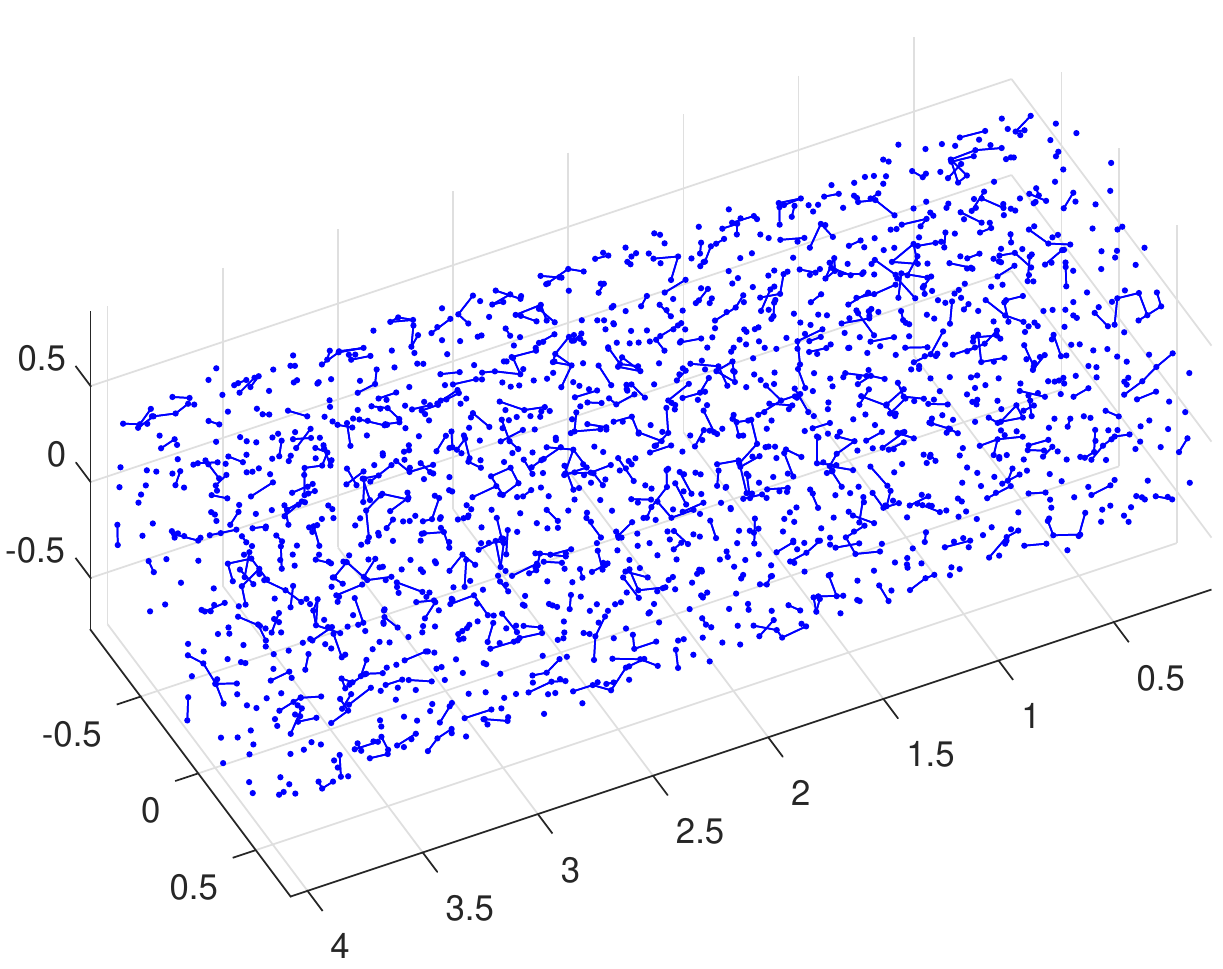}
    \end{minipage}\hfill
    \begin{minipage}{0.33\textwidth}
        \centering
\includegraphics[width=.7\textwidth]{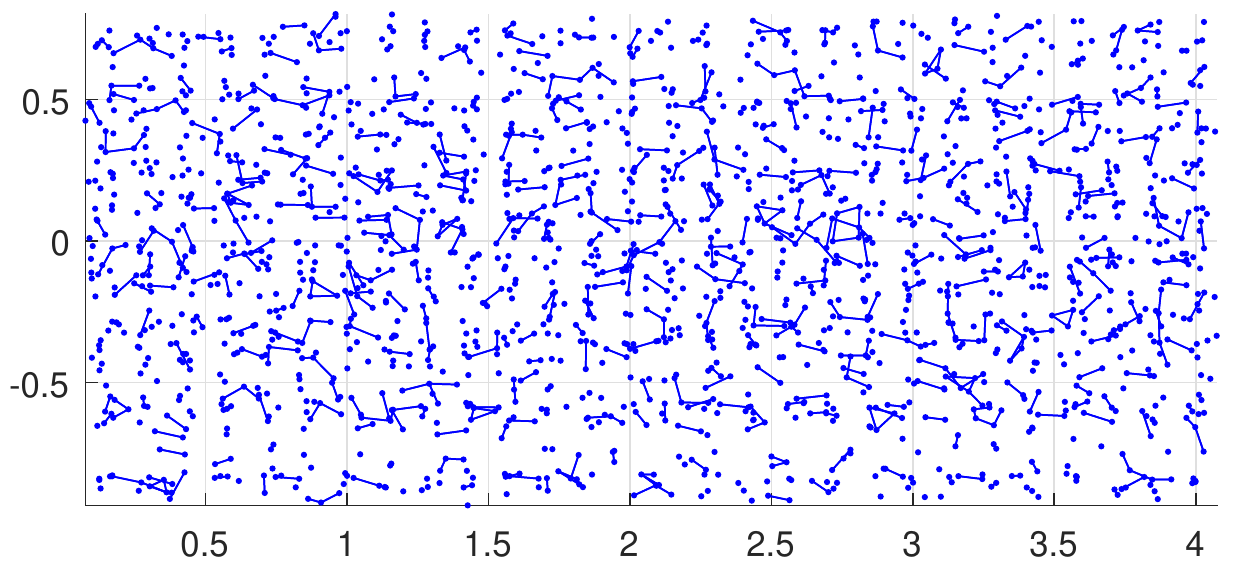}
    \end{minipage}\hfill
\begin{minipage}{0.33\textwidth}
        \centering
\includegraphics[width=.6\textwidth]{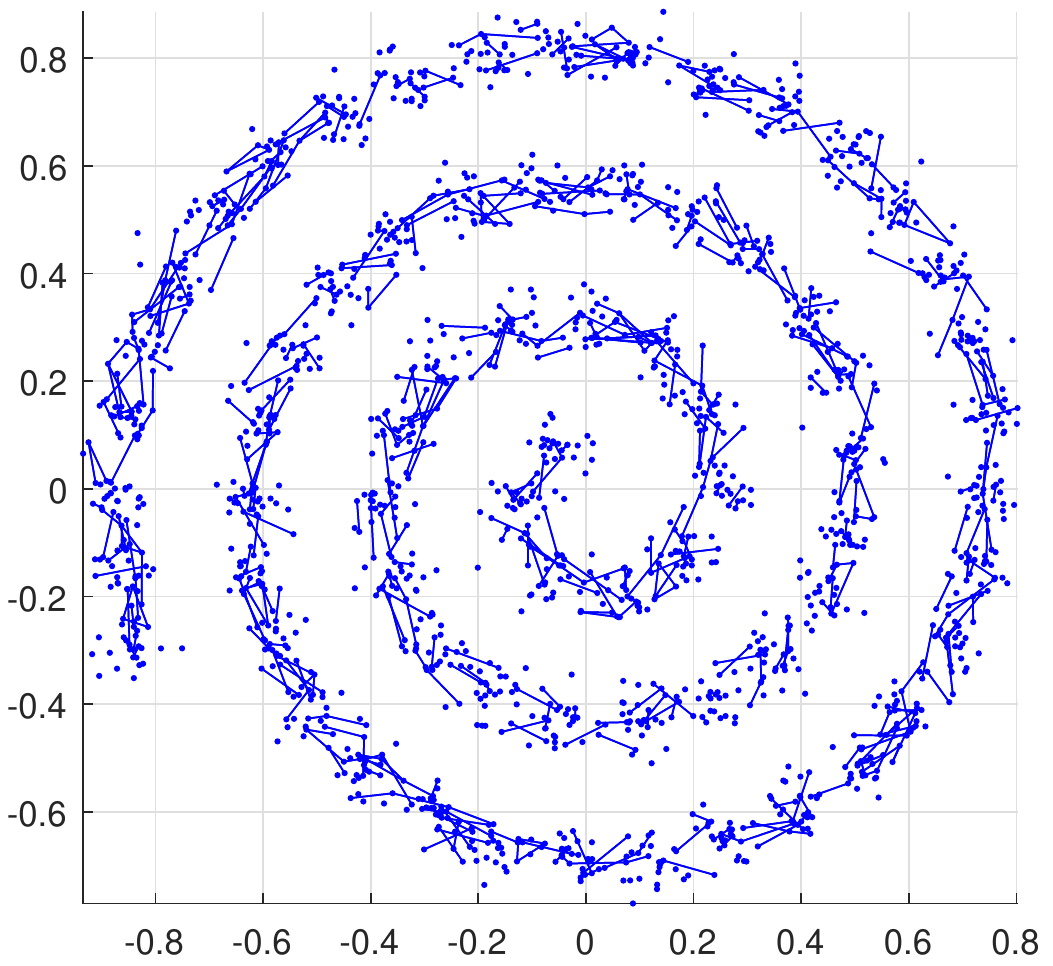}
    \end{minipage}\hfill
        
    \leftline{\hskip 0.00cm (b)} 
    \begin{minipage}{0.33\textwidth}
        \centering
\includegraphics[width=.7\textwidth]{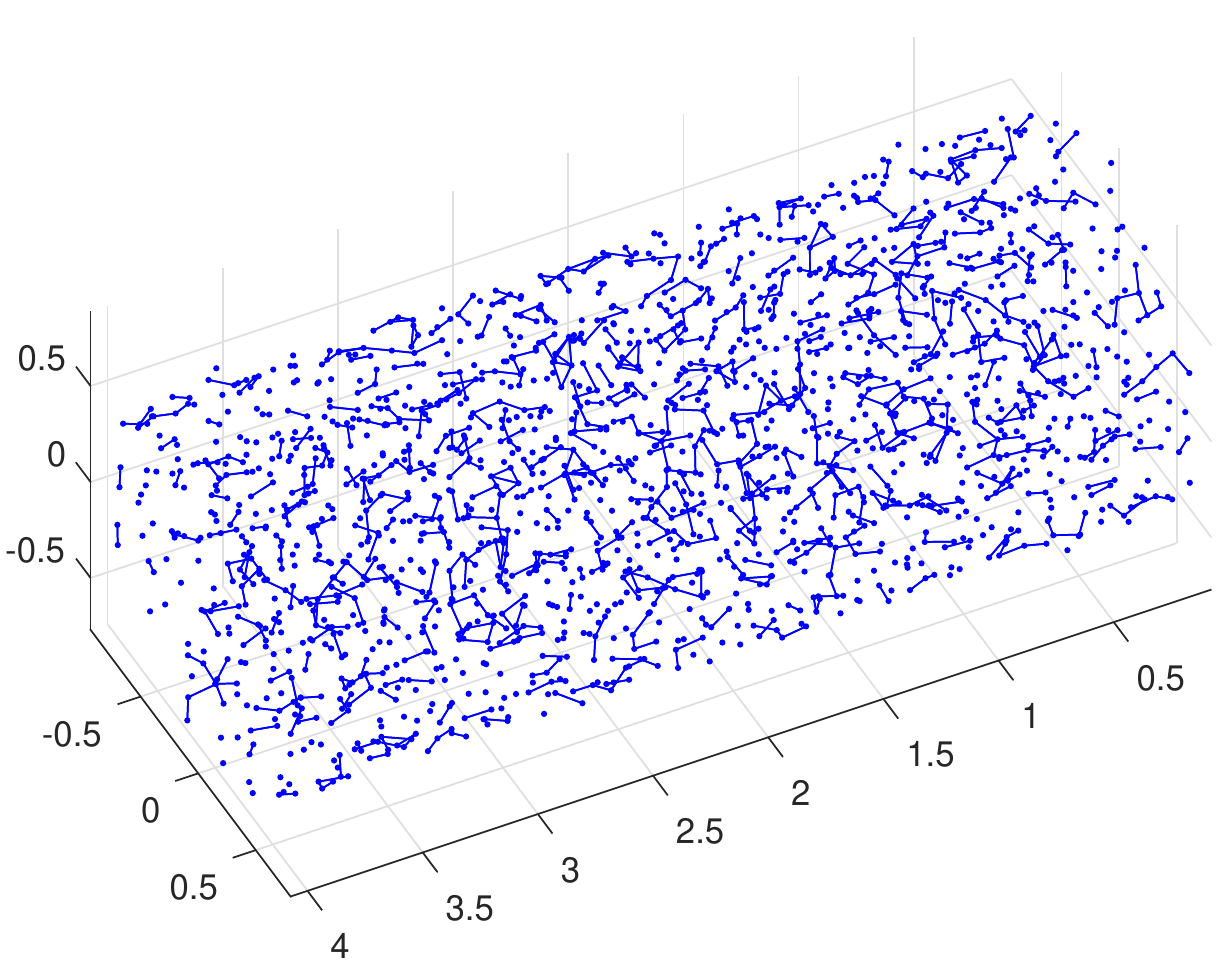}
    \end{minipage}\hfill
    \begin{minipage}{0.33\textwidth}
        \centering
\includegraphics[width=.7\textwidth]{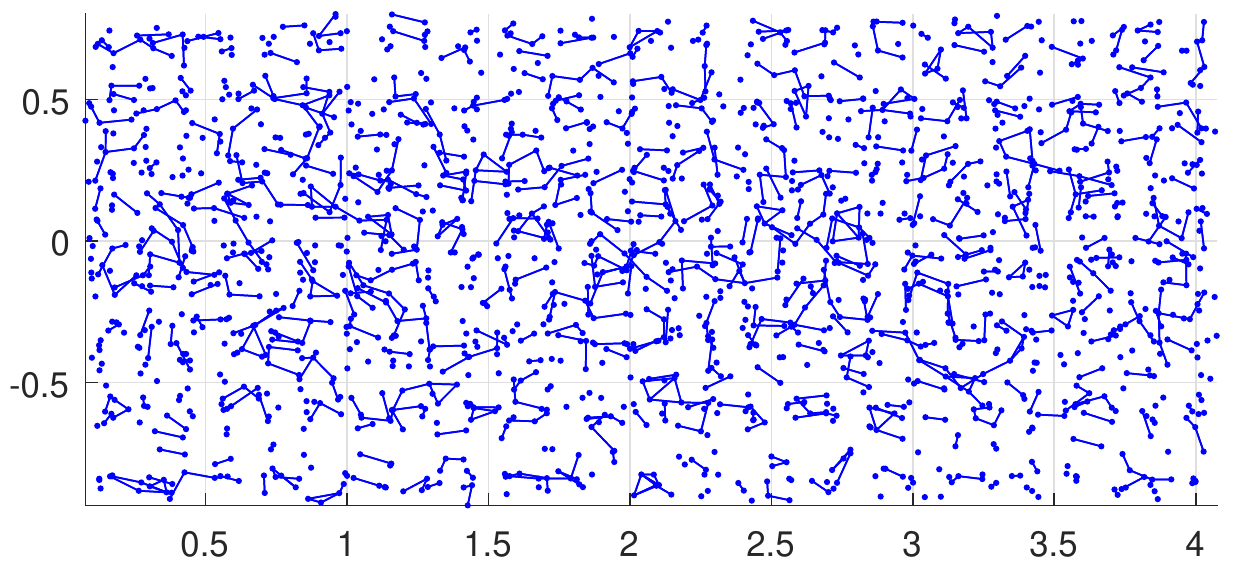}
    \end{minipage}\hfill
    \begin{minipage}{0.33\textwidth}
        \centering
\includegraphics[width=.6\textwidth]{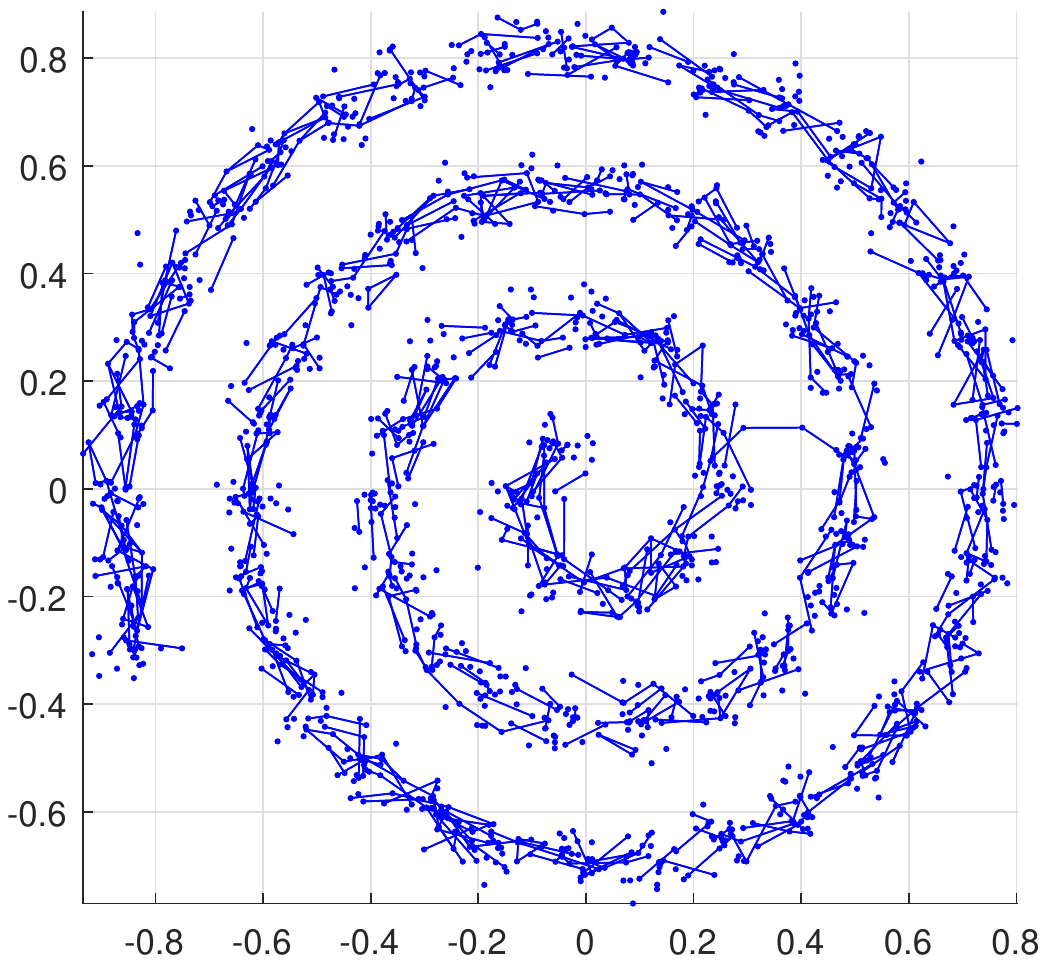}
    \end{minipage}\hfill
    
    \leftline{\hskip 0.00cm (c)} 
    \begin{minipage}{0.33\textwidth}
        \centering
\includegraphics[width=.7\textwidth]{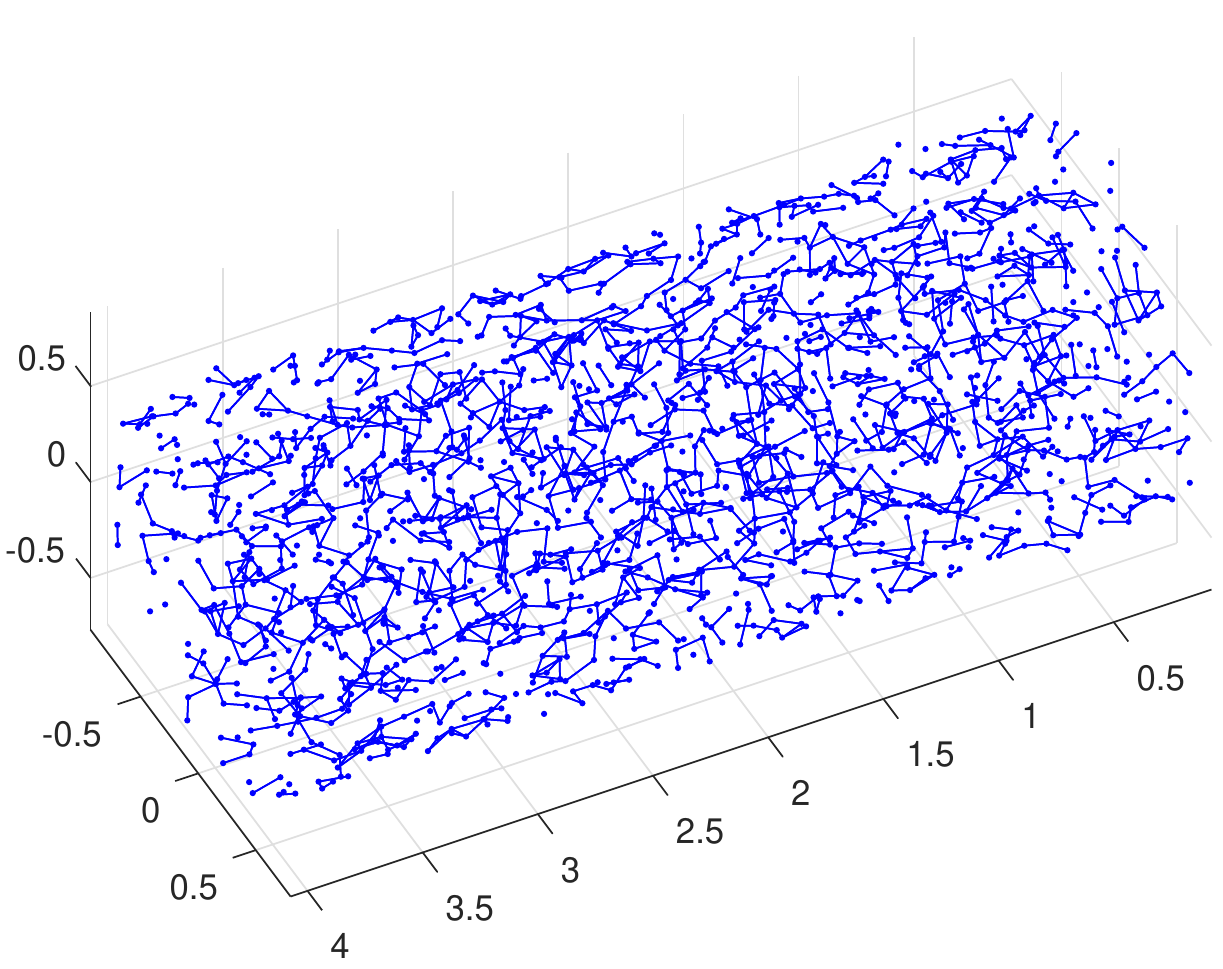}
    \end{minipage}\hfill
     \begin{minipage}{0.33\textwidth}
        \centering
\includegraphics[width=.7\textwidth]{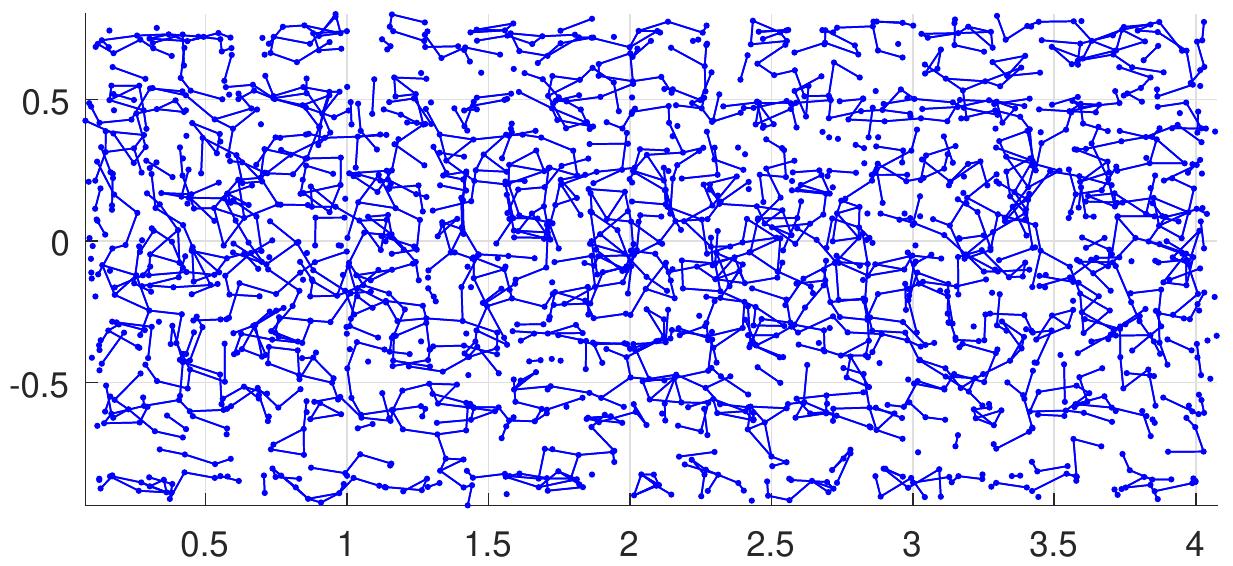}
    \end{minipage}\hfill
     \begin{minipage}{0.33\textwidth}
        \centering
\includegraphics[width=.6\textwidth]{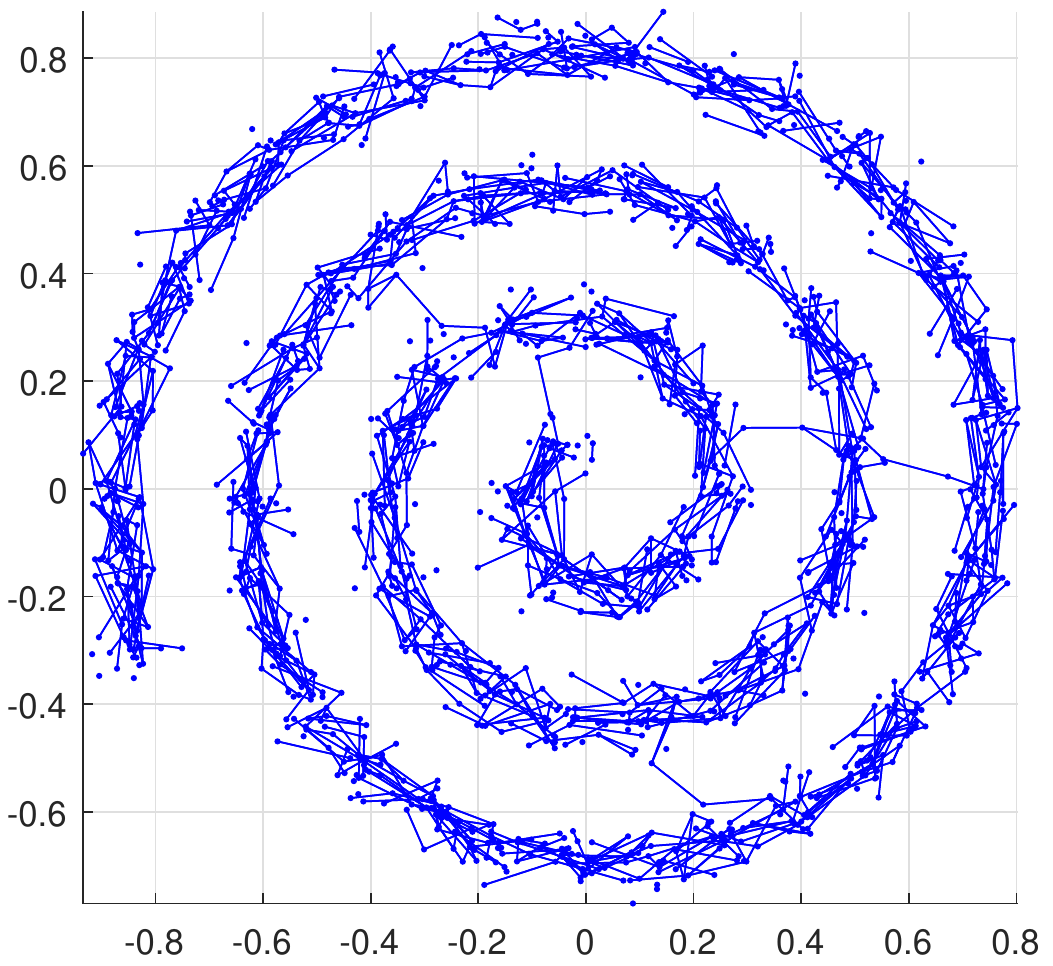}
    \end{minipage}\hfill
    
    \leftline{\hskip 0.00cm (d)} 
    \begin{minipage}{0.33\textwidth}
        \centering
\includegraphics[width=.7\textwidth]{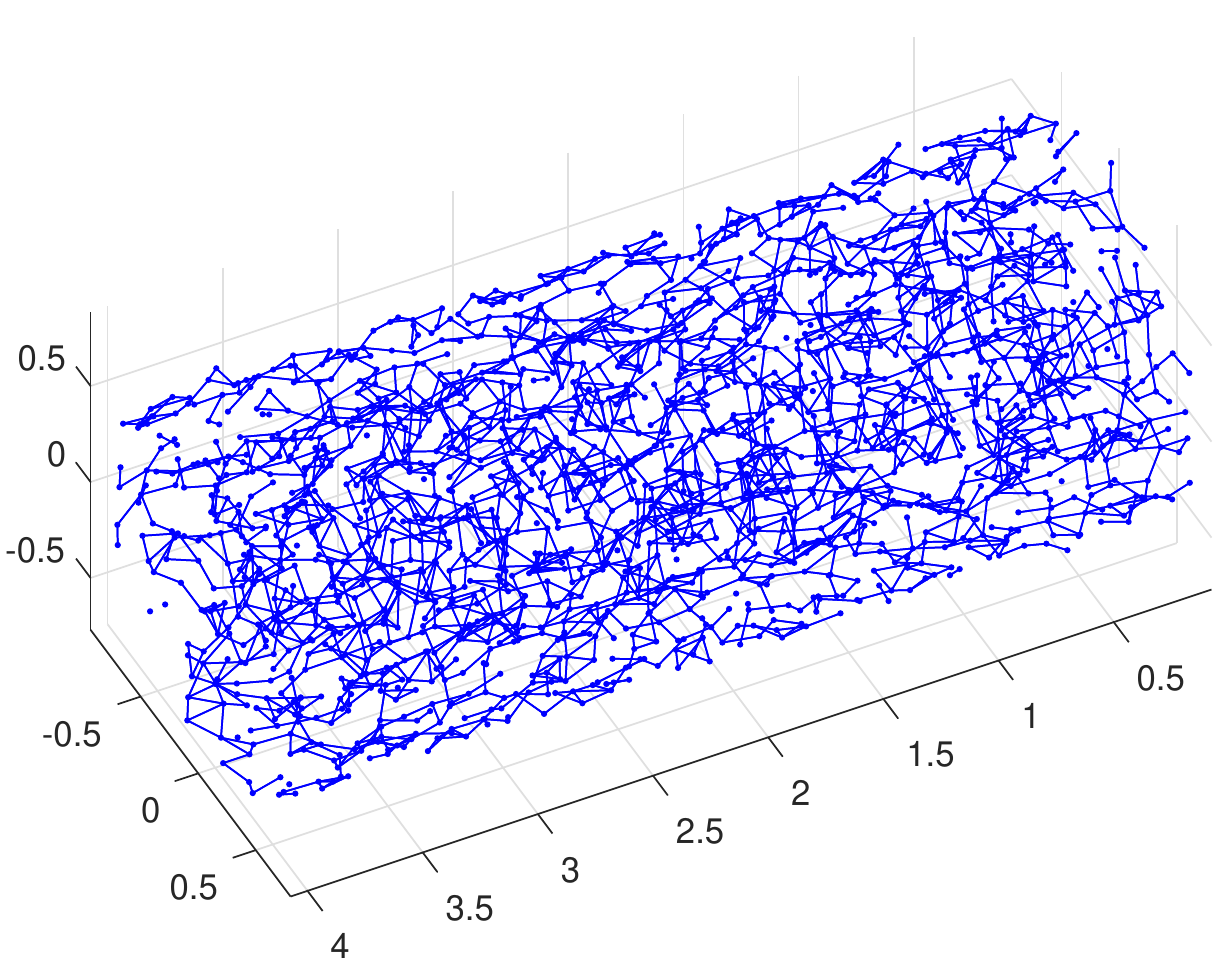}
    \end{minipage}\hfill
    \begin{minipage}{0.33\textwidth}
        \centering
\includegraphics[width=.7\textwidth]{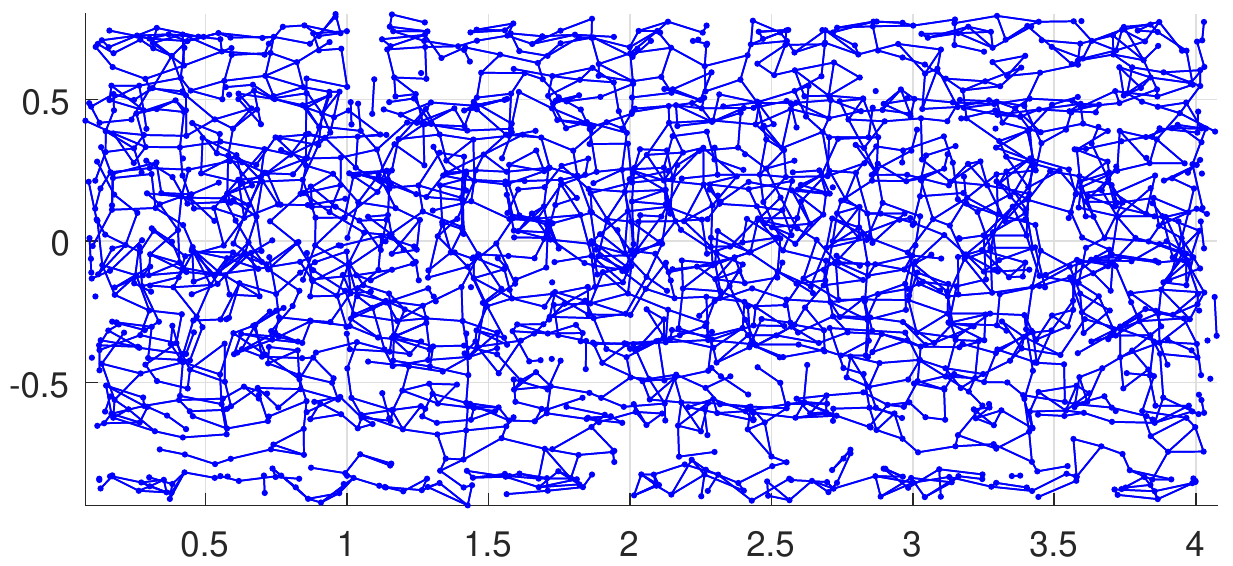}
    \end{minipage}\hfill
    \begin{minipage}{0.33\textwidth}
        \centering
\includegraphics[width=.6\textwidth]{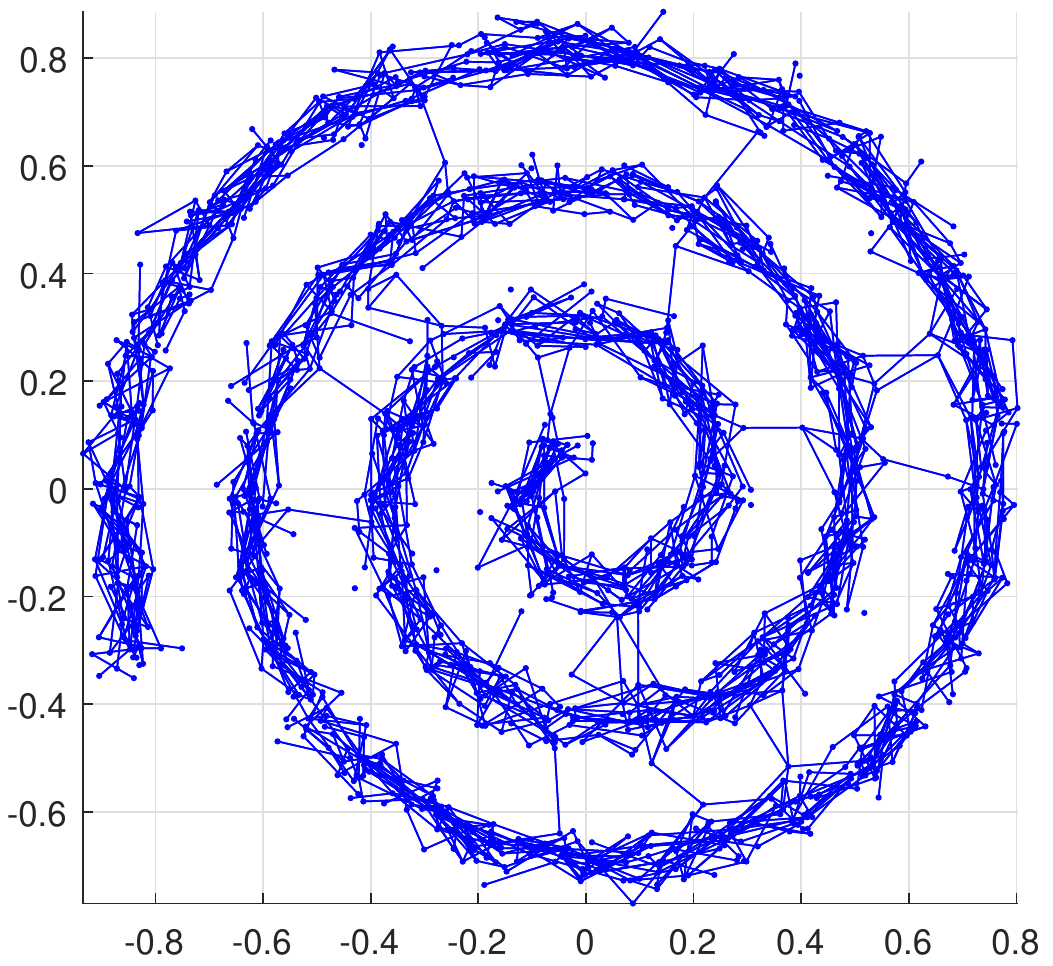}
    \end{minipage}\hfill
    
    \leftline{\hskip 0.00cm (e)} 
    \begin{minipage}{0.33\textwidth}
        \centering
\includegraphics[width=.7\textwidth]{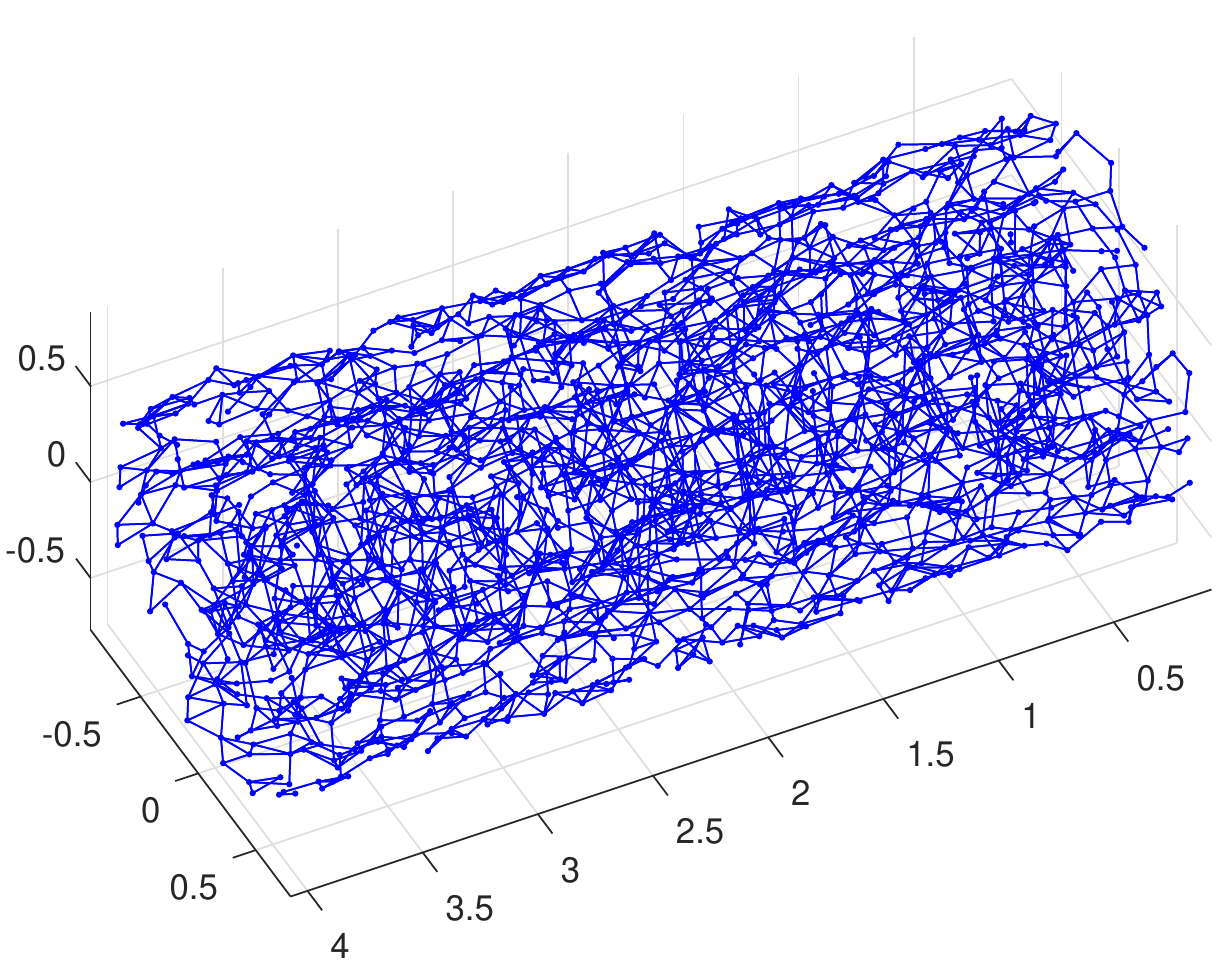}
    \end{minipage}\hfill
     \begin{minipage}{0.33\textwidth}
        \centering
\includegraphics[width=.7\textwidth]{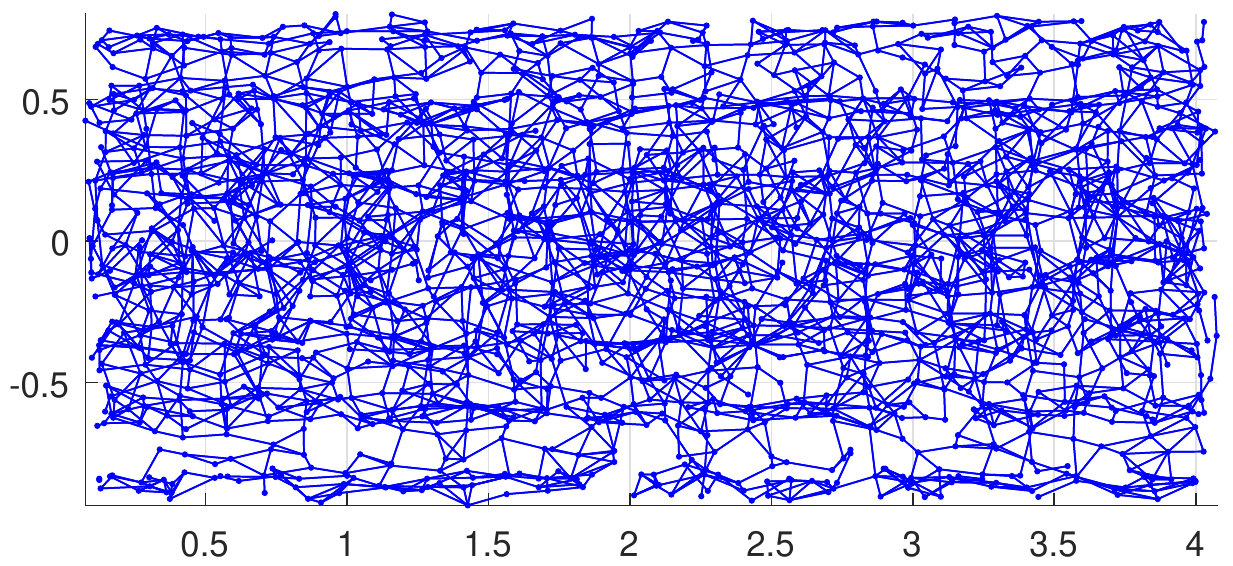}
    \end{minipage}\hfill
     \begin{minipage}{0.33\textwidth}
        \centering
\includegraphics[width=.6\textwidth]{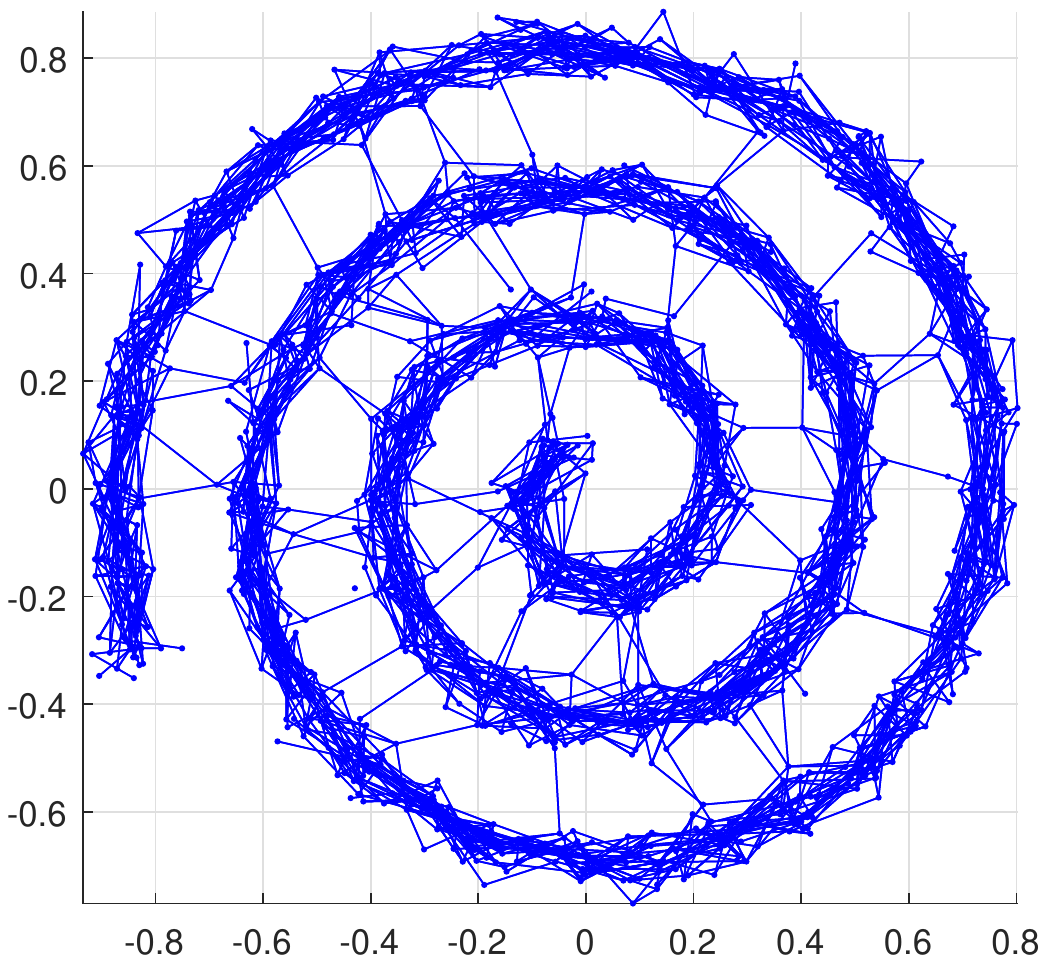}
    \end{minipage}\hfill
    
    \leftline{\hskip 0.00cm (f)} 
    \begin{minipage}{0.33\textwidth}
        \centering
\includegraphics[width=.7\textwidth]{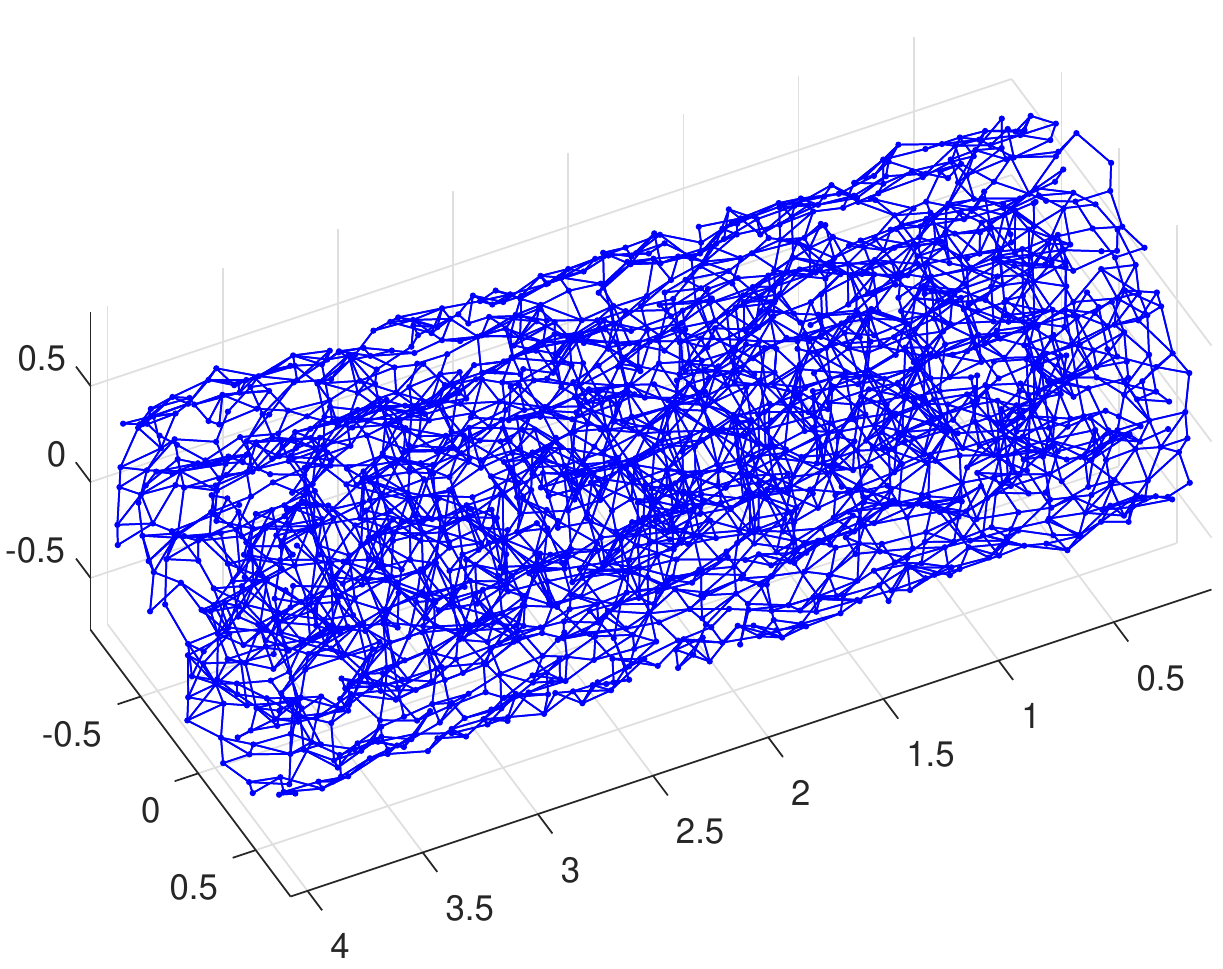}
    \end{minipage}\hfill
     \begin{minipage}{0.33\textwidth}
        \centering
\includegraphics[width=.7\textwidth]{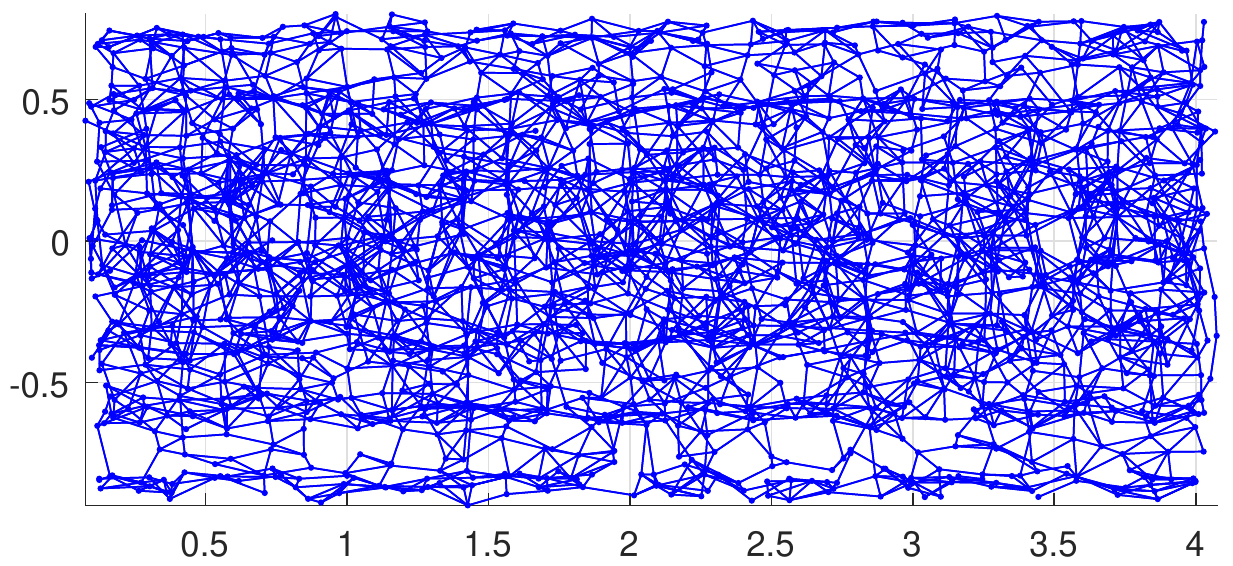}
    \end{minipage}\hfill
     \begin{minipage}{0.33\textwidth}
        \centering
\includegraphics[width=.6\textwidth]{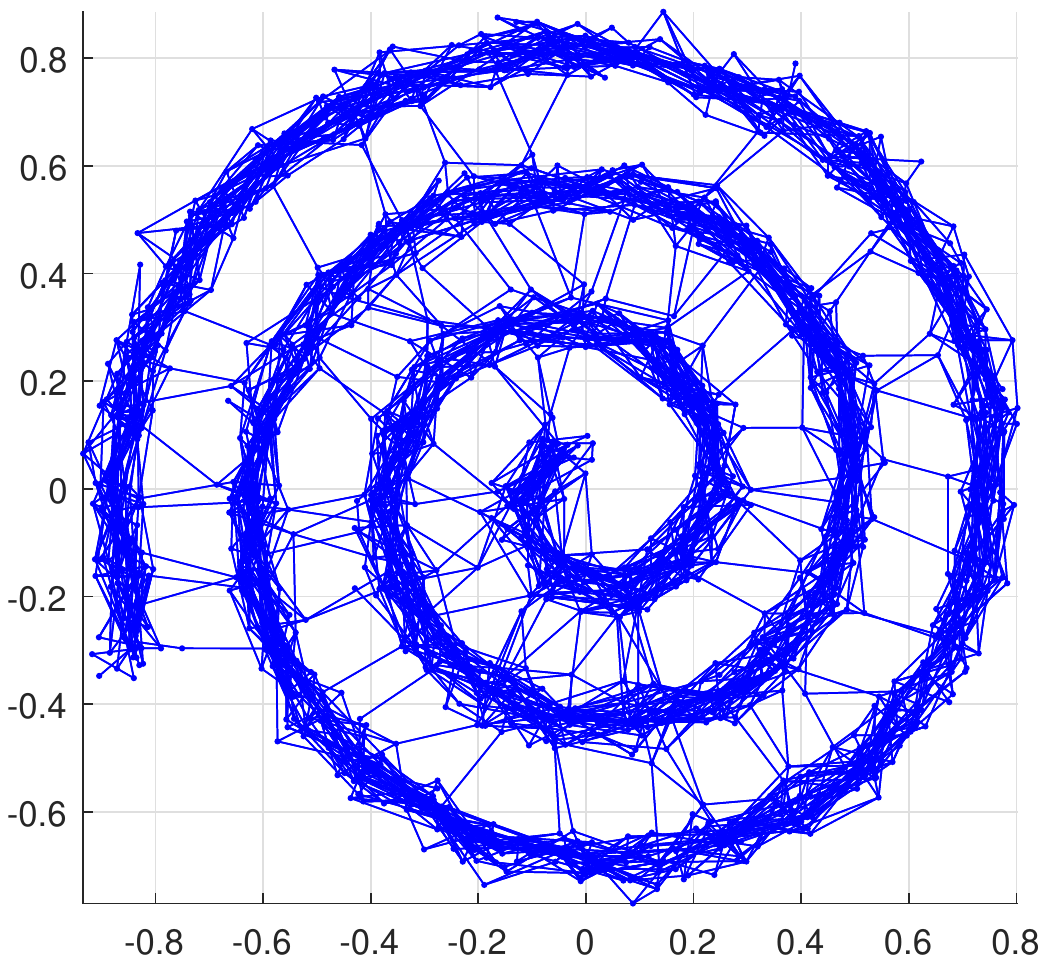}
    \end{minipage}\hfill
\caption{The $\epsilon$-neighbourhood graphs on a noisy point sample from a Swiss roll with $S/N=30$ for (a) $\epsilon=0.11$, (b) $\epsilon=0.12$, (c) $\epsilon=0.14$, (d) $\epsilon=0.16$, (e) $\epsilon=0.18$, and (f) $\epsilon=0.19$. (We show three views of each plot.) For small $\epsilon$, the $\epsilon$-neighbourhood graph does not adequately `cover' the underlying Swiss roll. As $\epsilon$ increases, noisy inter-sheet edges appear (for e.g.~$\epsilon=0.12$) before $\epsilon$ is large enough for the neighbourhood graph to adequately `cover' the underlying Swiss roll. This is an example of a data set for which Isomap cannot be used successfully as a manifold-learning technique with any choice of $\epsilon$.}
\label{epsilon_fail}
\end{figure}

\begin{figure}[H]
\centering
\leftline{\hskip 0.00cm (a) \hskip 7cm (b)} 
    \begin{minipage}{0.5\textwidth}
\includegraphics[scale=.5]{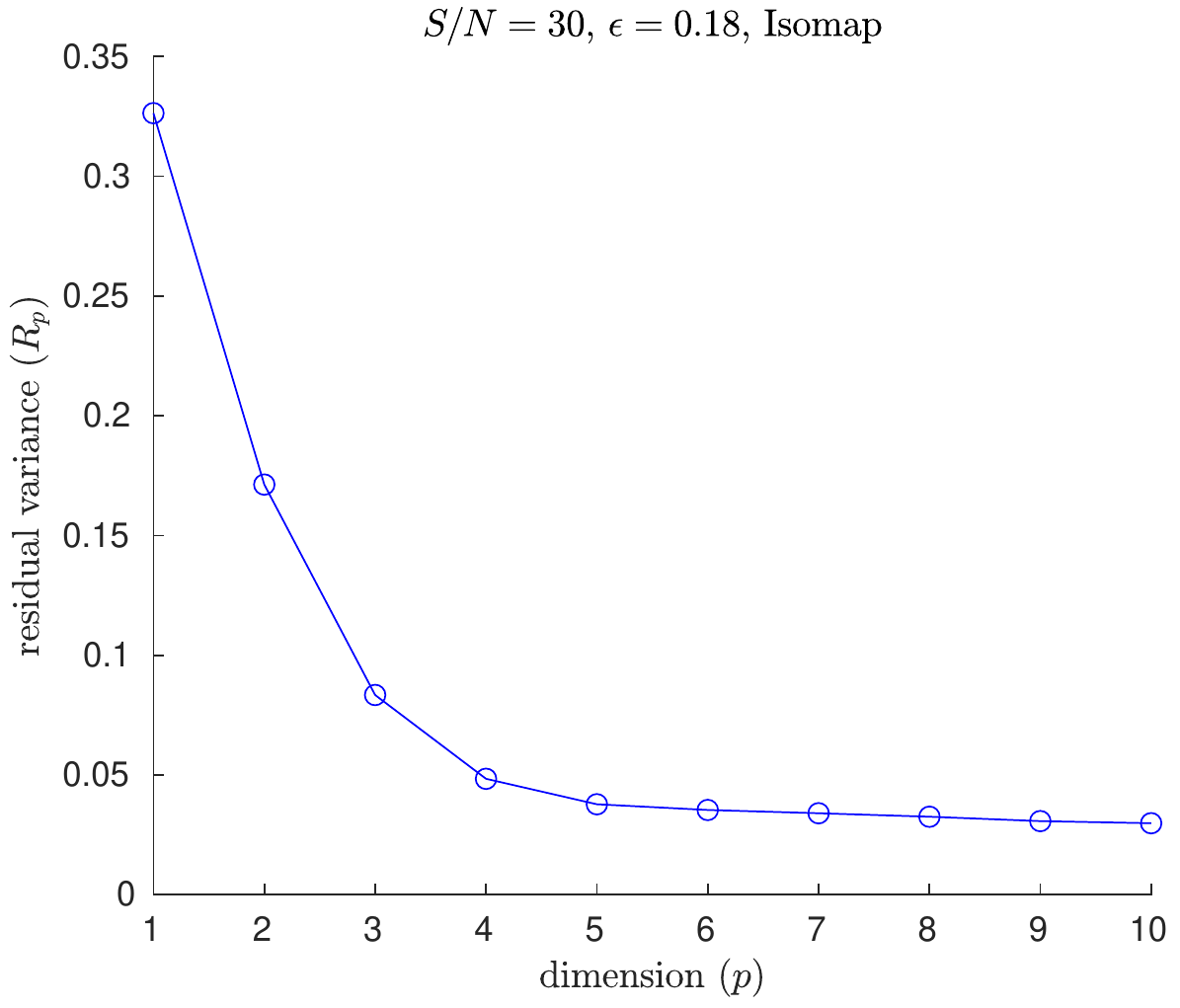}
    \end{minipage}\hfill
     \begin{minipage}{0.5\textwidth}
\includegraphics[scale=.5]{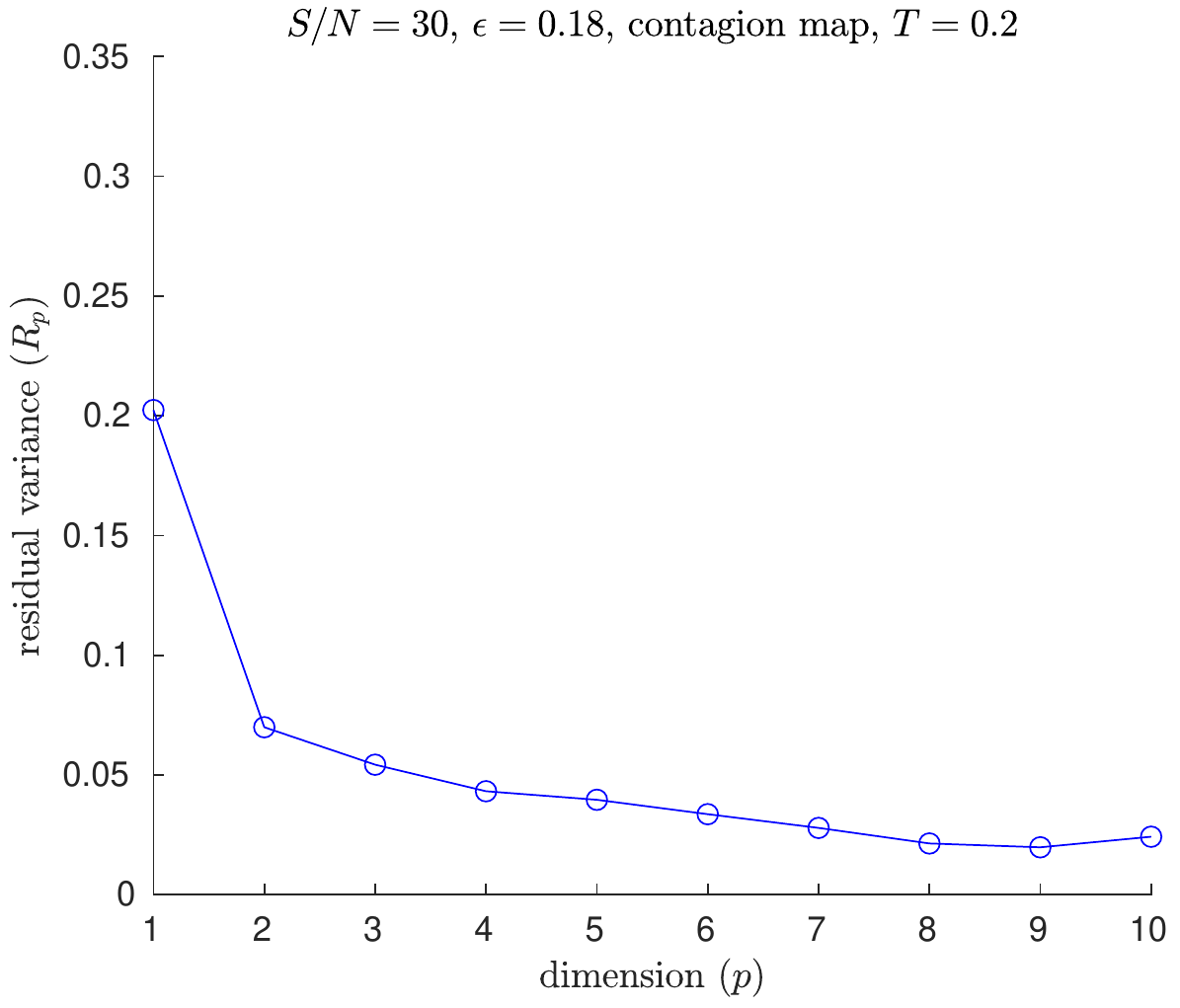}
    \end{minipage}\hfill
\caption{(a) Isomap and (b) contagion map with $T=0.2$ (both starting with a $0.18$-neighbourhood graph) on a noisy sample from a Swiss roll with $S/N=30$. }
\label{Barbara_Swiss_iso_cont}
\end{figure}

See Figures~\ref{epsilon_success} and \ref{epsilon_fail} for examples on the Swiss roll that illustrate the concepts in this paragraph. In particular, Figure~\ref{epsilon_success} shows an example of a data set for which a careful choice of $\epsilon$ generates a neighbourhood graph that both captures the underlying manifold and does not include `short-circuit' edges. By contrast, Figure~\ref{epsilon_fail} shows an example of a data set, for which no choice of $\epsilon$ produces a neighbourhood graph that accurately represents the underlying manifold. Figure~\ref{Barbara_Swiss_iso_cont} shows the residual variances of projecting this data set via MDS to dimensions $1$ to $10$ when based on Isompap, and when based on the contagion map with $T=0.2$ (both starting with a $0.18$-neighbourhood graph). For the contagion map, the residual variance plunges at dimension $2$, thereby correctly identifying the intrinsic dimension of the data. The residual variances for Isomap, on the other hand, only decrease slighty and continuosly across the increasing target dimensions. That is, Isomap fails to see the correct intrinsic structure of the data when starting with an $\epsilon$-neighbourhood graph for $\epsilon=0.18$ (or any other value of $\epsilon$), while contagion map --- with a suitable choice of $\epsilon$ and $T$ --- correctly identifies the underlying structure. For contagion maps to work in this case, the value of $\epsilon$ had to be chosen large enough for the neighbourhood graph to cover the underlying manifold; and the value of $T$ had to be picked small enough to carry the contagion and large enough to be resistant to the unavoidable noisy inter-sheet edges in the $\epsilon$-neighbourhood graph.

\pagebreak

\pagebreak
We now consider $2000$ points of the Swiss roll data set from \cite{Tenenbaum2000}. This data set consists of $20000$ points in total.

\begin{figure}[H]
\centering
     \leftline{\hskip 0.00cm (a) \hskip 7.00cm (b)} 
     \begin{minipage}{0.5\textwidth}
        \centering
\includegraphics[width=.8\textwidth]{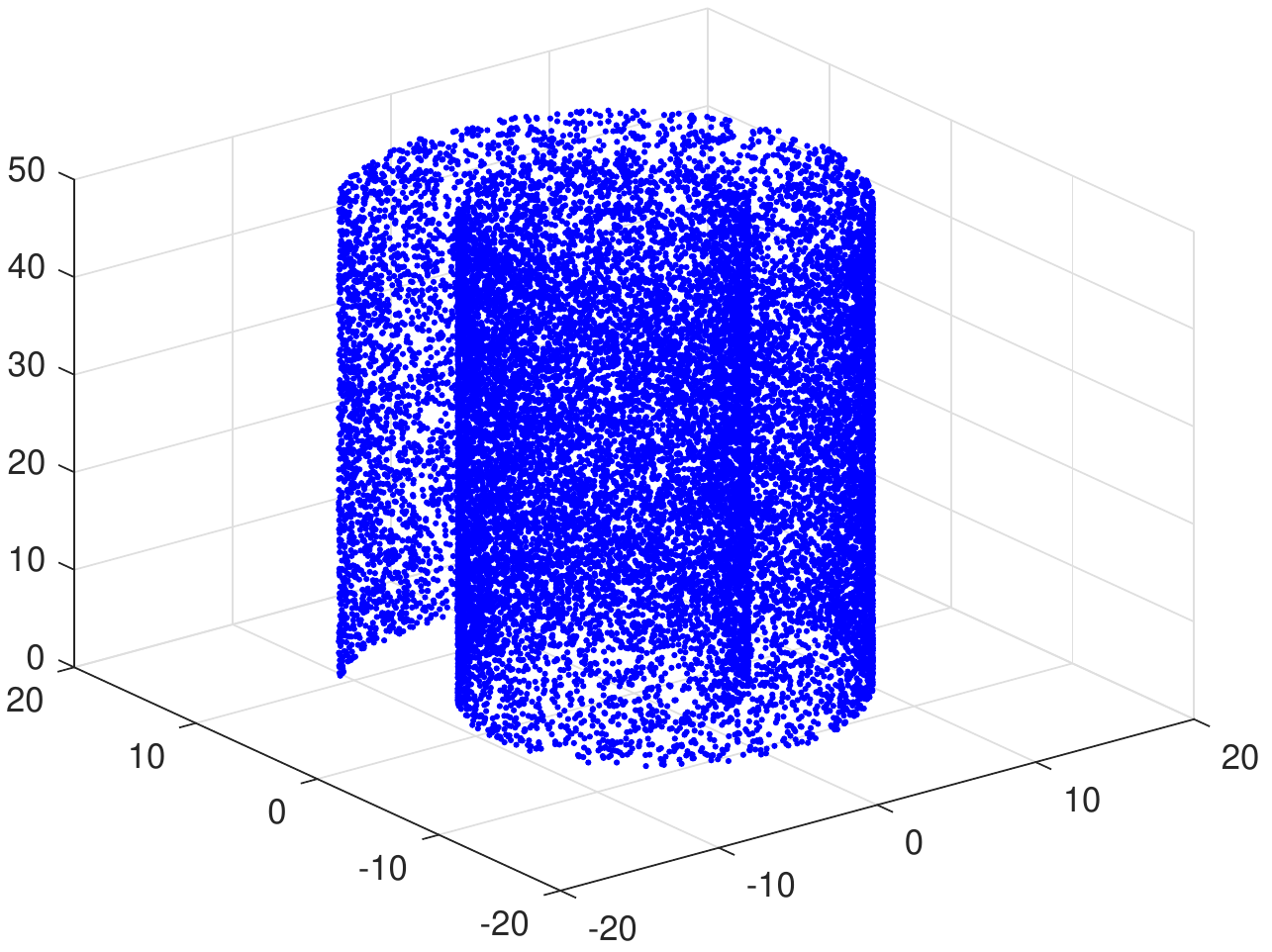}
    \end{minipage}\hfill
    \begin{minipage}{0.5\textwidth}
        \centering
\includegraphics[width=.8\textwidth]{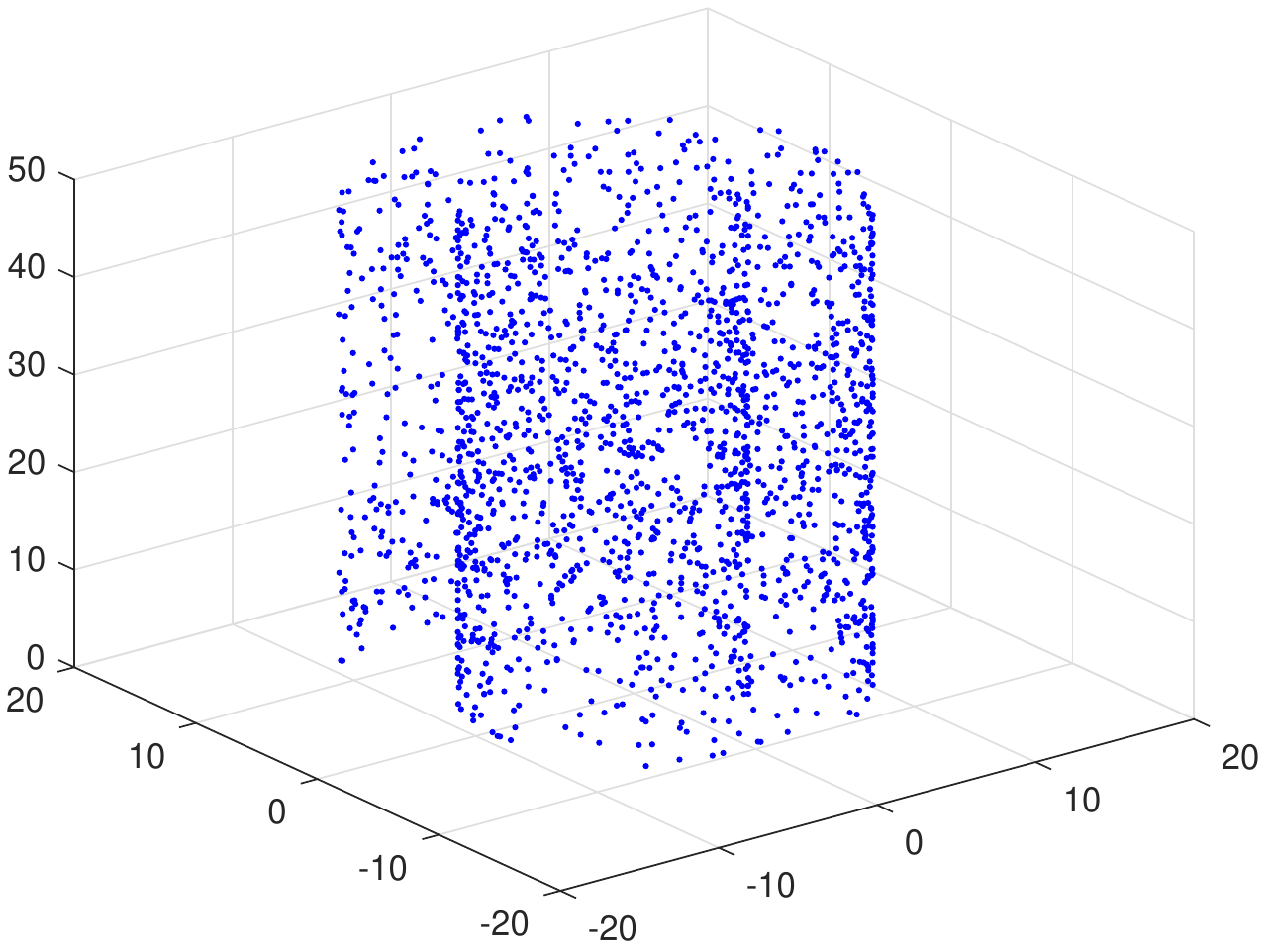}
\end{minipage}\hfill

\begin{minipage}{0.5\textwidth}
        \centering
\includegraphics[width=.8\textwidth]{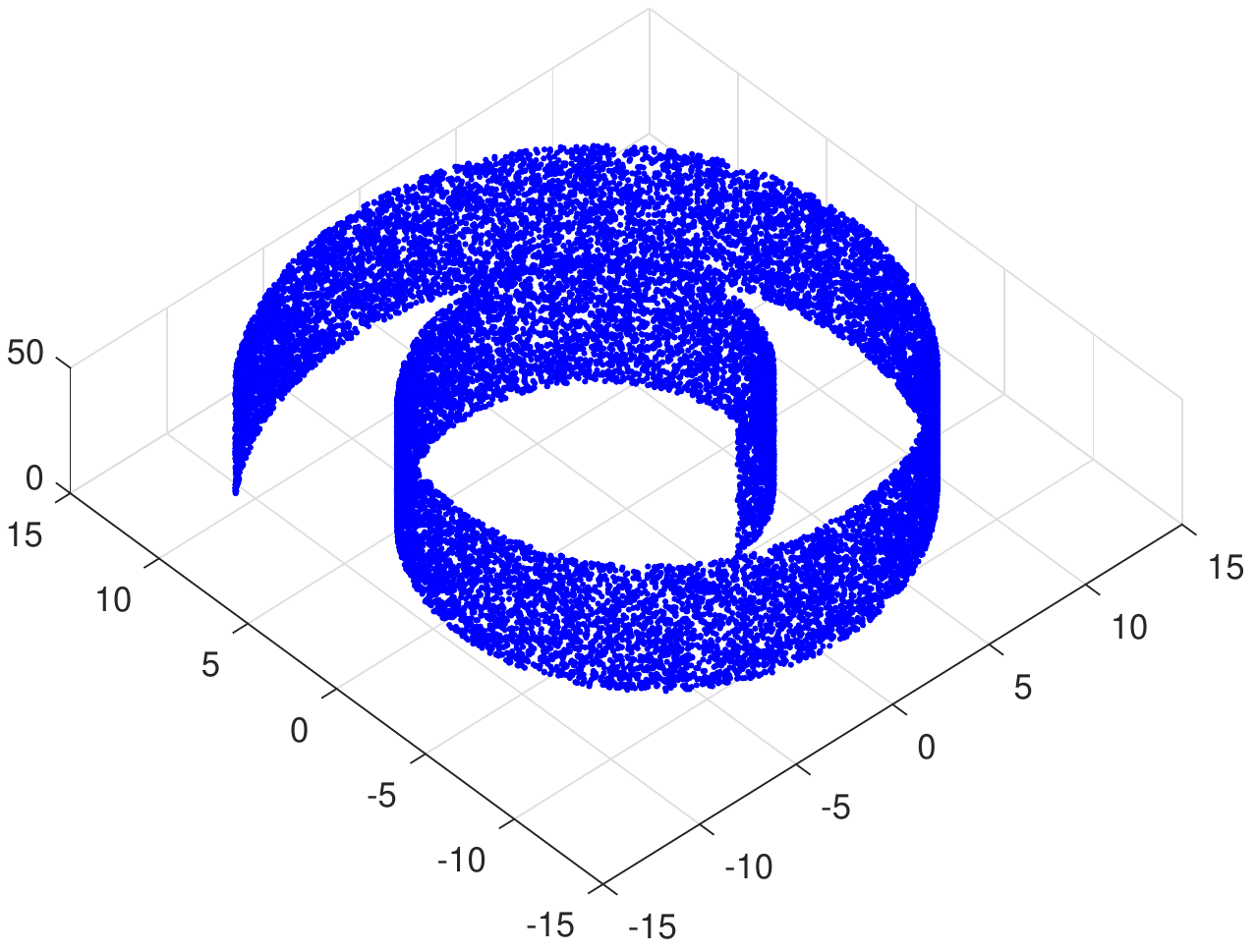}
    \end{minipage}\hfill
\begin{minipage}{0.5\textwidth}
        \centering
\includegraphics[width=.8\textwidth]{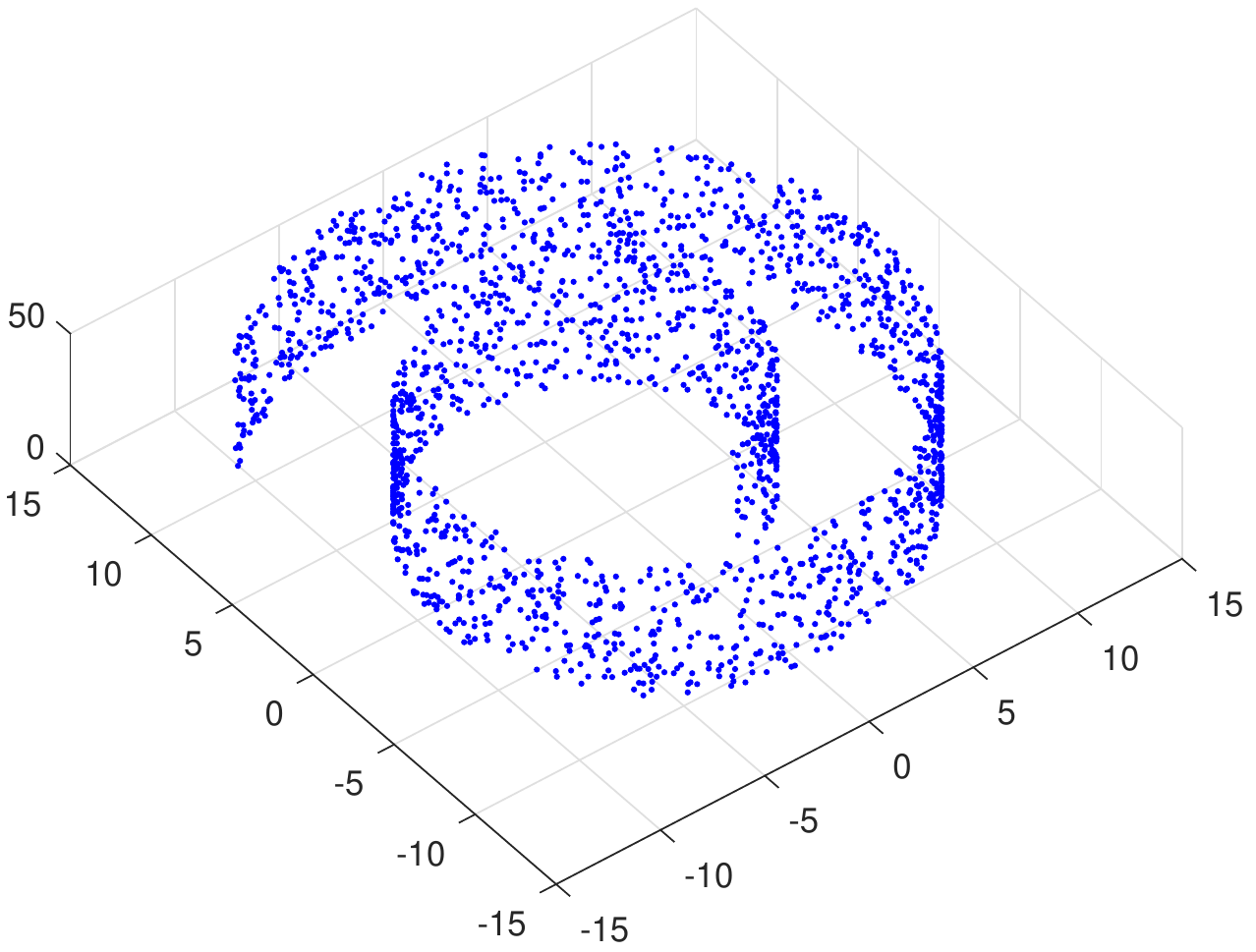}
\end{minipage}\hfill

\leftline{\hskip 0.00cm (c) \hskip 4.5cm (d) \hskip 4.5cm (e)} 
\centering
\begin{minipage}{0.3\textwidth}
\includegraphics[width=1.00\textwidth]{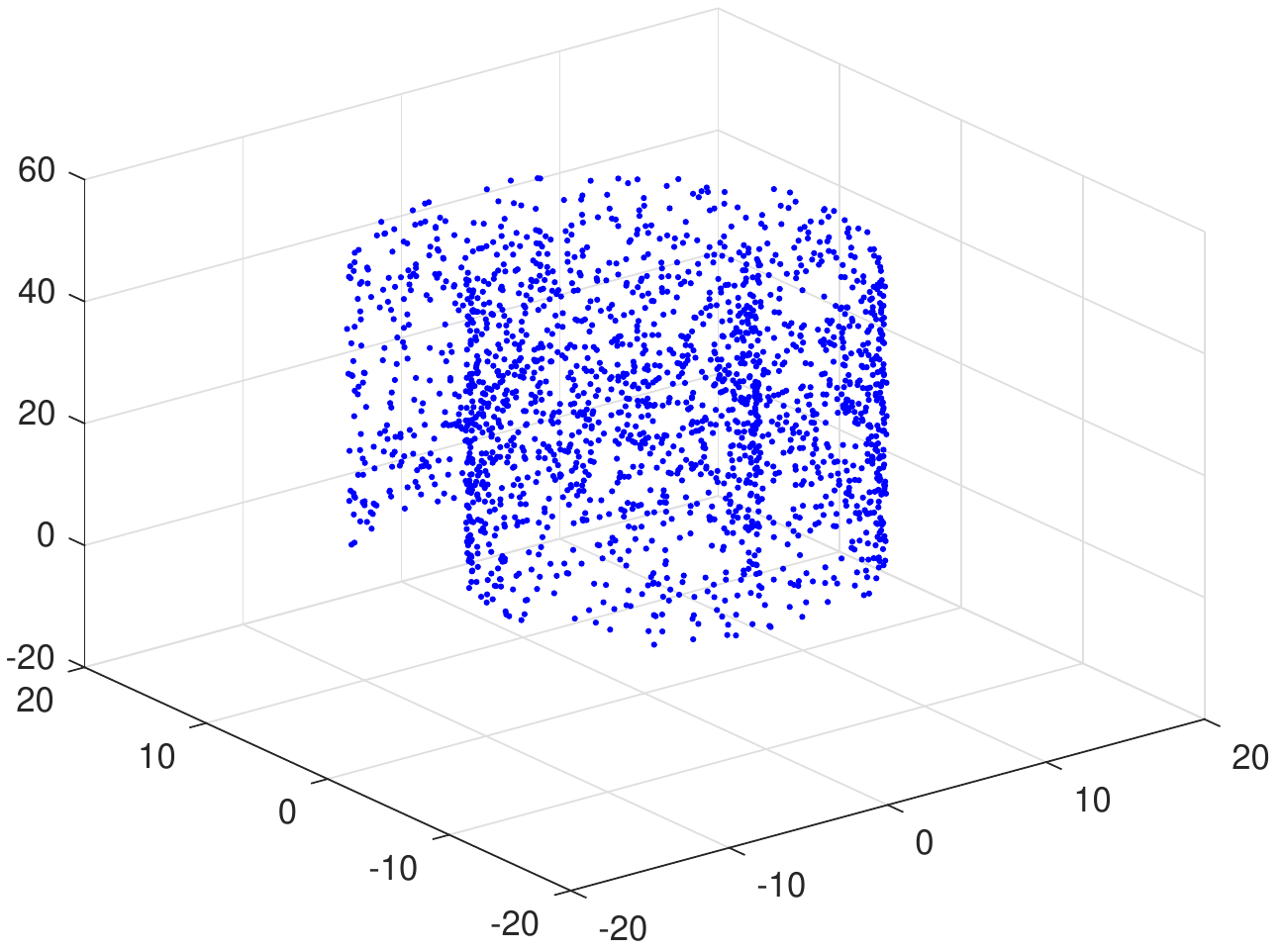}
\end{minipage}\hfill
\begin{minipage}{0.3\textwidth}
\includegraphics[width=1.00\textwidth]{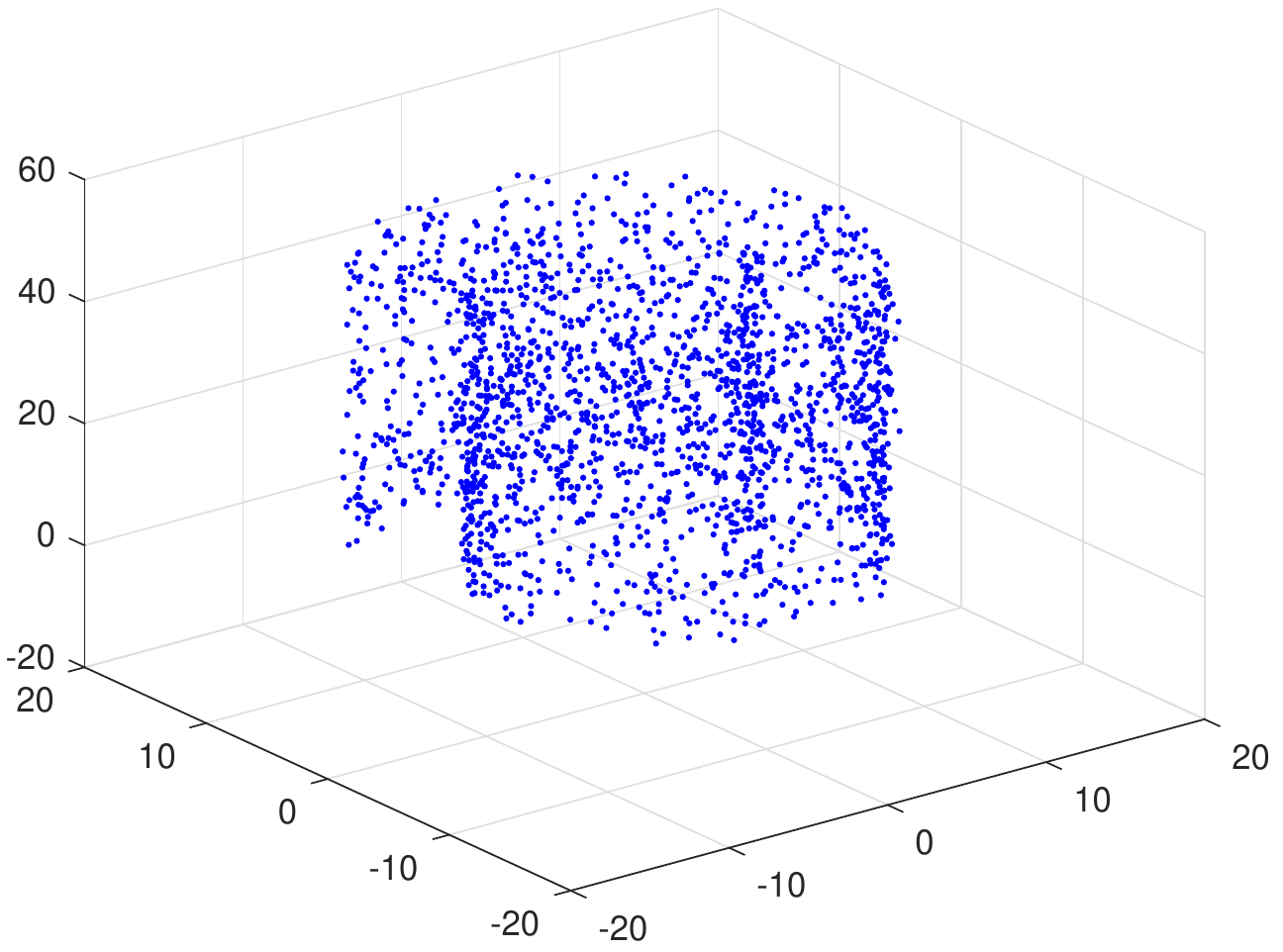}
\end{minipage}\hfill
\begin{minipage}{0.3\textwidth}
\includegraphics[width=1.00\textwidth]{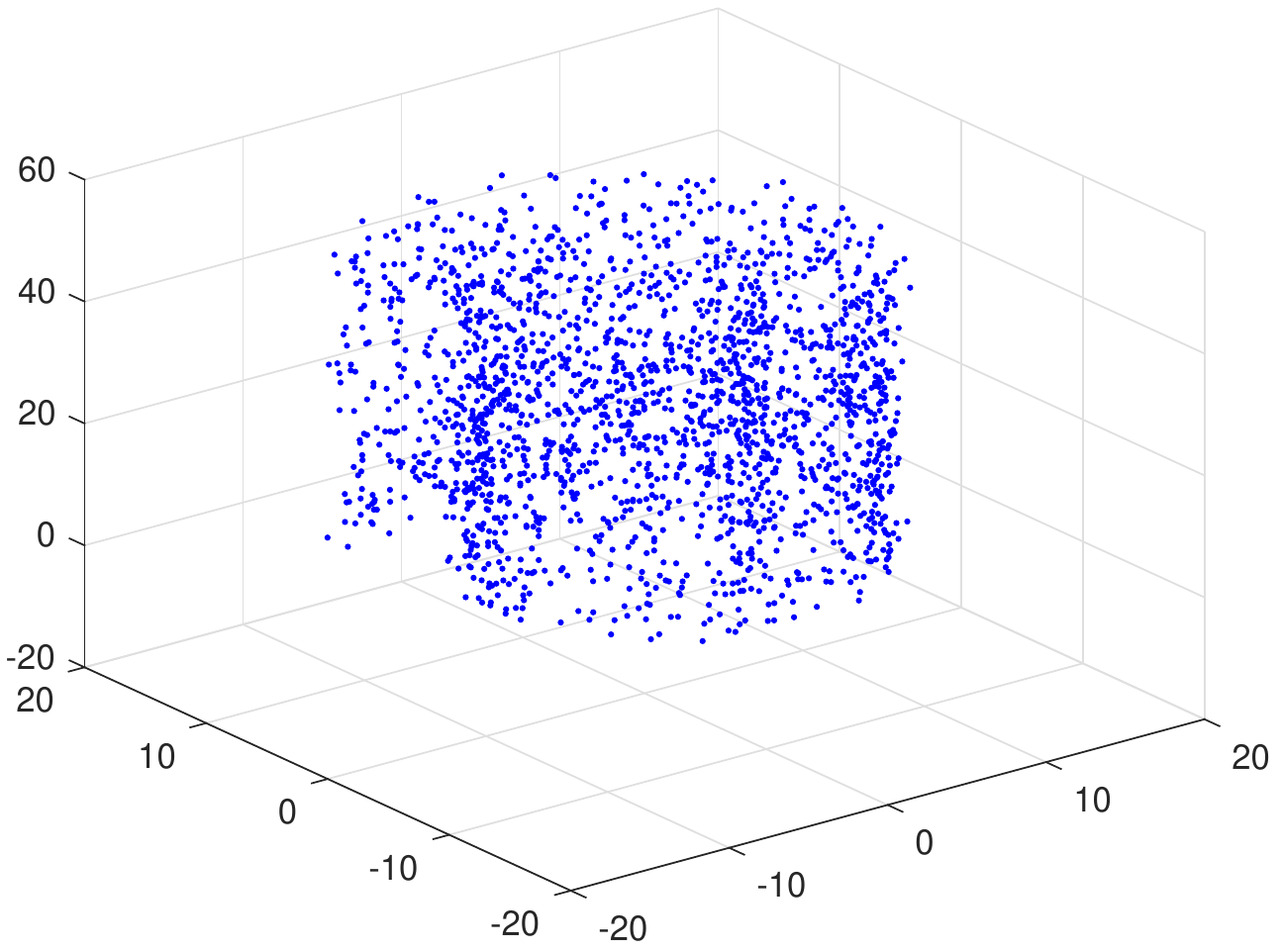}
\end{minipage}\hfill

\centering
\begin{minipage}{0.3\textwidth}
        \centering
\includegraphics[width=1.00\textwidth]{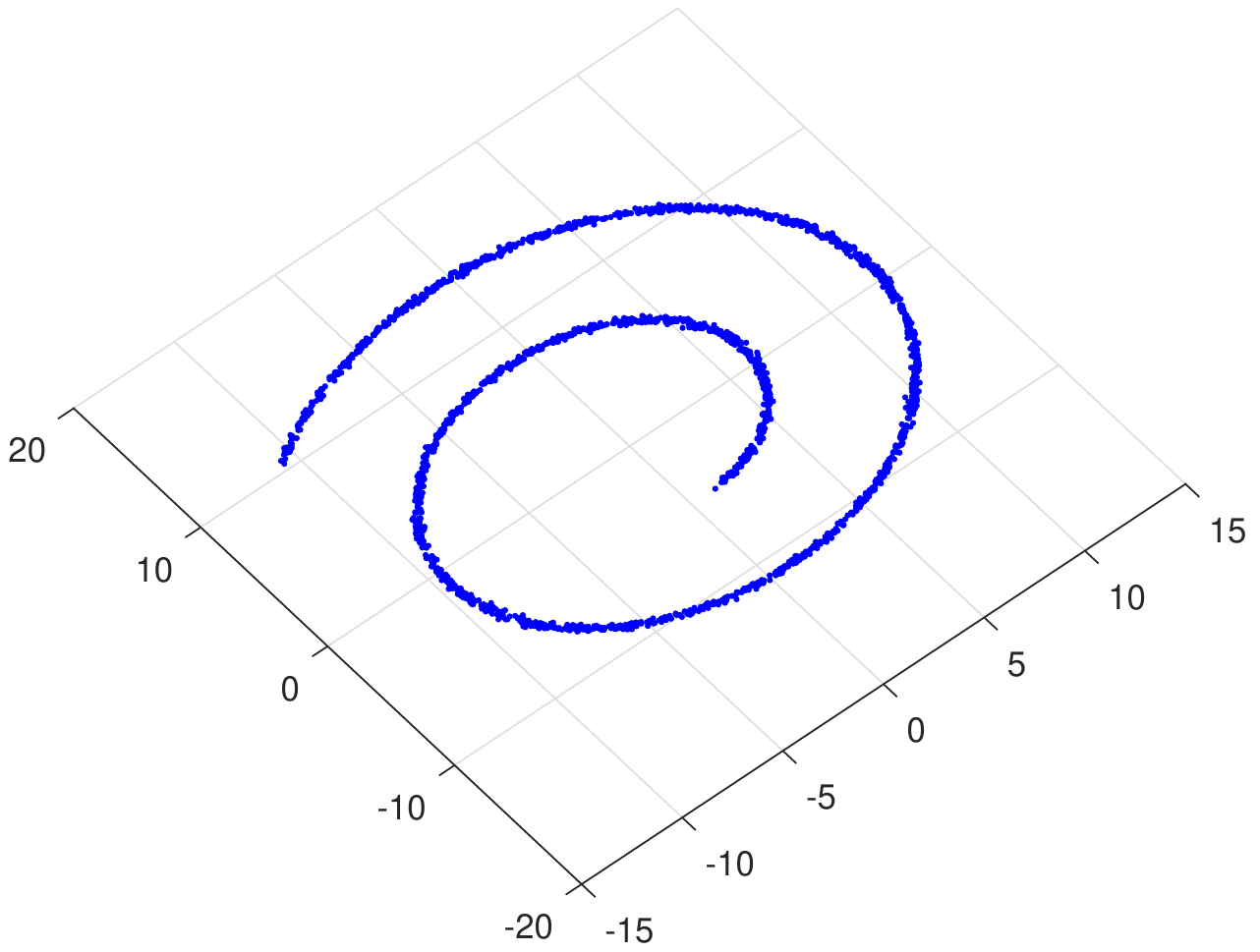}
 \end{minipage}\hfill     
\begin{minipage}{0.3\textwidth}
        \centering
\includegraphics[width=1.00\textwidth]{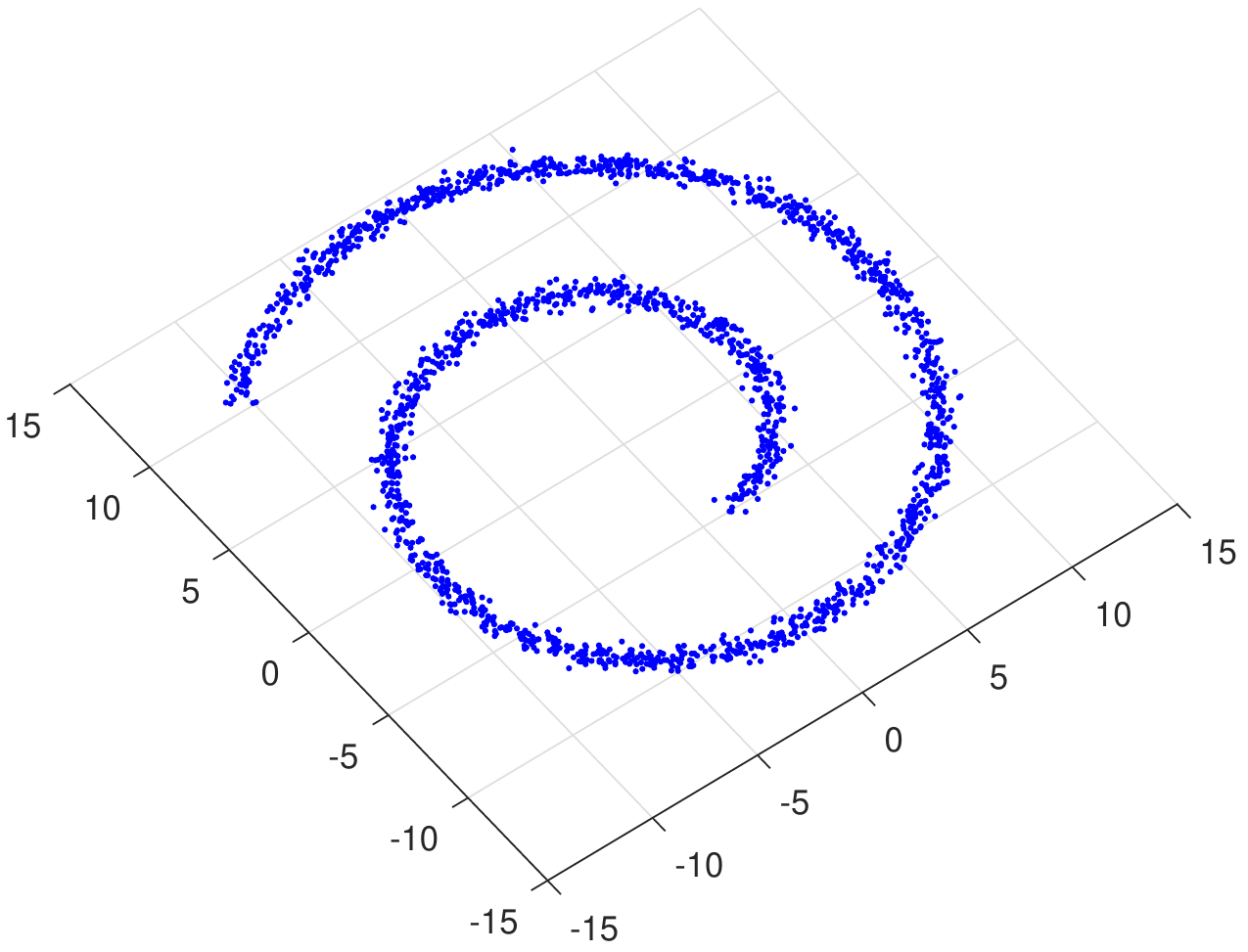}
 \end{minipage}\hfill
\begin{minipage}{0.3\textwidth}
        \centering
\includegraphics[width=1.00\textwidth]{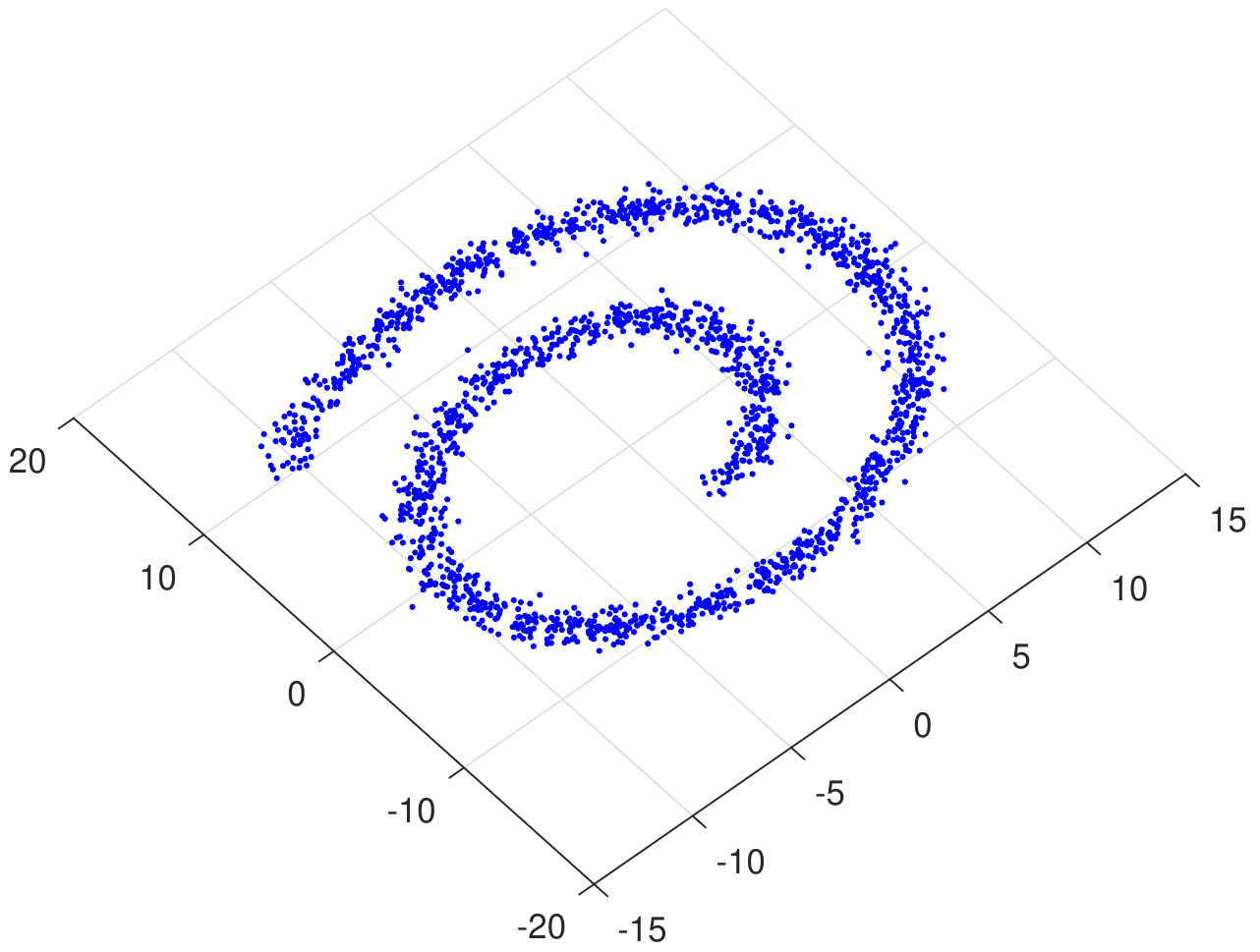}
 \end{minipage}\hfill
    
\caption{(a) All $20000$ points, and (b) the first $2000$ points in the Swiss roll data set from \cite{Tenenbaum2000}. We add white Gaussian noise to the $2000$ points with several different signal-to-noise ratios ($S/N$): 
(c) $S/N=20$, (d) $S/N=10$, (e) $S/N=5$. (We show two views of each plot.)}\label{Tenenbaum_Swiss_roll}
\end{figure} 

\begin{figure}[H]
\centering

\leftline{\hskip 0.00cm (a)} 
\centering
     \begin{minipage}{0.3\textwidth}
\includegraphics[scale=.35]{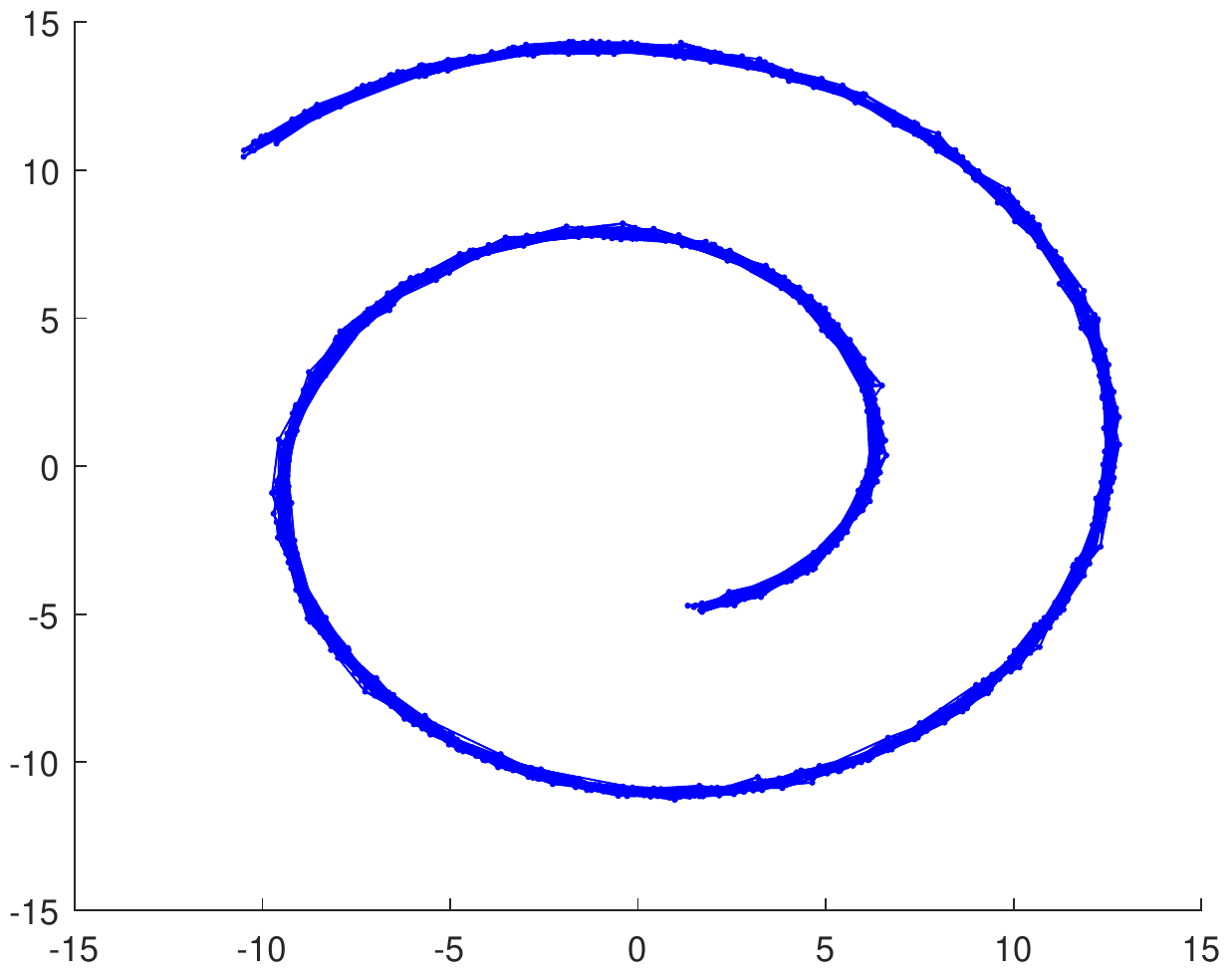}
\end{minipage}\hfill
\begin{minipage}{0.3\textwidth}
        \centering
\includegraphics[scale=.4]{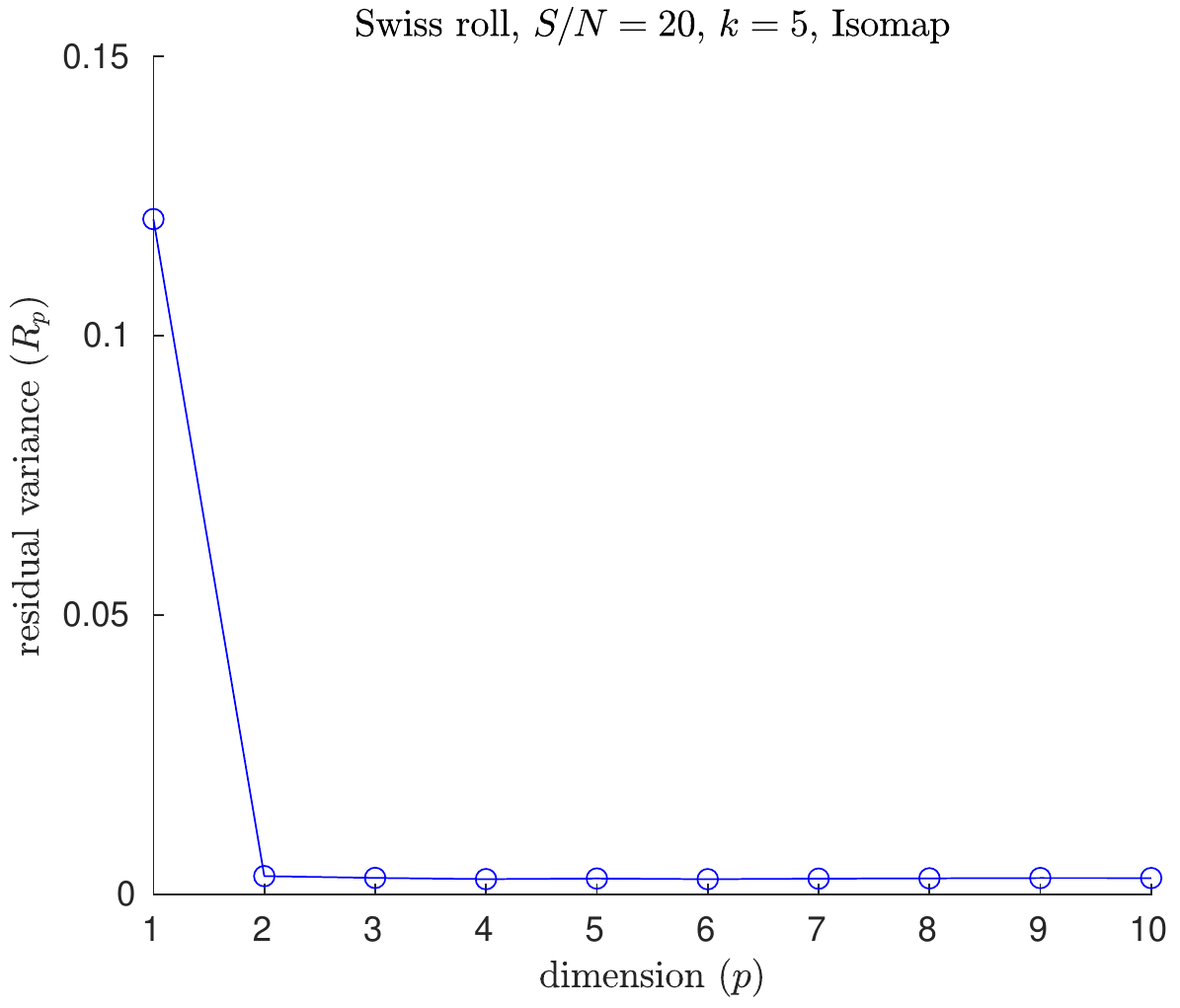}
\end{minipage}\hfill
\begin{minipage}{0.3\textwidth}
        \centering
        \includegraphics[scale=.4]{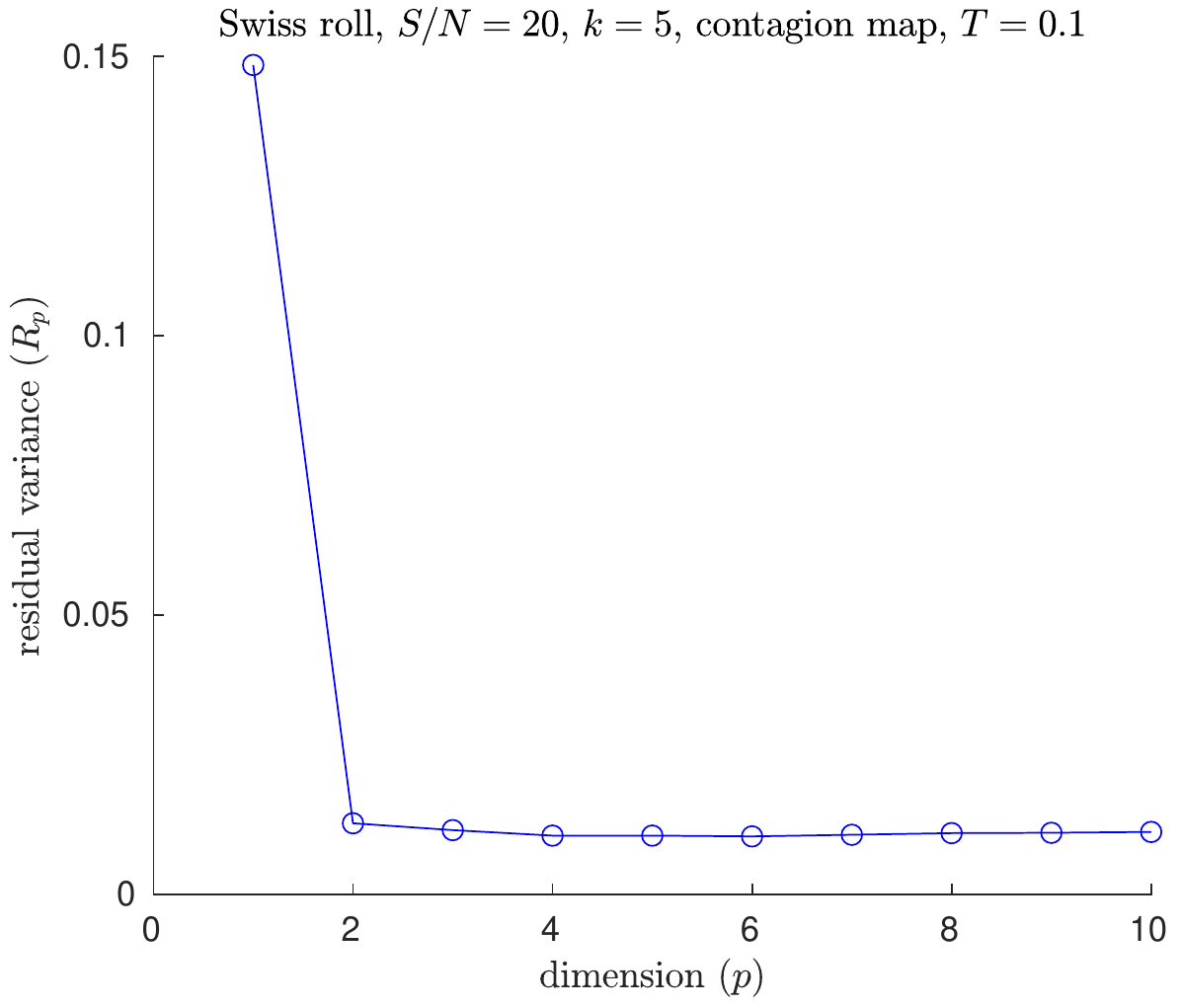}
\end{minipage}\hfill

\leftline{\hskip 0.00cm (b) } 
\centering
     \begin{minipage}{.3\textwidth}
\includegraphics[scale=.35]{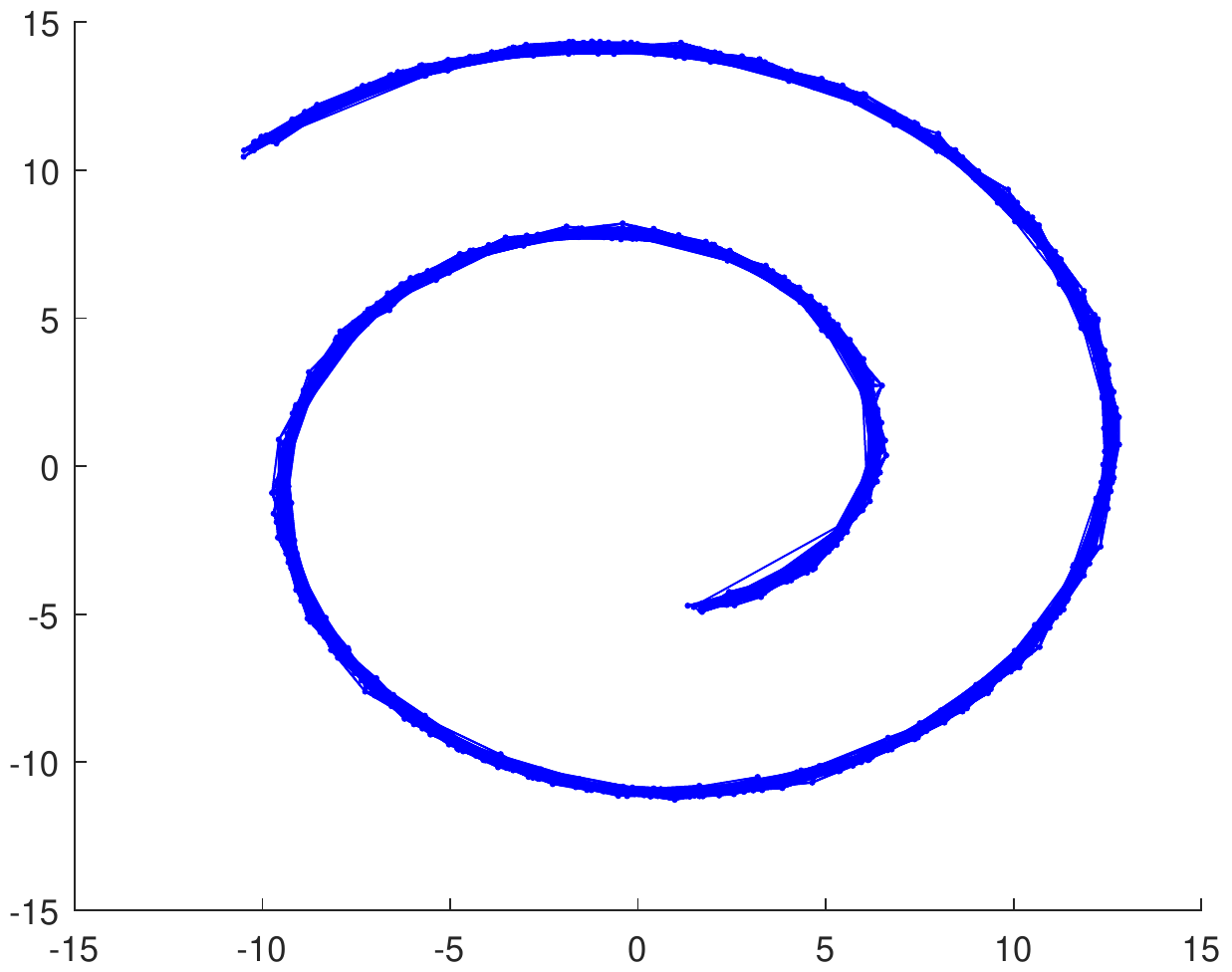}
\end{minipage}\hfill
\begin{minipage}{.3\textwidth}
        \centering
\includegraphics[scale=.4]{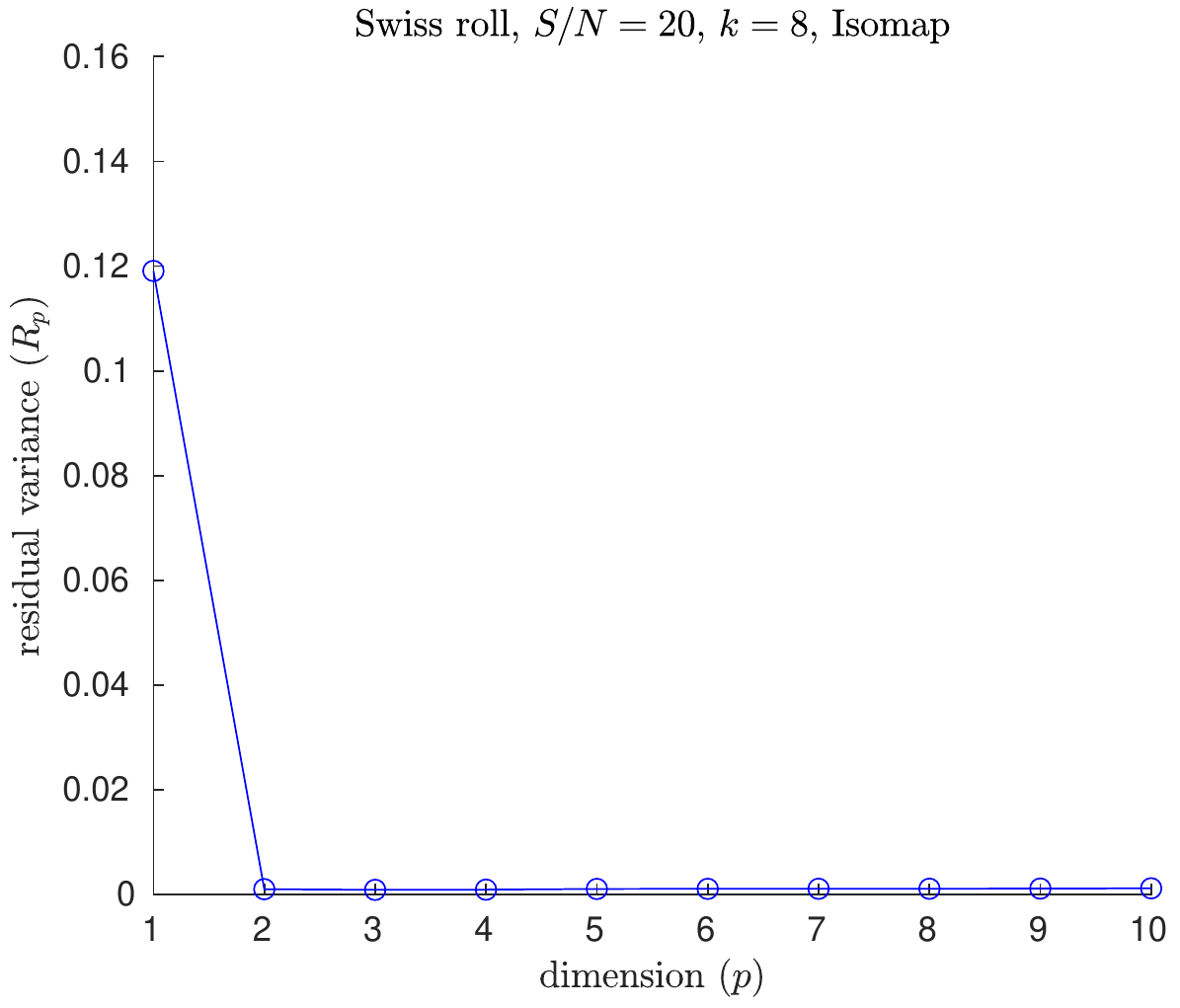}
\end{minipage}\hfill
\begin{minipage}{.3\textwidth}
        \centering
        \includegraphics[scale=.4]{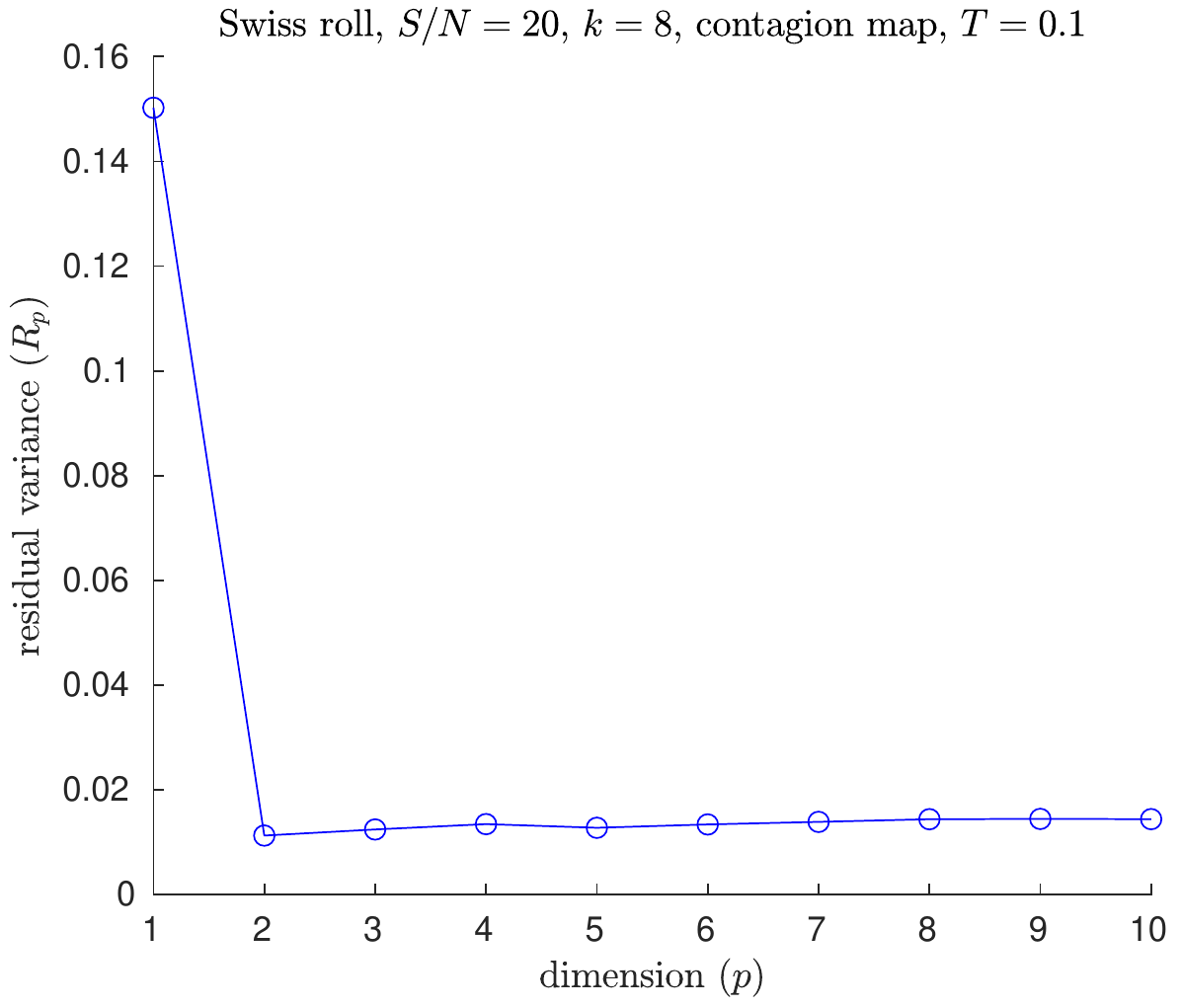}
\end{minipage}\hfill

 \leftline{\hskip 0.00cm (c) } 
 \centering
     \begin{minipage}{.3\textwidth}
\includegraphics[scale=.35]{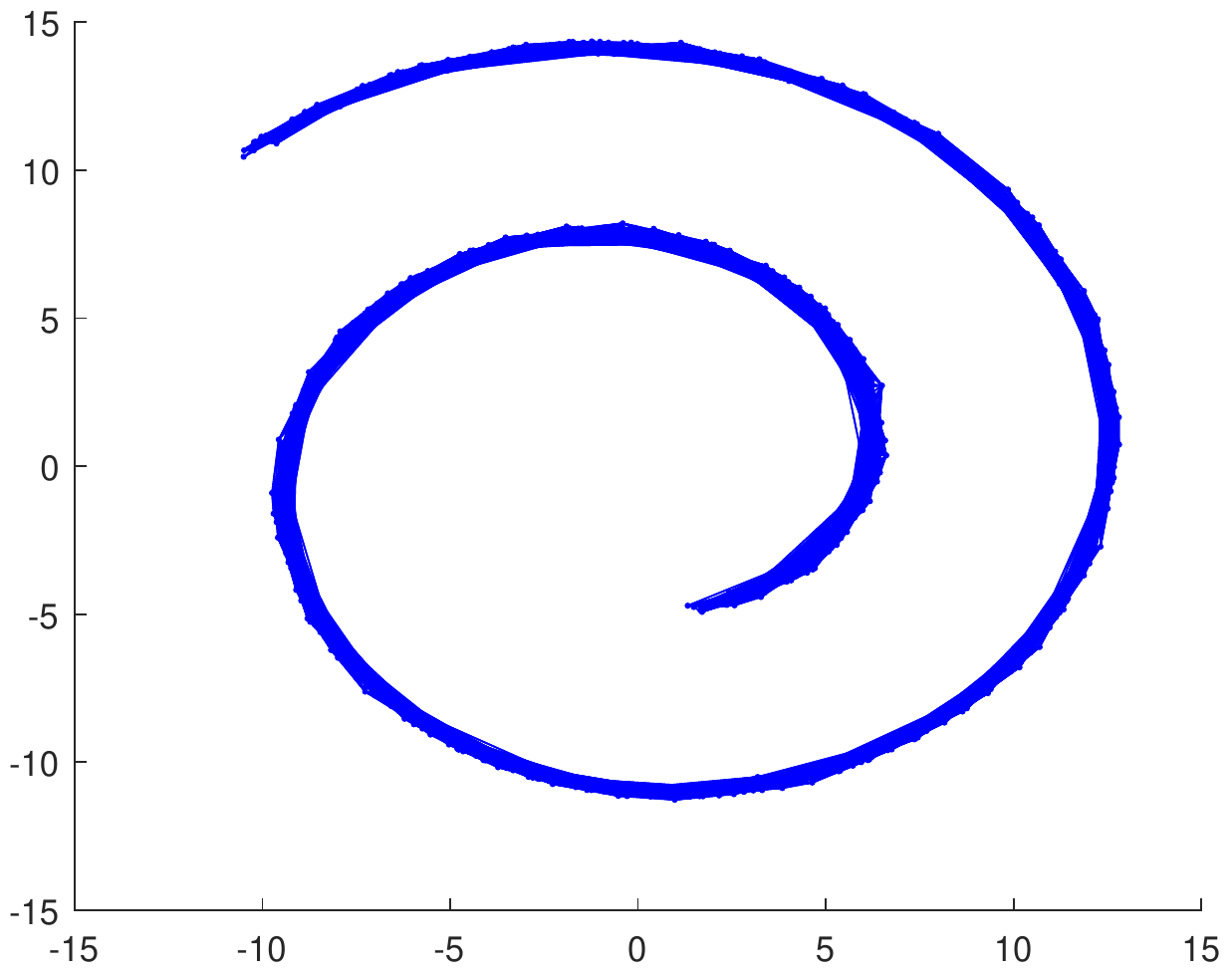}
\end{minipage}\hfill
\begin{minipage}{.3\textwidth}
        \centering
\includegraphics[scale=.4]{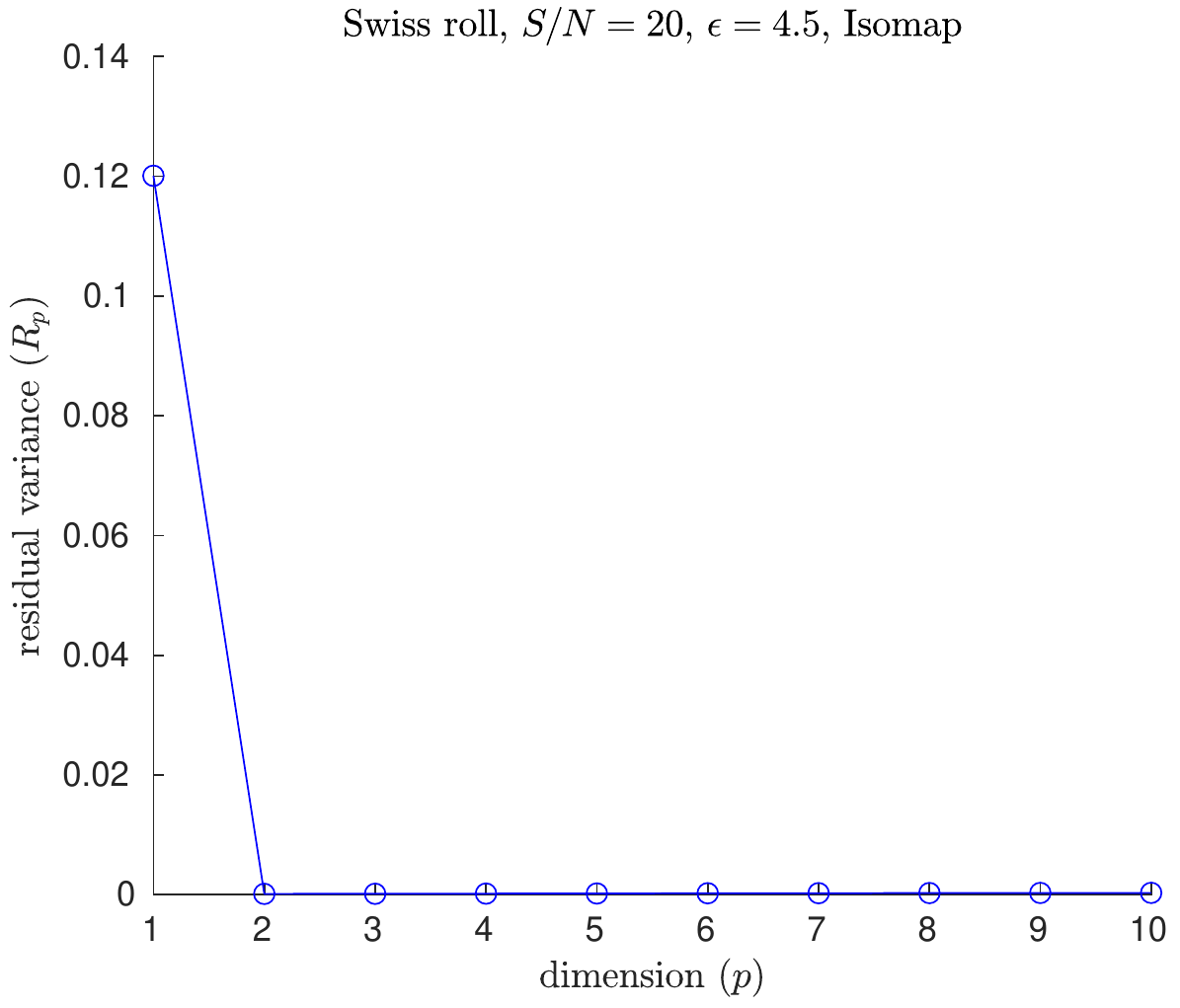}
\end{minipage}\hfill
\begin{minipage}{0.3\textwidth}
        \centering
        \includegraphics[scale=.4]{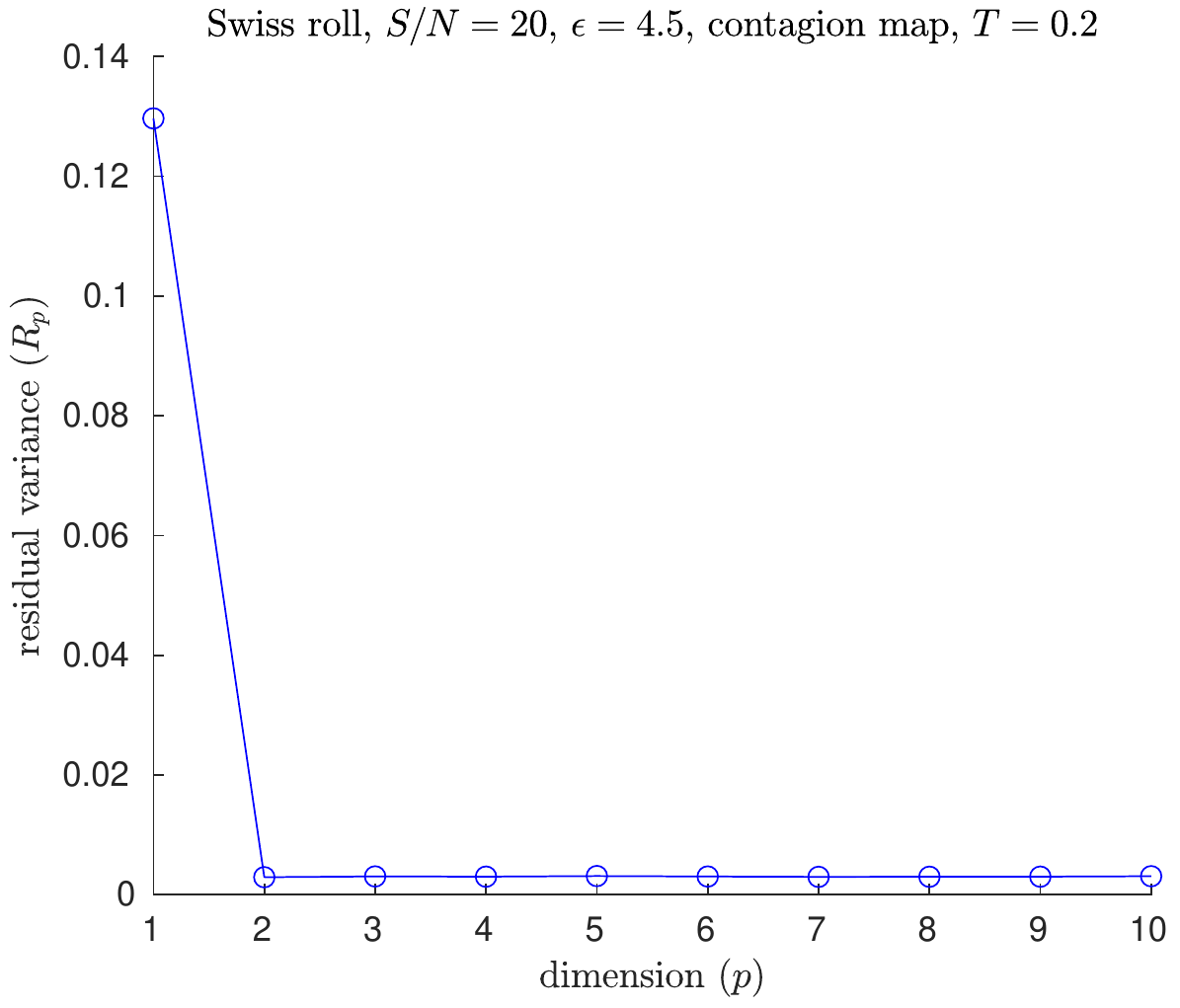}
\end{minipage}\hfill

 \leftline{\hskip 0.00cm (d)}
 \centering
     \begin{minipage}{.3\textwidth} 
\includegraphics[scale=.35]{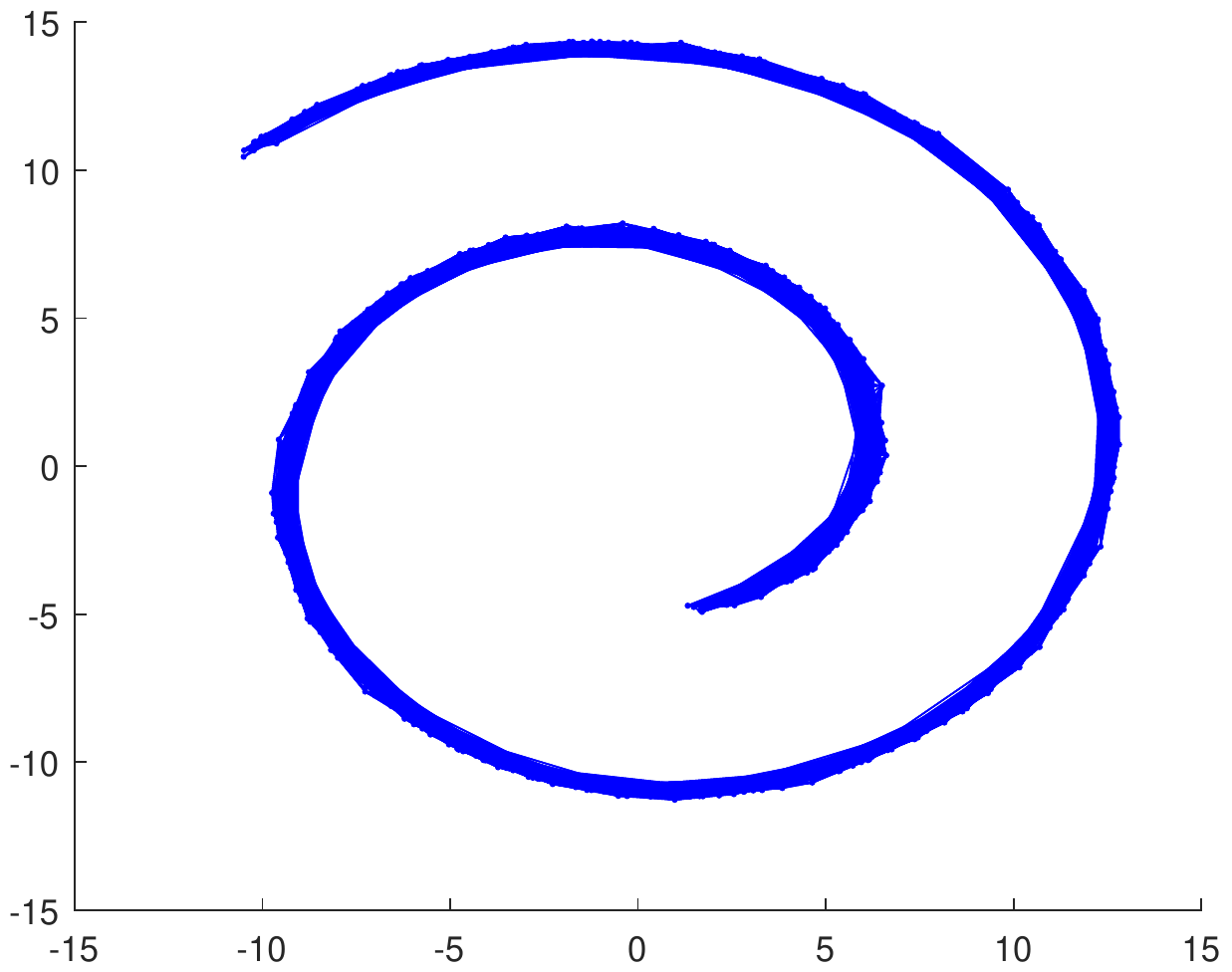}
\end{minipage}\hfill
\begin{minipage}{.3\textwidth}
        \centering
\includegraphics[scale=.4]{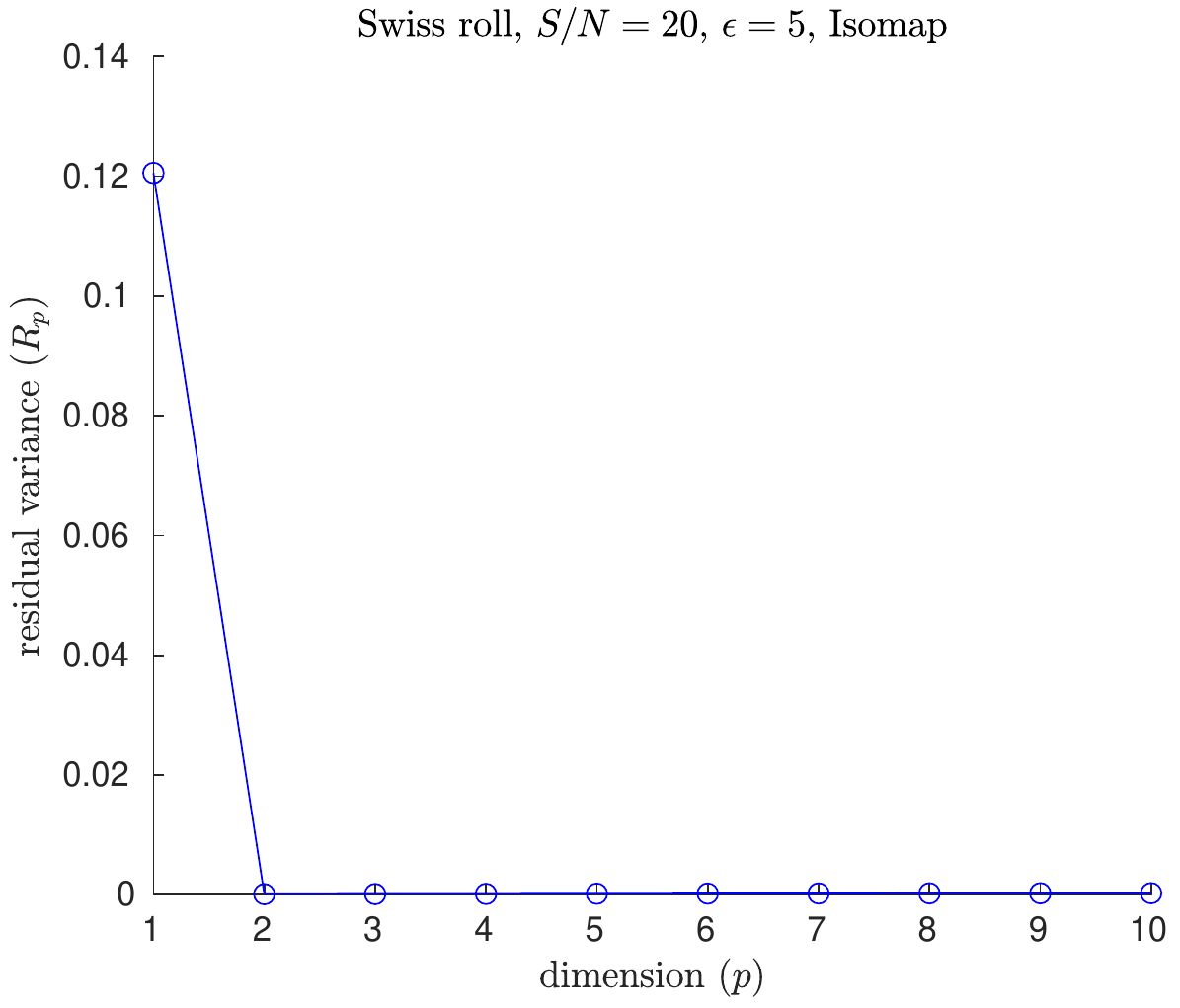}
\end{minipage}\hfill
\begin{minipage}{.3\textwidth}
        \centering
\includegraphics[scale=.4]{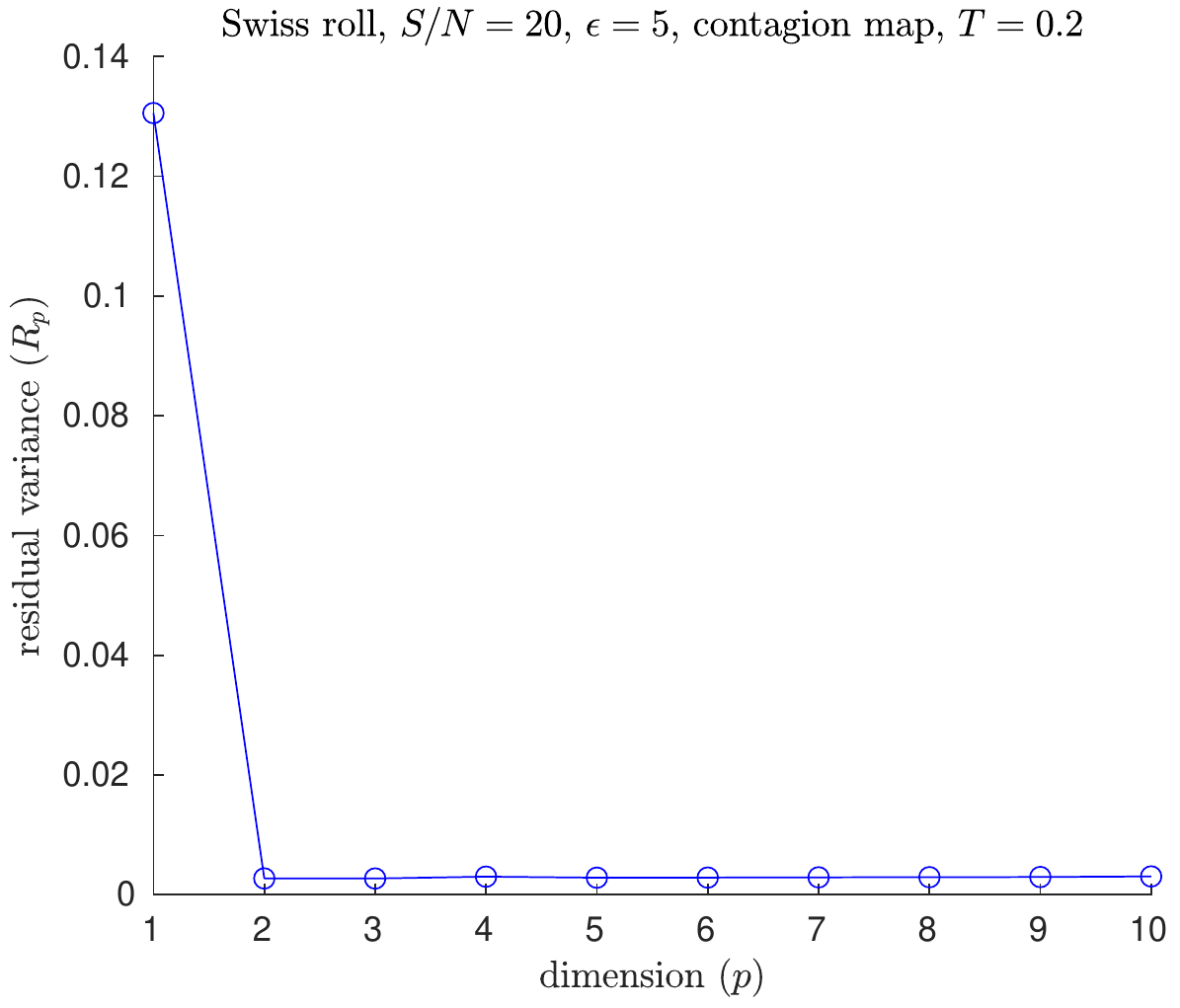}
\end{minipage}\hfill
\caption{Neighbourhood graphs on $2000$ points of the Swiss roll data set from \cite{Tenenbaum2000} with white Gaussian noise added with $S/N=20$ (left column). The residual variances of MDS projections to dimensions $1$ to $10$ resulting from (middle column) the approximate pairwise geodesic distances between nodes based on shortest paths on the weighted graph (i.e.~Isomap) and (right column) the activation times in a threshold contagion on the unweighted graph (i.e.~contagion maps). The neighbourhood graphs are (a) the $5$-nearest-neighbour graph, (b) the $8$-nearest-neighbour graph, (c) the $4.5$-neighbourhood graph, and (d) the $5$-neighbourhood graph.}
\label{Tenenbaum_Swiss_roll_neighbourhood_noise20}
\end{figure}

Both Isomap and contagion maps perform adequately for the high signal-to-noise ration of $S/N=20$ (i.e.~low noise level) with all four examined versions of neighbourhood graphs (see Figure~\ref{Tenenbaum_Swiss_roll_neighbourhood_noise20}).

\begin{figure}[H]
\leftline{\hskip 0.00cm (a)} 
\centering
     \begin{minipage}{0.3\textwidth} 
\includegraphics[scale=.35]{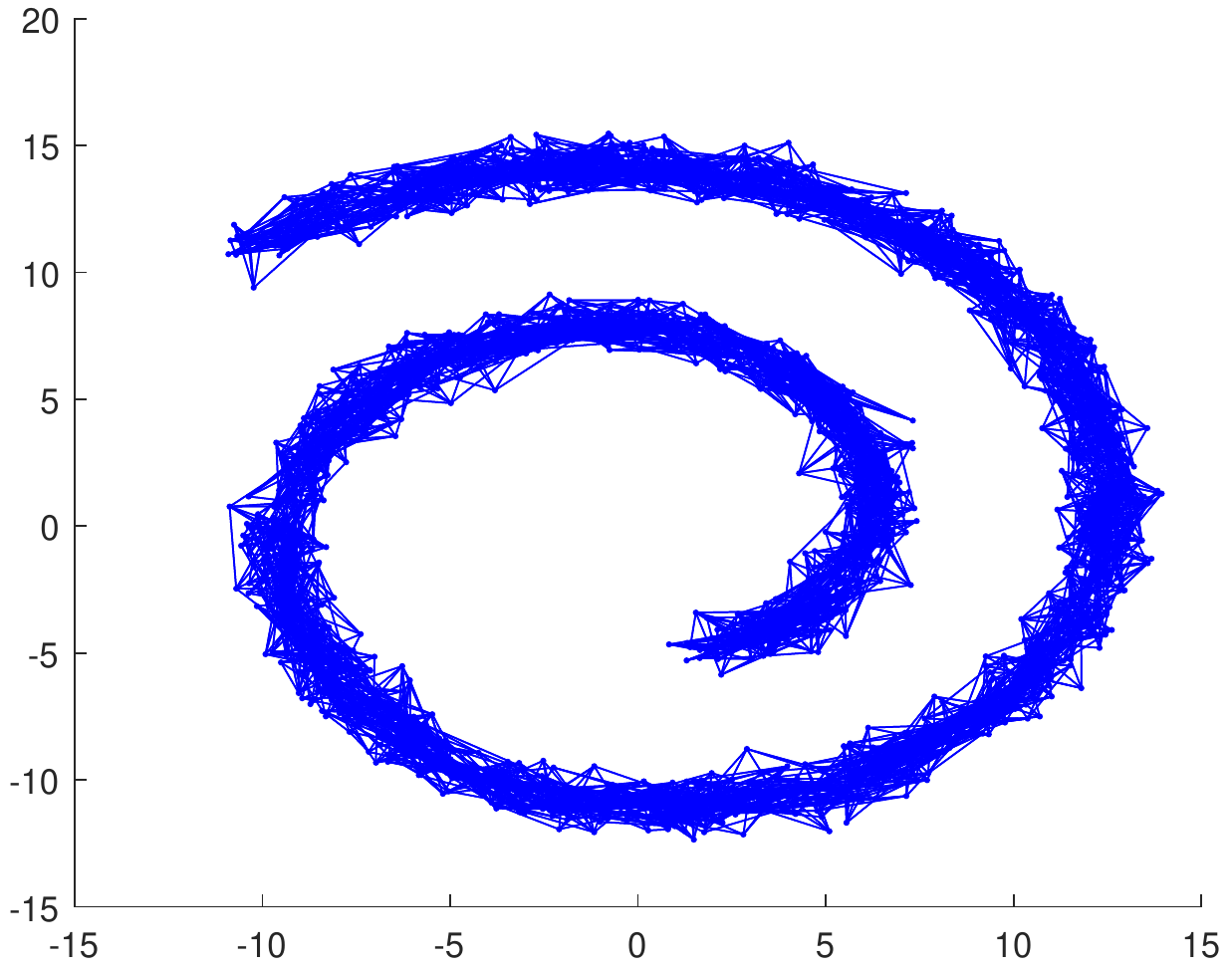}
\end{minipage}\hfill
\begin{minipage}{0.3\textwidth}
        \centering
\includegraphics[scale=.4]{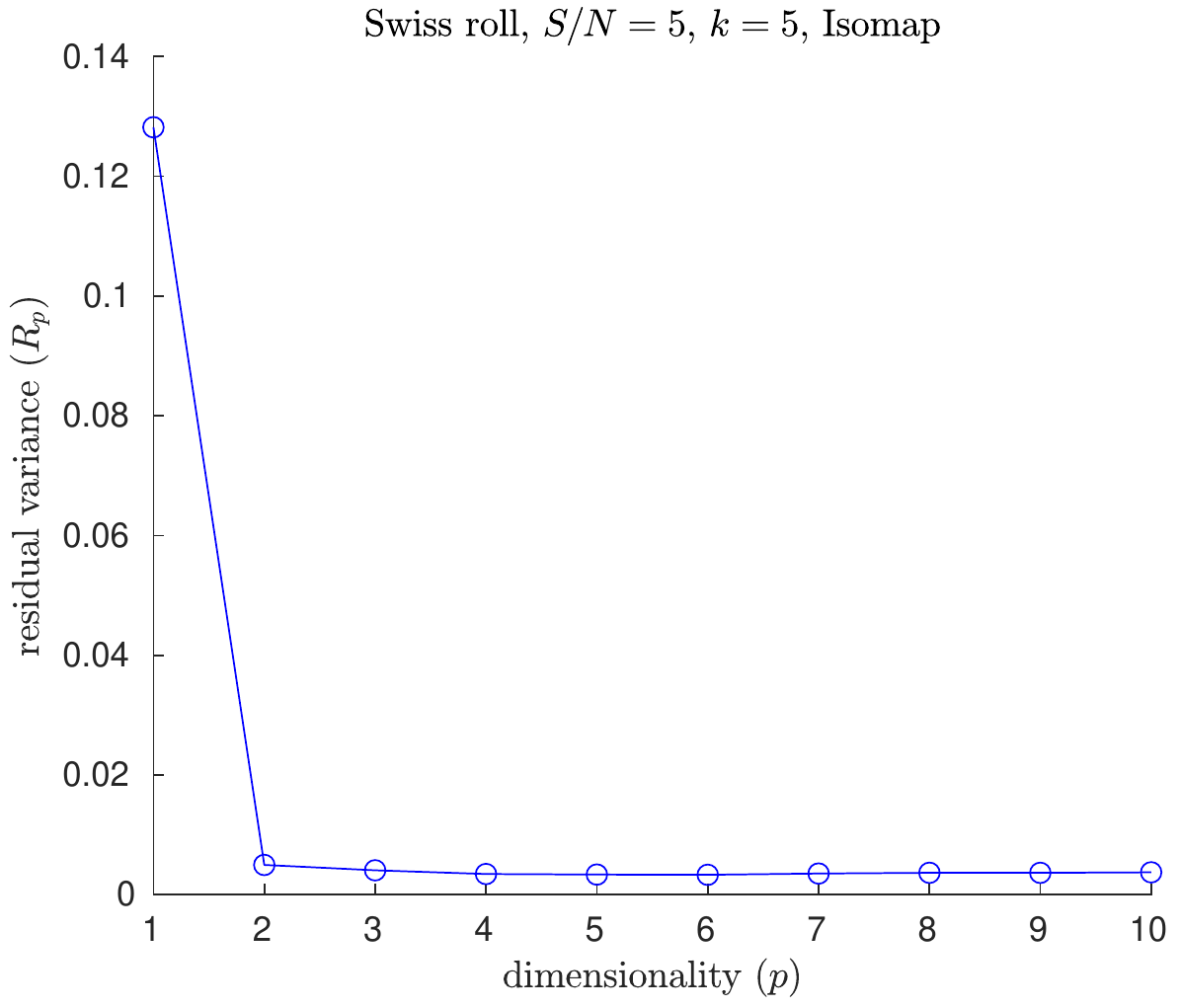}
\end{minipage}\hfill
\begin{minipage}{0.3\textwidth}
        \centering
\includegraphics[scale=.4]
{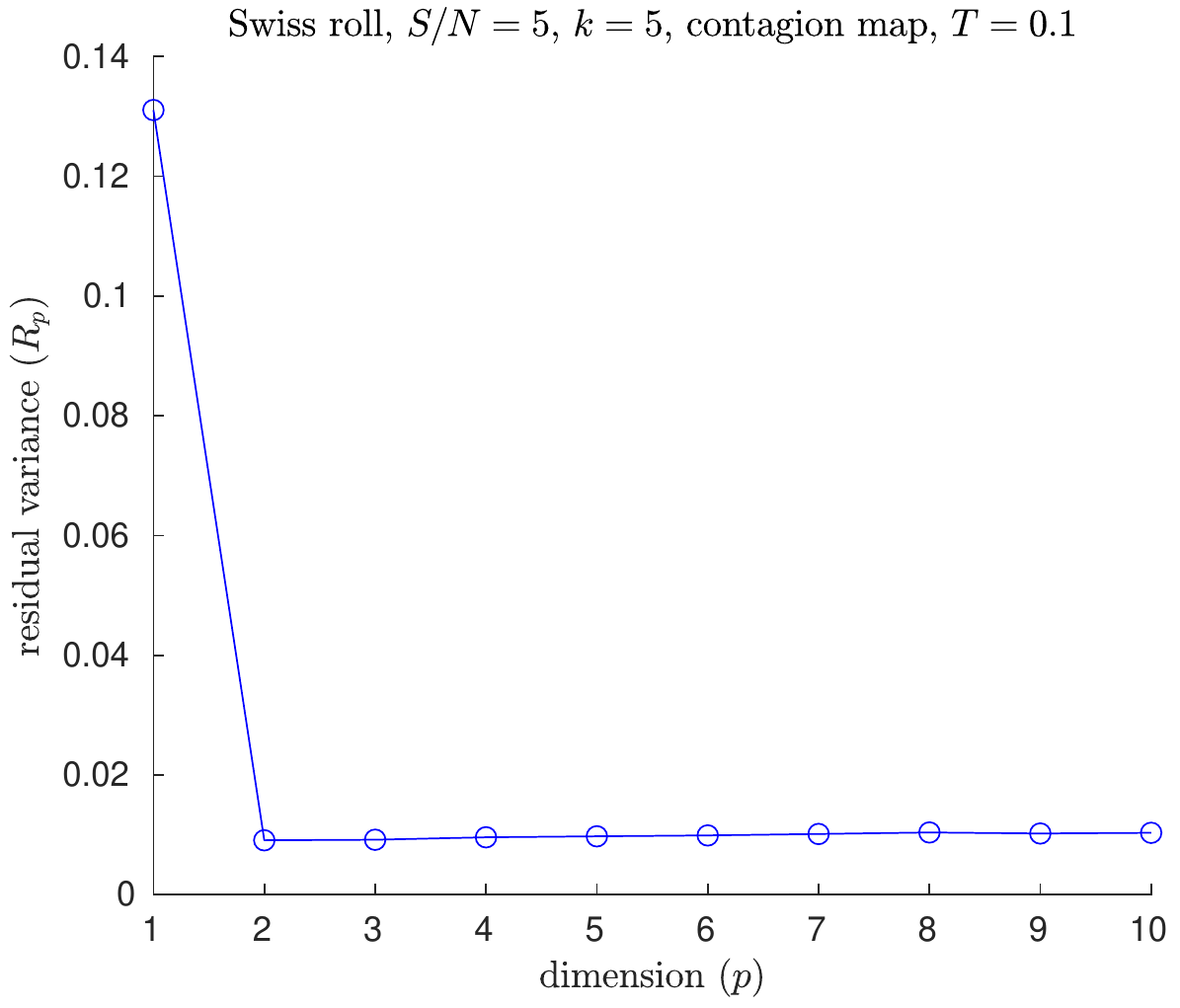}
\end{minipage}\hfill

   \leftline{\hskip 0.00cm (b)} 
   \centering
     \begin{minipage}{0.3\textwidth} 
\includegraphics[scale=.35]{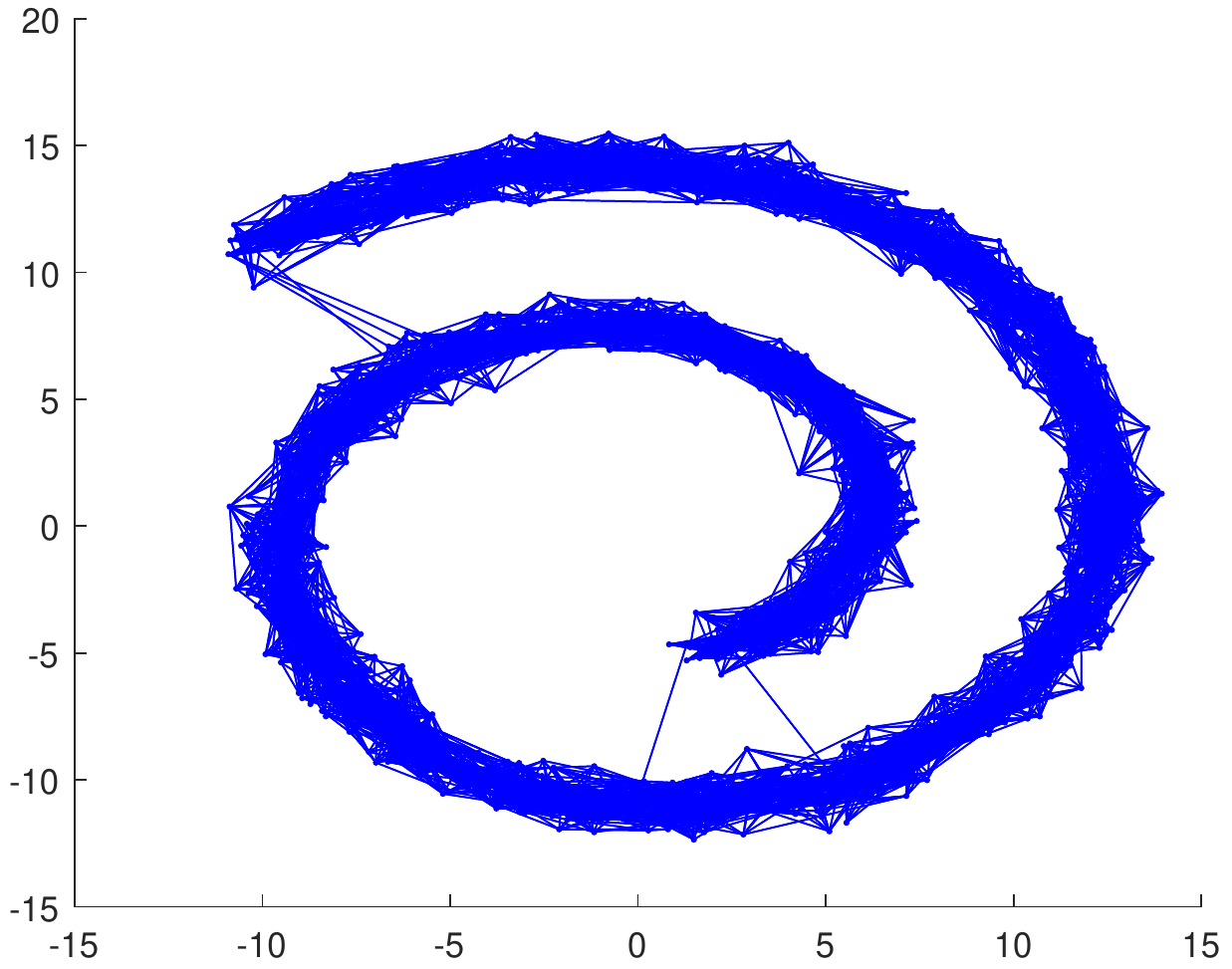}
\end{minipage}\hfill
\begin{minipage}{0.3\textwidth}
        \centering
\includegraphics[scale=.4]{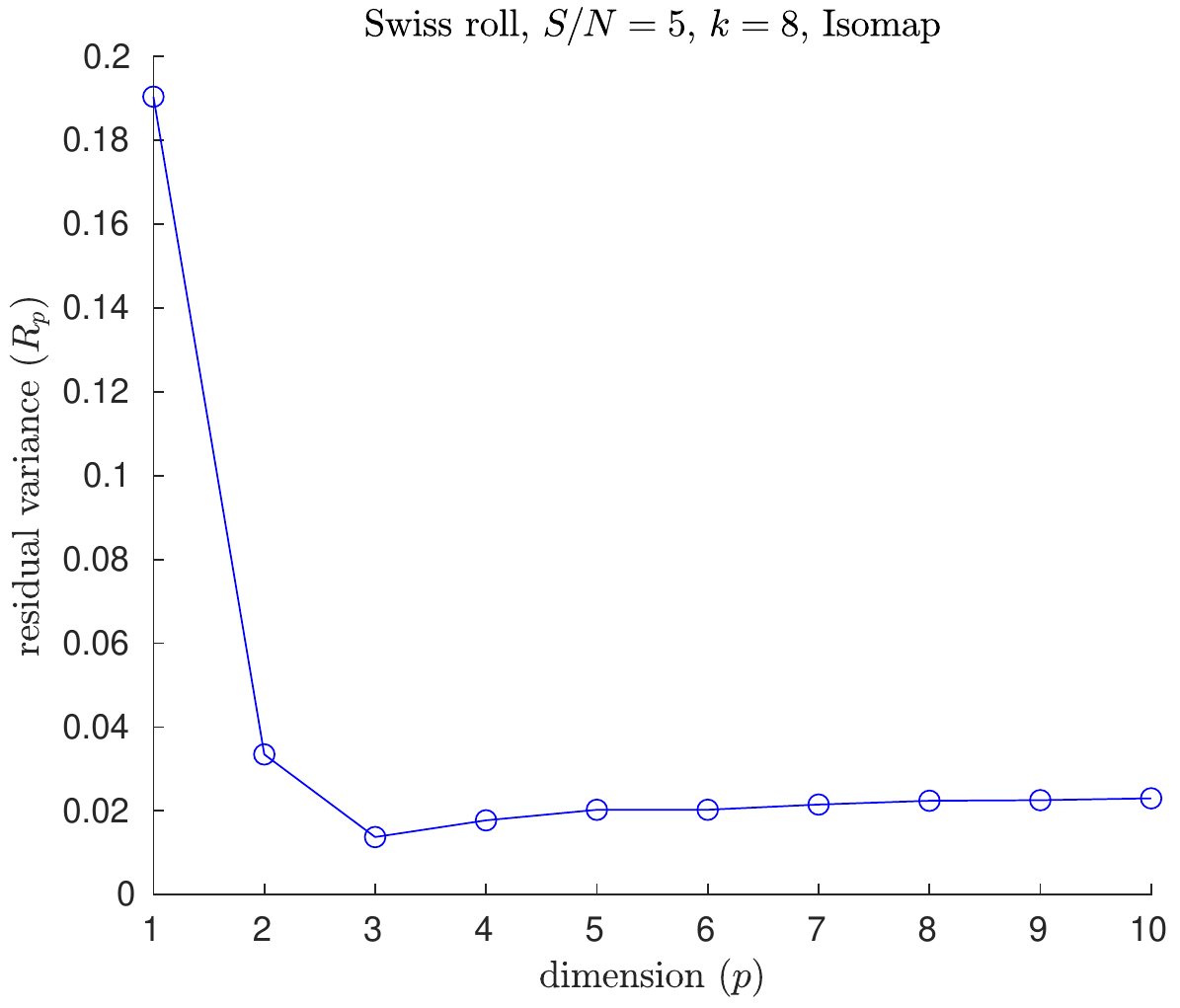}
\end{minipage}\hfill
\begin{minipage}{0.3\textwidth}
   \centering
\includegraphics[scale=.4]{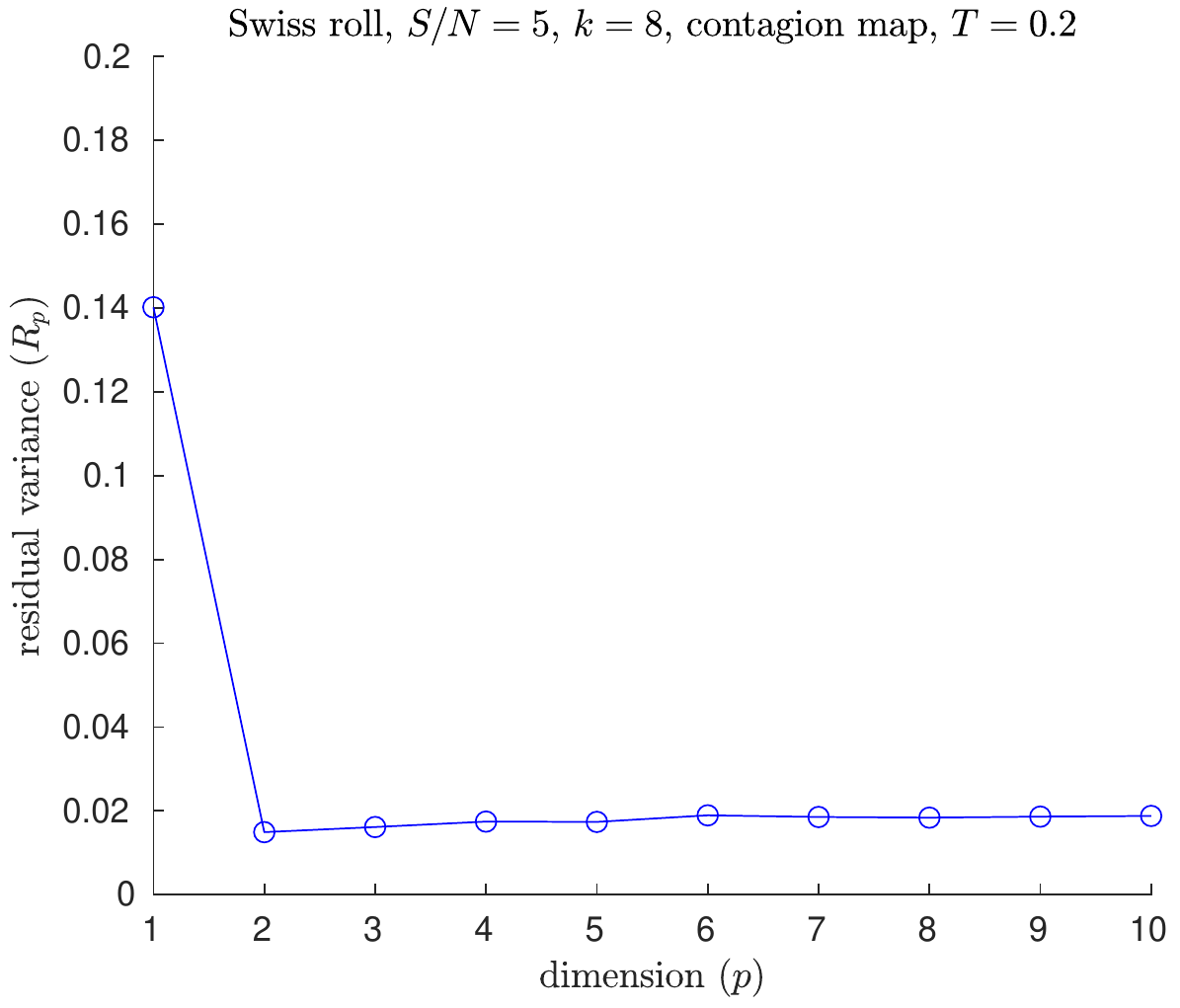}
\end{minipage}\hfill

 \leftline{\hskip 0.00cm (c)} 
 \centering
     \begin{minipage}{0.3\textwidth} 
\includegraphics[scale=.35]{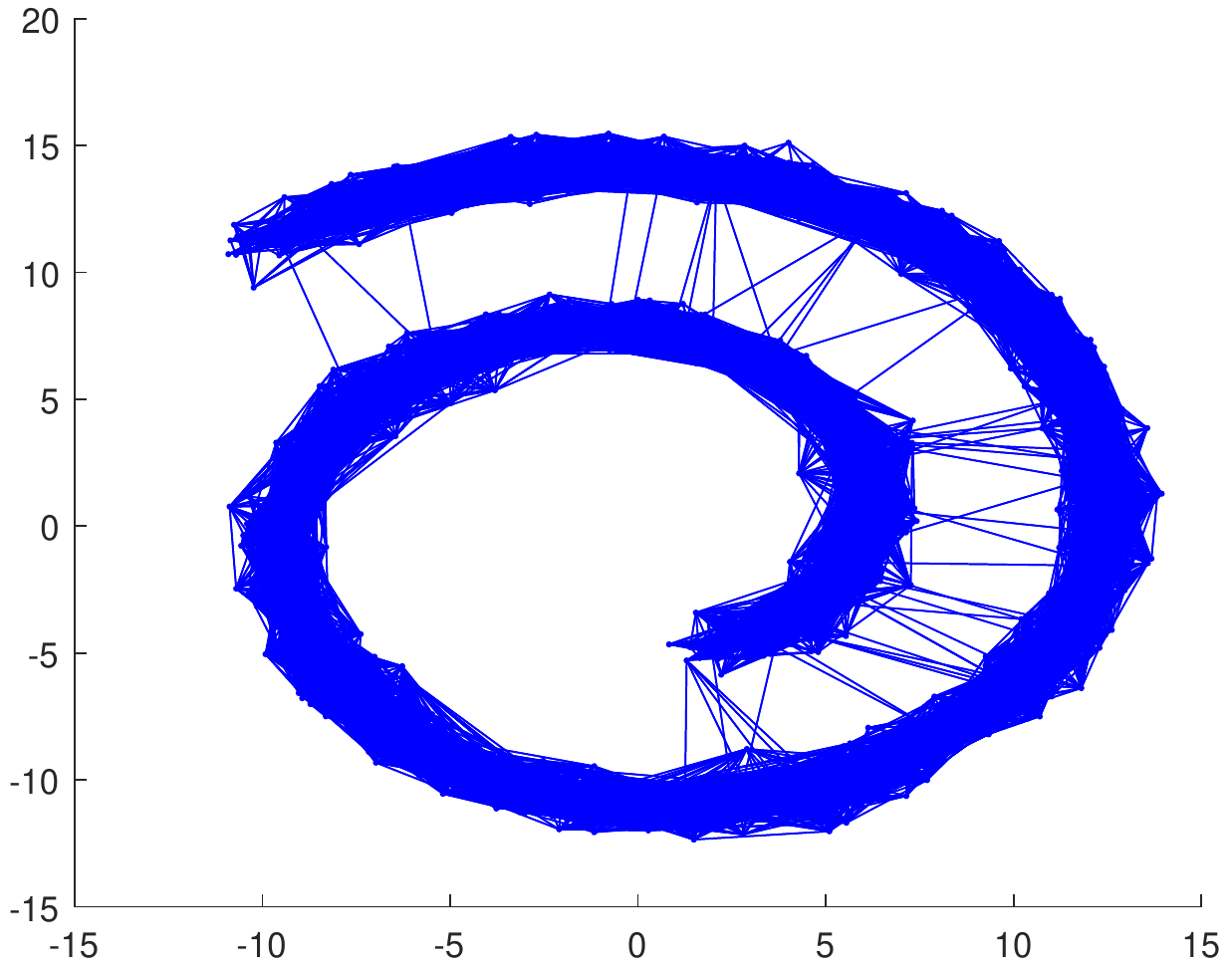}
\end{minipage}\hfill
\begin{minipage}{0.3\textwidth}
        \centering
\includegraphics[scale=.4]{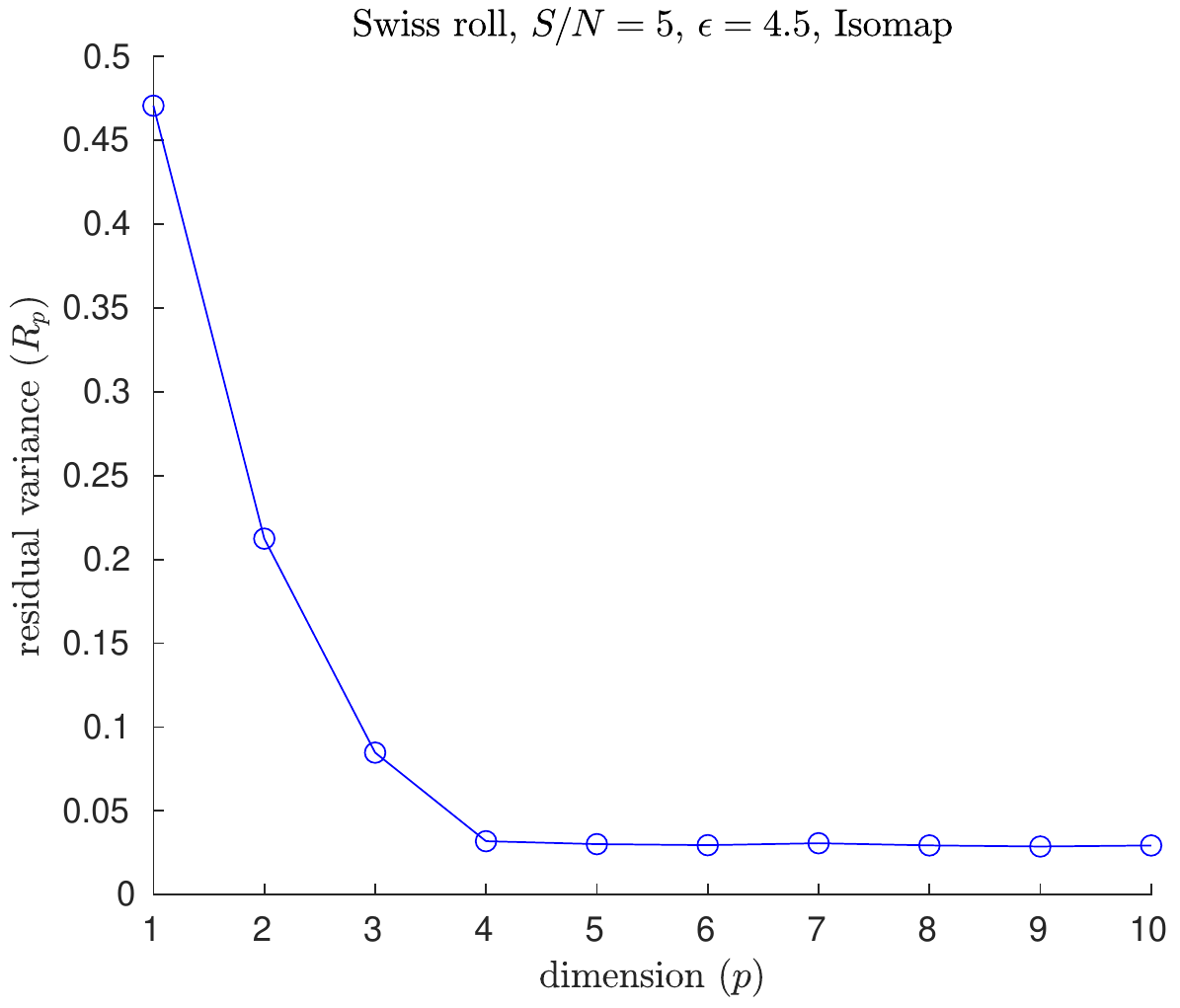}
\end{minipage}\hfill
\begin{minipage}{0.3\textwidth}
   \centering
\includegraphics[scale=.4]{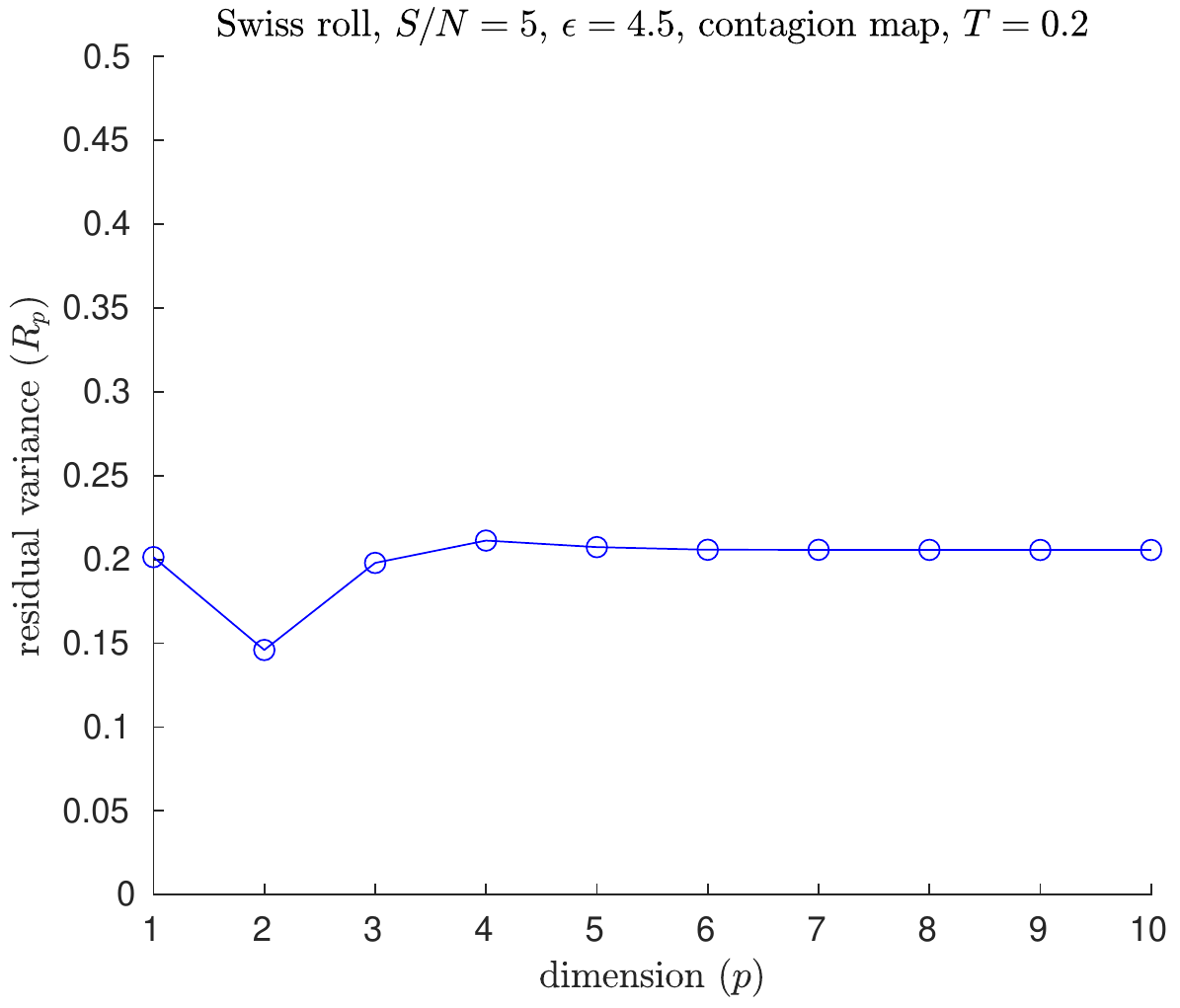}
\end{minipage}\hfill

   \leftline{\hskip 0.00cm (d)}
   \centering
     \begin{minipage}{0.3\textwidth} 
\includegraphics[scale=.35]{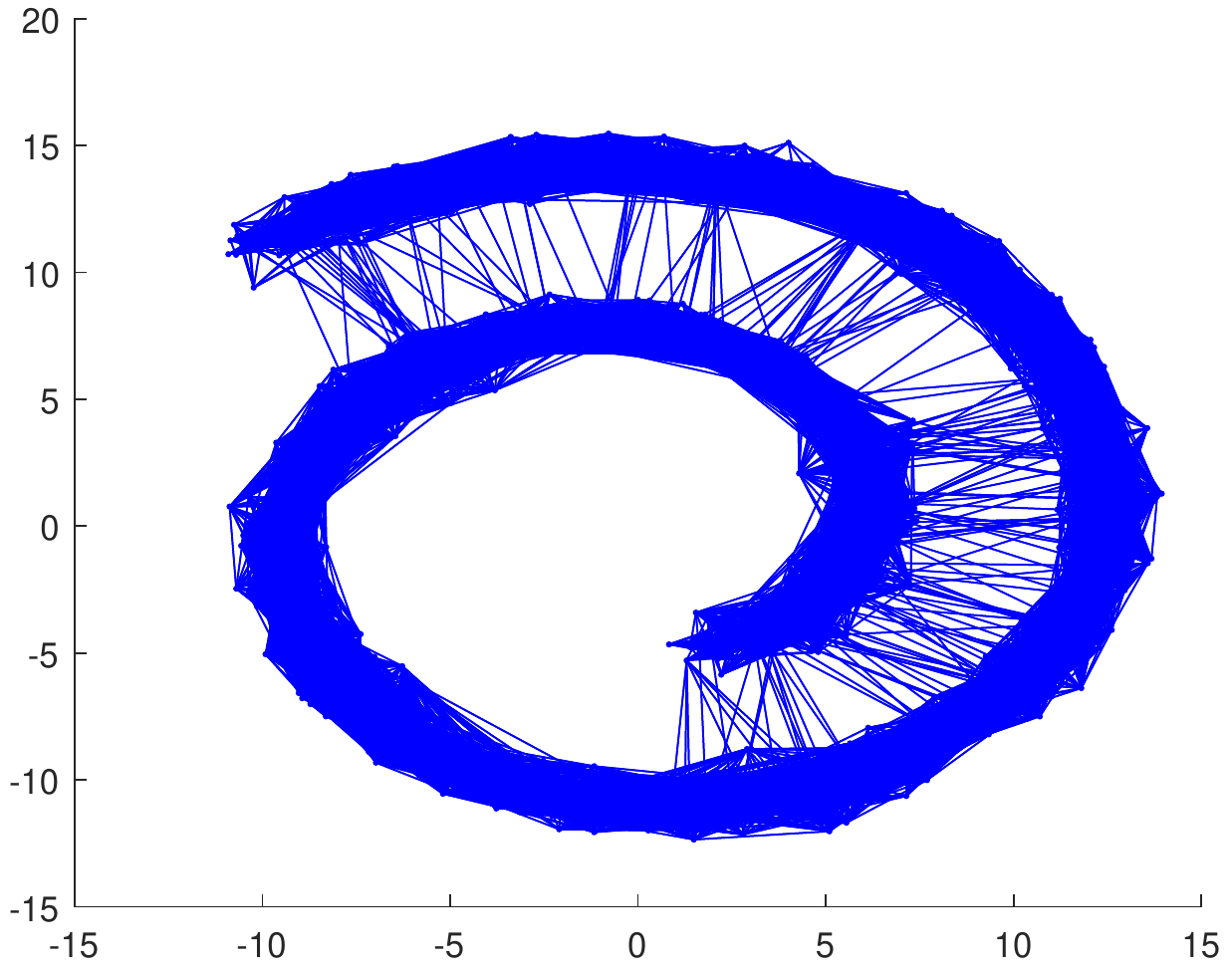}
\end{minipage}\hfill
\begin{minipage}{0.3\textwidth}
        \centering
\includegraphics[scale=.4]{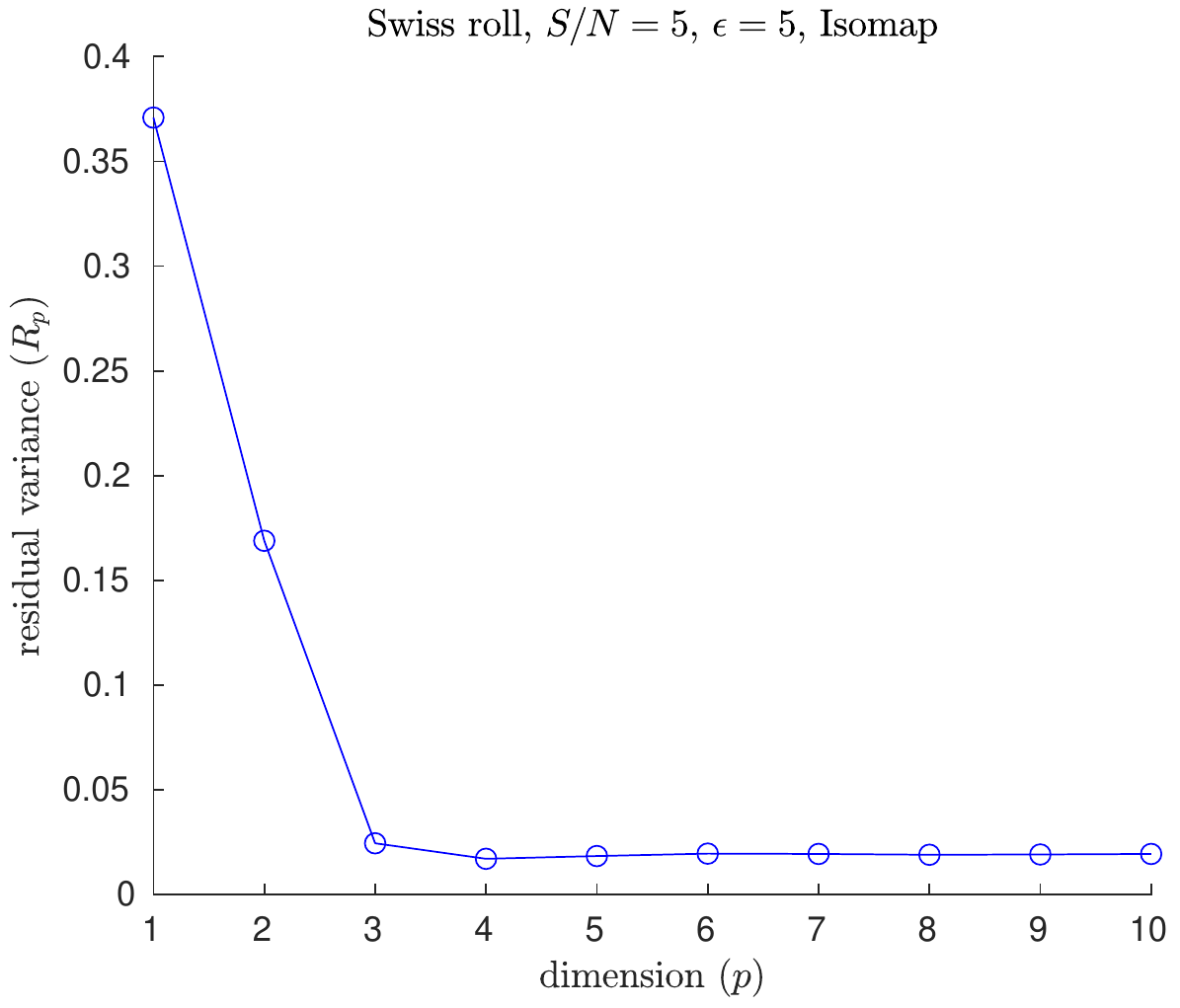}
\end{minipage}\hfill
\begin{minipage}{0.3\textwidth}
   \centering
\includegraphics[scale=.4]{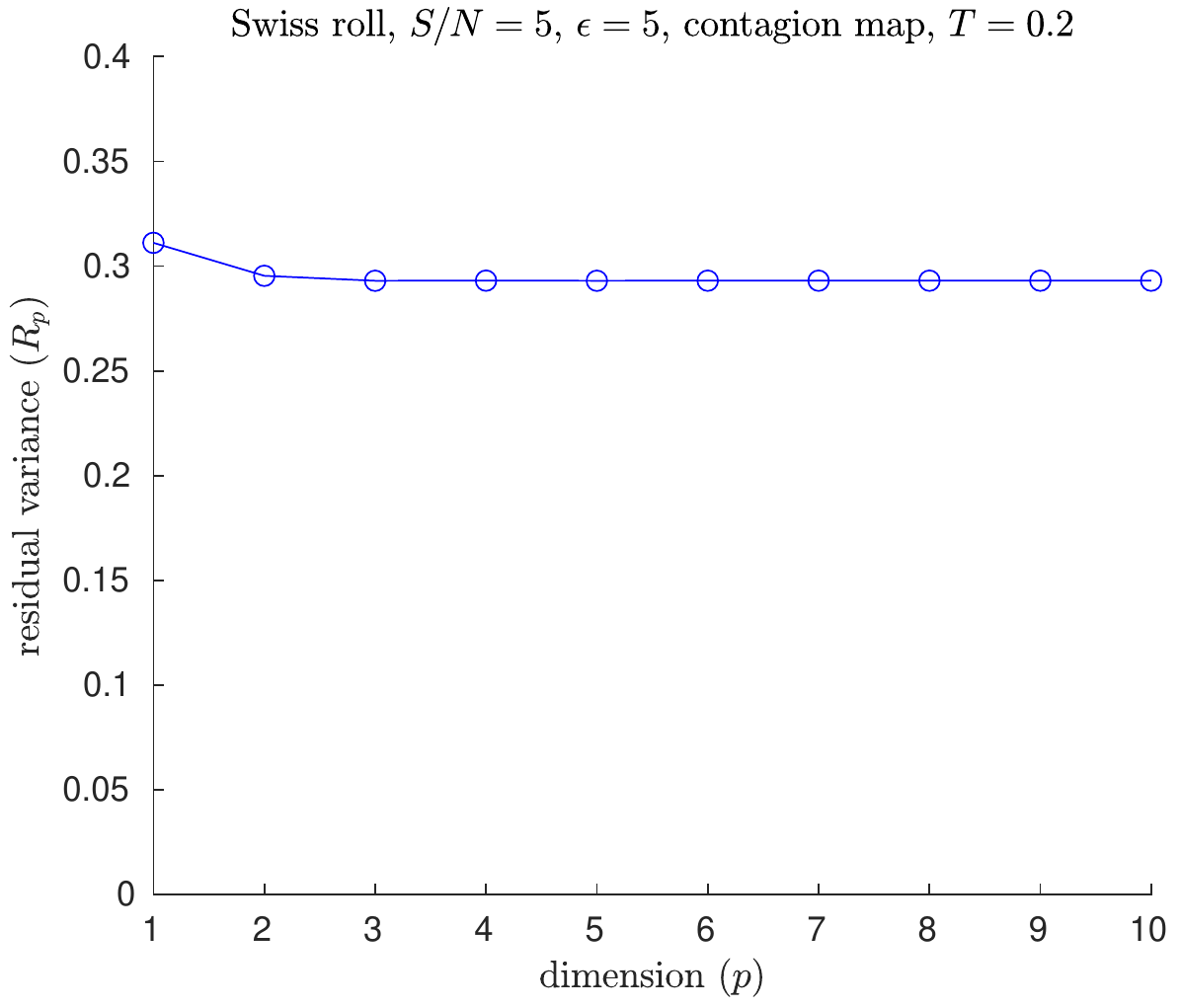}
\end{minipage}\hfill

\caption{Neighbourhood graphs on $2000$ points of the Swiss roll data set from \cite{Tenenbaum2000} with white Gaussian noise added with $S/N=5$ (left column). The residual variances of MDS projections to dimensions $1$ to $10$ resulting from (middle column) the approximate pairwise geodesic distances between nodes based on shortest path on the weighted graph (i.e.~Isomap) and (right column) the activation times in a threshold contagion on the unweighted graph (i.e.~contagion maps). The neighbourhood graphs are (a) the $5$-nearest neighbour graph, (b) the $8$-nearest neighbour graph, (c) the $4.5$-neighbourhood graph, and (d) the $5$-neighbourhood graph. }\label{Tenenbaum_Swiss_roll_neighbourhood_noise5}
\end{figure}

For the lowest signal-to-noise ratio (i.e.~greatest noise level) that we consider --- namely $S/N=5$ --- the $8$-nearest-neighbour graph includes noisy inter-sheet edges. As a result, Isomap does not detect the intrinsic dimension $2$ of the Swiss roll when using the $8$-nearest neighbour graph, and neither does the contagion map with $T=0$ or $T=0.1$, thresholds for which the activation times of our threshold contagion are close to the shortest paths in the unweighted neighbourhood graph (see Figure~\ref{Tenenbaum_Swiss_roll_neighbourhood_noise5}). With $T=0.2$, however, the contagion map does correctly recover the intrinsic dimension $2$, as this threshold is just large enough to be robust to the occurring noisy edges. For thresholds larger than $T=0.2$, many of the realizations of our threshold contagion leave nodes in the neighbourhood graph inactive (recorded as `infinite' activation times), and, as a result, the residual variances of the embeddings based on these activation times are large for all considered dimensions ($1$ to $10$). This illustrates that, while contagion maps can be a powerful tool when dealing with such noisy edges, a suitable choice of threshold can be a delicate matter.

\begin{figure}[H]
\centering
   \leftline{\hskip 0.00cm (a)} 
   \centering
     \begin{minipage}{.3\textwidth} 
\includegraphics[scale=.35]{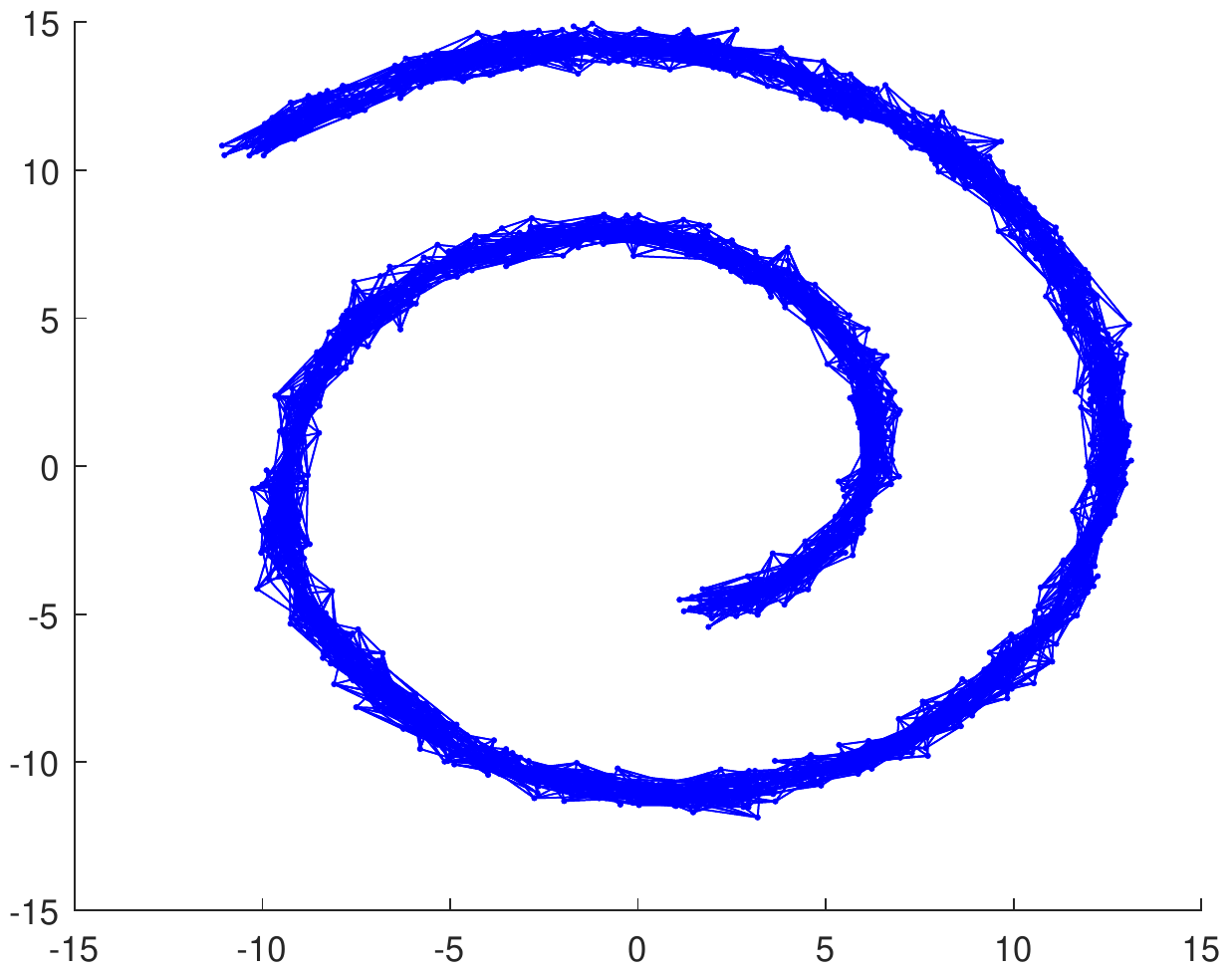}
\end{minipage}\hfill
\begin{minipage}{0.3\textwidth}
        \centering
\includegraphics[scale=.4]{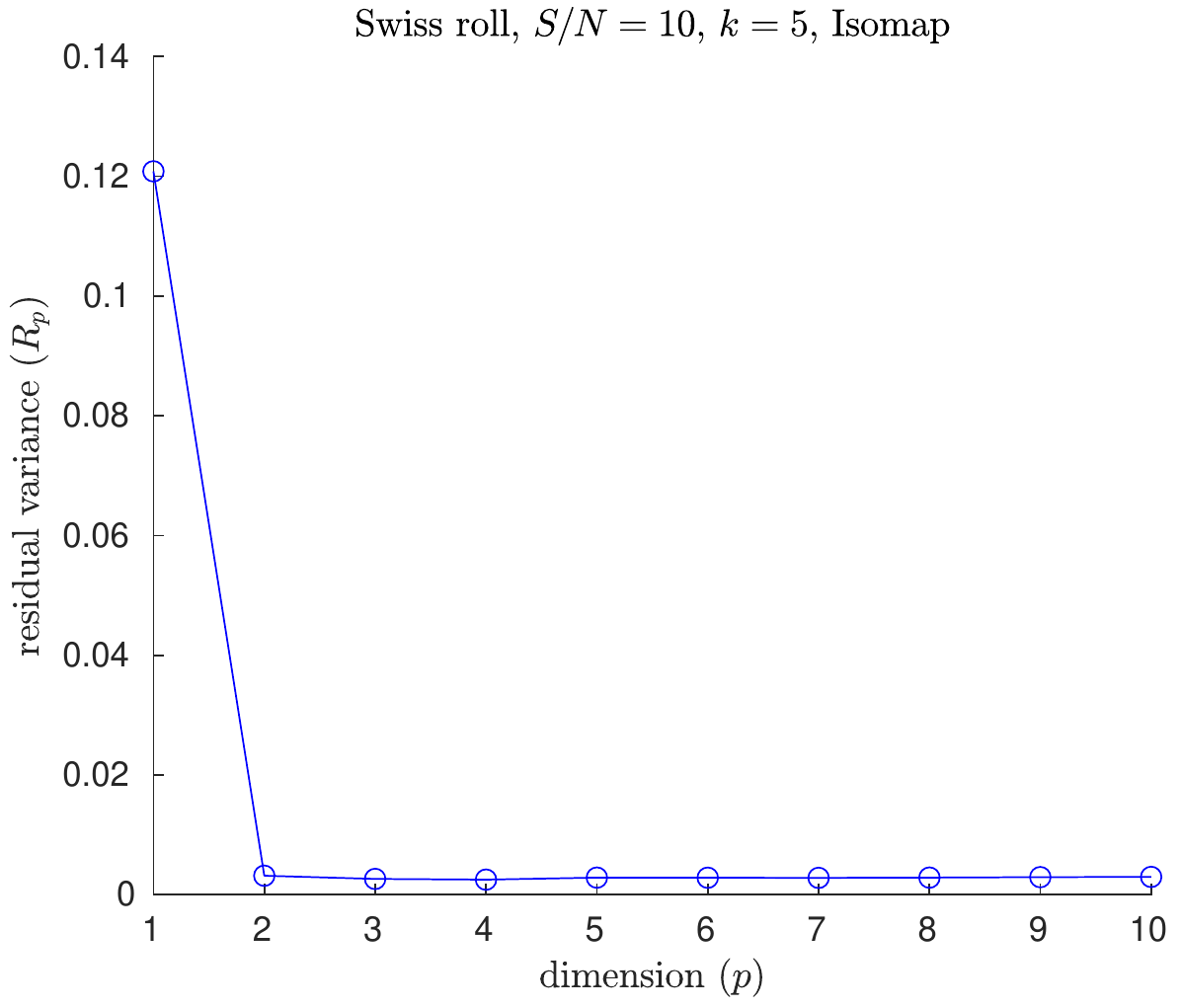}
\end{minipage}\hfill
\begin{minipage}{0.3\textwidth}
        \centering
\includegraphics[scale=.4]{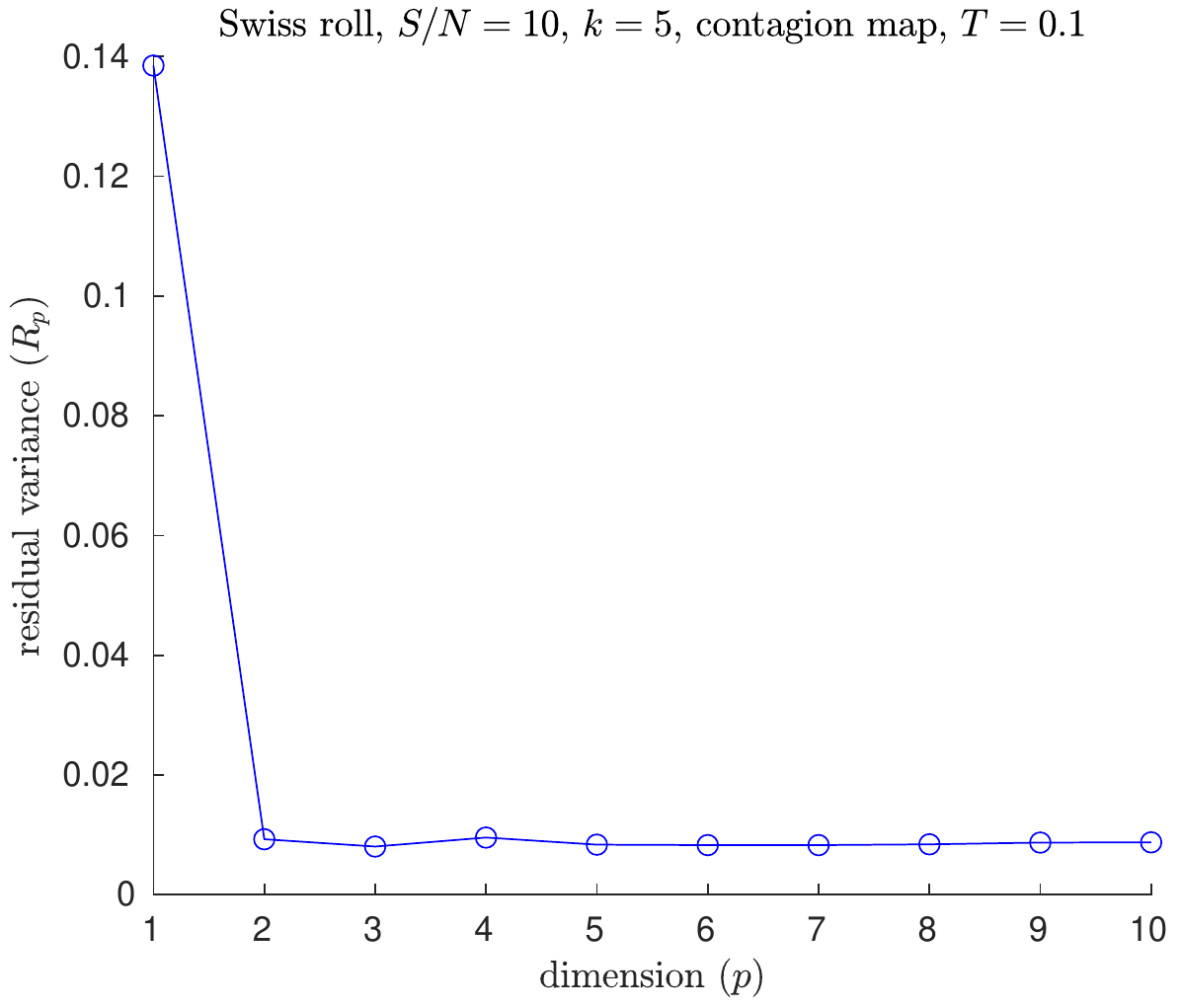}
\end{minipage}\hfill

   \leftline{\hskip 0.00cm (b) } 
   \centering
     \begin{minipage}{0.3\textwidth} 
\includegraphics[scale=.35]{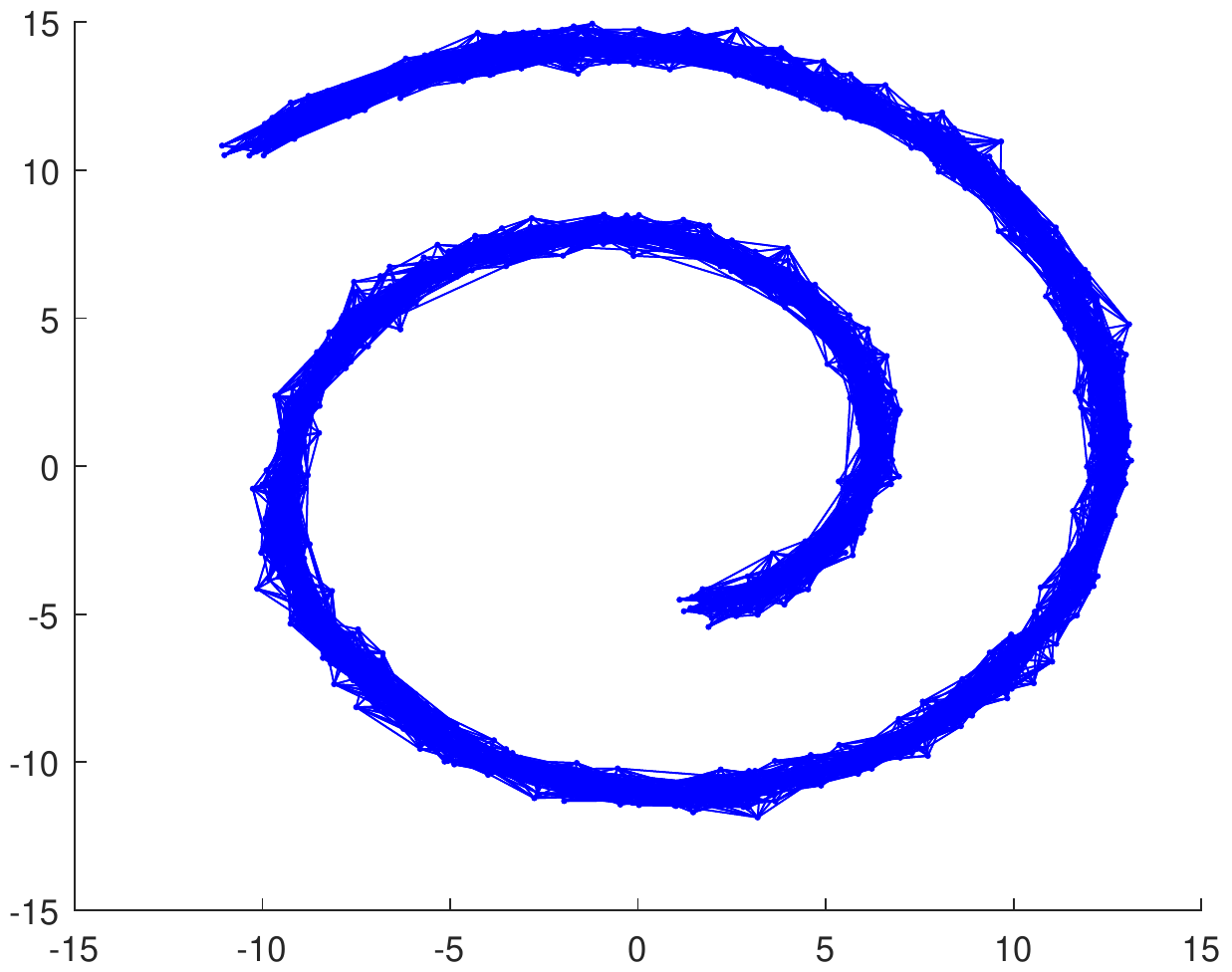}
\end{minipage}\hfill
\begin{minipage}{0.3\textwidth}
        \centering
\includegraphics[scale=.4]{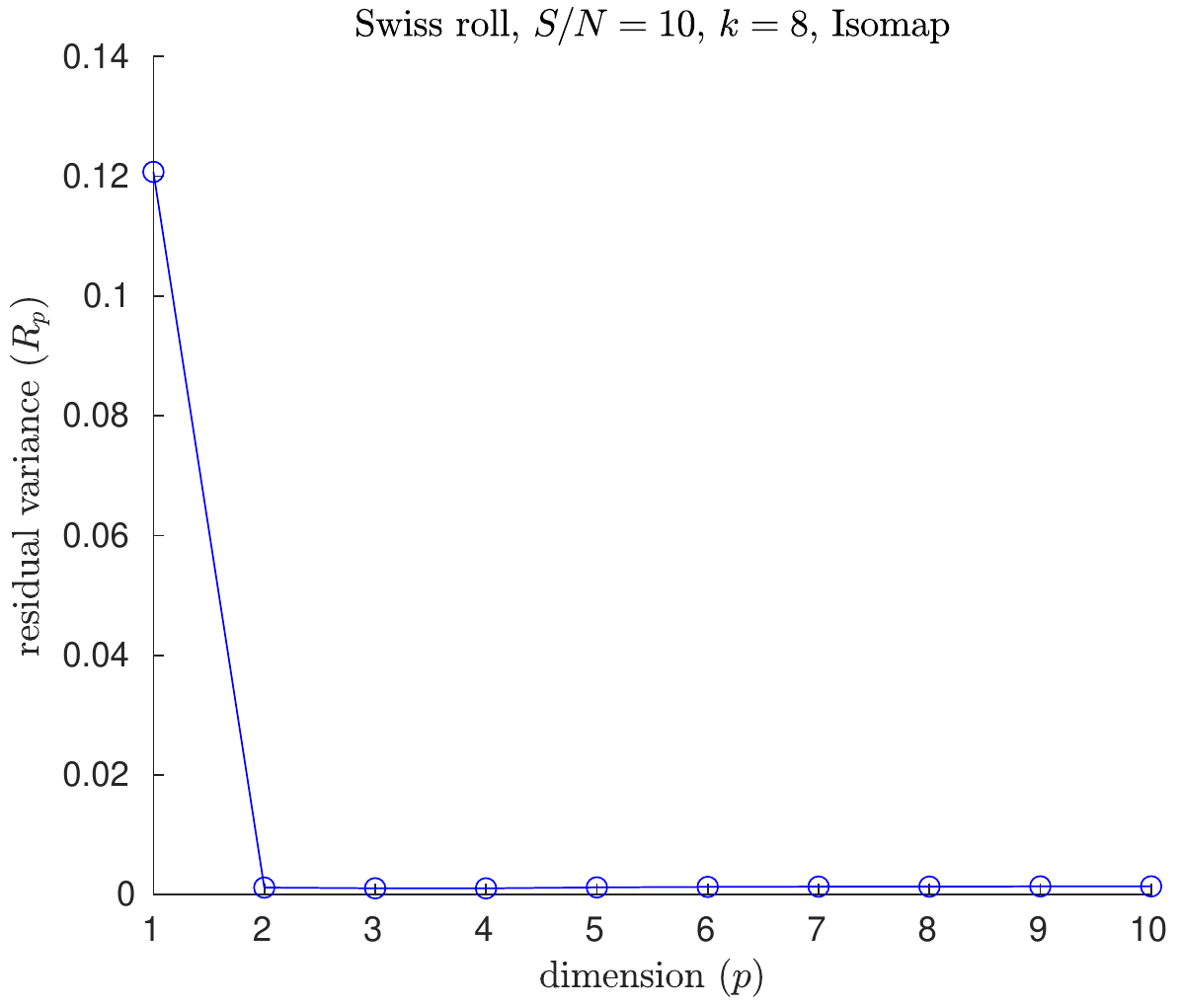}
\end{minipage}\hfill
\begin{minipage}{0.3\textwidth}
        \centering
\includegraphics[scale=.4]{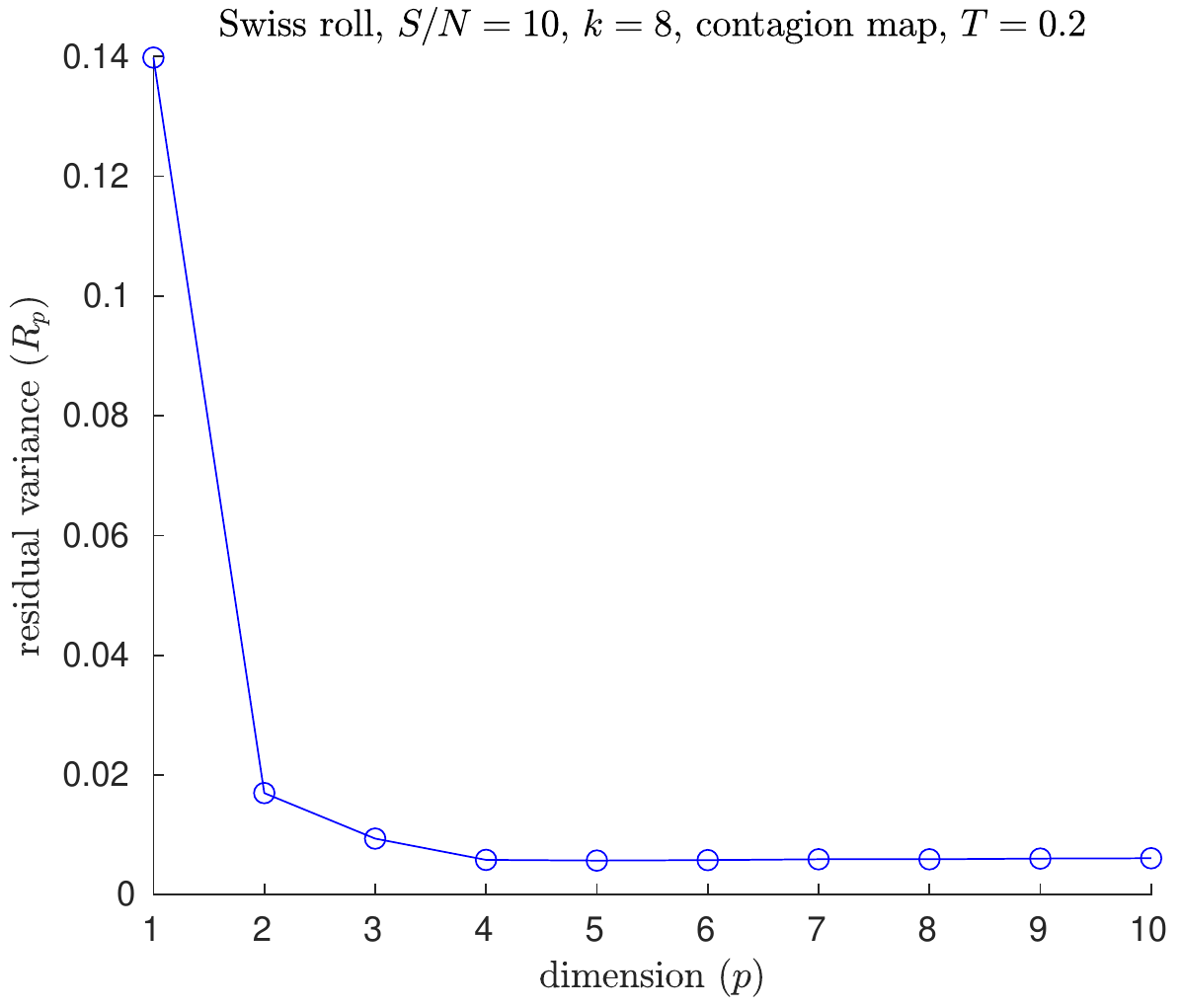}
\end{minipage}\hfill

 \leftline{\hskip 0.00cm (c)} 
 \centering
     \begin{minipage}{0.3\textwidth} 
\includegraphics[scale=.35]{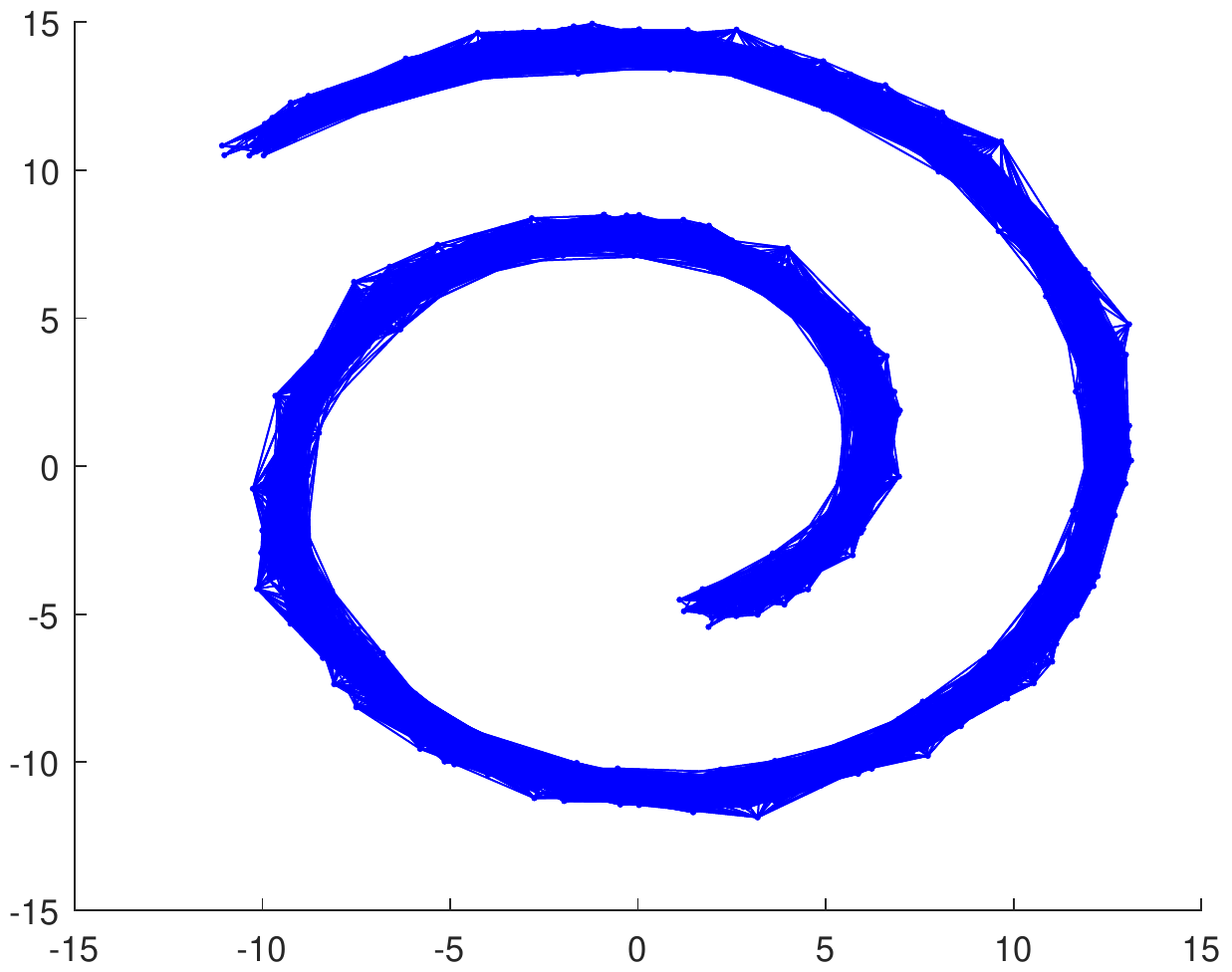}
\end{minipage}\hfill
\begin{minipage}{0.3\textwidth}
        \centering
\includegraphics[scale=.4]{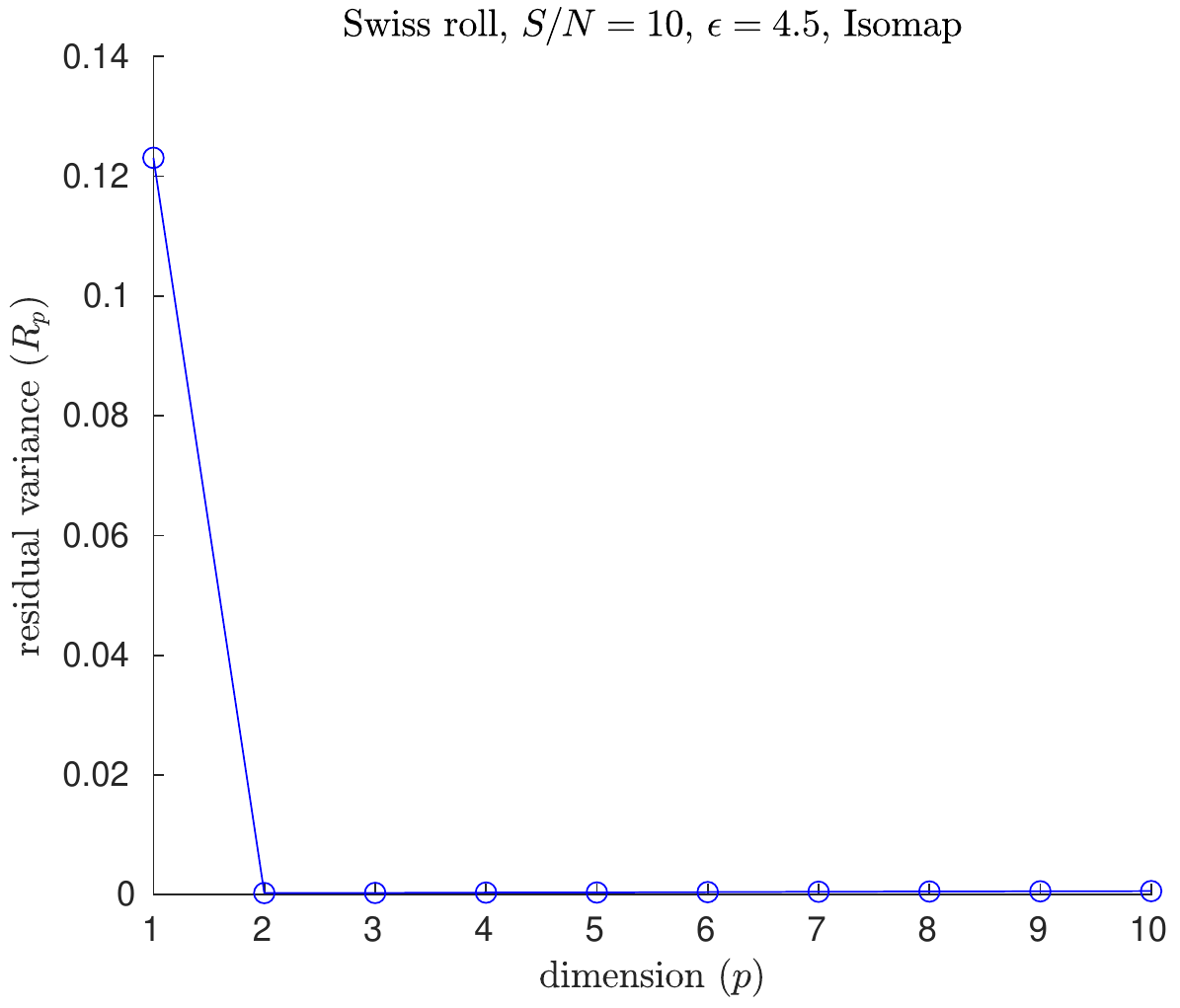}
\end{minipage}\hfill
\begin{minipage}{0.3\textwidth}
        \centering
\includegraphics[scale=.4]
{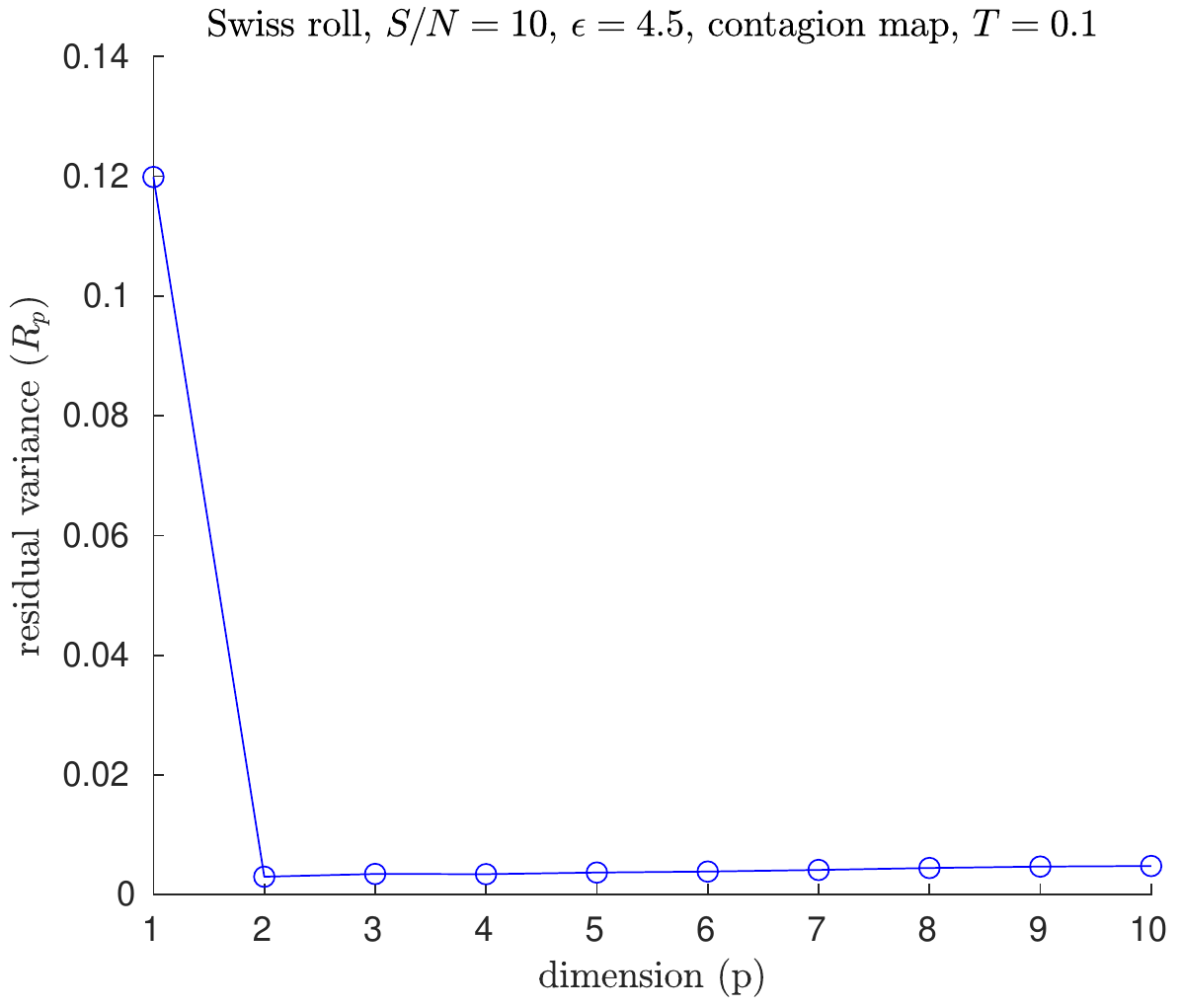}
\end{minipage}\hfill

   \leftline{\hskip 0.00cm (d)}
   \centering
     \begin{minipage}{0.3\textwidth} 
\includegraphics[scale=.35]{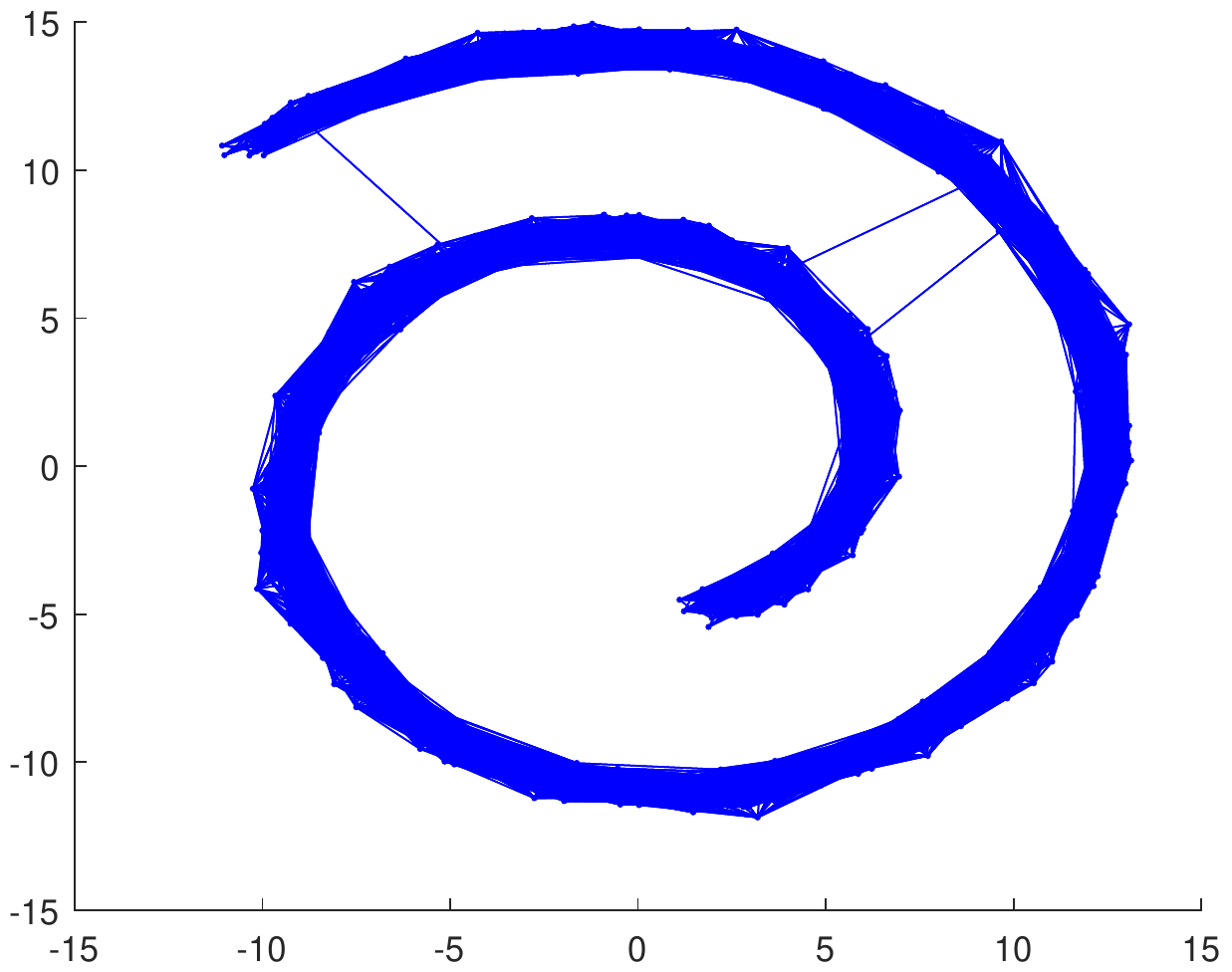}
\end{minipage}\hfill
\begin{minipage}{0.3\textwidth}
        \centering
\includegraphics[scale=.4]{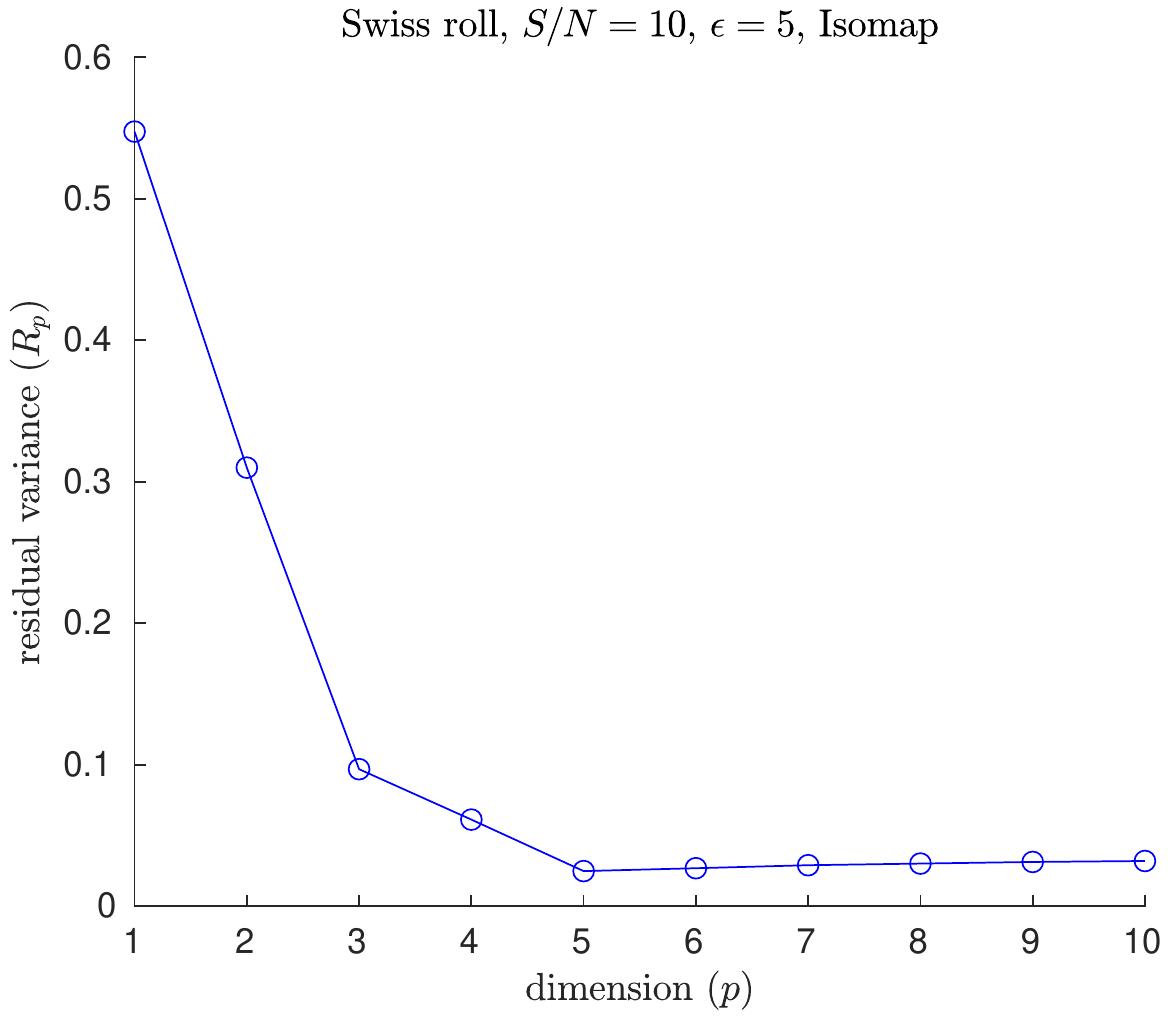}
\end{minipage}\hfill
\begin{minipage}{0.3\textwidth}
   \centering
\includegraphics[scale=.4]{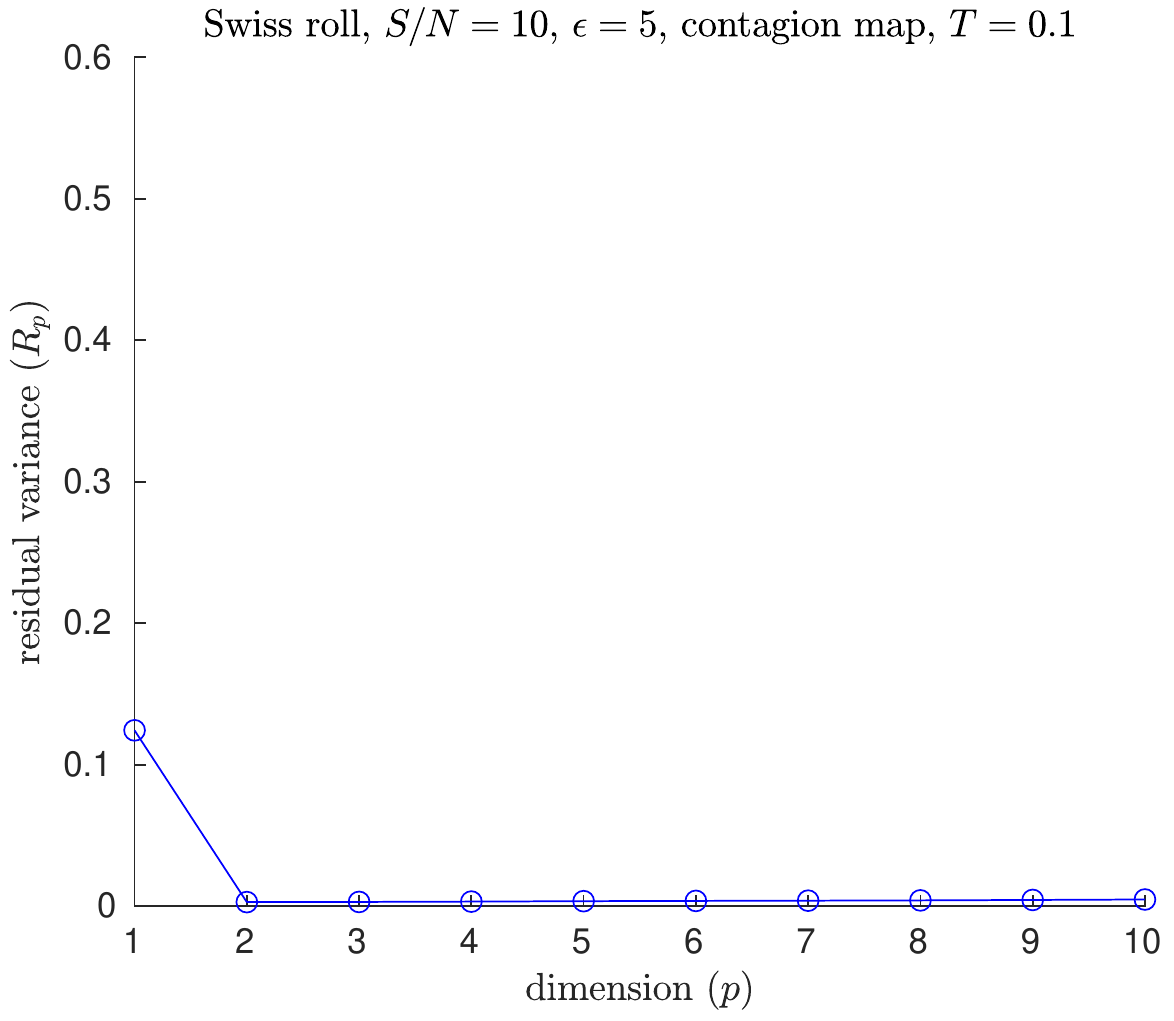}
\end{minipage}\hfill
\caption{Neighbourhood graphs on $2000$ points of the Swiss roll data set from \cite{Tenenbaum2000} with white Gaussian noise added with $S/N=10$ (left column). The residual variances of MDS projections to dimensions $1$ to $10$ resulting from (middle column) the approximate pairwise geodesic distances between nodes based on shortest paths on the weighted graph (i.e.~Isomap) and (right column) the activation times in a threshold contagion on the unweighted graph (i.e.~contagion maps). The neighbourhood graphs are (a) the $5$-nearest neighbour graph, (b) the $8$-nearest neighbour graph, (c) the $4.5$-neighbourhood graph, and (d) the $5$-neighbourhood graph.}
\label{Tenenbaum_Swiss_roll_neighbourhood_noise10}
\end{figure}

Similarly, for signal-to-noise ratio $S/N=5$, Isomap fails when based on the $4.5$-neighbourhood graph or the $5$-neighbourhood graph, but the contagion map for a threshold of $T=0.2$ successfully identifies the intrinsic dimension $2$ for both examined values of the neighbourhood parameter $\epsilon$. 
Furthermore, when based on a $5$-neighbourhood graph, Isomap fails even for the higher $S/N=10$, as the $5$-neighbourhood graph includes noisy inter-sheet edges even for this lower noise level (see Figure~\ref{Tenenbaum_Swiss_roll_neighbourhood_noise10}). 

Note that our measures for topology and geometry are not useful for this data set, as the underlying manifold has no non-trivial topological features, and there is no base-geometry provided.   

\subsection{Torus-based networks}
\begin{figure}[H]
\centering
\includegraphics[width=.3\textwidth]{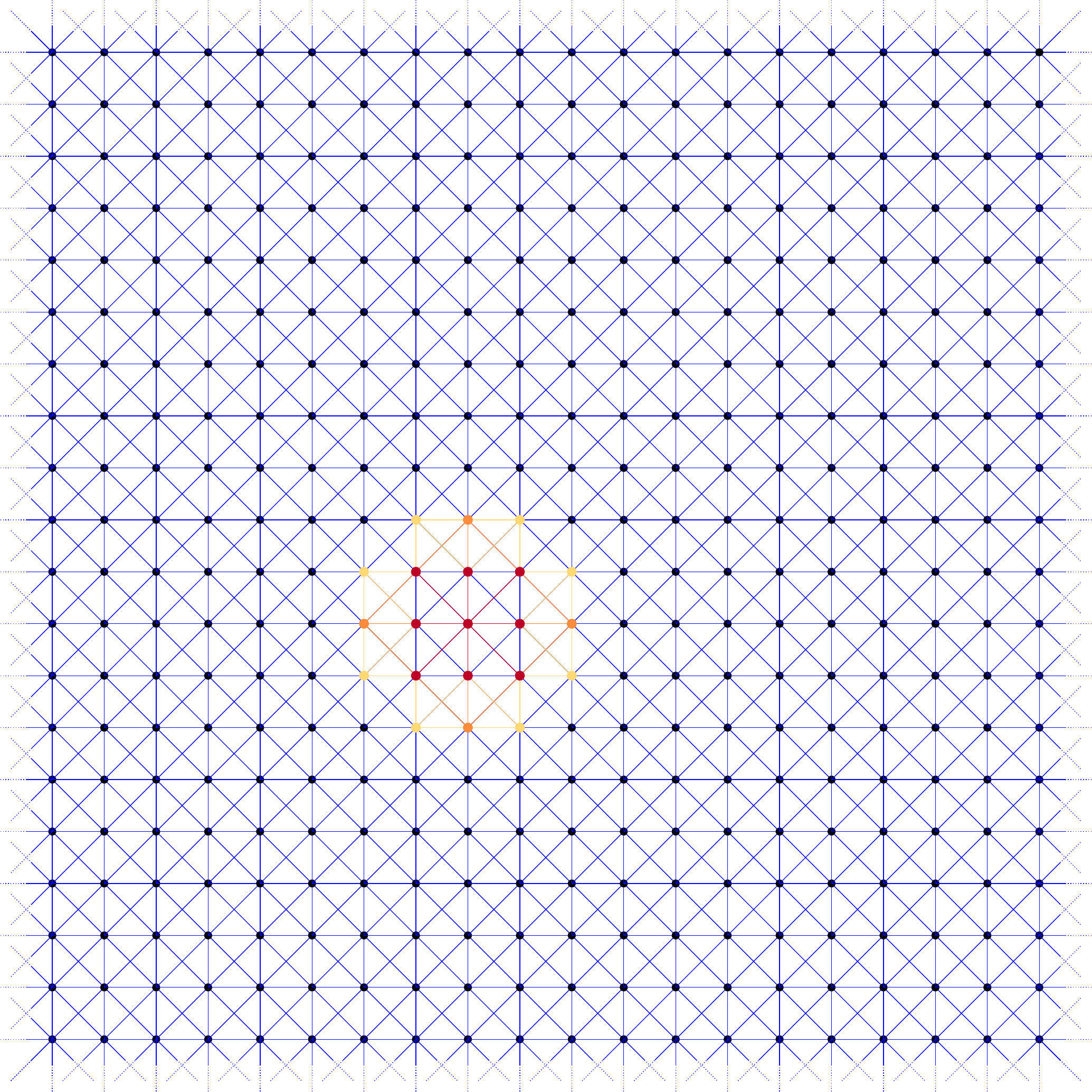}
\caption{Illustration of the purely geometric base network of the networks that we consider in this section. This network is similar to  Kleinberg's small-world like network \cite{Kleinberg2000small}. Each node has $8$ geometric neighbours; nodes on opposite sides of the illustration are adjacent. The set of dark red nodes consists of one node and its immediate neighbours. We consider realizations of our threshold contagion that are seeded at such a set of nodes. The set of medium colour (respectively, light colour) nodes consists of the nodes that are activated in the first (respectively, second) time step when $T<\frac{3}{8}$. We consider this network (on $2500$ nodes) with $2$ and $4$ non-geometric edges per node added uniformly at random.}
\label{eight_base}
\end{figure}

We consider the torus-based model described in \cite{Mahler2020} with $N=2500$ nodes and a geometric degree of $d^{\rm{G}}=8$ (see Figure~\ref{eight_base}). Networks in this model are similar to Kleinberg's small-world like network \cite{Kleinberg2000small}. They consist of a periodic grid of nodes that are connected via \emph{geometric edges} (i.e.~edges between neighbouring nodes) in a regular manner, and to which \emph{non-geometric edges} are added according to a probability distribution. We add first $2$ and then $4$ non-geometric edges per node uniformly at random (i.e.~$d^{\rm{NG}}=2$ or $4$, and $\gamma=0$, using the parameters described in \cite{Mahler2020}) and apply versions of both Isomap and contagion maps (with thresholds $T=0, 0.1, \dots, 1$) to the resulting networks. Namely, we calculate the shortest-path lengths between pairs of nodes in these two unweighted networks, as well as the activation times in all realizations of our threshold contagion that are seeded at the direct neighbourhoods of individual nodes (the red nodes in Figure~\ref{eight_base}). We thus obtain two dissimilarity matrices, one holding the shortest-path lengths ($D_{\rm iso}$) and one holding the symmetrized activation times ($D_{\rm cont}$). We analyse the information held in these two dissimilarity matrices geometrically, topologically, and in terms of dimensionality in two ways each. We first analyse the estimated pairwise geodesic distances held in the dissimilarity matrices directly, and we then consider the point cloud that results from taking columns (or, equivalently, rows) of each dissimilarity matrices as the coordinate vectors of points in $\mathbb{R}^{2500}$. 

In detail, we perform the following analyses: 

\begin{itemize}
\item In terms of dimensionality: We perform MDS based on the entries in each dissimilarity matrix to dimensions $1$ to $10$ and record the residual variance for each dimension. We also perform MDS on the point cloud that results from taking columns (or, equivalently, rows) of each dissimilarity matrix as the coordinate vectors of points in $\mathbb{R}^{2500}$. In both cases, we identify the approximate embedding dimension as the lowest dimension such that the residual variance when projecting down to that dimension via MDS is below 5\%.

\item Topologically: We build Vietoris\textendash Rips filtrations based on the approximations to the pairwise geodesic distances (i.e.~the entries in each dissimilarity matrix) and compute their persistent homologies. We also build Vietoris--Rips filtrations on the points cloud that results from taking columns (or, equivalently, rows) of each dissimilarity matrix as the coordinate vectors of points in $\mathbb{R}^{2500}$ and compute their persistent homologies. 

\item Geometrically: We calculate the Pearson correlation coefficient between the entries in each dissimilarity matrix and the corresponding pairwise distances between regularly spaced points on a torus:  

\begin{equation}\label{reg_points}
	\begin{split}
		\frac{1}{2 \pi} \left( \cos \frac{2 \pi x}{n}, \sin \frac{2 \pi x}{n}, \cos \frac{2 \pi y}{n}, \sin \frac{2 \pi y}{n} \right) \in \mathbb{T} = \frac{1}{2 \pi} \mathbb{S}^1 \times \frac{1}{2 \pi} \mathbb{S}^1 \subset  \mathbb{R}^4 \,, \\
		x,y \in \{0,1, \dots, n-1\} \, .
	\end{split}
\end{equation}

We also calculate the Pearson correlation coefficient between the pairwise distances between points in the point cloud that results from taking columns (or, equivalently, rows) of each dissimilarity matrix as the coordinate vectors of points in $\mathbb{R}^{2500}$ and the corresponding pairwise distances between the regularly spaced points on a torus \eqref{reg_points}. 
\end{itemize}

We find that Isomap is unable to infer the underlying torus structure from these networks. Contagion map, however, detects the characteristics of the torus when using a threshold of $T\approx 0.2$.  
The results in this section illustrate the utility of contagion maps for spatial network data that incorporates noisy edges, and its potential to outperform Isomap in such scenarios, but they also highlight that a careful choice of $T$ is critical.

\subsubsection{Dimensionality}

\begin{figure}[H]
\centering
\leftline{\hskip 0.00cm (a) \hskip 7cm (f) } 
\begin{minipage}{0.5\textwidth}
\includegraphics[width=.8\textwidth]{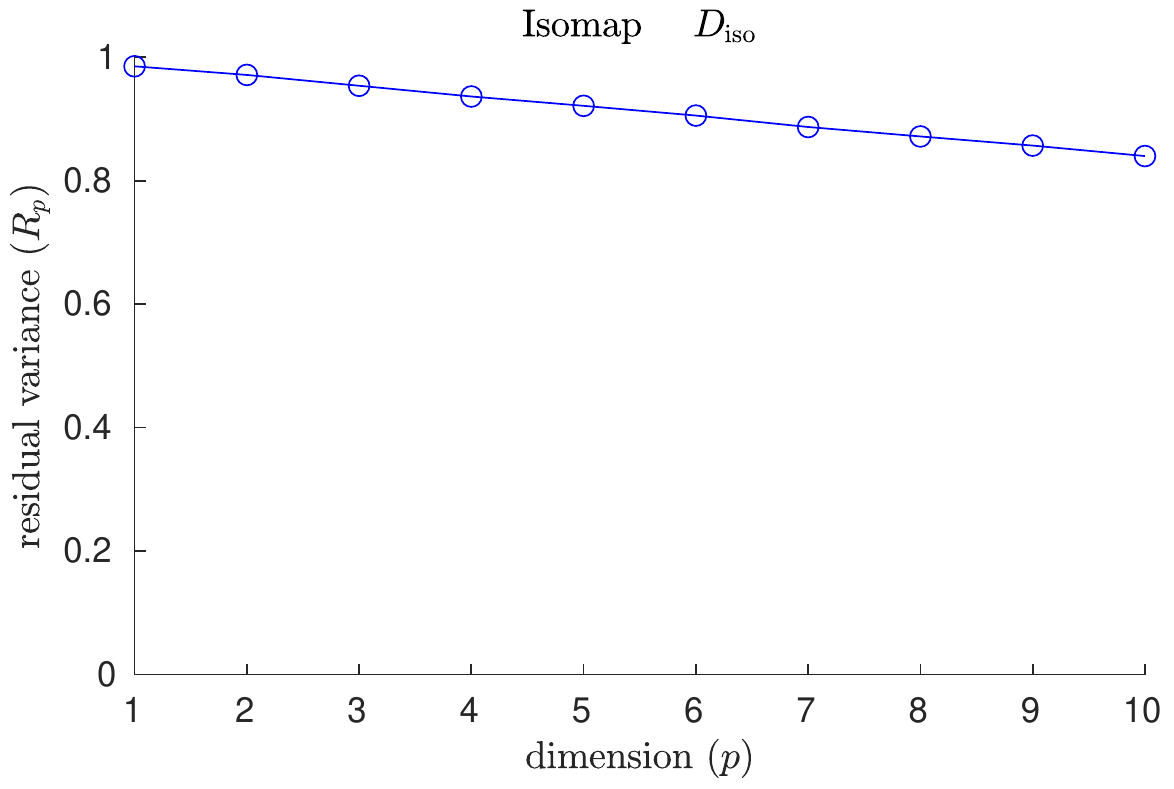}
\end{minipage}\hfill
\begin{minipage}{0.5\textwidth}
\includegraphics[width=.8\textwidth]{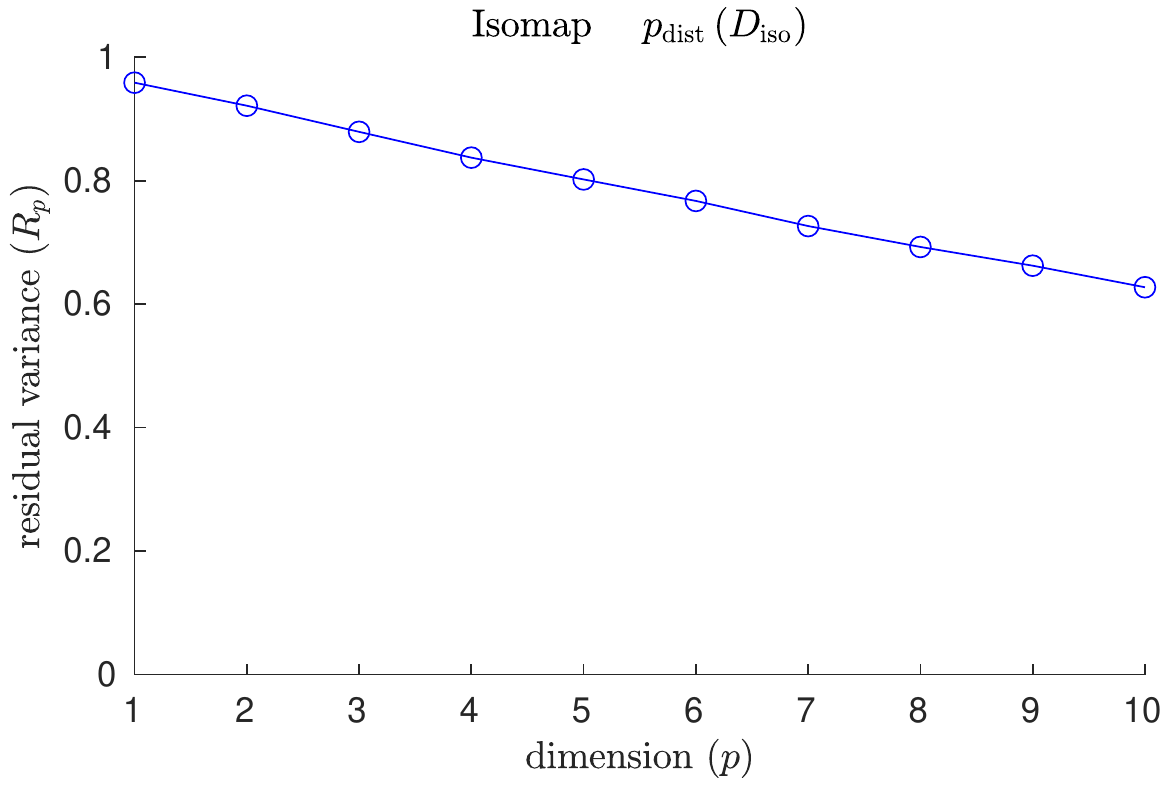}\end{minipage}\hfill

\leftline{\hskip 0.00cm (b) \hskip 7cm (g) } 
\begin{minipage}{0.5\textwidth}
\includegraphics[width=.7\textwidth]{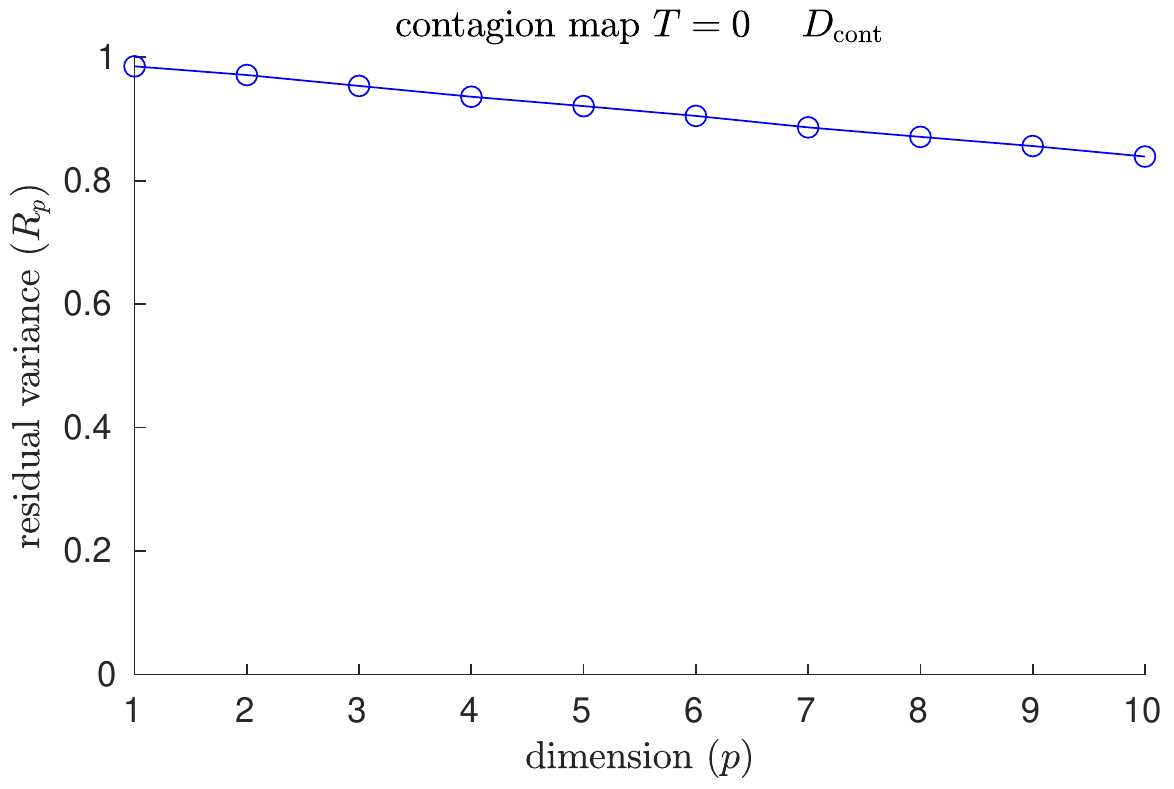}
\end{minipage}\hfill
\begin{minipage}{0.5\textwidth}
\includegraphics[width=.7\textwidth]{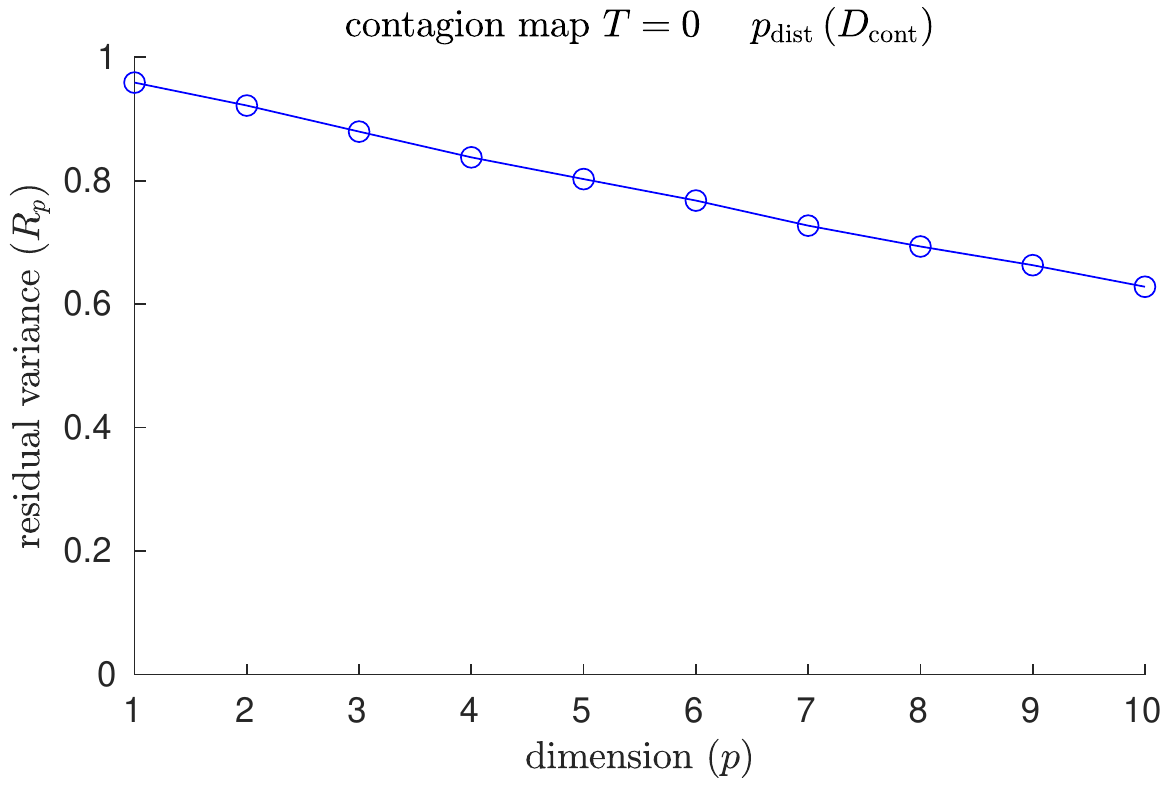}
\end{minipage}\hfill

\leftline{\hskip 0.00cm (c) \hskip 7cm (h) } 
\begin{minipage}{0.5\textwidth}
\includegraphics[width=.7\textwidth]{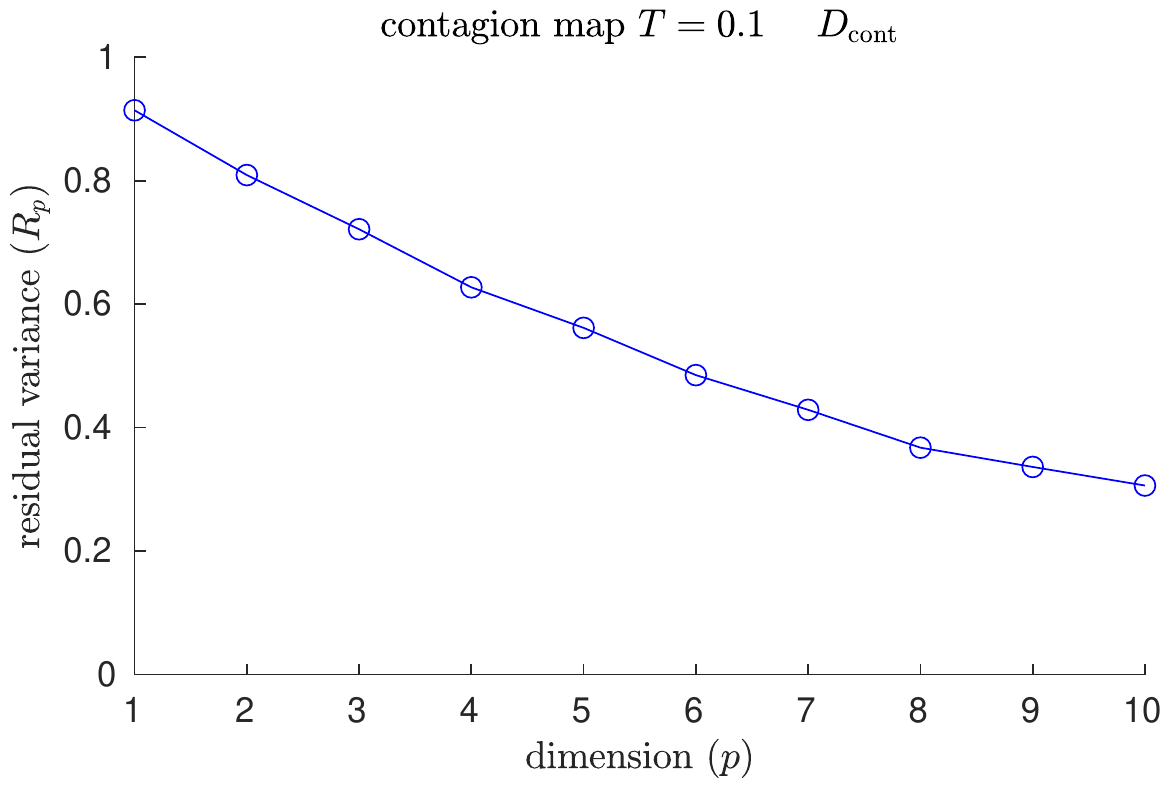}
\end{minipage}\hfill
\begin{minipage}{0.5\textwidth}
\includegraphics[width=.7\textwidth]{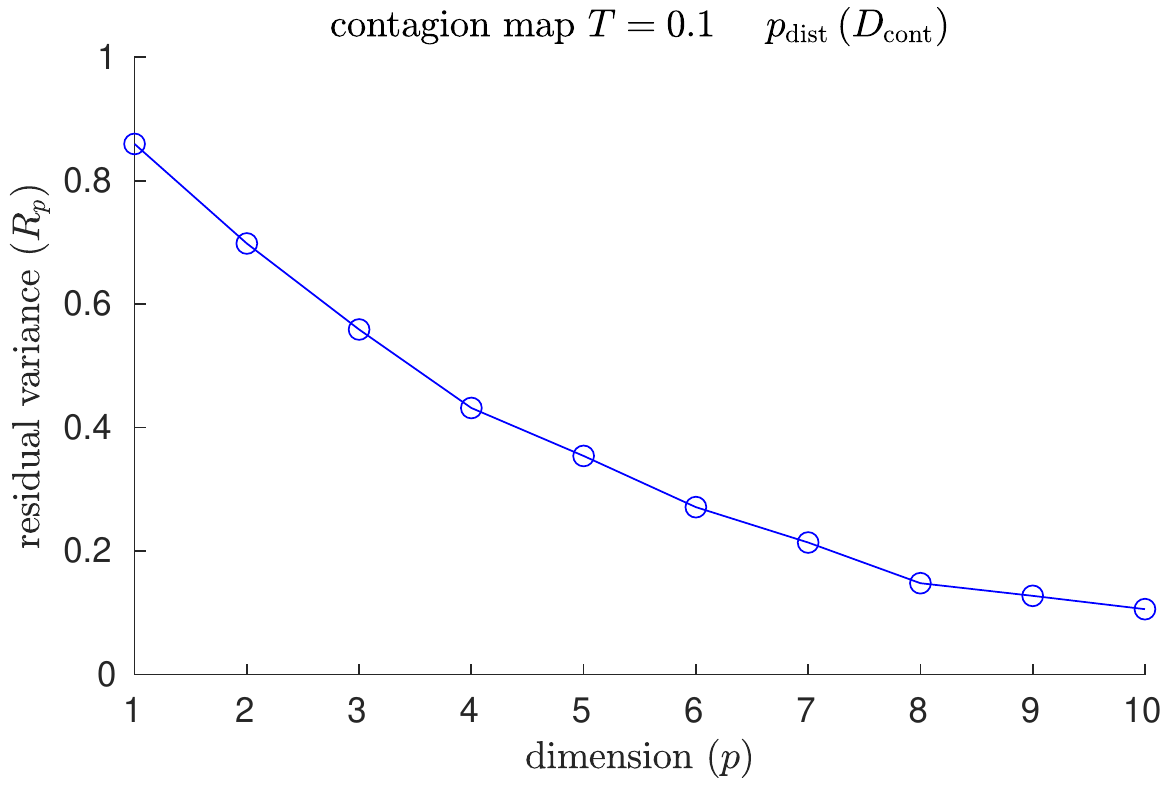}
\end{minipage}\hfill

\leftline{\hskip 0.00cm (d) \hskip 7cm (i) } 
\begin{minipage}{0.5\textwidth}
\includegraphics[width=.7\textwidth]{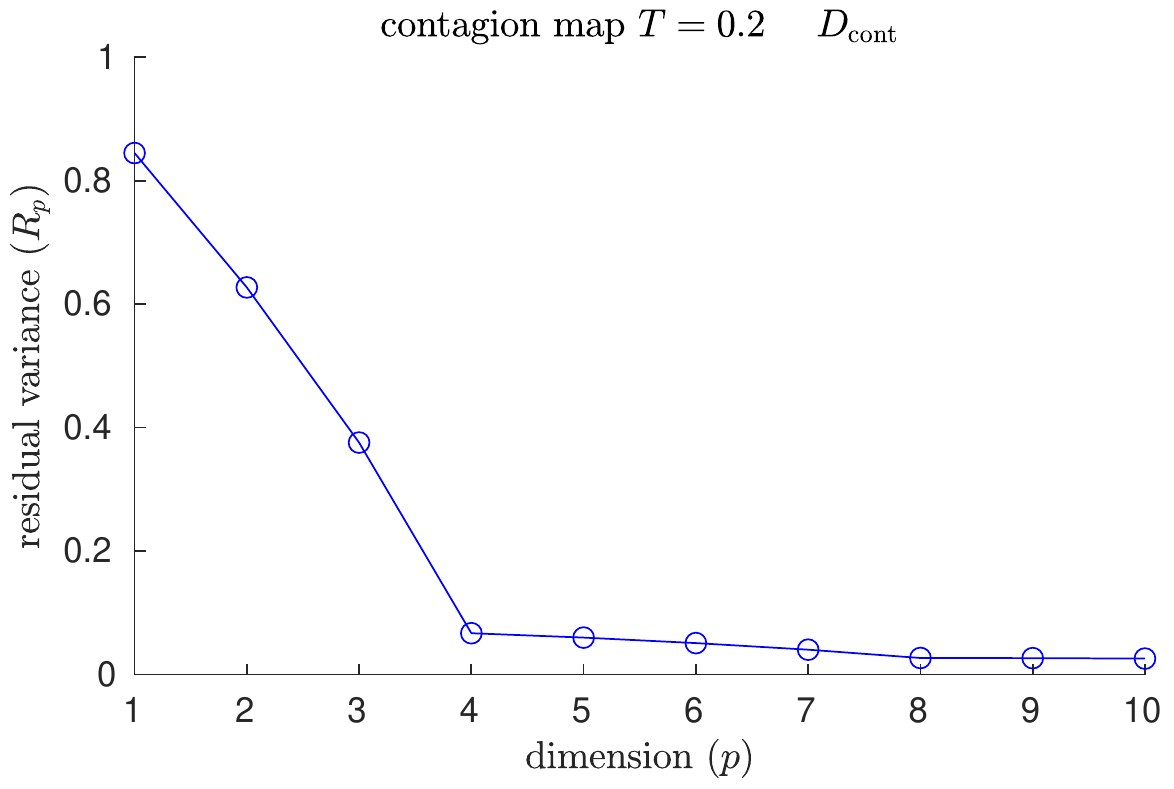}
\end{minipage}\hfill
\begin{minipage}{0.5\textwidth}
\includegraphics[width=.7\textwidth]{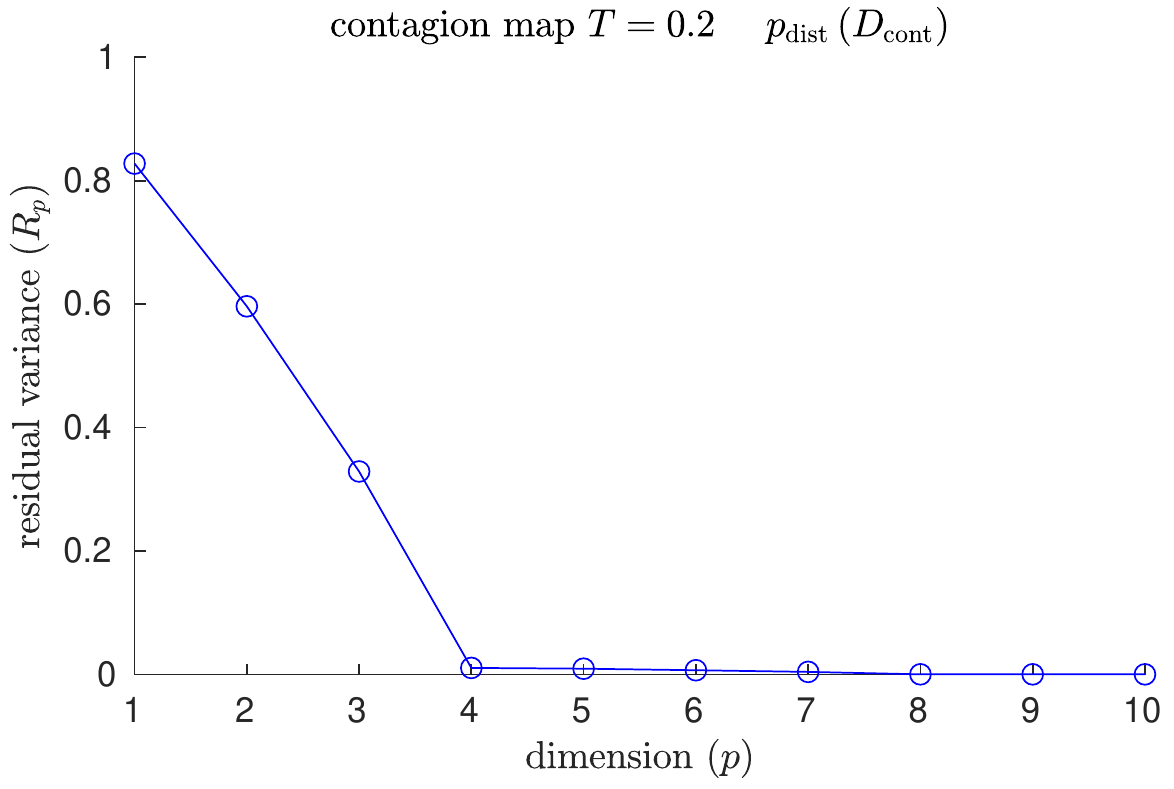}
\end{minipage}\hfill

\leftline{\hskip 0.00cm (e) \hskip 7cm (j) } 
\begin{minipage}{0.5\textwidth}
\includegraphics[width=.7\textwidth]{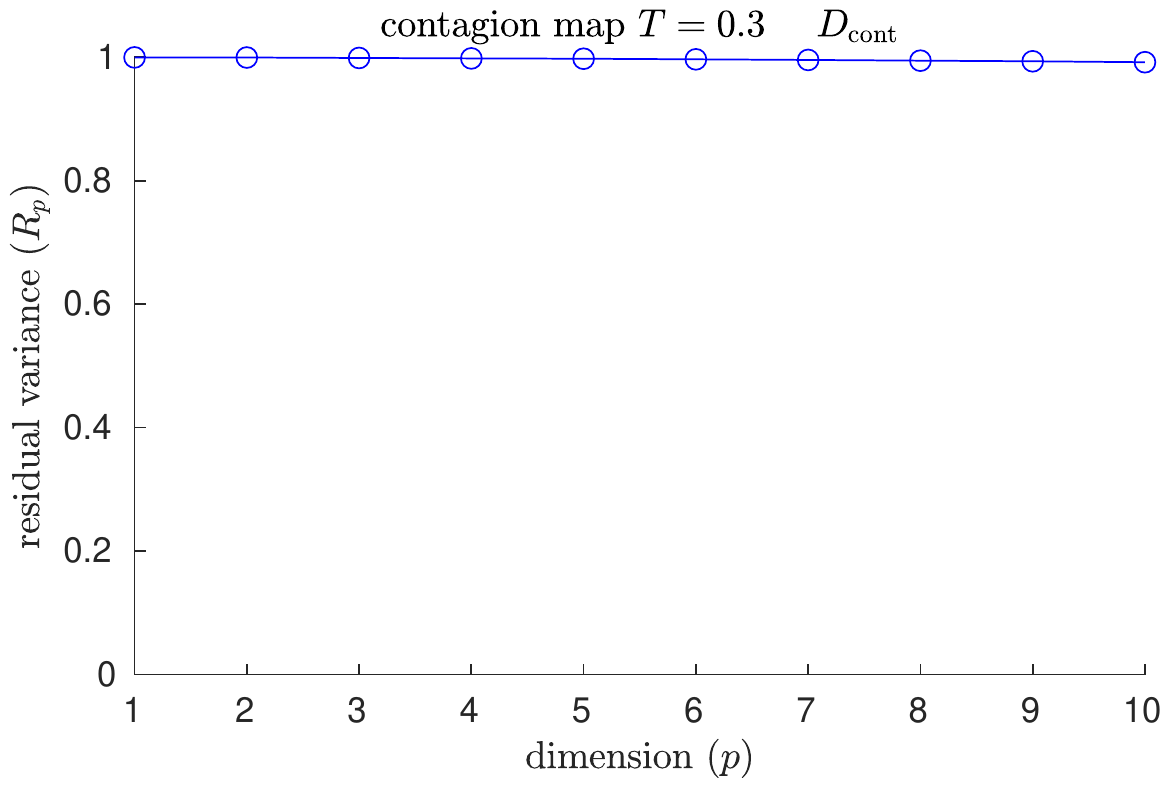}
\end{minipage}\hfill
\begin{minipage}{0.5\textwidth}
\includegraphics[width=.7\textwidth]{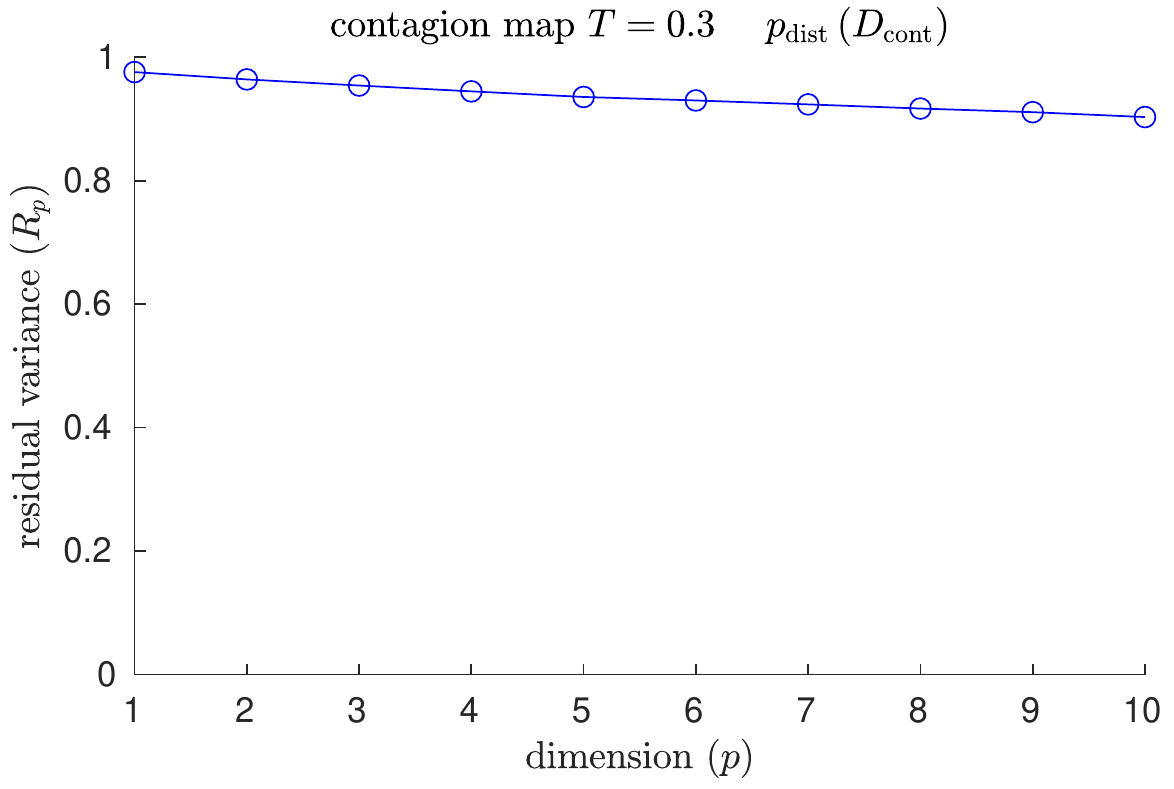}
\end{minipage}\hfill
\caption{Dimensionality results on our torus-based network with $d^{\rm{NG}}=2$. (Left column) Residual variances of MDS based on the estimated geodesic distances (i.e.~the entries in $D_{\rm iso}$ and $D_{\rm cont}$) according to (a) Isomap, and according to contagion maps with thresholds (b) $T=0$, (c) $T=0.1$, (d) $T=0.2$, and (e) $T=0.3$. (Right column) Residual variances of MDS based on the point cloud (i.e.~the rows of the $D_{\rm iso}$ and $D_{\rm cont}$) according to (f) Isomap, and according to contagion maps with thresholds (g) $T=0$, (h) $T=0.1$, (i) $T=0.2$, and (j) $T=0.3$. }
\label{torus_MDS_d_ng_2}
\end{figure}

\begin{figure}[H]
\centering
\leftline{\hskip 0.00cm (a) \hskip 7cm (f) } 
\begin{minipage}{0.5\textwidth}
\includegraphics[width=.7\textwidth]{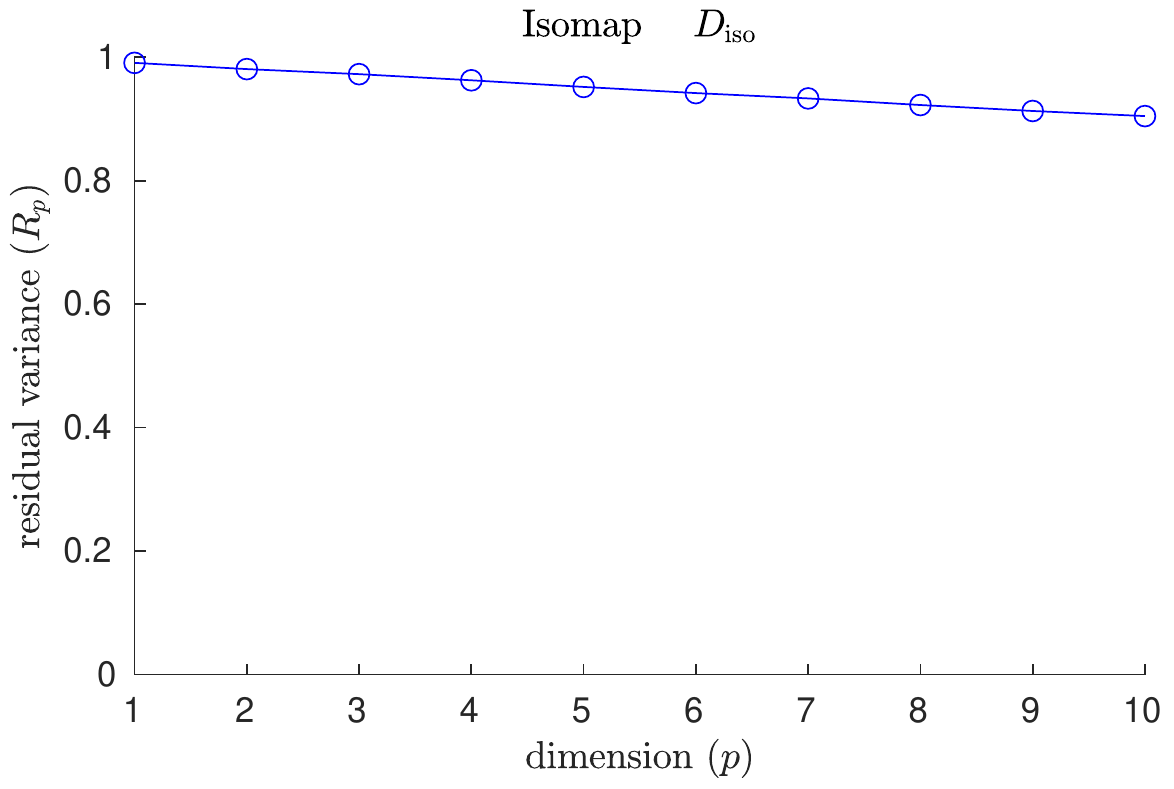}
\end{minipage}\hfill
\begin{minipage}{0.5\textwidth}
\includegraphics[width=.7\textwidth]{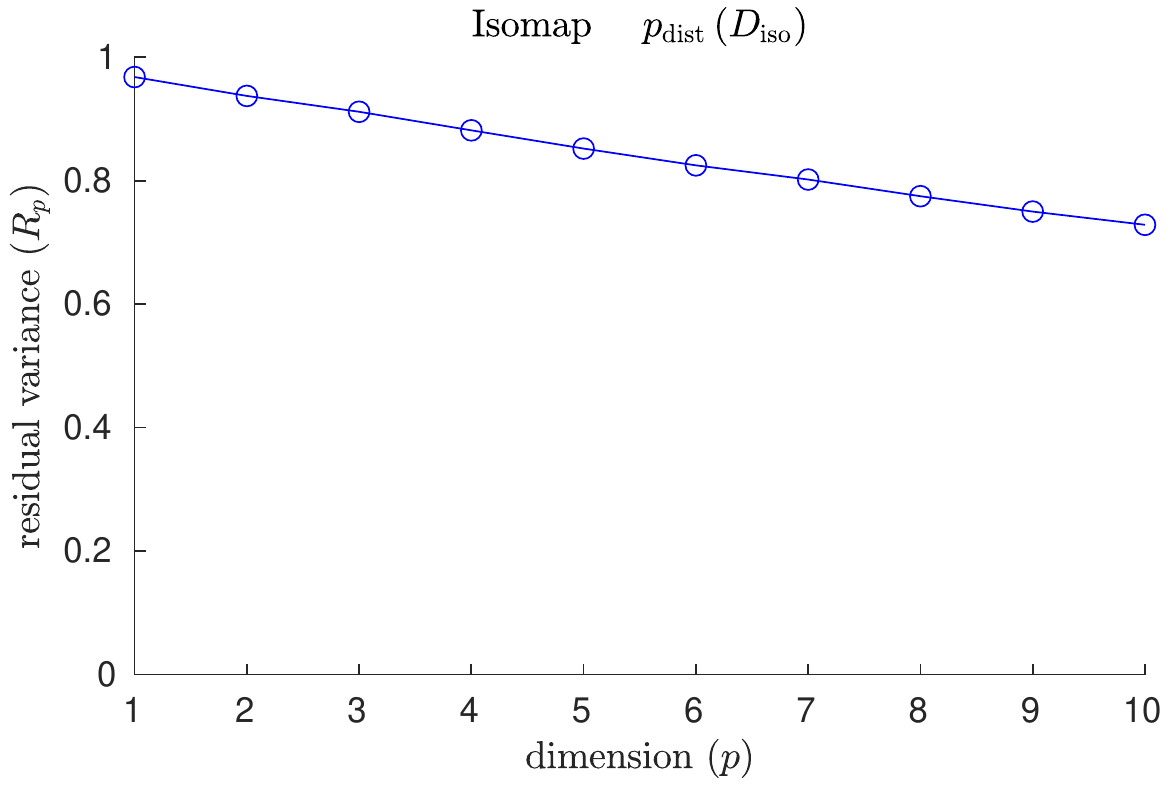}
\end{minipage}\hfill

\leftline{\hskip 0.00cm (b) \hskip 7cm (g) } 
\begin{minipage}{0.5\textwidth}
\includegraphics[width=.7\textwidth]{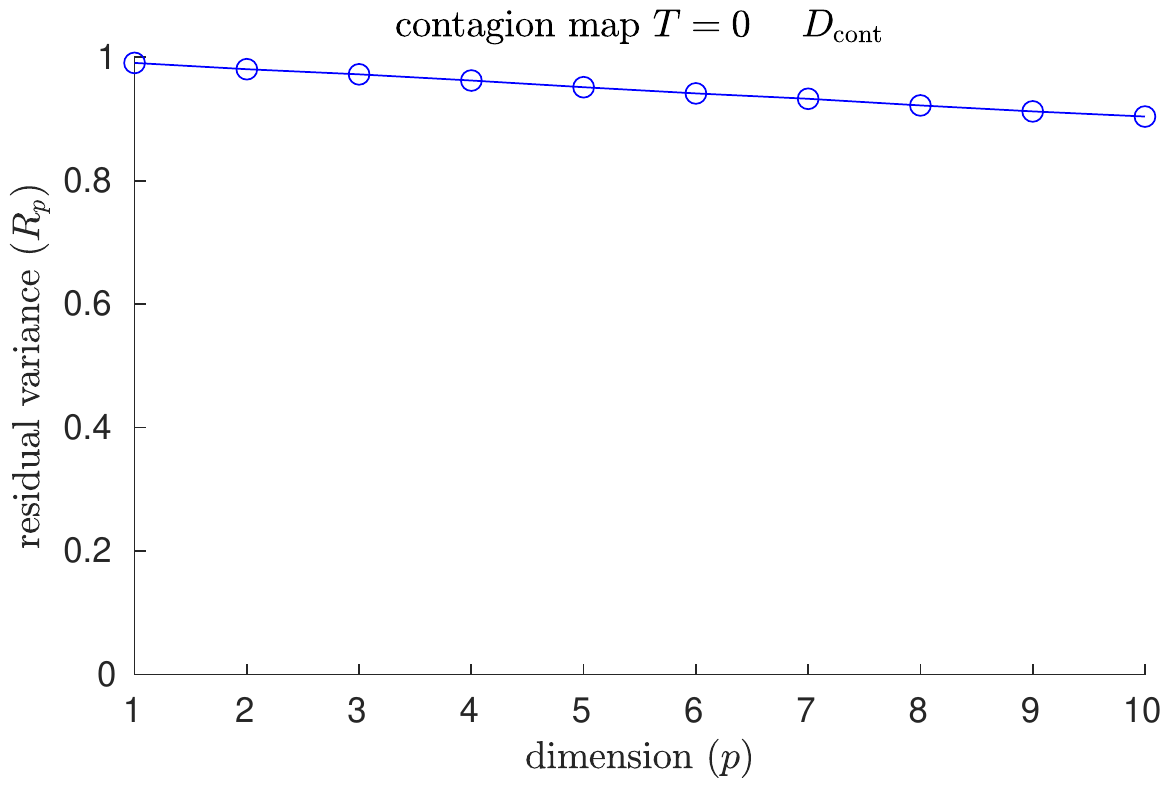}
\end{minipage}\hfill
\begin{minipage}{0.5\textwidth}
\includegraphics[width=.7\textwidth]{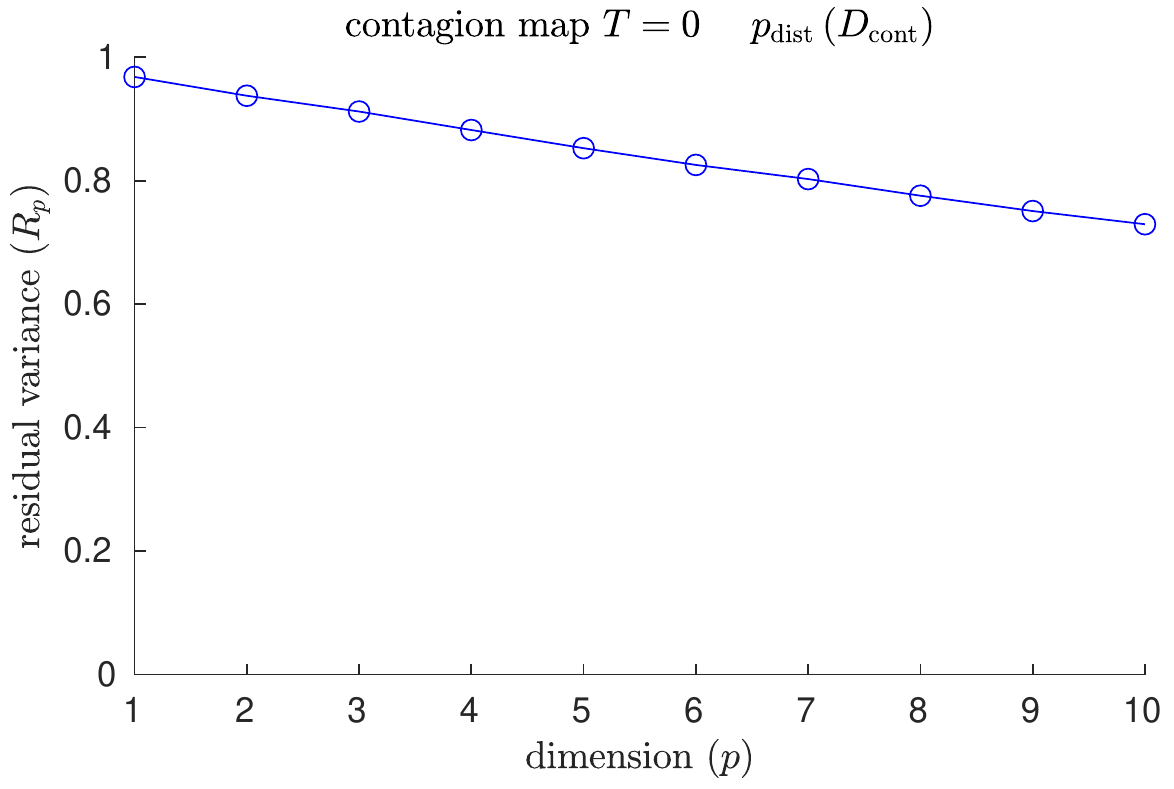}
\end{minipage}\hfill

\leftline{\hskip 0.00cm (c) \hskip 7cm (h) } 
\begin{minipage}{0.5\textwidth}
\includegraphics[width=.7\textwidth]{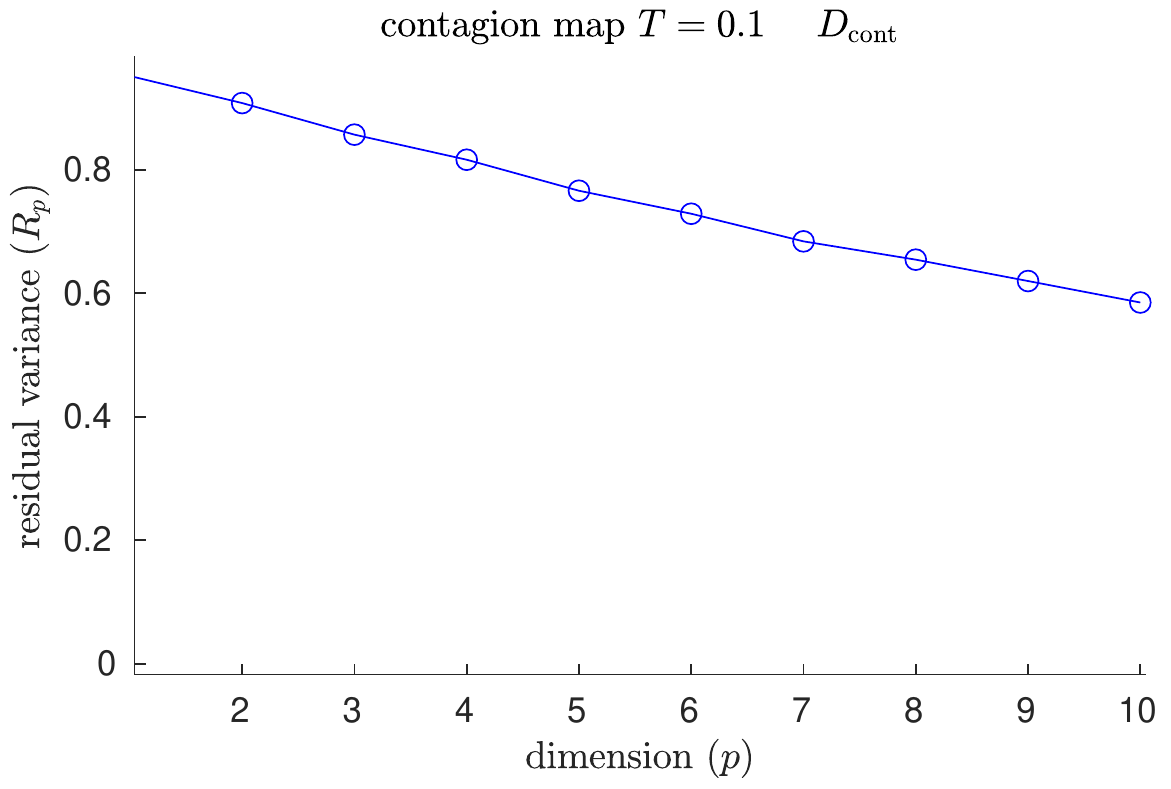}
\end{minipage}\hfill
\begin{minipage}{0.5\textwidth}
\includegraphics[width=.7\textwidth]{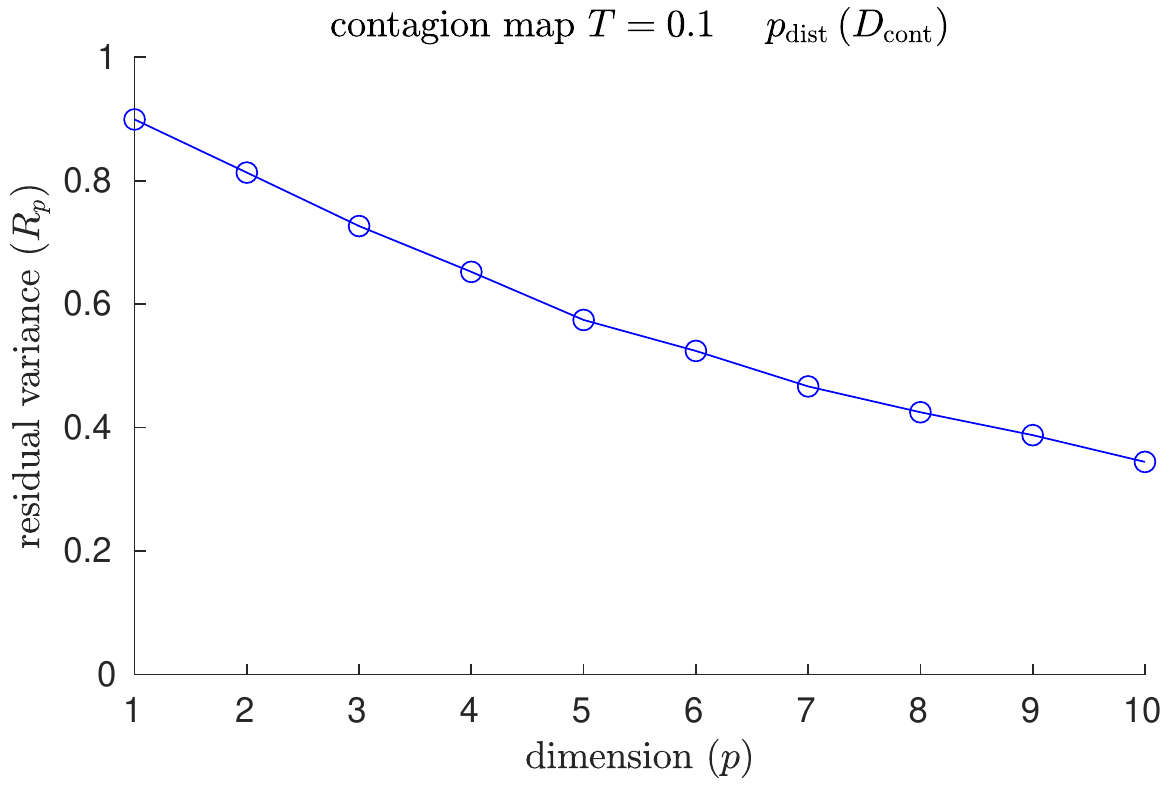}
\end{minipage}\hfill

\leftline{\hskip 0.00cm (d) \hskip 7cm (i) } 
\begin{minipage}{0.5\textwidth}
\includegraphics[width=.7\textwidth]{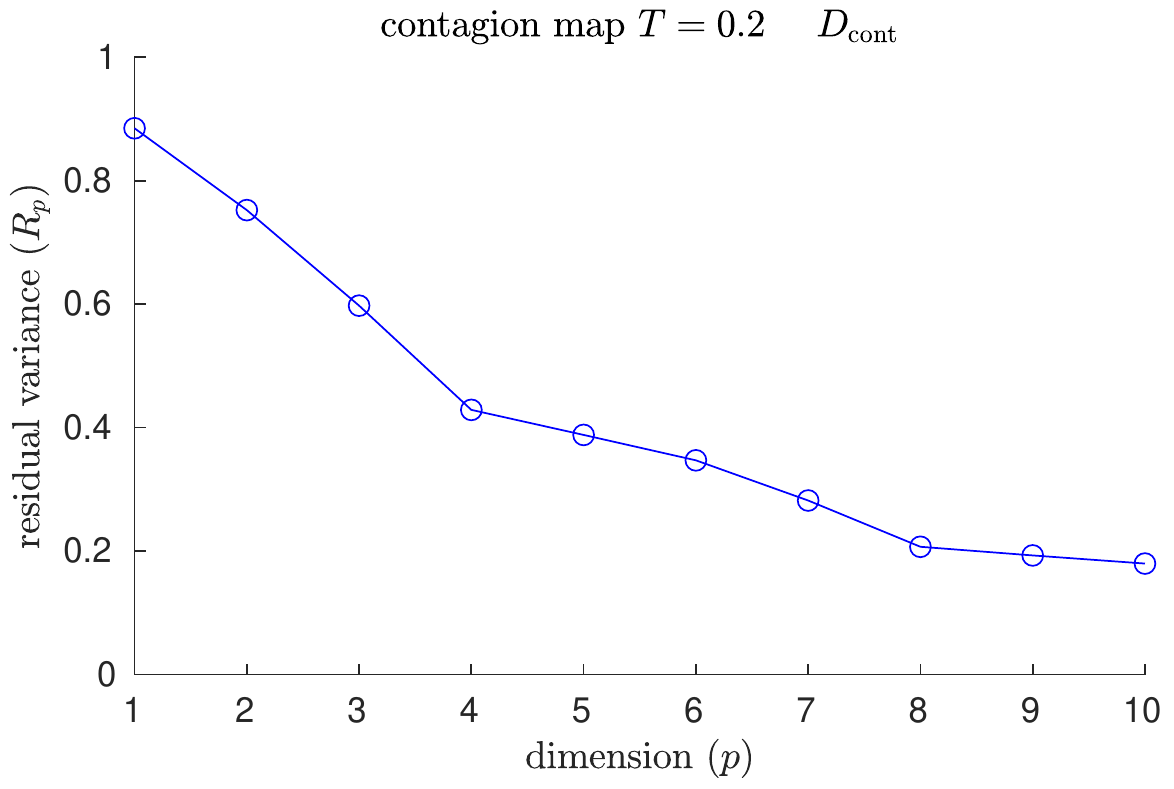}
\end{minipage}\hfill
\begin{minipage}{0.5\textwidth}
\includegraphics[width=.7\textwidth]{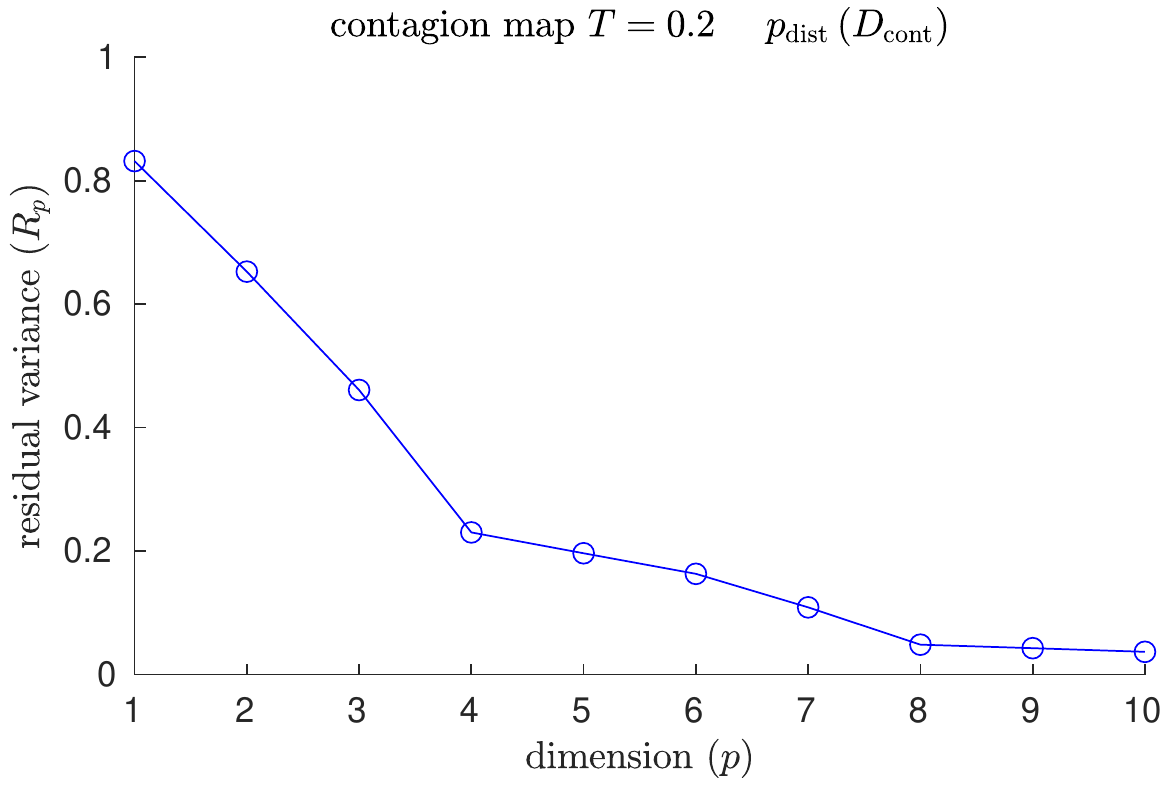}
\end{minipage}\hfill

\leftline{\hskip 0.00cm (e) \hskip 7cm (j) } 
\begin{minipage}{0.5\textwidth}
\includegraphics[width=.7\textwidth]{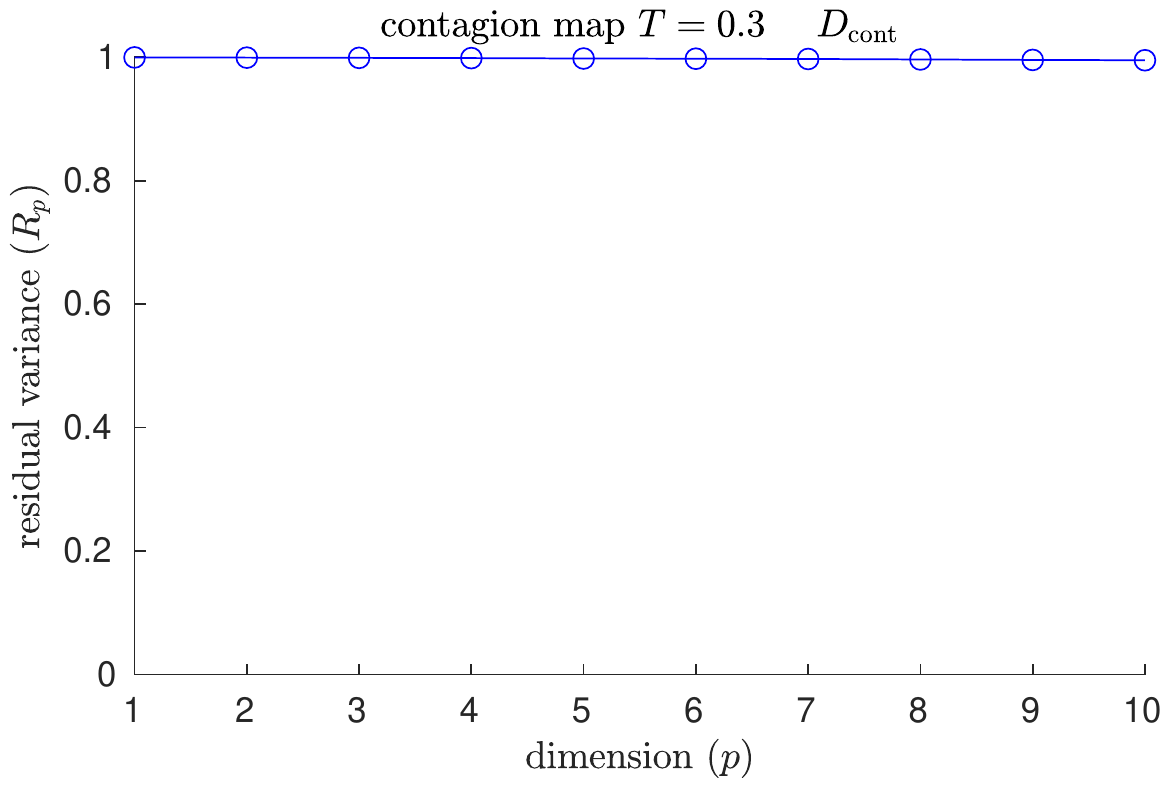}
\end{minipage}\hfill
\begin{minipage}{0.5\textwidth}
\includegraphics[width=.7\textwidth]{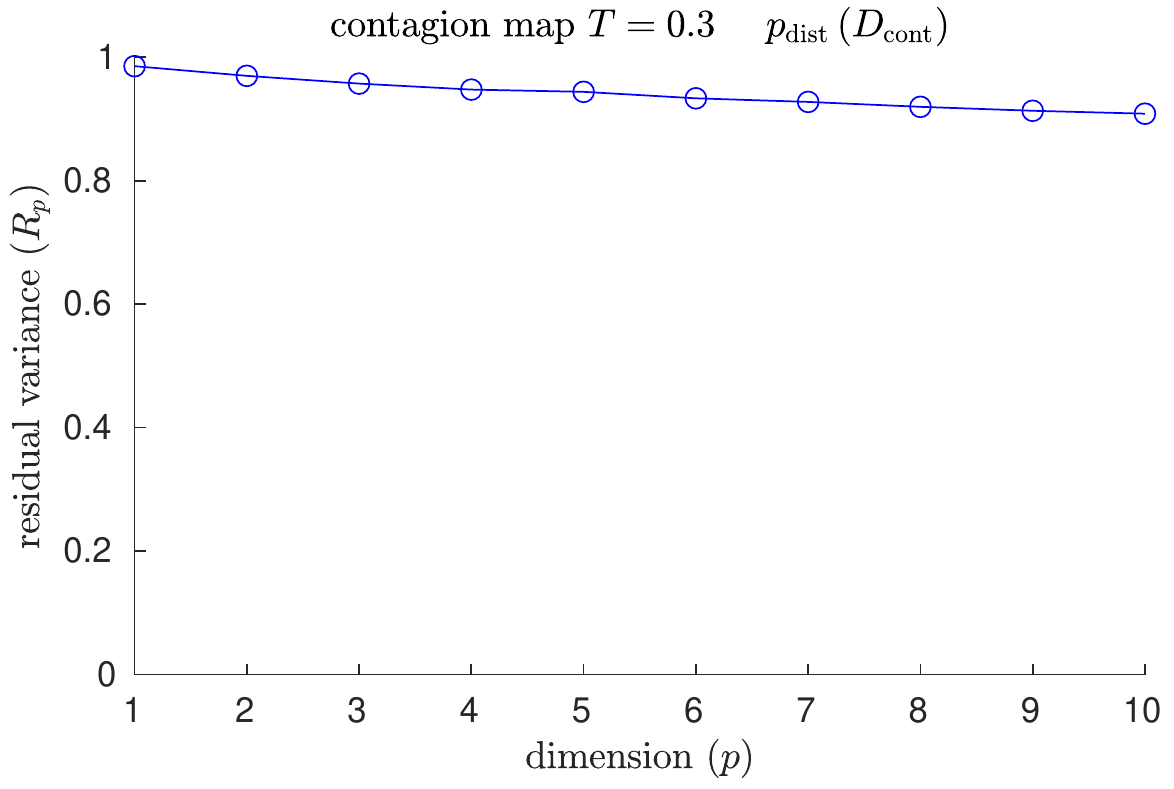}
\end{minipage}\hfill
\caption{Dimensionality results on our torus-based network with $d^{\rm{NG}}=4$. (Left column) Residual variances of MDS based on the estimated geodesic distances (i.e.~the entries in $D_{\rm iso}$ and $D_{\rm cont}$) according to (a) Isomap, and according to contagion maps with thresholds (b) $T=0$, (c) $T=0.1$, (d) $T=0.2$, and (e) $T=0.3$. (Right column) Residual variances of MDS based on the point cloud (i.e.~the rows of the $D_{\rm iso}$ and $D_{\rm cont}$) according to (f) Isomap, and according to contagion maps with thresholds (g) $T=0$, (h) $T=0.1$, (i) $T=0.2$, and (j) $T=0.3$. }
\label{torus_MDS_d_ng_4}
\end{figure}

MDS based on Isomap does not identify the embedding dimension $4$ of a torus for either $d^{\rm{NG}}=2$ or $d^{\rm{NG}}=4$ (see Figure~\ref{torus_MDS_d_ng_2}(a,f) and Figure~\ref{torus_MDS_d_ng_4}(a,f)). 
The residual variance based on contagion maps does have a dip at dimension $4$ for the threshold $T=0.2$ (see Figure~\ref{torus_MDS_d_ng_2}(d,i) and Figure~\ref{torus_MDS_d_ng_4}(d,i)), but does not when the threshold is $T=0.1$ or $T=0.3$.
Note that for Isomap, as well as contagion maps with all considered thresholds, the residual variances are smaller for all considered dimensions when the analysis is done on the point cloud, making the results for contagion map with $T=0.2$ look sharper. 

\begin{figure}[H]
\centering
\leftline{\hskip 0.00cm (a) \hskip 7cm (b)} 
\begin{minipage}{.5\textwidth}
\includegraphics[width=1\textwidth]{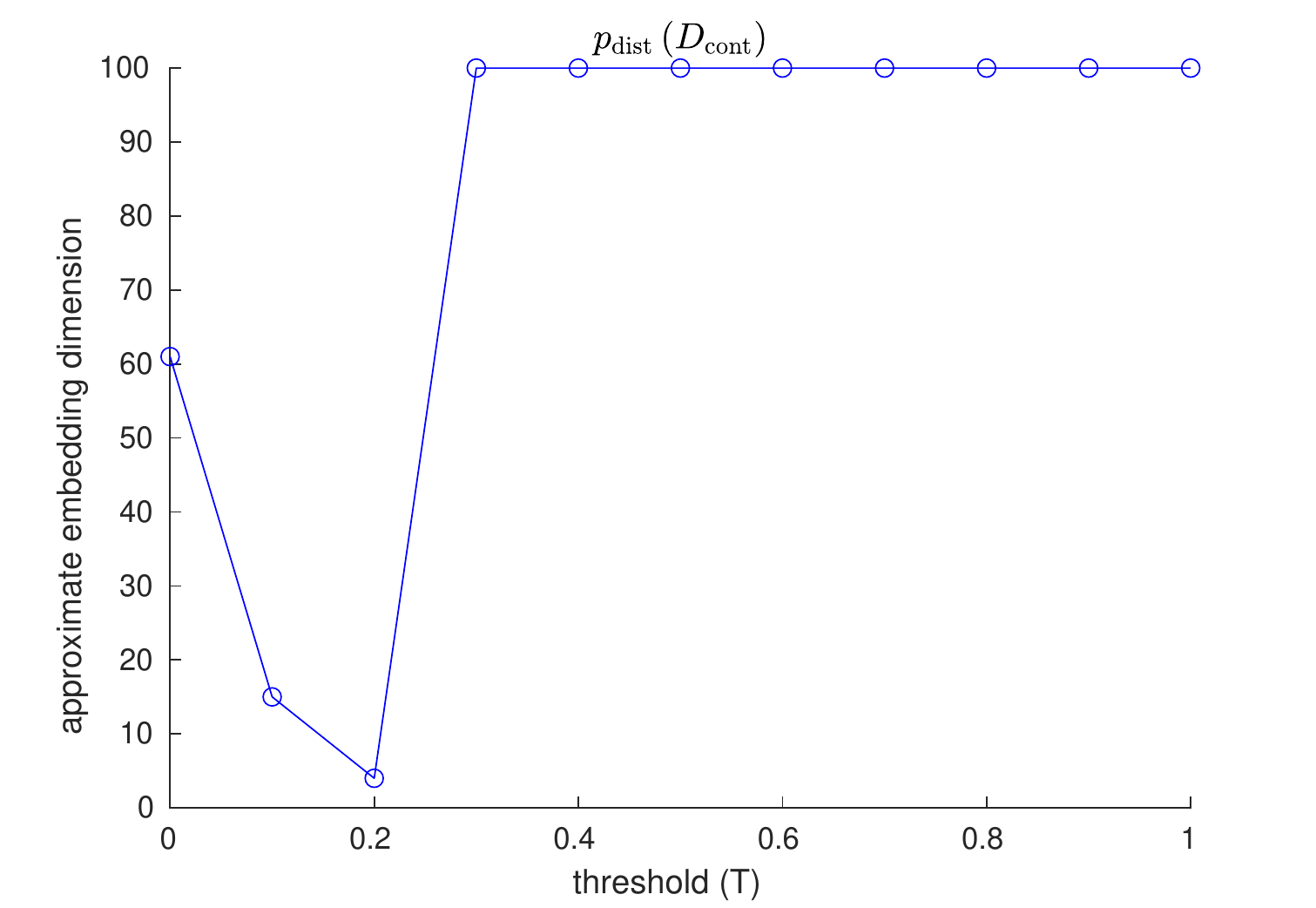}
\end{minipage}\hfill
\begin{minipage}{.5\textwidth}
\includegraphics[width=1\textwidth]{dimensionality_torus_d_g8_d_ng2_gamma0_contagion_map-eps-converted-to.pdf}
\end{minipage}\hfill

\caption{Dimensionality results on our torus-based network with $d^{\rm{NG}}=2$. Approximate embedding dimension according to the contagion map with different values of threshold $T$ and critical value 5\% using (a) the dissimilarity matrix $D_{\rm cont}$, and (b) the point cloud whose coordinate vectors are the row in $D_{\rm cont}$. The results for Isomap are practically identical to those for contagion map with $T=0$: The approximate embedding dimension is at least $100$ (which is the dimension at which we cap our computations) when based on the entries in $D_{\rm iso}$; and it is $61$ when based on the point cloud whose coordinate vectors are given by the rows of $D_{\rm iso}$. The approximate embedding dimension according to contagion maps reaches a minimum value for $T=0.2$ both when working with $D_{\rm cont}$ and when working with $p_{\rm dist}D_{\rm cont}$. This minimal value is $7$ in the former case (see (a)), and is $4$ in the latter case (see (b)). }
\label{torus_dimensionality_dng2}
\end{figure}

\begin{figure}[H]
\centering
\leftline{\hskip 0.00cm (a) \hskip 7cm (b)} 
\begin{minipage}{.5\textwidth}
\includegraphics[width=1\textwidth]{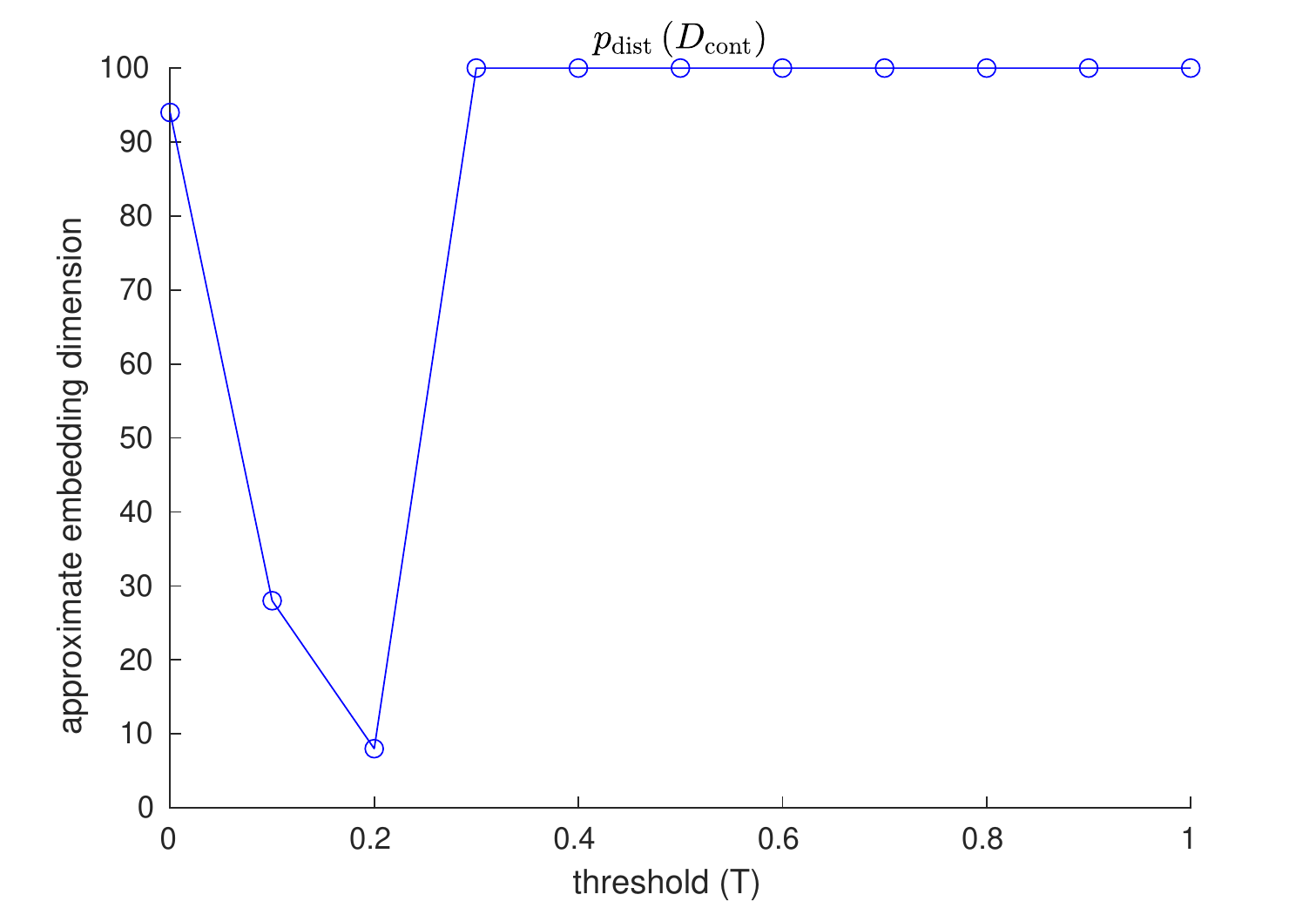}
\end{minipage}\hfill
\begin{minipage}{.5\textwidth}
\includegraphics[width=1\textwidth]{dimensionality_torus_d_g8_d_ng4_gamma0_contagion_map-eps-converted-to.pdf}
\end{minipage}\hfill

\caption{Dimensionality results on our torus-based network with $d^{\rm{NG}}=4$. Approximate embedding dimension according to the contagion map with different values of threshold $T$ and critical value 5\% using (a) the dissimilarity matrix $D_{\rm cont}$, and (b) the point cloud whose coordinate vectors are the row in $D_{\rm cont}$. The results for Isomap are practically identical to those for contagion map with $T=0$: The approximate embedding dimension is at least $100$ (which is the dimension at which we cap out computations) when based on the entries in $D_{\rm iso}$; and it is $95$ when based on the point cloud whose coordinate vectors are given by the rows of $D_{\rm iso}$. The approximate embedding dimension according to contagion maps reaches a minimum value of $8$ for $T=0.2$ when working with $p_{\rm dist}D_{\rm cont}$ (see (b)). }
\label{torus_dimensionality_dng4}
\end{figure}

We identify the approximate embedding dimensions for Isomap and contagion maps with thresholds $T=0,0.1, \dots,1$ and show the results in Figure~\ref{torus_dimensionality_dng2} (for the torus-based network with $d^{\rm{NG}}=2$) and Figure~\ref{torus_dimensionality_dng4} (for the torus-based network with $d^{\rm{NG}}=4$). Both versions of Isomap (the one performing MDS based on the entries in $D_{\rm iso}$ and the one performing MDS based on the distances between the rows of $D_{\rm iso}$) vastly overestimate the embedding dimension for the torus-based network with $d^{\rm{NG}}=2$ and the one with $d^{\rm{NG}}=4$. When performing MDS based on the entries in $D_{\rm iso}$, the approximate embedding dimension is at least $100$ (which is the dimension at which we cap our computations). When performing MDS based on the distances between the rows of $D_{\rm iso}$, the embedding dimensions are still very large: It is $61$ for the network with $d^{\rm{NG}}=2$, and $95$ for the network with $d^{\rm{NG}}=4$. Note that these results are practically identical to those for contagion map with $T=0$, as we are working with the same unweighted graphs in both Isomap and contagion maps. For contagion maps with varying thresholds $T$, the approximate embedding dimension has a dip around $T=0.2$ (see Figure~\ref{torus_dimensionality_dng2} (b) and Figure~\ref{torus_dimensionality_dng4} (b)), except when performing MDS based on the entries in $D_{\rm cont}$ for the network with $d^{\rm{NG}}=4$, in which case contagion maps return an embedding dimension of at least $100$ for all thresholds (see Figure~\ref{torus_dimensionality_dng4} (a)). This suggests that, for thresholds close to $0.2$, the contagion spreads predominantly via WFP along the underlying torus, making it possible to identify the underlying low-dimensional structure. 

\pagebreak

\pagebreak

\subsubsection{Topology}

\begin{figure}[H]
\centering
\leftline{\hskip 0.00cm (a) \hskip 7cm (f) } 
\begin{minipage}{0.5\textwidth}
\includegraphics[width=.7\textwidth]{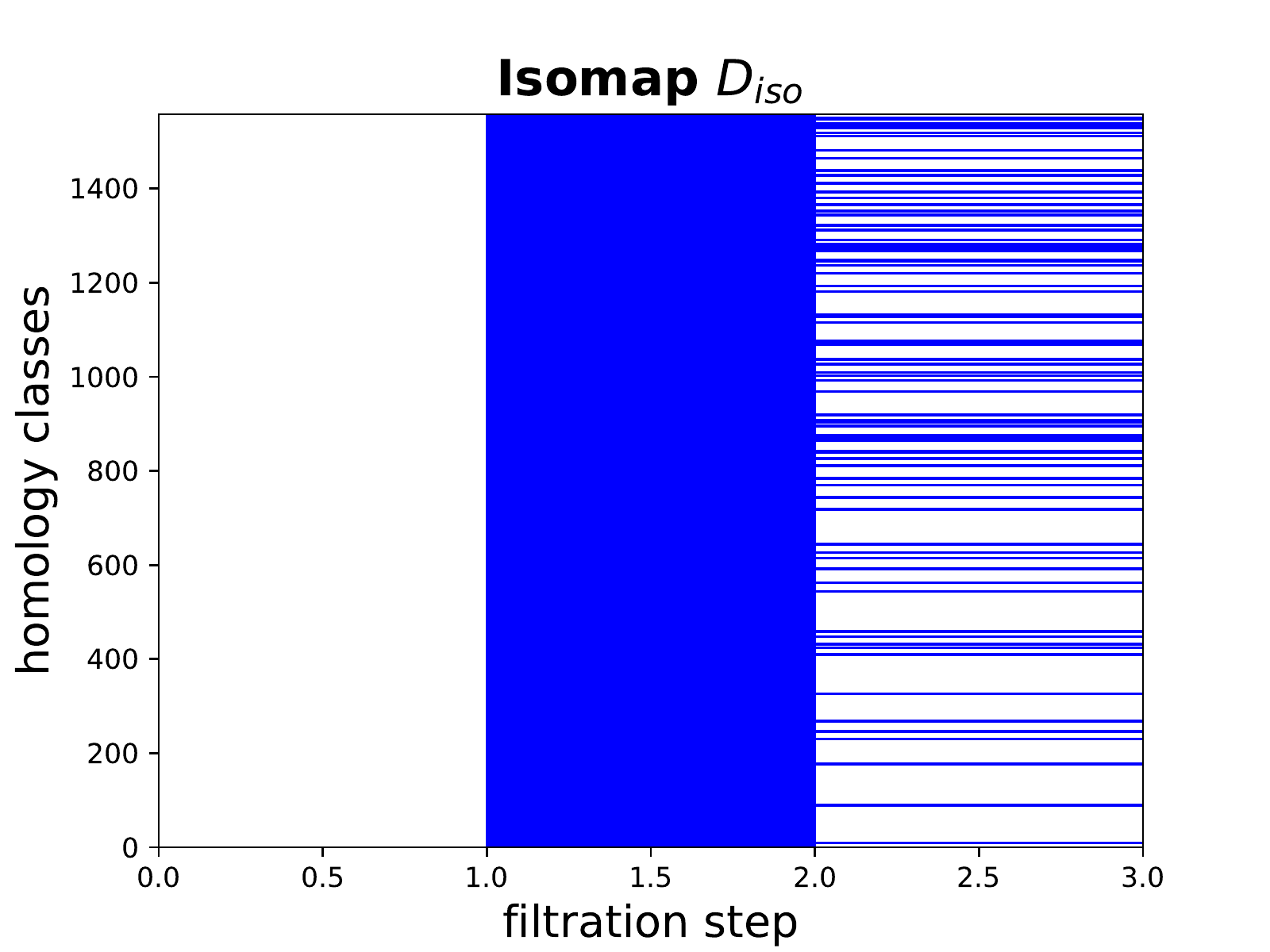}
\end{minipage}\hfill
\begin{minipage}{0.5\textwidth}
\includegraphics[width=.7\textwidth]{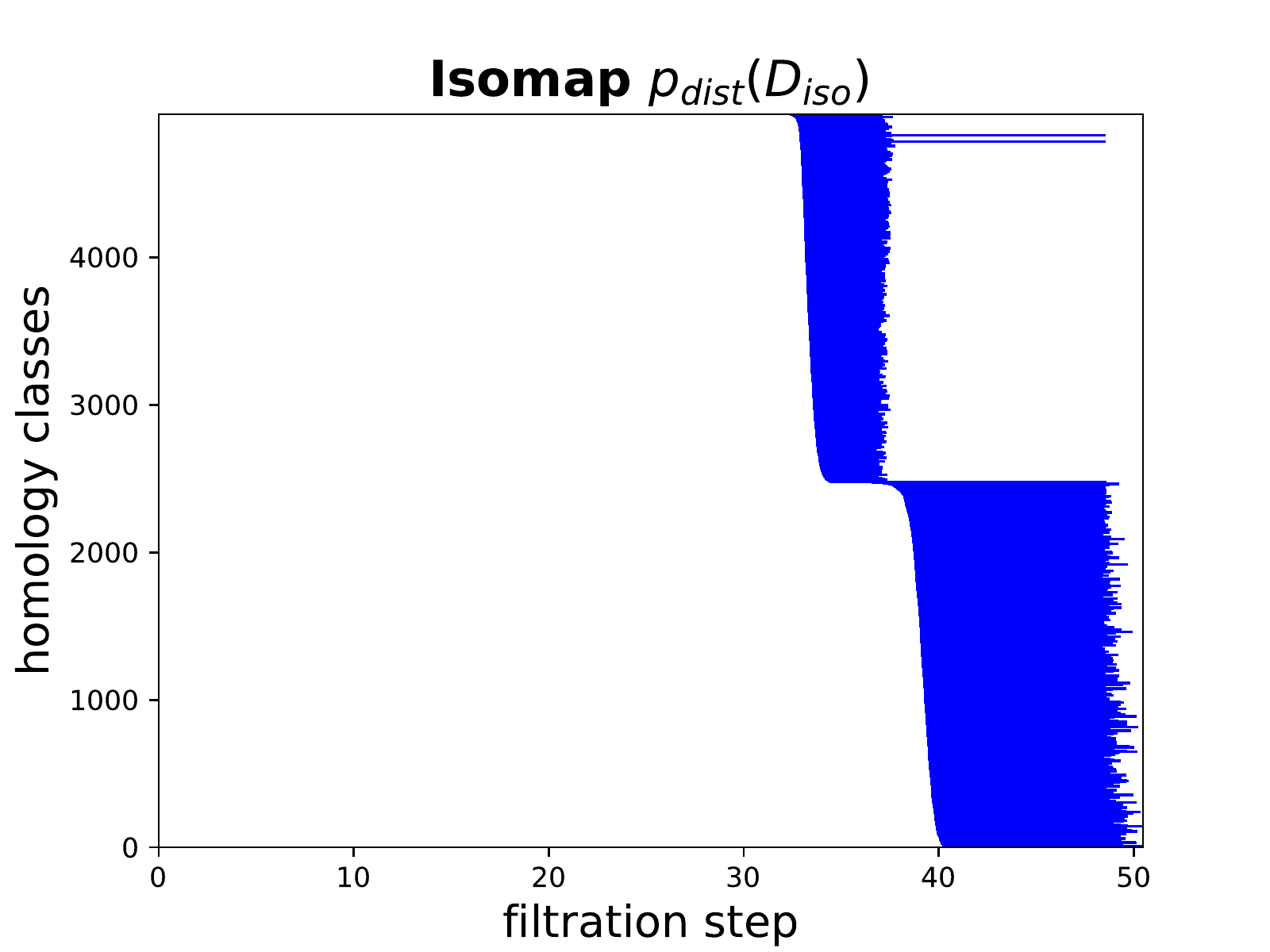}
\end{minipage}\hfill

\leftline{\hskip 0.00cm (b) \hskip 7cm (g) } 
\begin{minipage}{0.5\textwidth}
\includegraphics[width=.7\textwidth]{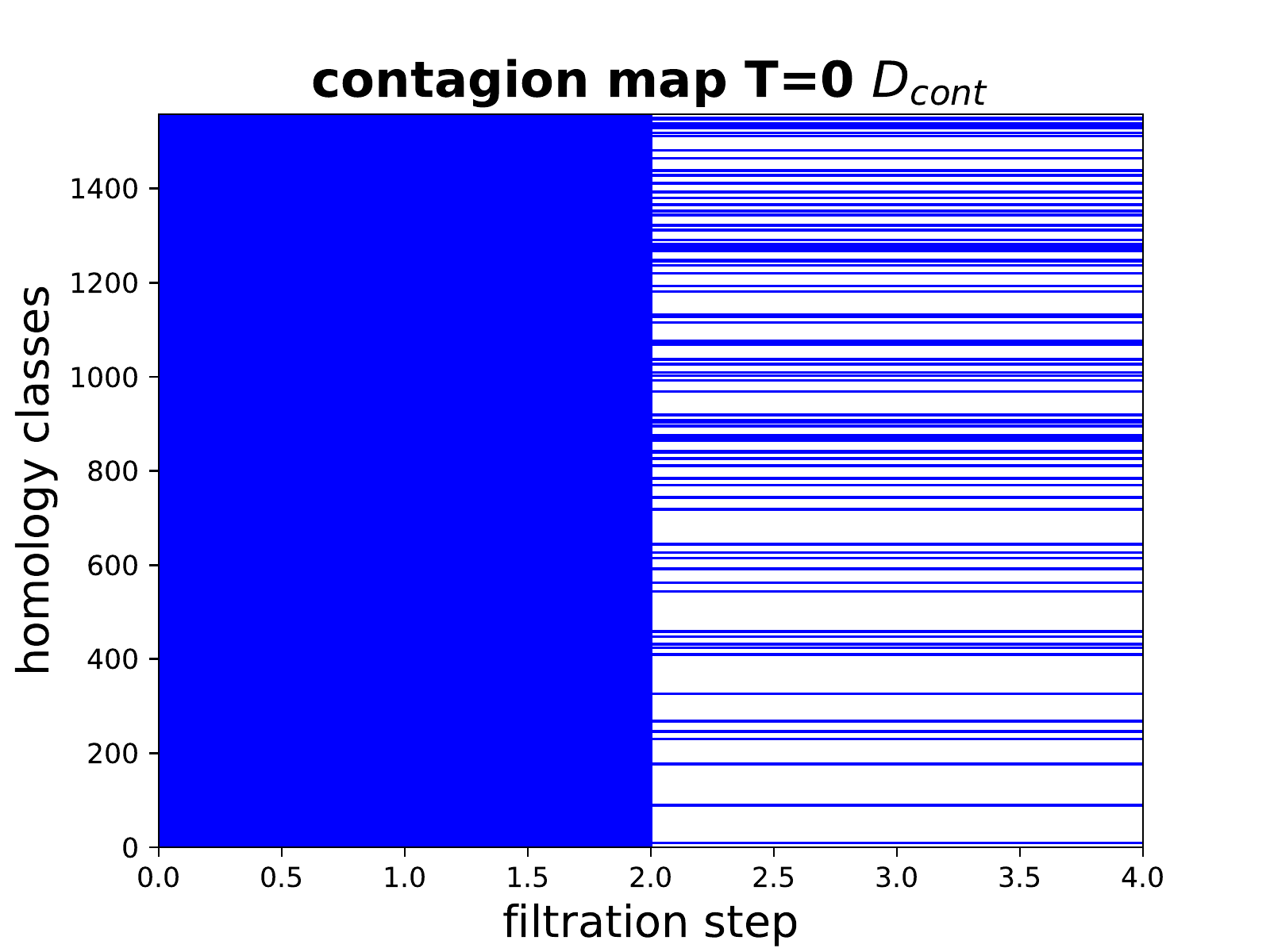}
\end{minipage}\hfill
\begin{minipage}{0.5\textwidth}
\includegraphics[width=.7\textwidth]{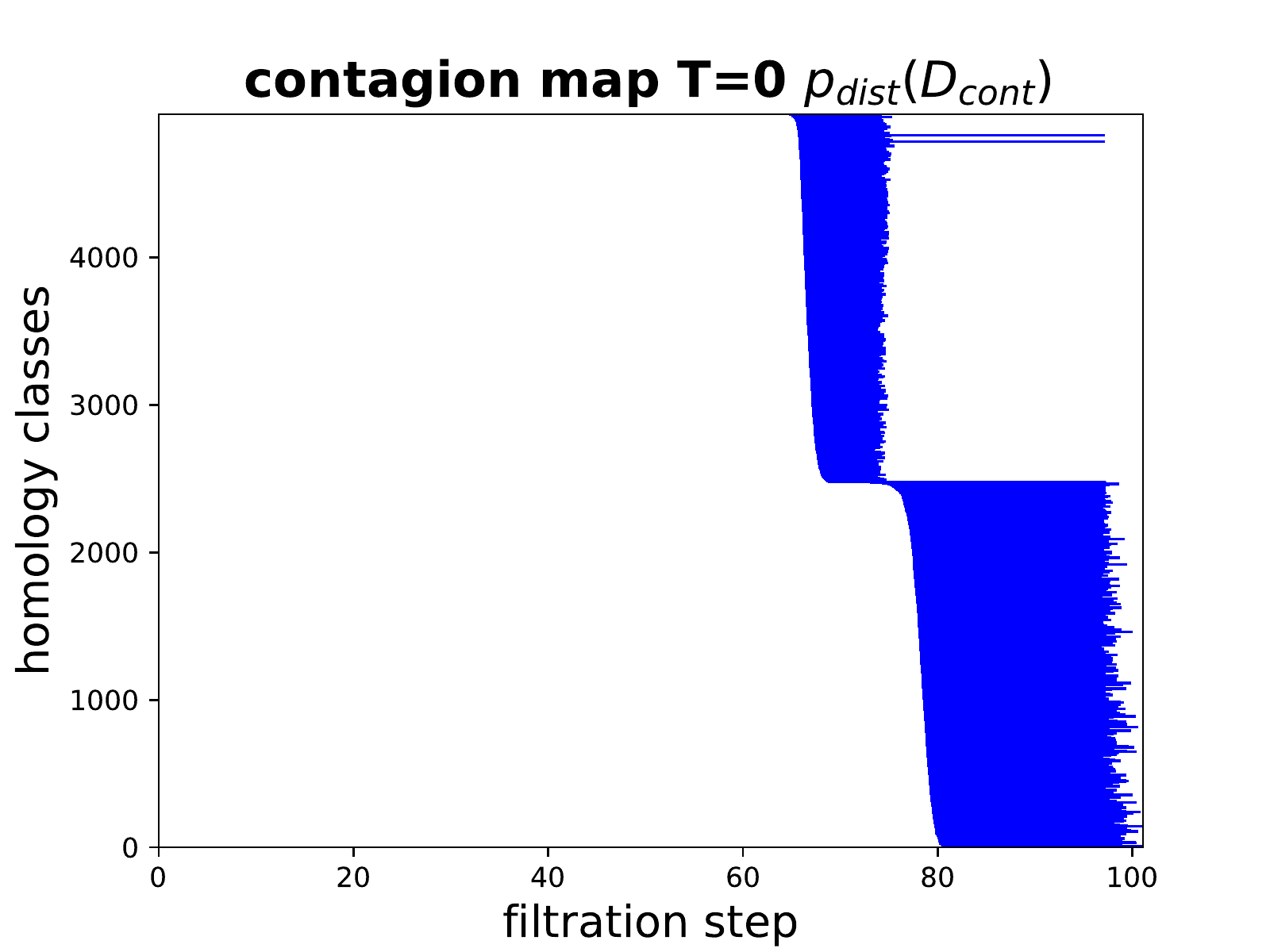}
\end{minipage}\hfill

\leftline{\hskip 0.00cm (c) \hskip 7cm (h) } 
\begin{minipage}{0.5\textwidth}
\includegraphics[width=.7\textwidth]{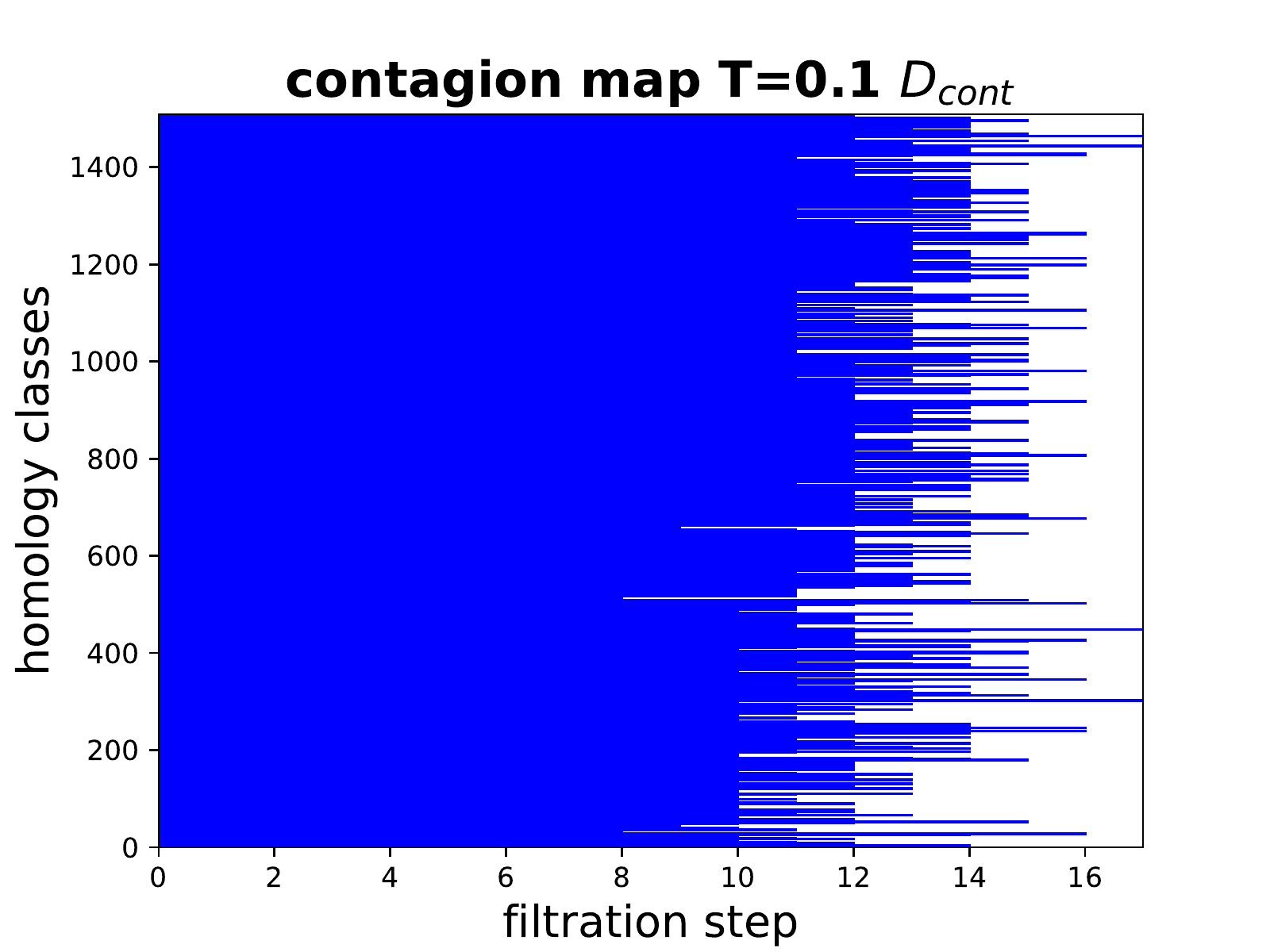}
\end{minipage}\hfill
\begin{minipage}{0.5\textwidth}
\includegraphics[width=.7\textwidth]{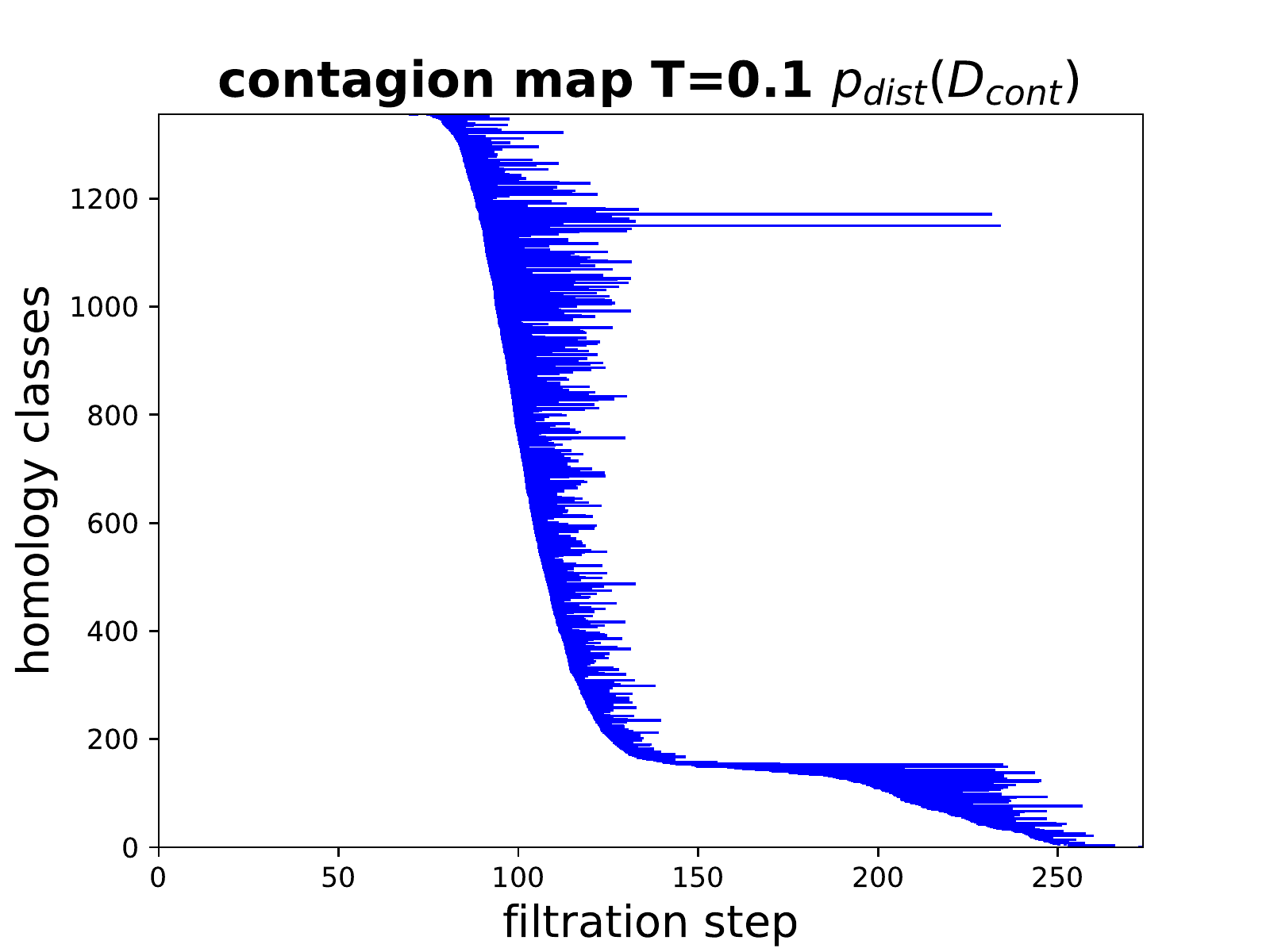}
\end{minipage}\hfill

\leftline{\hskip 0.00cm (d) \hskip 7cm (i) } 
\begin{minipage}{0.5\textwidth}
\includegraphics[width=.7\textwidth]{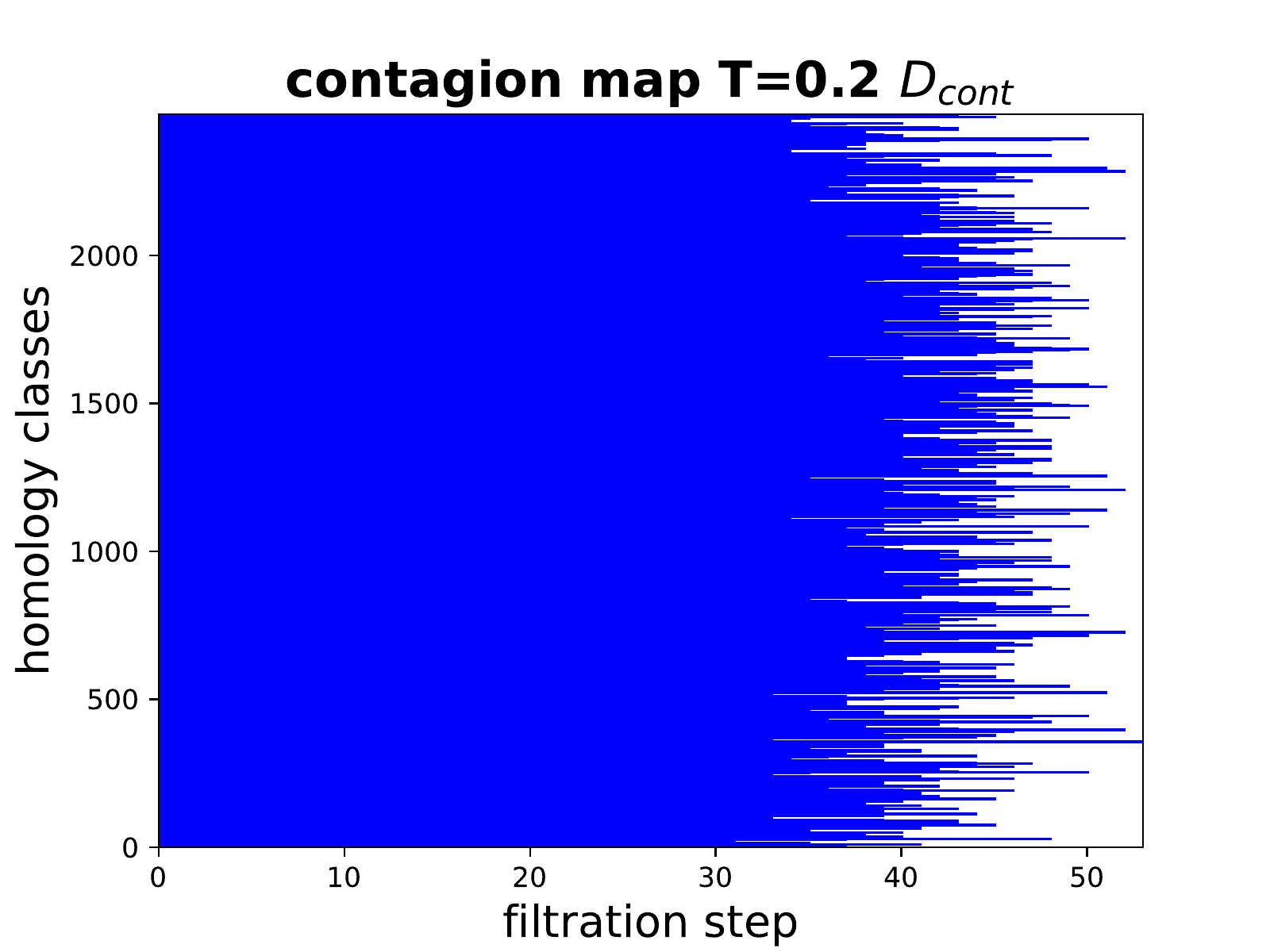}
\end{minipage}\hfill
\begin{minipage}{0.5\textwidth}
\includegraphics[width=.7\textwidth]{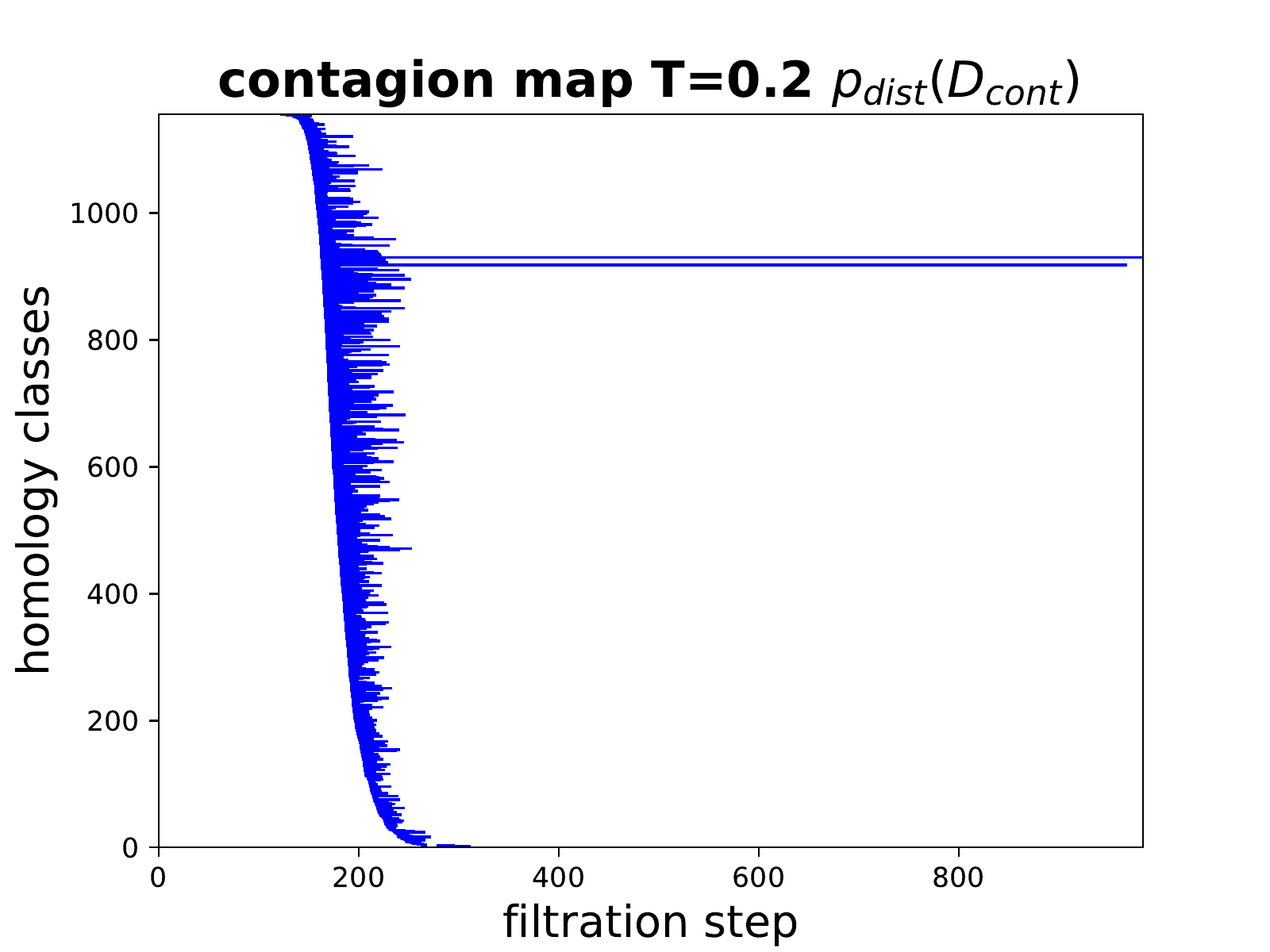}
\end{minipage}\hfill

\leftline{\hskip 0.00cm (e) \hskip 7cm (j) } 
\begin{minipage}{0.5\textwidth}
\includegraphics[width=.7\textwidth]{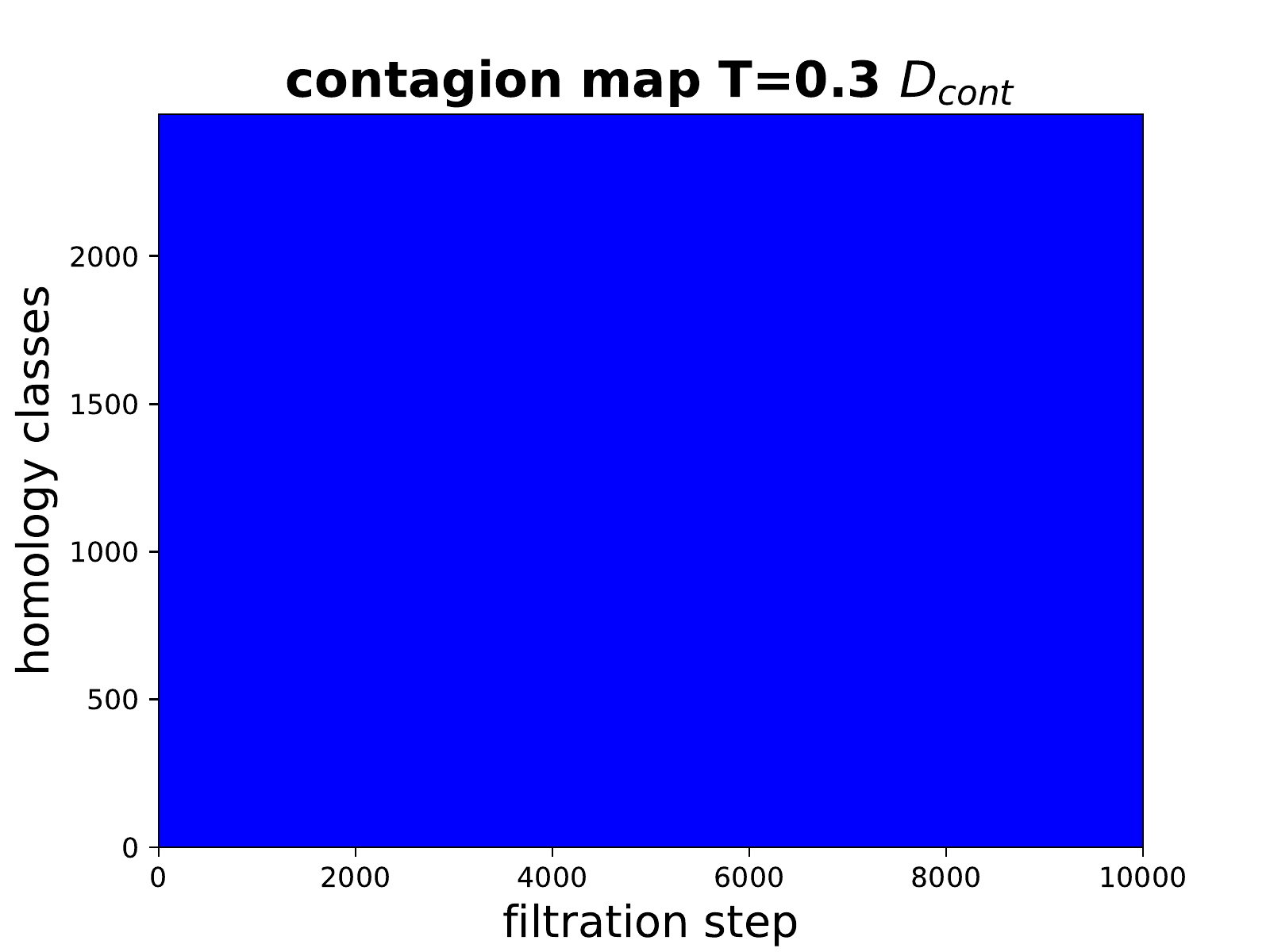}
\end{minipage}\hfill
\begin{minipage}{0.5\textwidth}
\includegraphics[width=.7\textwidth]{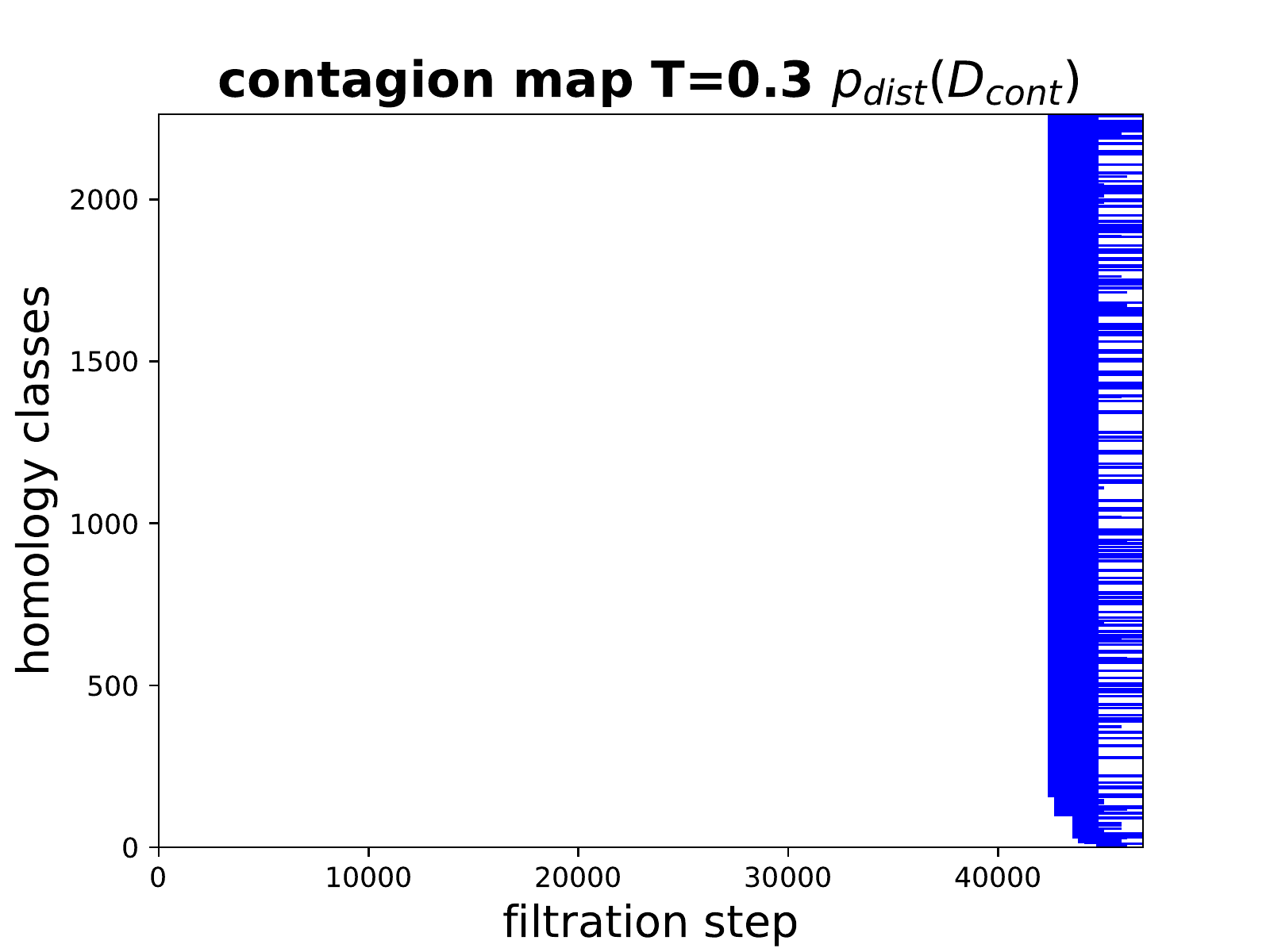}
\end{minipage}\hfill
\caption{Topology results on our torus-based network with $d^{\rm{NG}}=2$. (Left column) Dimension $1$ barcodes of the Vietoris\textendash Rips filtrations based on the estimated geodesic distances (i.e.~the entries in $D_{\rm iso}$ and $D_{\rm cont}$) according to on (a) Isomap, and according to contagion maps with (b) $T=0$, (c) $T=0.1$, (d) $T=0.2$, and (e) $T=0.3$. (Right column) Barcodes of the Vietoris\textendash Rips filtrations on the point clouds according to (f) Isomap, and according to contagion maps with (g) $T=0$, (h) $T=0.1$, (i) $T=0.2$, and (j) $T=0.3$. 
}\label{torus_Ripser_d_ng_2}
\end{figure}

\begin{figure}[H]
\centering
\leftline{\hskip 0.00cm (a) \hskip 7cm (f) } 
\begin{minipage}{0.5\textwidth}
\includegraphics[width=.7\textwidth]{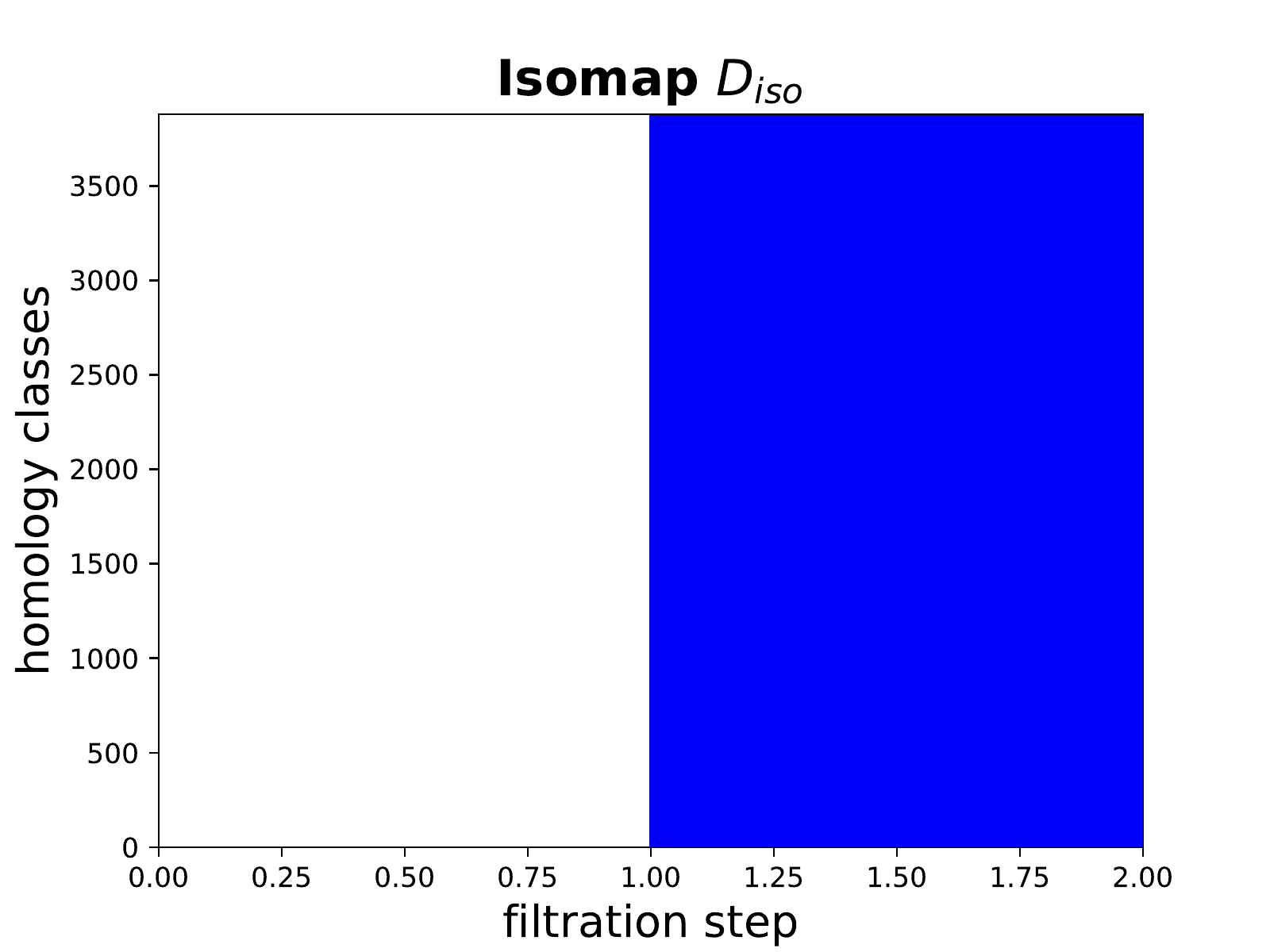}
\end{minipage}\hfill
\begin{minipage}{0.5\textwidth}
\includegraphics[width=.7\textwidth]{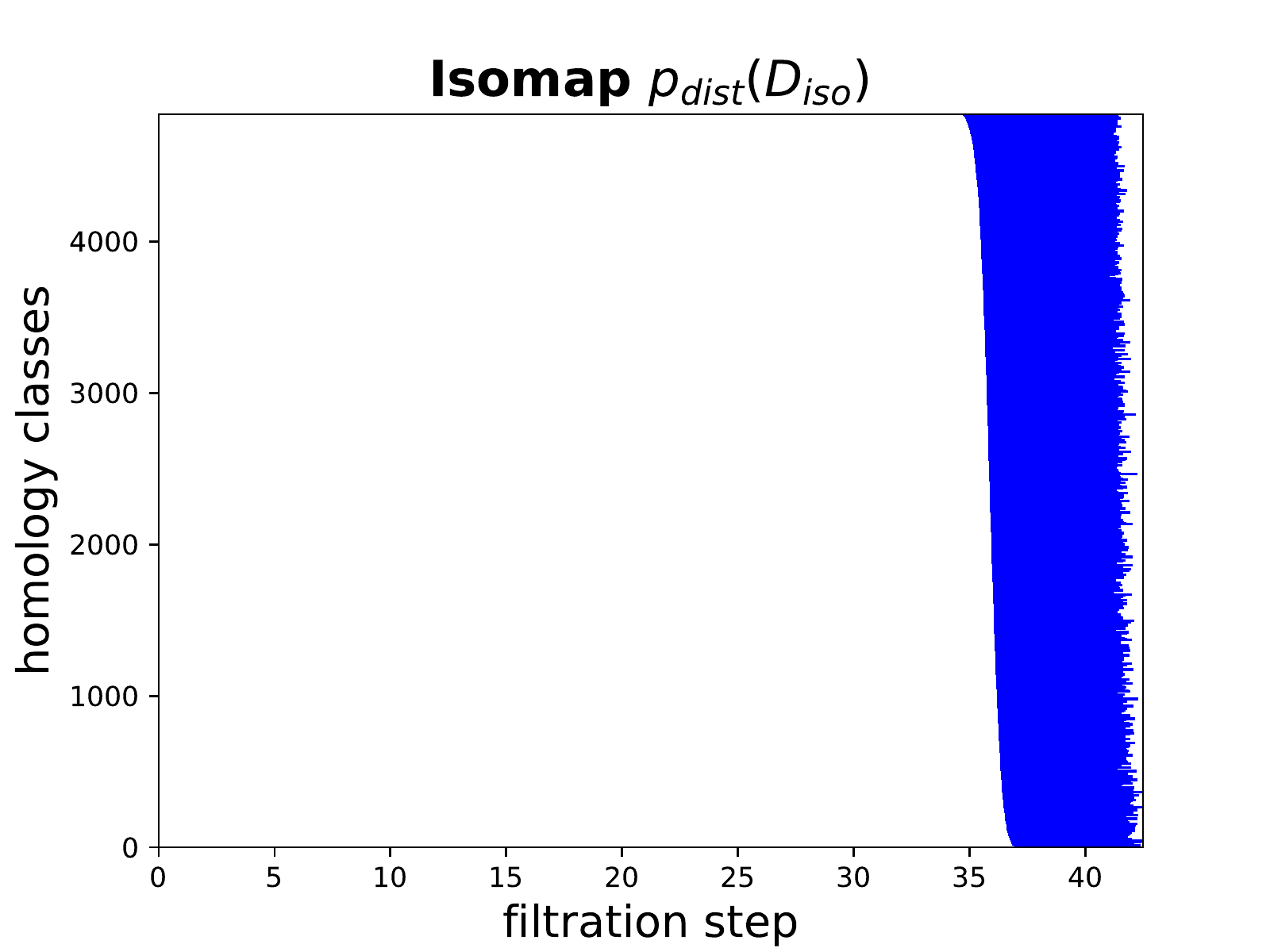}
\end{minipage}\hfill

\leftline{\hskip 0.00cm (b) \hskip 7cm (g) } 
\begin{minipage}{0.5\textwidth}
\includegraphics[width=.7\textwidth]{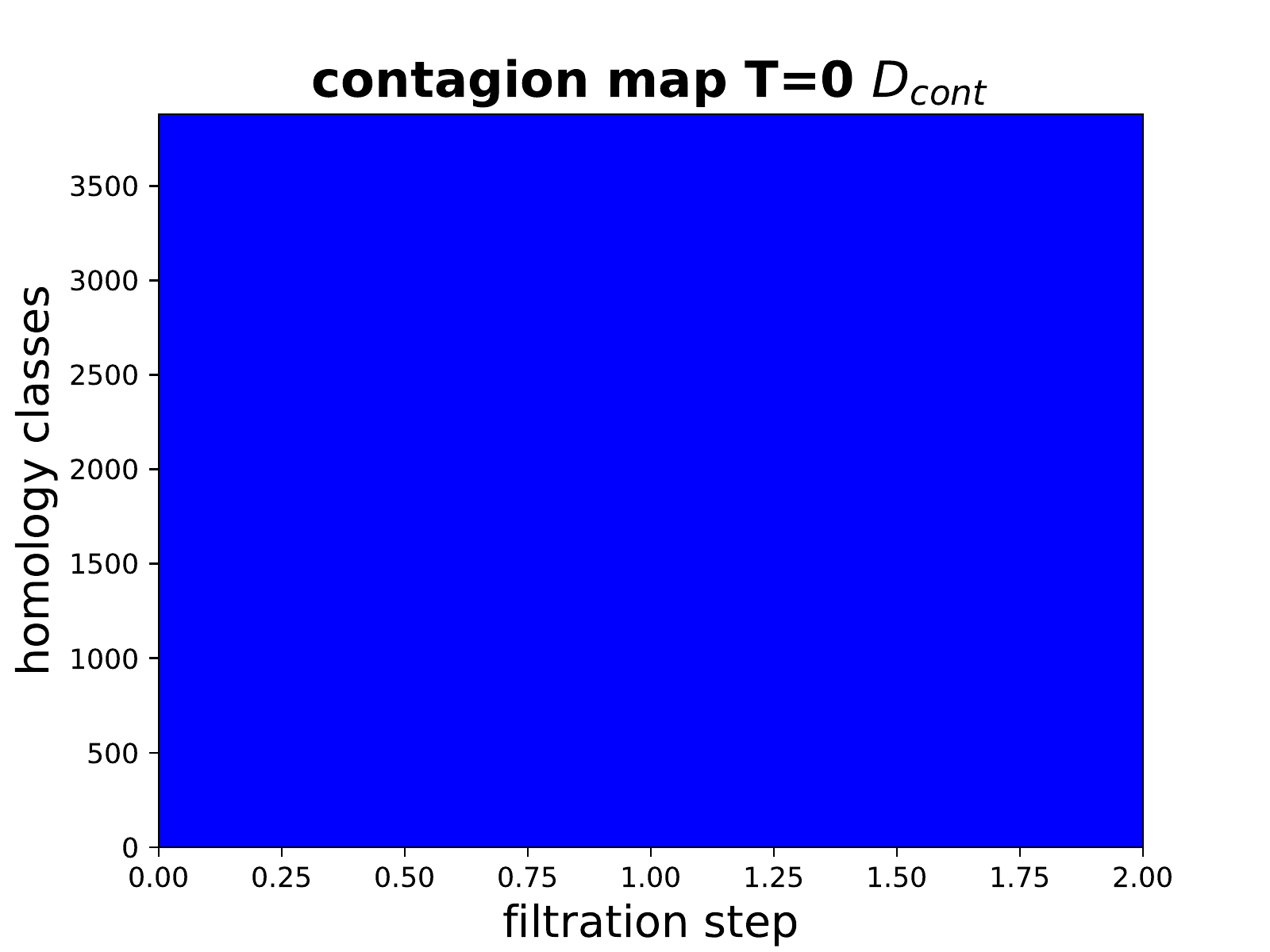}
\end{minipage}\hfill
\begin{minipage}{0.5\textwidth}
\includegraphics[width=.7\textwidth]{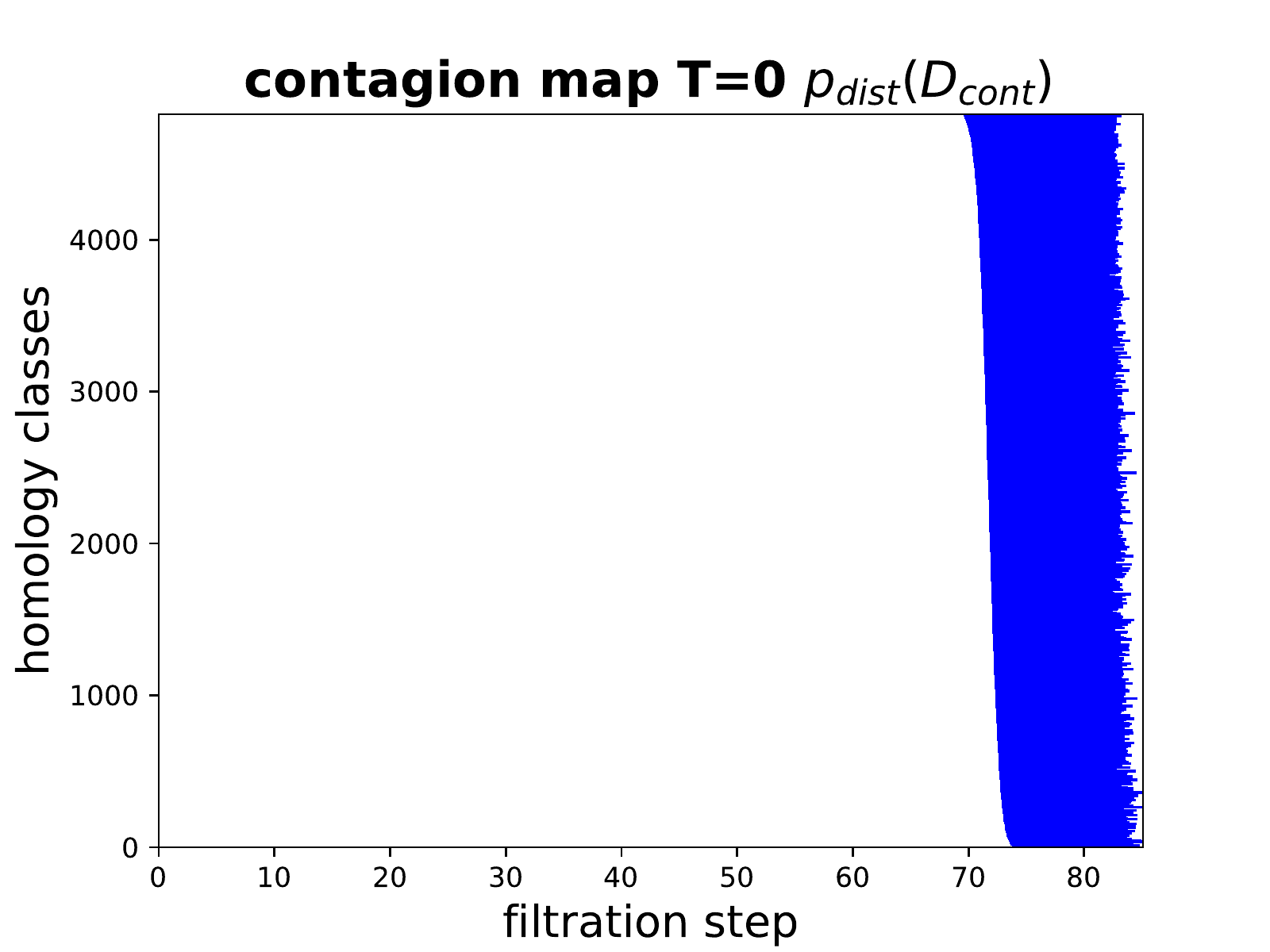}
\end{minipage}\hfill

\leftline{\hskip 0.00cm (c) \hskip 7cm (h) } 
\begin{minipage}{0.5\textwidth}
\includegraphics[width=.7\textwidth]{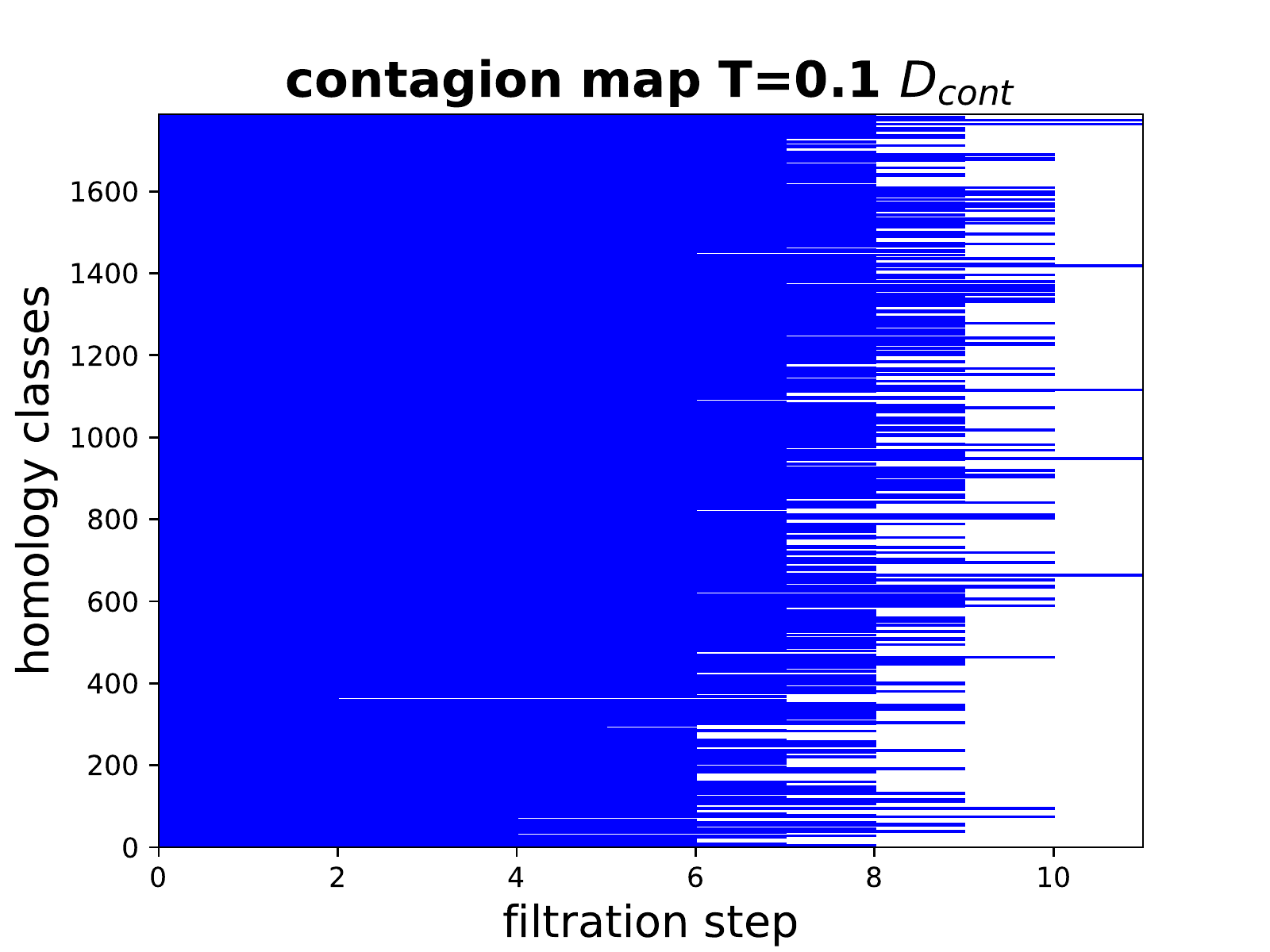}
\end{minipage}\hfill
\begin{minipage}{0.5\textwidth}
\includegraphics[width=.7\textwidth]{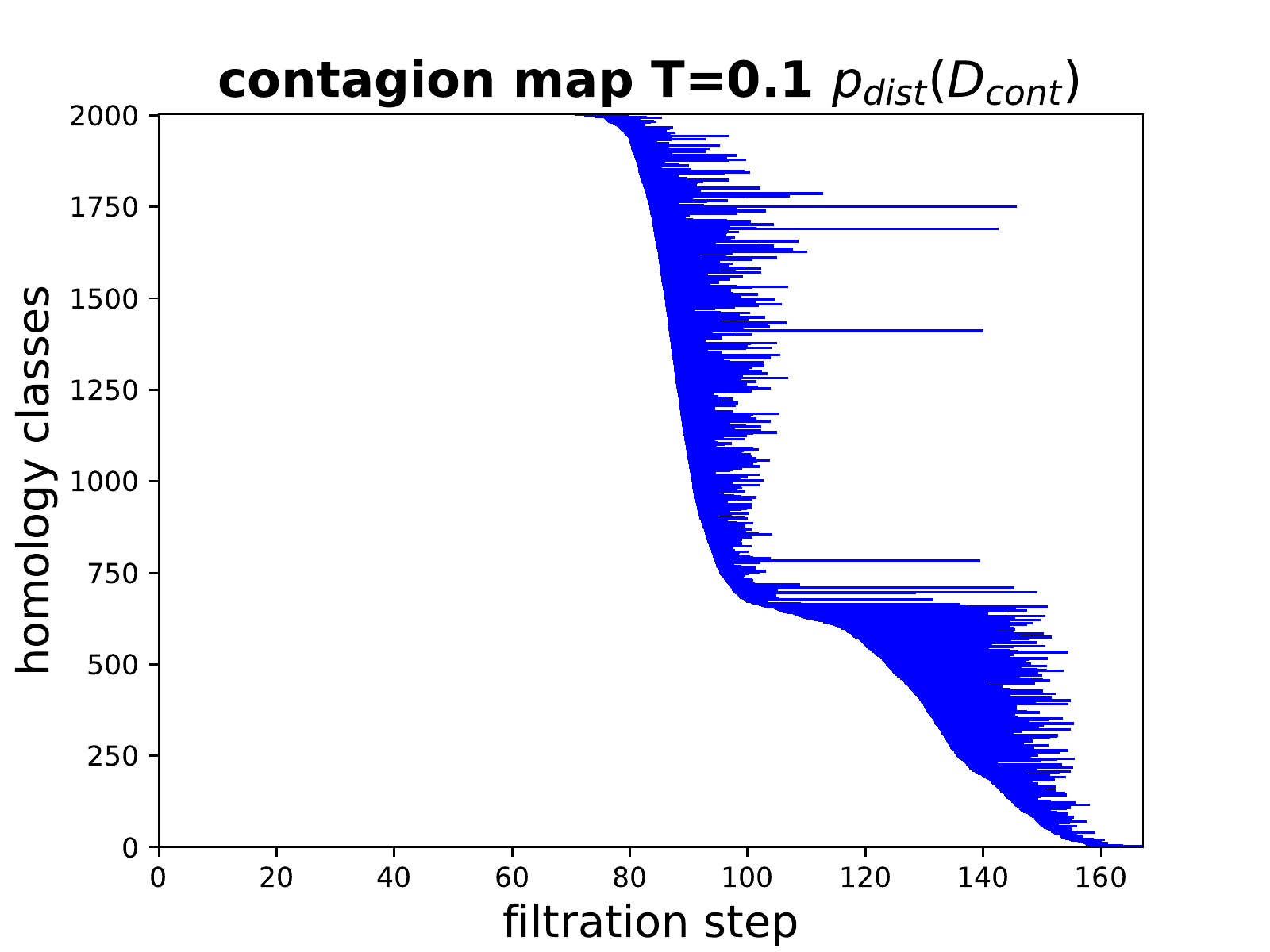}
\end{minipage}\hfill

\leftline{\hskip 0.00cm (d) \hskip 7cm (i) } 
\begin{minipage}{0.5\textwidth}
\includegraphics[width=.7\textwidth]{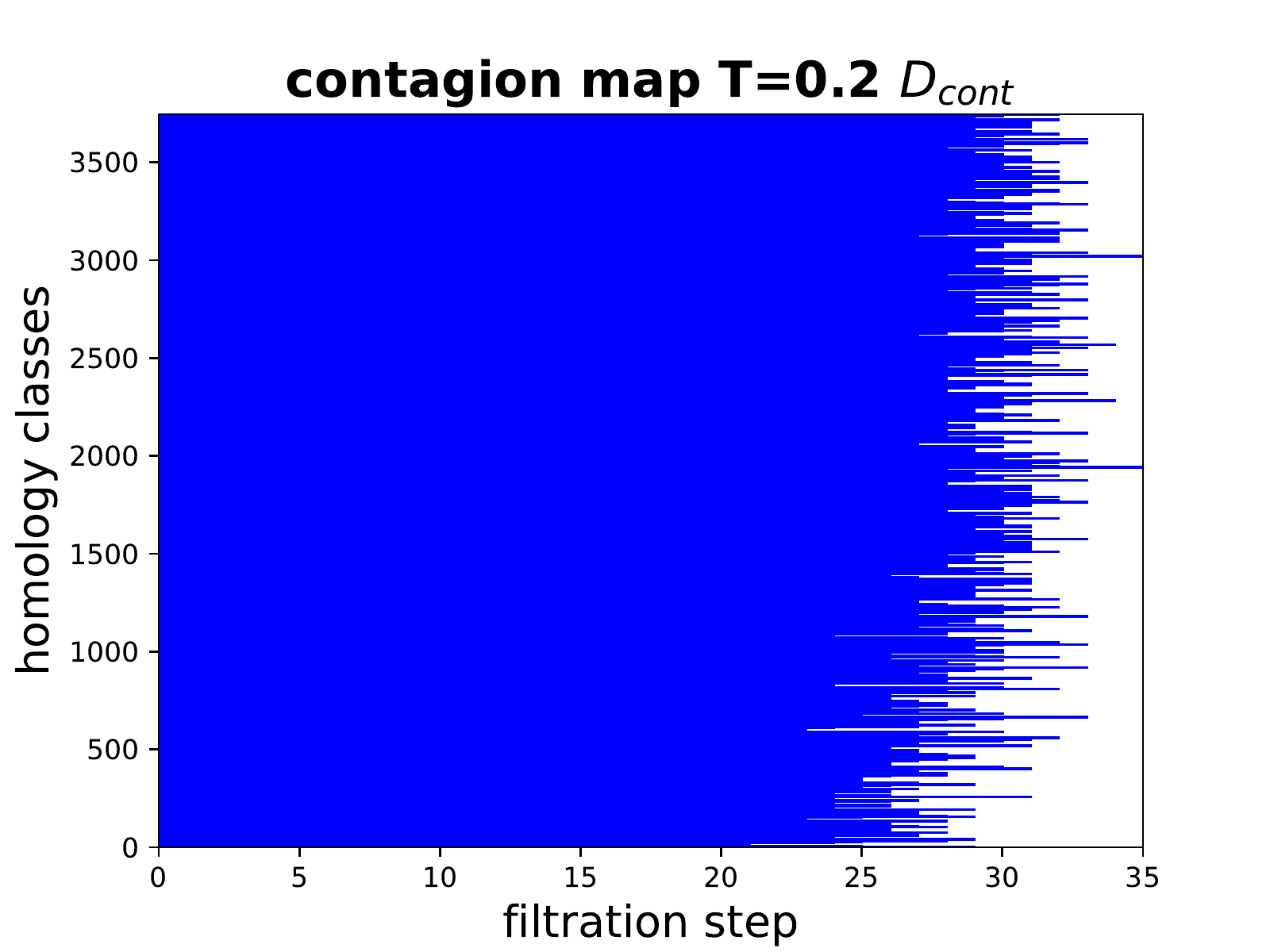}
\end{minipage}\hfill
\begin{minipage}{0.5\textwidth}
\includegraphics[width=.7\textwidth]{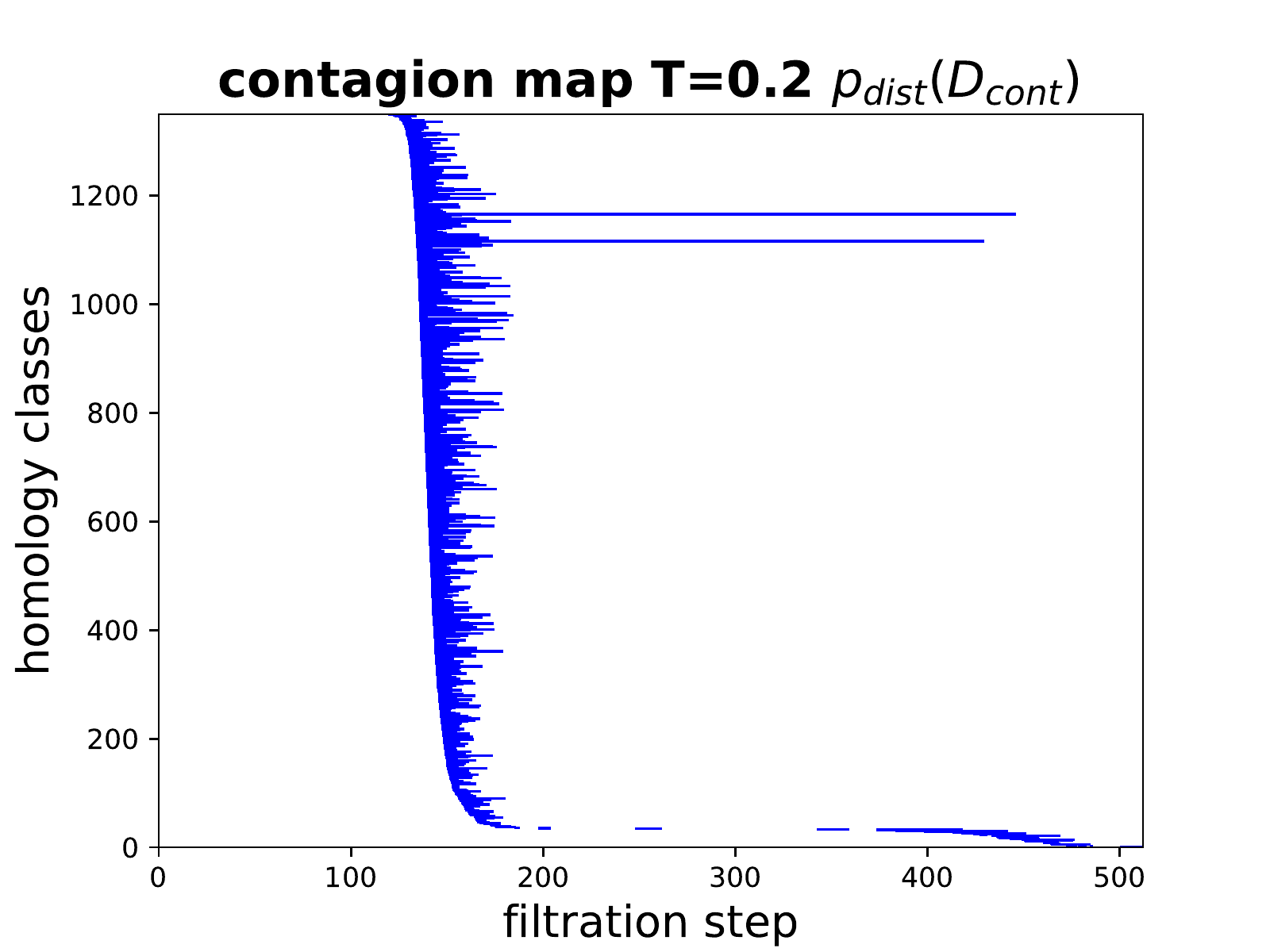}
\end{minipage}\hfill

\leftline{\hskip 0.00cm (e) \hskip 7cm (j) } 
\begin{minipage}{0.5\textwidth}
\includegraphics[width=.7\textwidth]{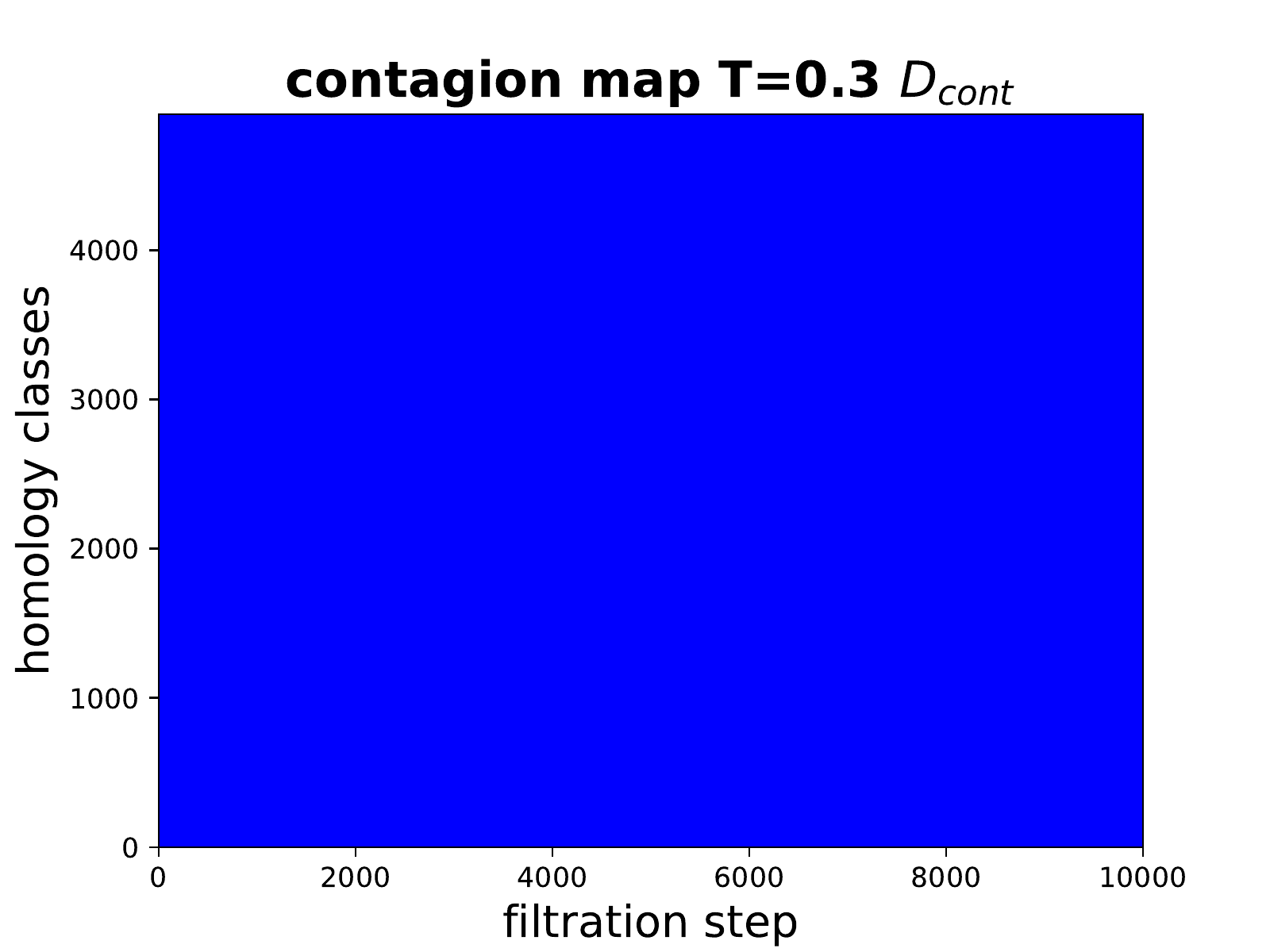}
\end{minipage}\hfill
\begin{minipage}{0.5\textwidth}
\includegraphics[width=.7\textwidth]{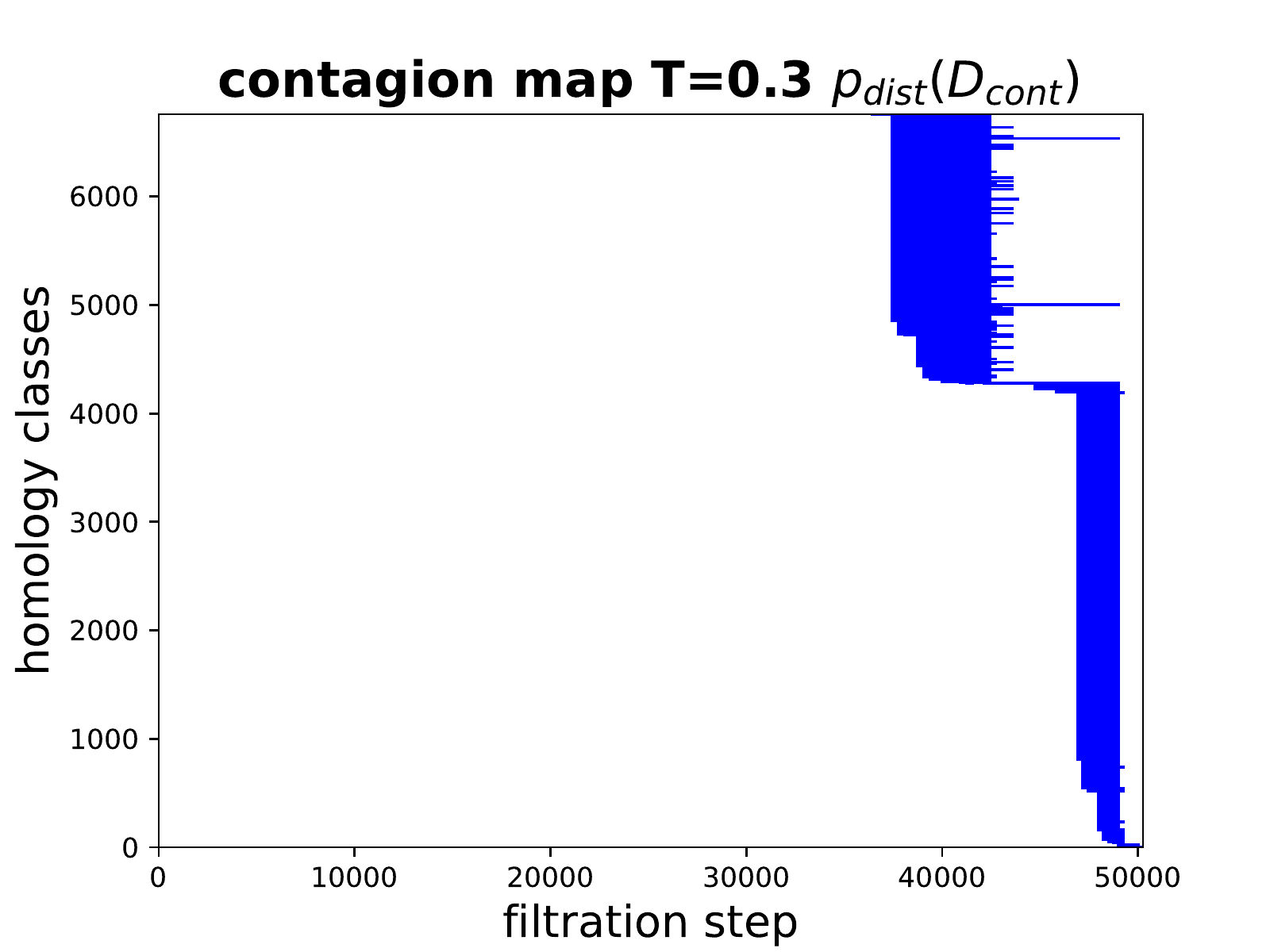}
\end{minipage}\hfill
\caption{Topology results on our torus-based network with $d^{\rm{NG}}=4$. (Left column) Dimension $1$ barcodes of the Vietoris\textendash Rips filtrations based on the estimated geodesic distances (i.e.~the entries in $D_{\rm iso}$ and $D_{\rm cont}$) according to on (a) Isomap, and according to contagion maps with (b) $T=0$, (c) $T=0.1$, (d) $T=0.2$, and (e) $T=0.3$. (Right column) Barcodes of the Vietoris\textendash Rips filtrations on the point clouds according to (f) Isomap, and according to contagion maps with (g) $T=0$, (h) $T=0.1$, (i) $T=0.2$, and (j) $T=0.3$. }
\label{torus_Ripser_d_ng_4}
\end{figure}

Figures~\ref{torus_Ripser_d_ng_2} and \ref{torus_Ripser_d_ng_4} show the barcodes corresponding to the persistent homology in dimension $1$ of the Vietoris--Rips filtrations built according to the different versions of Isomap and contagion maps.  
The barcodes in dimension $1$ of the Vietoris\textendash Rips filtration based on the estimated geodesic distances (i.e.~the entries in $D_{\rm iso}$ and $D_{\rm cont}$) do not seem to reveal any significant features for either Isomap or contagion maps. (See panels (a--e) of Figures~\ref{torus_Ripser_d_ng_2} and \ref{torus_Ripser_d_ng_4}). 
The barcode in dimension $1$ of the Vietoris\textendash Rips filtration on the point cloud based on Isomap (i.e.~given by the rows of $D_{\rm iso}$) on the network with $d^{\rm{NG}}=2$ does feature two dominant bars (see Figure~\ref{torus_Ripser_d_ng_2}(f)), as do the barcodes corresponding to point clouds based on contagion maps for $T=0$, $T=0.1$, and $T=0.2$ (i.e.~given by the rows of $D_{\rm cont}$) (see Figure~\ref{torus_Ripser_d_ng_2}~(g--i)). For the network with $d^{\rm{NG}}=4$, however, the barcode in dimension $1$ of the Vietoris\textendash Rips filtration on the point cloud based on Isomap does not have any dominant bars, whereas the ones based on contagion maps for $T=0.2$ and $T=0.3$ do (see Figure~\ref{torus_Ripser_d_ng_4}~(i,j)). 

\pagebreak

\pagebreak

\subsubsection{Geometry}

\begin{figure}[H]
\centering
\leftline{\hskip 0.00cm (a) \hskip 7cm (b)}
\begin{minipage}{.5\textwidth}
\includegraphics[width=1\textwidth]{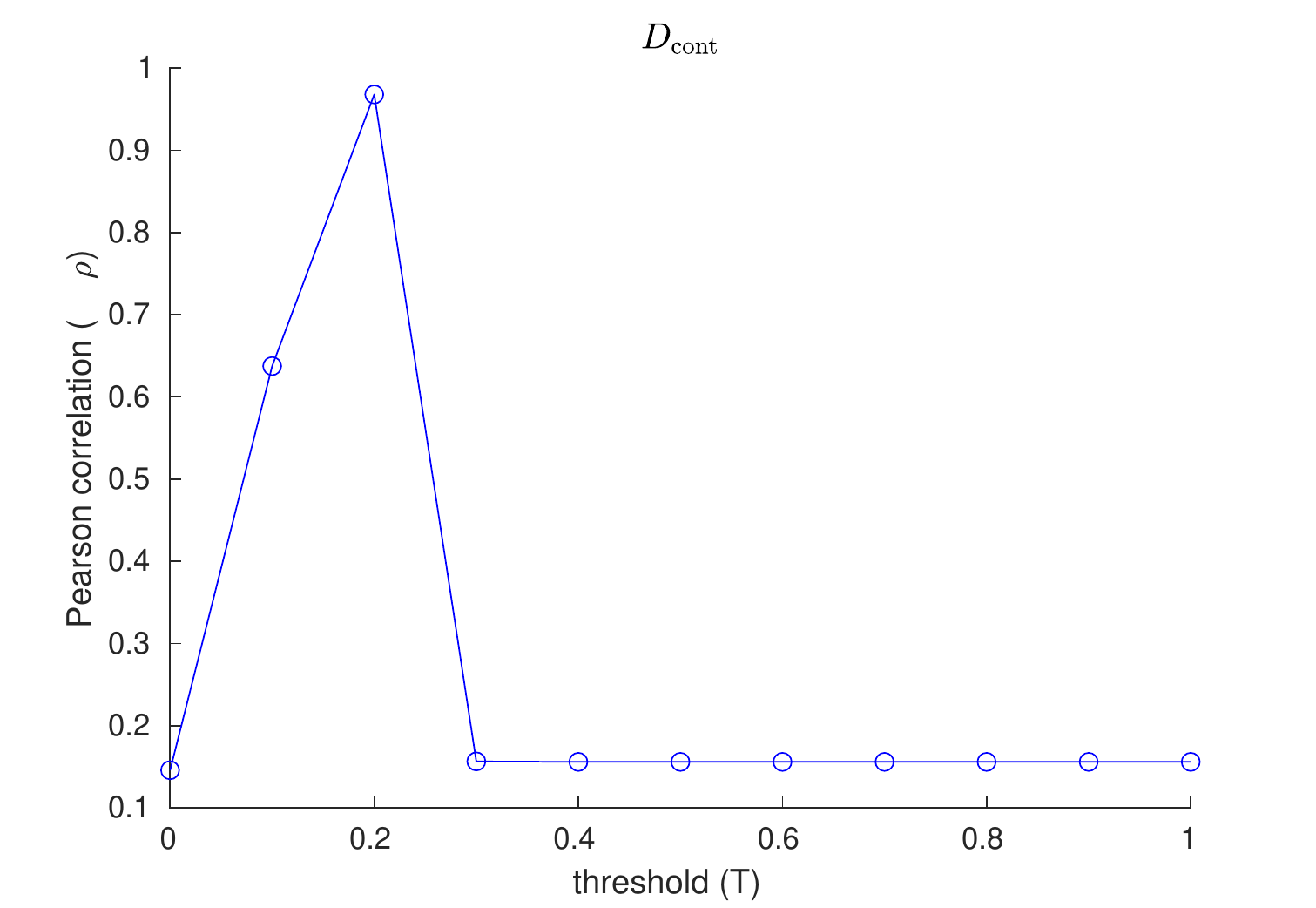}
\end{minipage}\hfill
\begin{minipage}{.5\textwidth}
\includegraphics[width=1\textwidth]{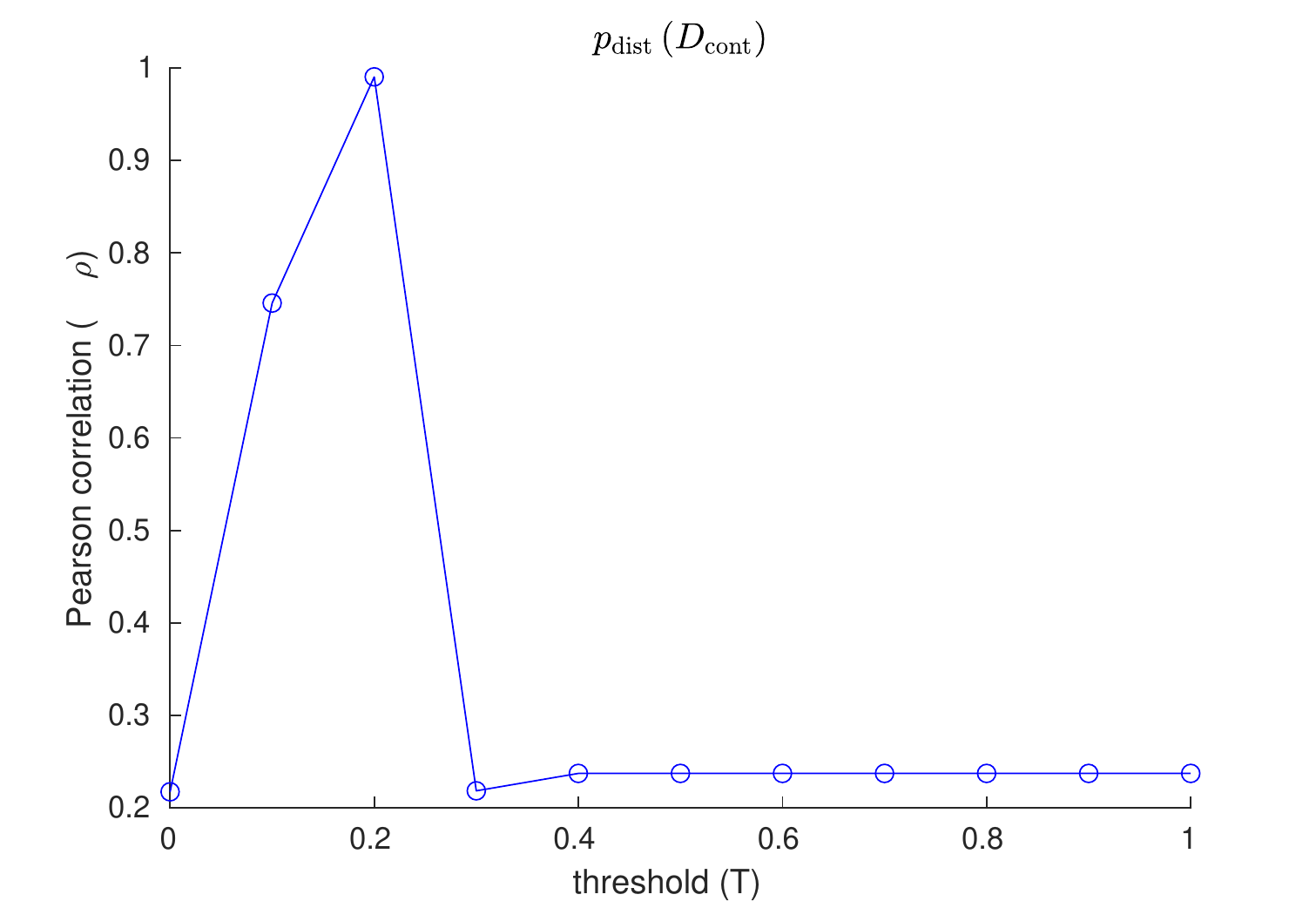}
\end{minipage}\hfill
\caption{Geometry results on our torus-based network with $d^{\rm{NG}}=2$. Pearson correlation coefficient between the pairwise distances between regularly spaced points on a torus \eqref{reg_points} and the following sets: (a) The estimated geodesic distances (i.e.~the entries in $D_{\rm cont}$) according to contagion maps with thresholds $T=0,0.1,\dots,1$. (b) The pairwise distances between points in $\mathbb{R}^{2500}$ whose coordinate vectors are the rows of $D_{\rm cont}$ according to contagion maps with thresholds $T=0,0.1,\dots,1$. The results for Isomap are practically identical to those for contagion map with $T=0$: The Pearson correlation is $0.1458$ when based on the entries in $D_{\rm iso}$; and it is $0.2175$ when based on the point cloud whose coordinate vectors are given by the rows of $D_{\rm iso}$. }
\label{torus_geometry_dng2}
\end{figure}

\begin{figure}[H]
\centering
\leftline{\hskip 0.00cm (a) \hskip 7cm (b)}
\begin{minipage}{.5\textwidth}
\includegraphics[width=1\textwidth]{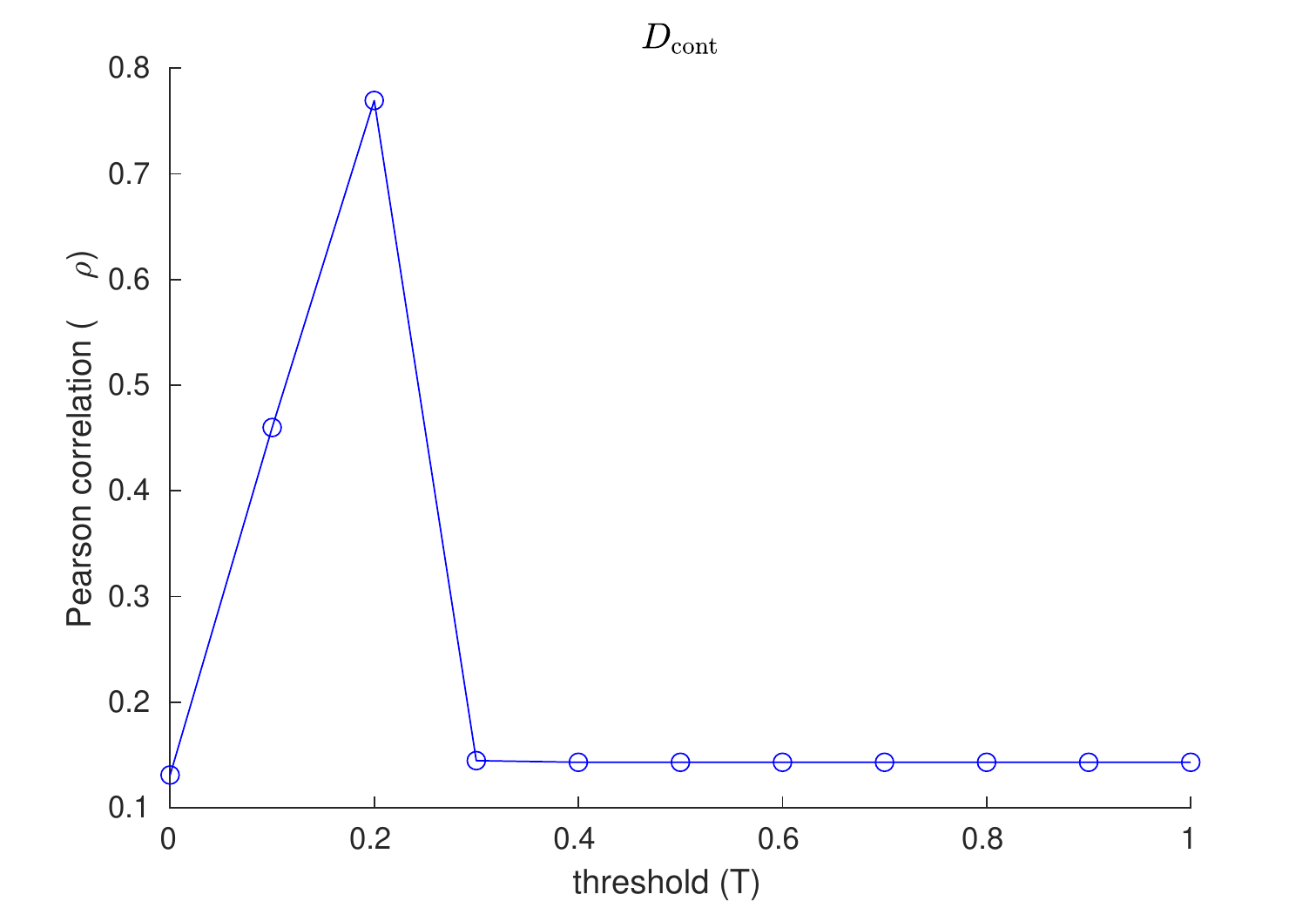}
\end{minipage}\hfill
\begin{minipage}{.5\textwidth}
\includegraphics[width=1\textwidth]{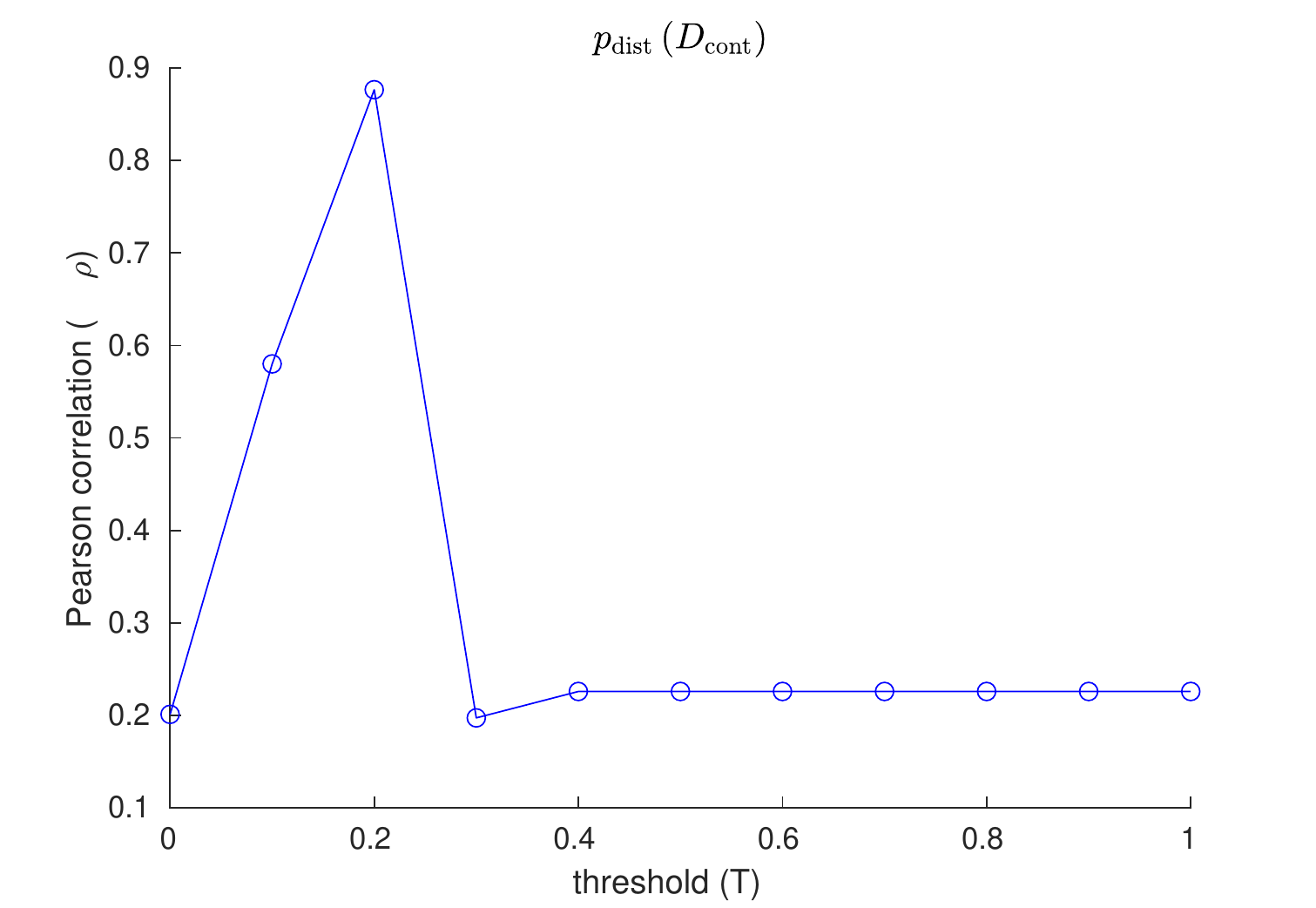}
\end{minipage}\hfill

\caption{Geometry results on our torus-based network with $d^{\rm{NG}}=4$. Pearson correlation coefficient between the pairwise distances between regularly spaced points on a torus \eqref{reg_points} and the following sets: (a) The estimated geodesic distances (i.e.~the entries in $D_{\rm cont}$) according to contagion maps with thresholds $T=0,0.1,\dots,1$. (b) The pairwise distances between points in $\mathbb{R}^{2500}$ whose coordinate vectors are the rows of $D_{\rm cont}$ according to contagion maps with thresholds $T=0,0.1,\dots,1$. The results for Isomap are practically identical to those for contagion map with $T=0$: The Pearson correlation is $0.1311$ when based on the entries in $D_{\rm iso}$; and it is $0.2013$ when based on the point cloud whose coordinate vectors are given by the rows of $D_{\rm iso}$.}
\label{torus_geometry_dng4}
\end{figure}

We examine the geometry of our Isomap and contagion map results by comparing the entries of $D_{\rm iso}$ and $D_{\rm cont}$, as well as the point clouds given by the rows of these dissimilarity matrices, to the regularly spaced points on a torus \eqref{reg_points} via the Pearson correlation coefficient. See Figure~\ref{torus_geometry_dng2} for the results for the torus-based network with non-geometric degree $d^{\rm NG}=2$ and Figure~\ref{torus_geometry_dng4} for the results for the torus-based network with non-geometric degree $d^{\rm NG}=4$. Isomap returns a low Pearson correlation in all cases, suggesting that the shortest-path distances are (as expected, given the large number of non-geometric edges) not good estimates to the distances along the torus that underlies these networks. Note that these results are practically identical to those for contagion map with $T=0$, as we are working with the same unweighted graphs in both Isomap and contagion maps. For contagion maps with varying thresholds $T$, the Pearson correlation coefficient peaks around $T=0.2$, suggesting that, for thresholds close to $T=0.2$, the contagion spreads predominantly via WFP, making the activation times good estimates to the distance along the torus that underlies these networks.

\subsection{Conformation space of the cyclo-octane molecule}
The cyclo-octane molecule $(CH_2)_8$ consists of a ring of eight carbon atoms, each bonded with two hydrogen atoms. 
\begin{figure}[H]
\centering
     \leftline{\hskip 0.5cm (a) \hskip 6.5cm (b)} 
\includegraphics[scale=.3]{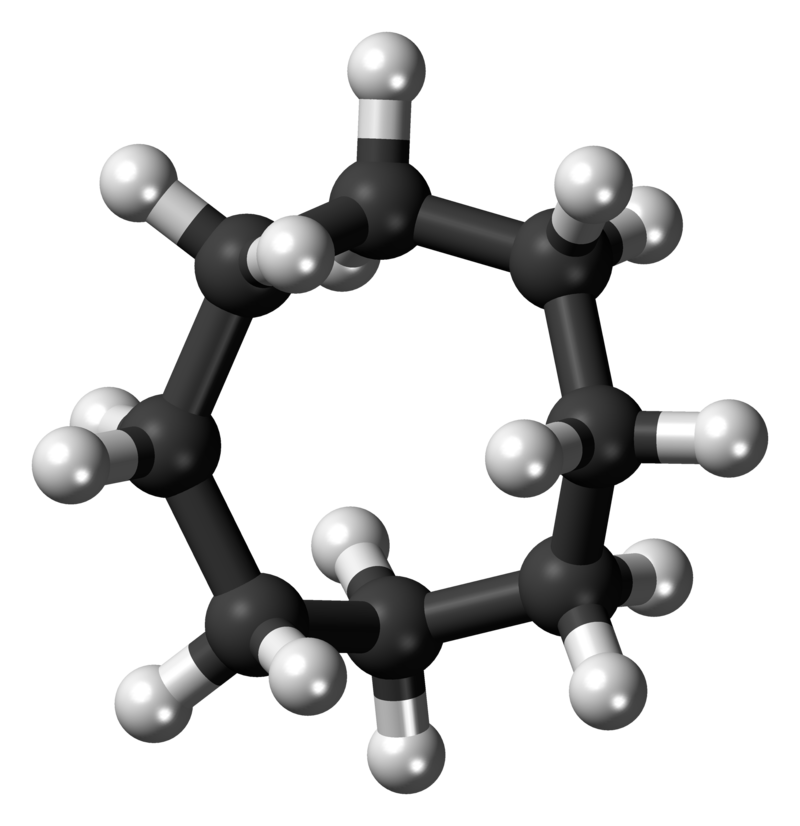}
\includegraphics[scale=.3]{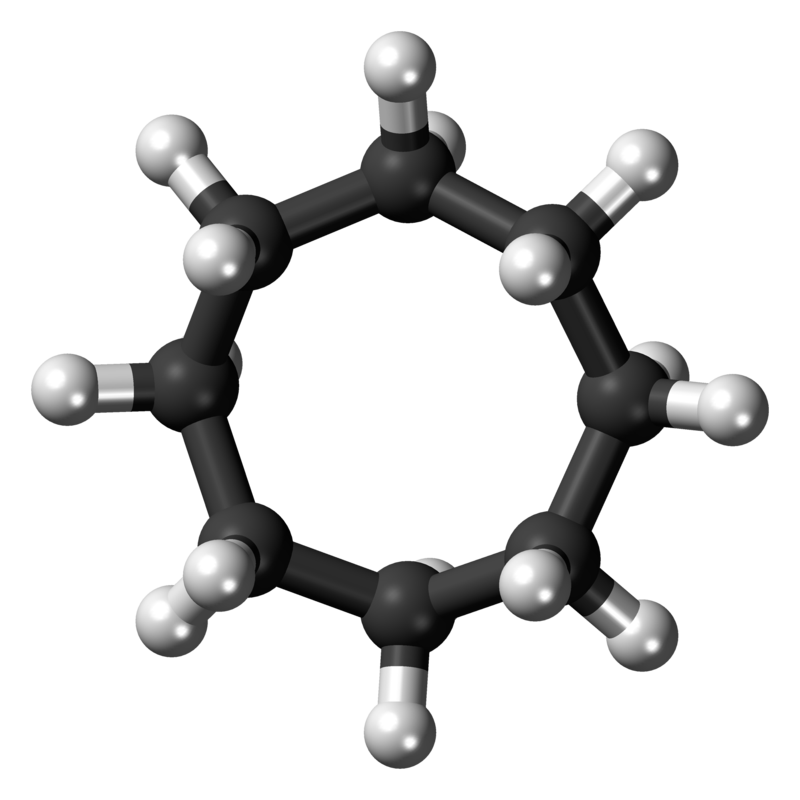}
\caption{Ball-and-stick model representations of the cyclo-octane molecule in (a) its boat-chair conformation and (b) its crown conformation \cite{Pakes1981}. (Figure taken from \cite{cyclopic}.) }\label{cyclooctane}
\end{figure}

A \emph{conformation} of a molecule is a possible spatial arrangement of its atoms (modulo rotation and translation) \cite{moss1996basic}. The conformation of a molecule can be specified by the coordinates of each of its atoms in three-dimensional space, giving a point in $\mathbb{R}^{3a}$, where $a$ is the number of atoms in the molecule. (In this case, each three-dimensional coordinate of each atom is a feature and $\mathbb{R}^{3a}$ is the feature space.) The set of such points for all conformations of a molecule is called its \emph{conformation space}. Each conformation is accompanied by a state of potential energy of the molecule, and a conformation is more likely to occur the lower its associated potential energy. The cyclo-octane molecule has many conformations of comparable potential energy, and its conformation space has been studied in computational chemistry for over 50 years \cite{Hendrickson1967,Pakes1981}. Given the locations of the eight carbon atoms in a conformation of the cyclo-octane molecule, the locations of the hydrogen atoms are determined to minimize energy: The two covalent hydrogen atoms of each carbon atom are positioned to form a tetrahedral arrangement with the two neighbouring carbon atoms that minimizes the potential energy of that subunit of the molecule. The conformation space of cyclo-octane thus lies in $\mathbb{R}^{3\times 8}=\mathbb{R}^{24}$. It is generally assumed that conformation spaces form low-dimensional manifolds, so identifying the structure of the conformation space of a molecule is essentially a manifold-learning problem. The conformation space of cyclo-octane has been shown to be the union of a sphere with a Klein bottle intersecting in two circles of singularities \cite{Brown2008,Martin2010}, forming a two-dimensional manifold with singularities. 

Martin \emph{et al.} \cite{Martin2010} analysed a data set of $6040$ points in the conformation space of cyclo-octane, subsampled from a larger data set consisting of $1031644$ cyclo-octane conformations. This data set is publicly available as part of the {\sc javaPlex} software package\cite{Javaplex}. To visualize this set of points, Martin \emph{et al.} mapped the points from $\mathbb{R}^{24}$ to $\mathbb{R}^{3}$ via Isomap. We explore different versions of both Isomap and contagion maps on it. 

\begin{figure}[H]
\centering
     \leftline{\hskip 0.00cm (a) \hskip 7cm (b)} 
     \begin{minipage}{0.45\textwidth}
\includegraphics[scale=.4]{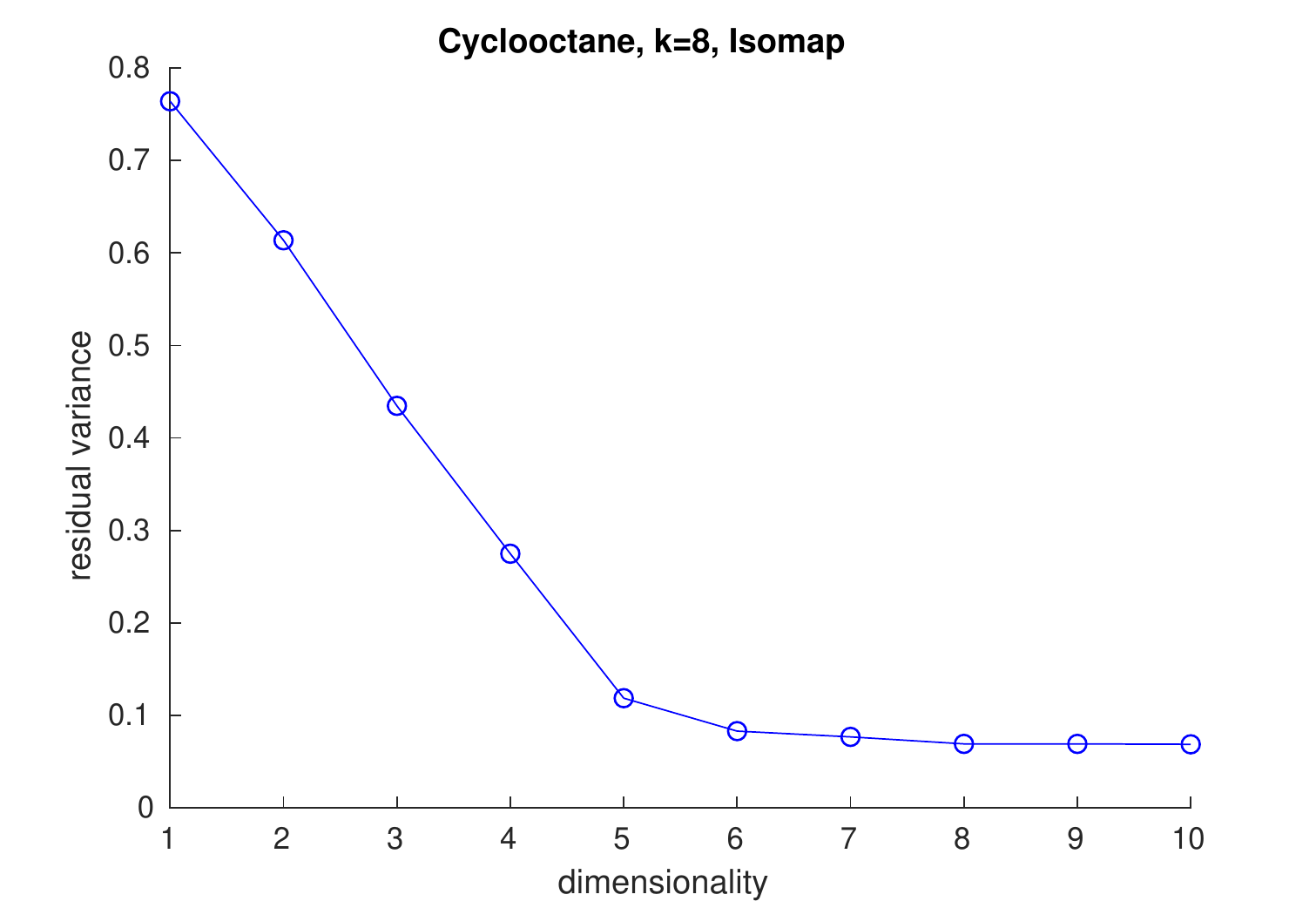}
\end{minipage}
\begin{minipage}{0.45\textwidth}
\includegraphics[scale=.4]{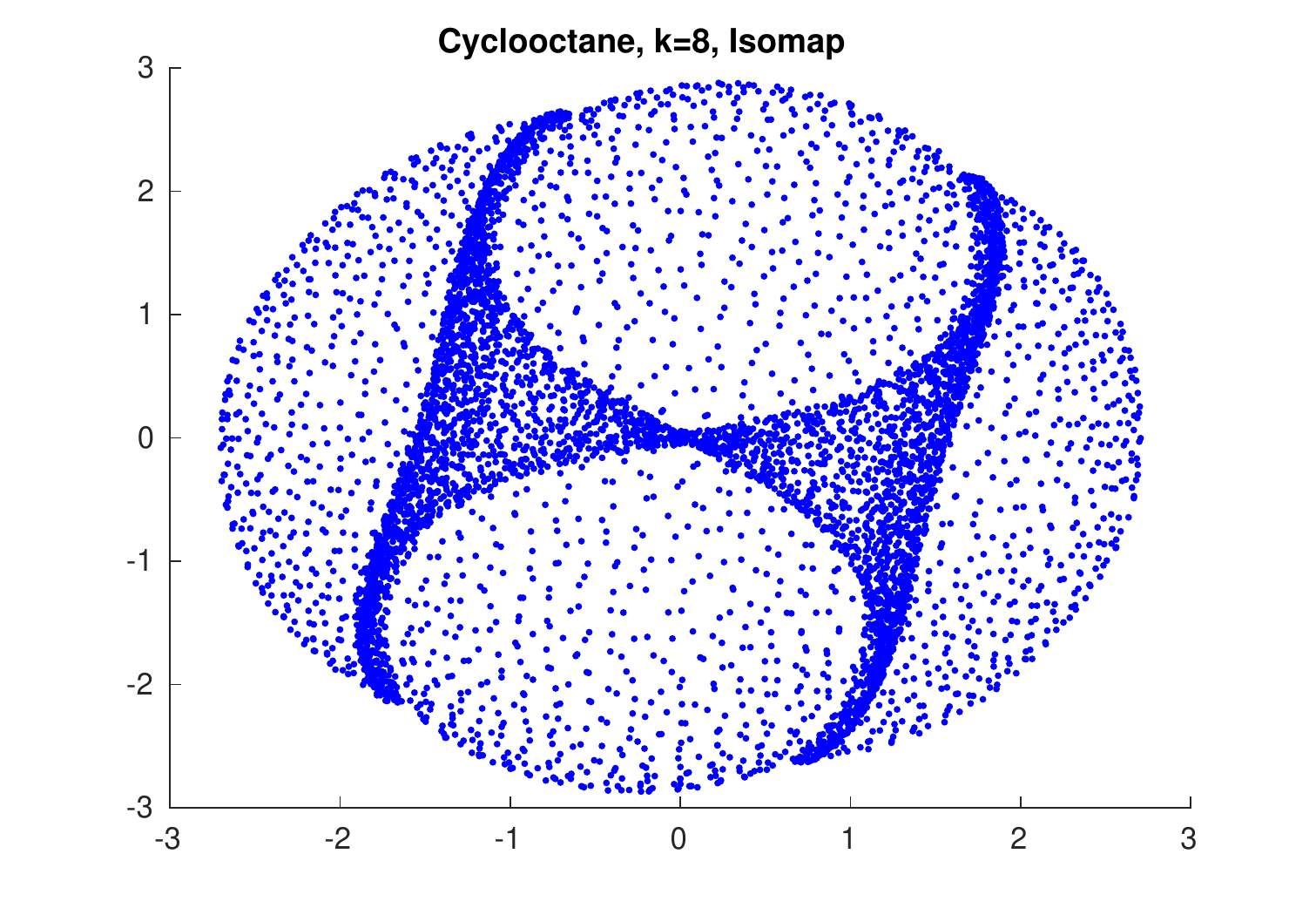}
\end{minipage}
\caption{Results for Isomap on the data set of 6040 points on the conformation space of cyclo-octane (using an $8$-nearest neighbour neighbourhood graph). (a) Residual variances for projections via MDS onto dimensions 1 to 10 in the original Isomap algorithm. (b) Visualization of Isomap in 3D.}\label{cycloisomap}
\end{figure}

\begin{figure}[H]
\leftline{\hskip 0.00cm (a) \hskip 7cm (e)} 
     \begin{minipage}{0.45\textwidth}
\includegraphics[scale=.4]{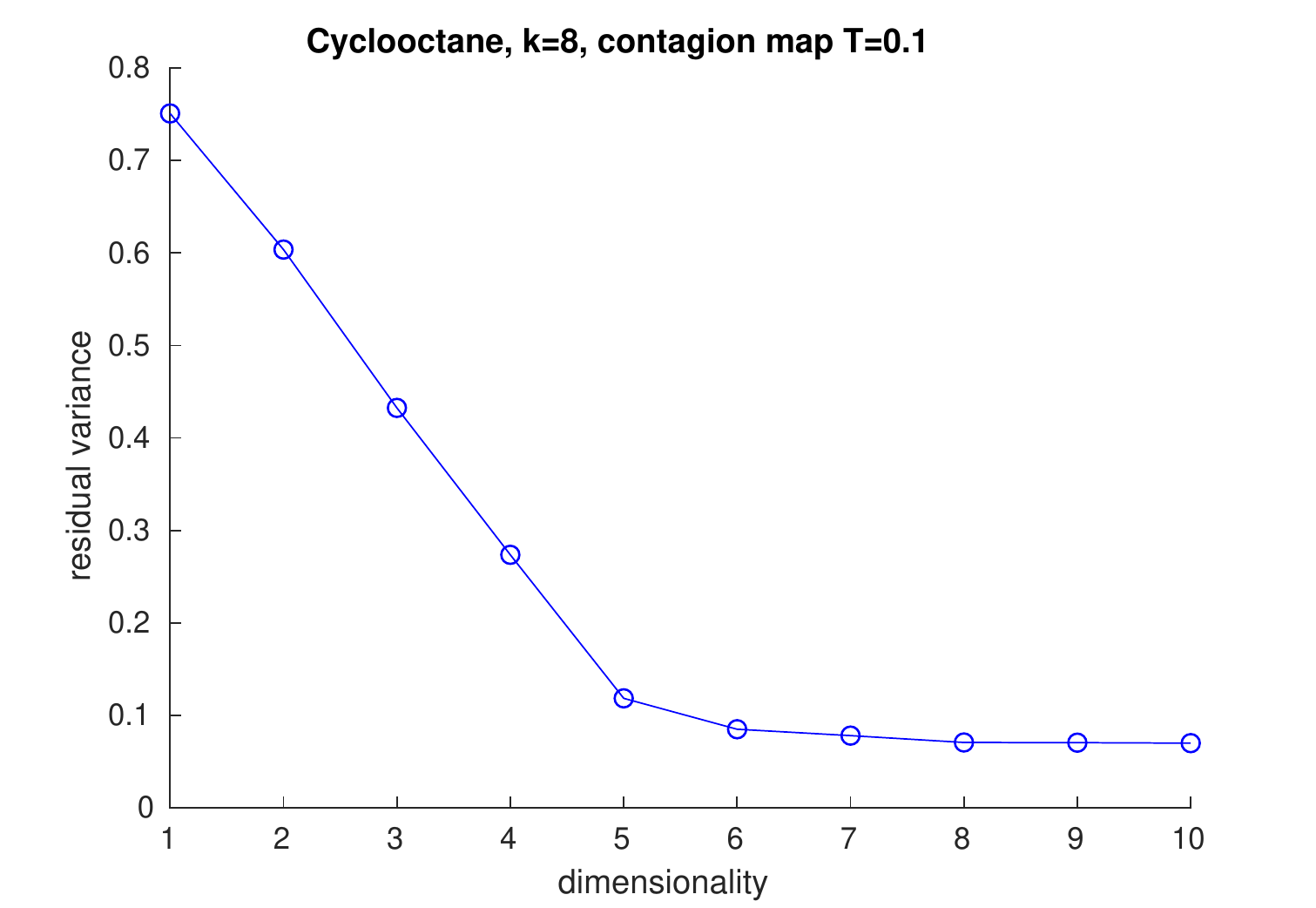}
\end{minipage}
\begin{minipage}{0.45\textwidth}
\includegraphics[scale=.4]{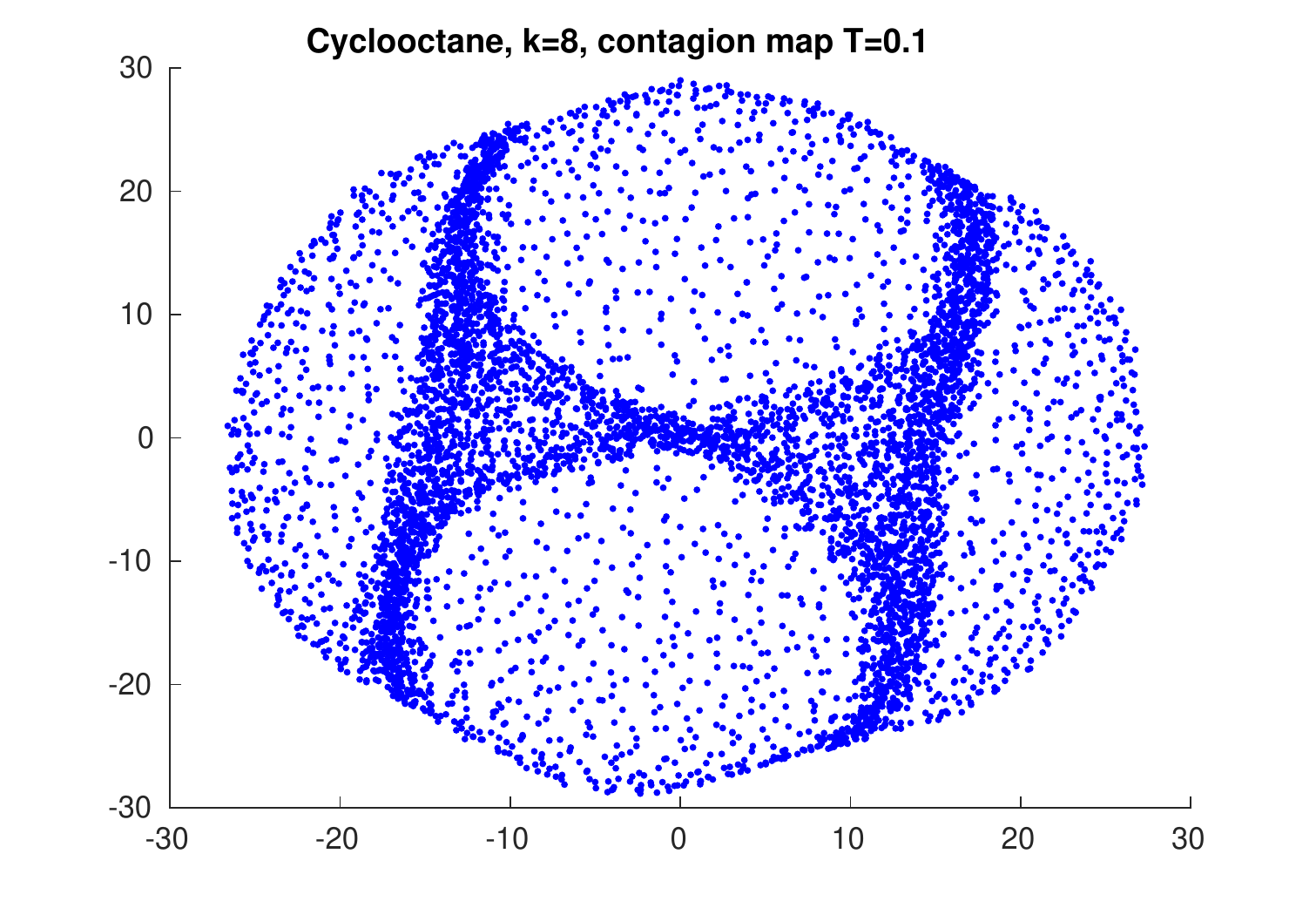}
\end{minipage}

     \leftline{\hskip 0.00cm (b) \hskip 7cm (f)} 
     \begin{minipage}{0.45\textwidth}
\includegraphics[scale=.4]{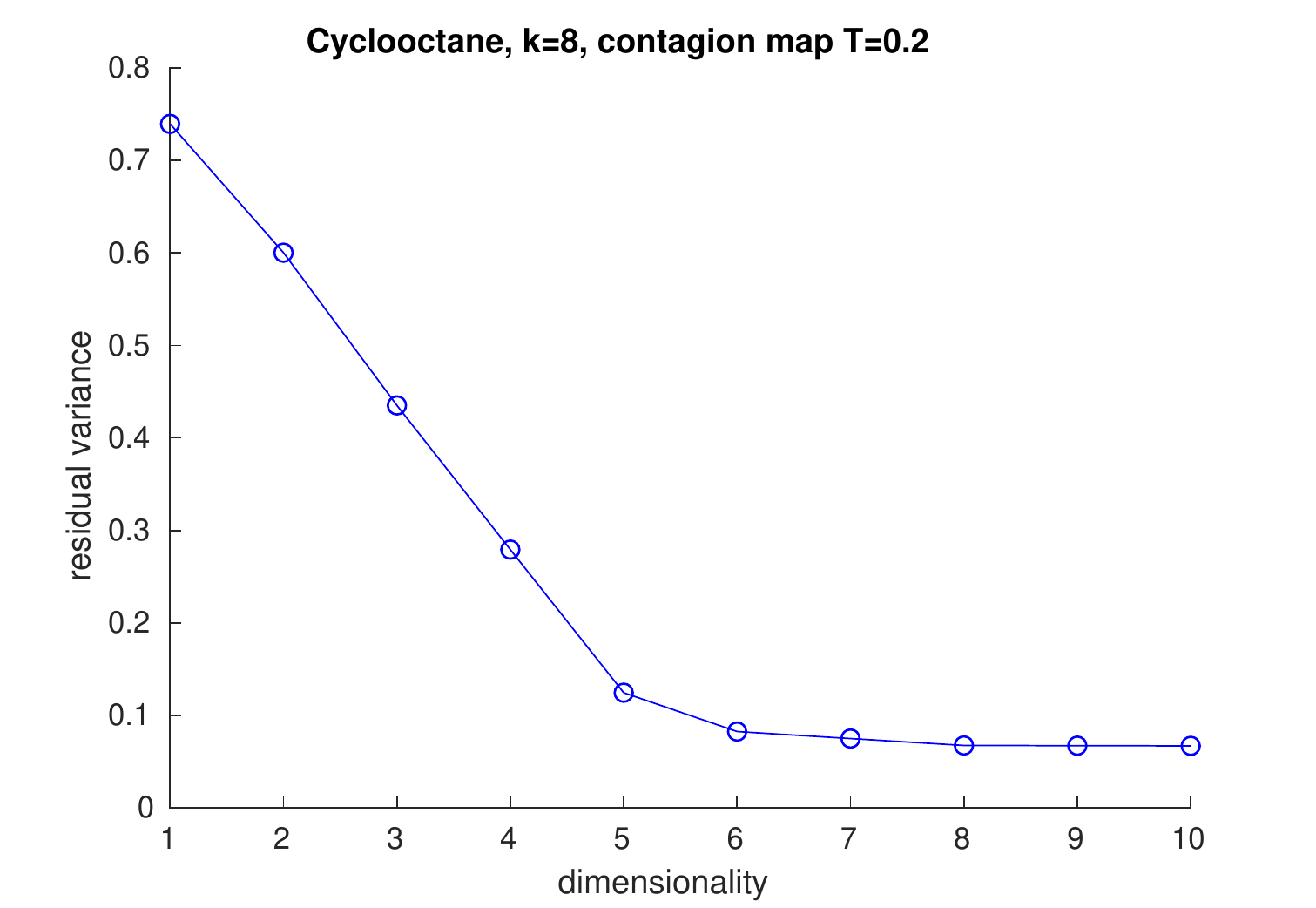}
\end{minipage}
\begin{minipage}{0.45\textwidth}
\includegraphics[scale=.4]{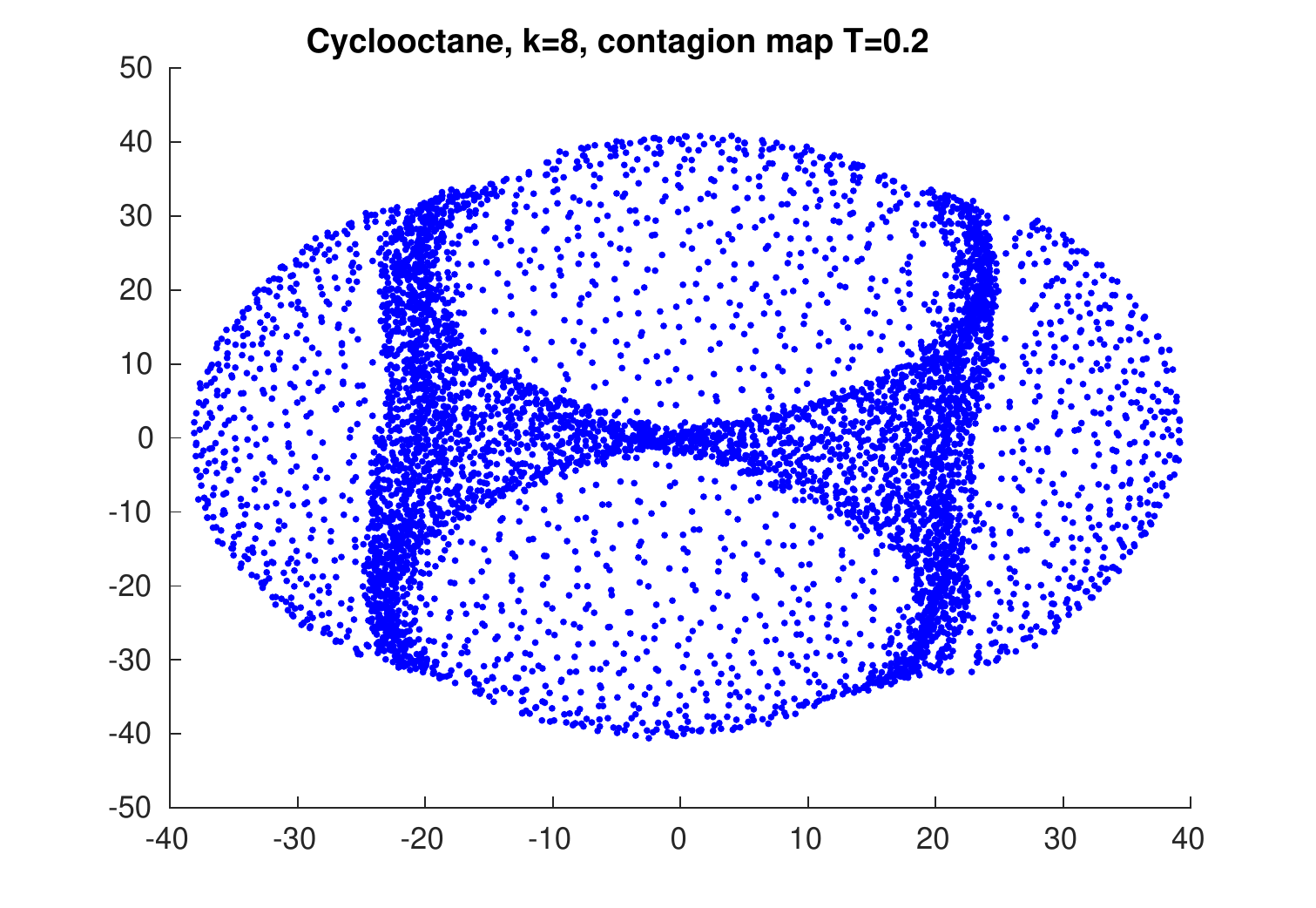}
\end{minipage}

\leftline{\hskip 0.00cm (c) \hskip 7cm (g)} 
     \begin{minipage}{0.45\textwidth}
\includegraphics[scale=.4]{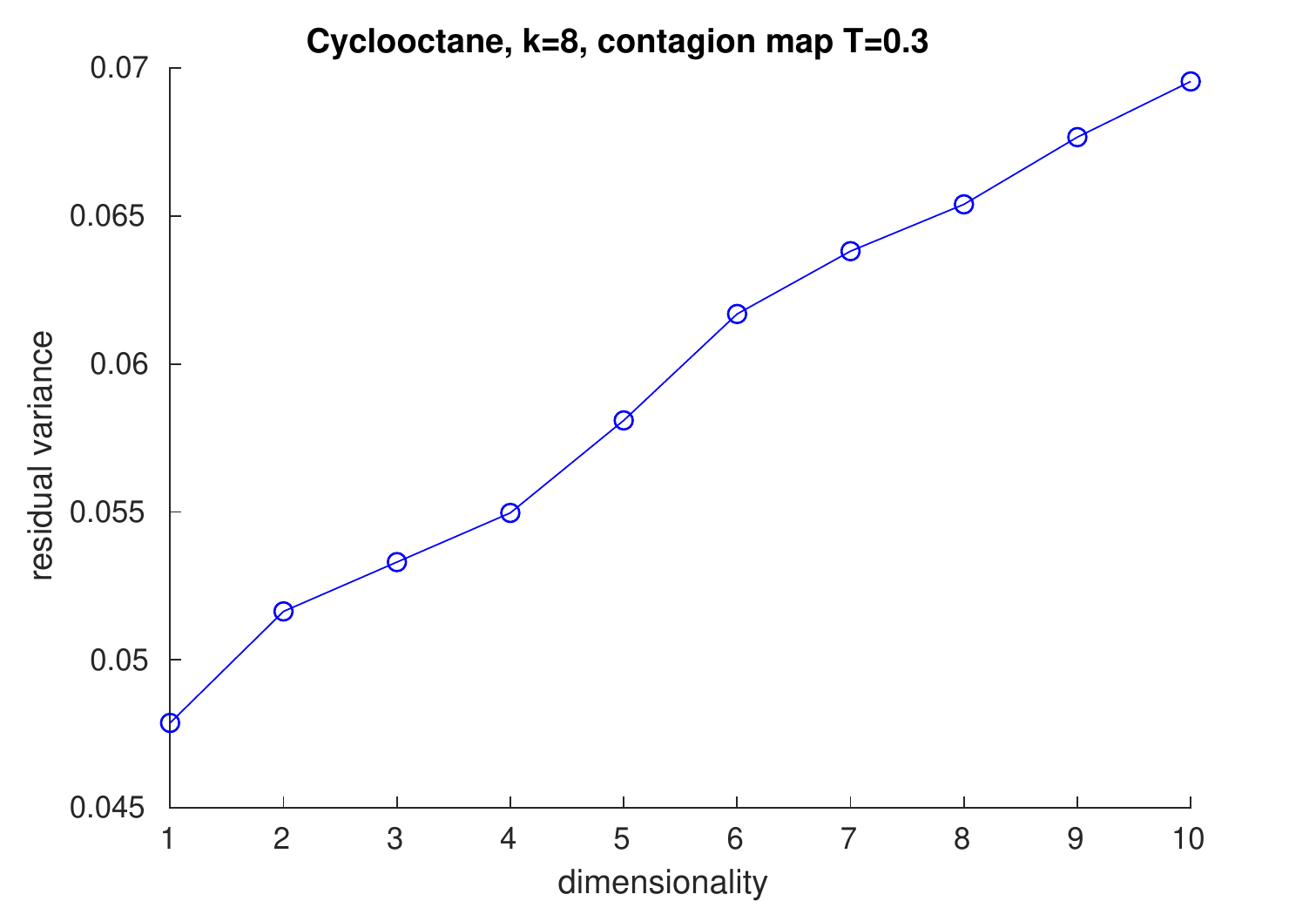}
\end{minipage}
\begin{minipage}{0.45\textwidth}
\includegraphics[scale=.4]{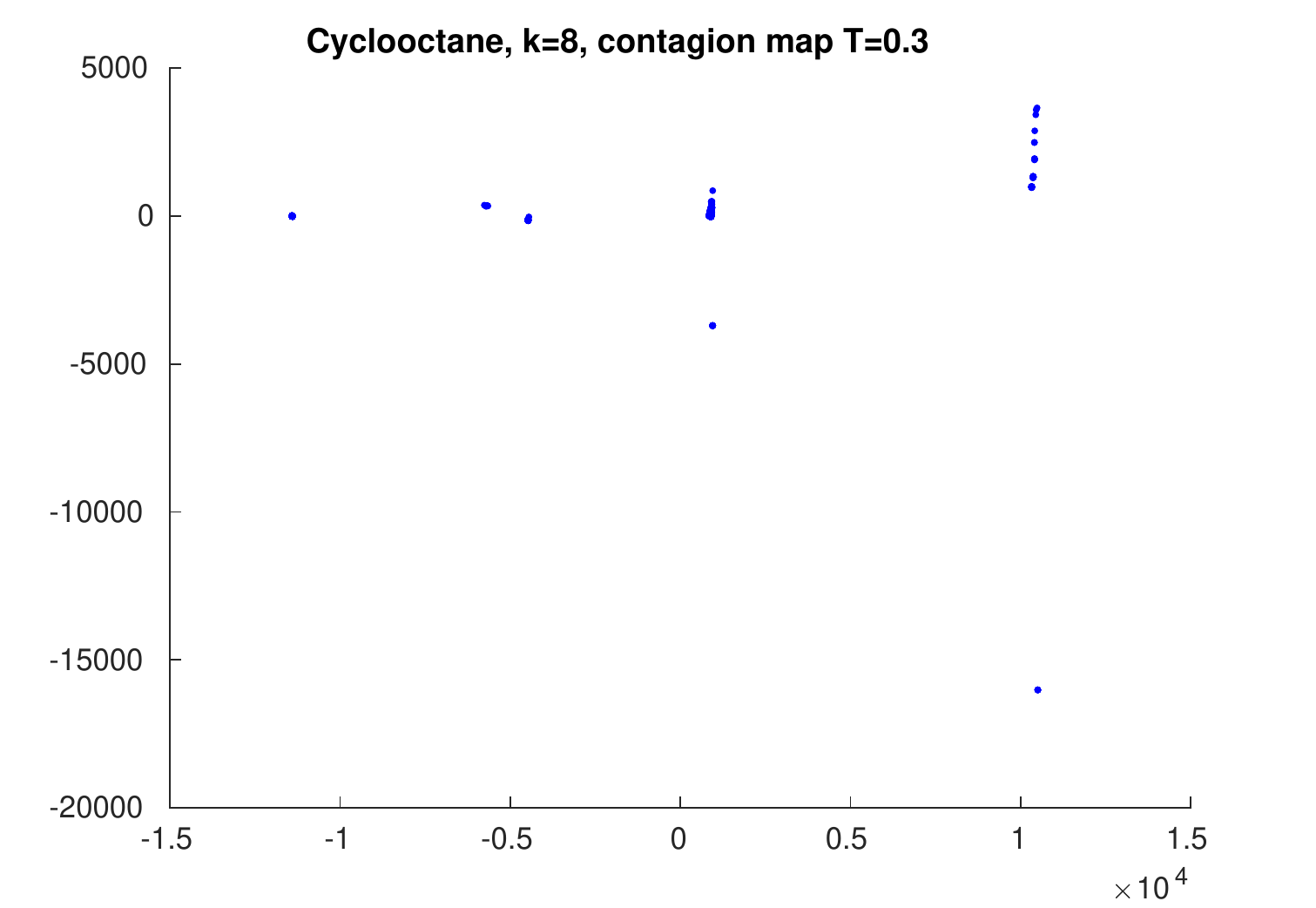}
\end{minipage}

\leftline{\hskip 0.00cm (d) \hskip 7cm (h)} 
     \begin{minipage}{0.45\textwidth}
\includegraphics[scale=.4]{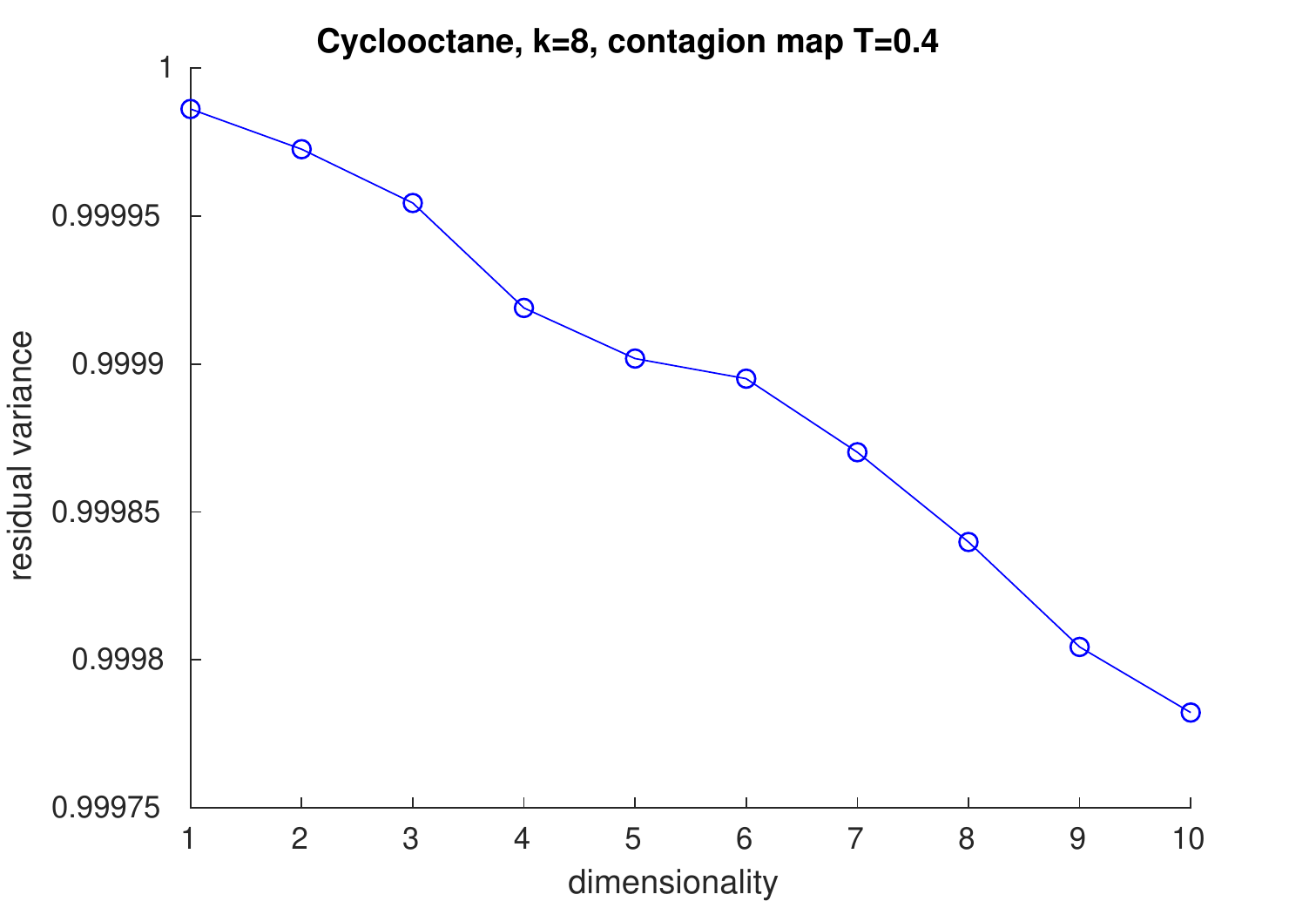}
\end{minipage}
\begin{minipage}{0.45\textwidth}
\includegraphics[scale=.4]{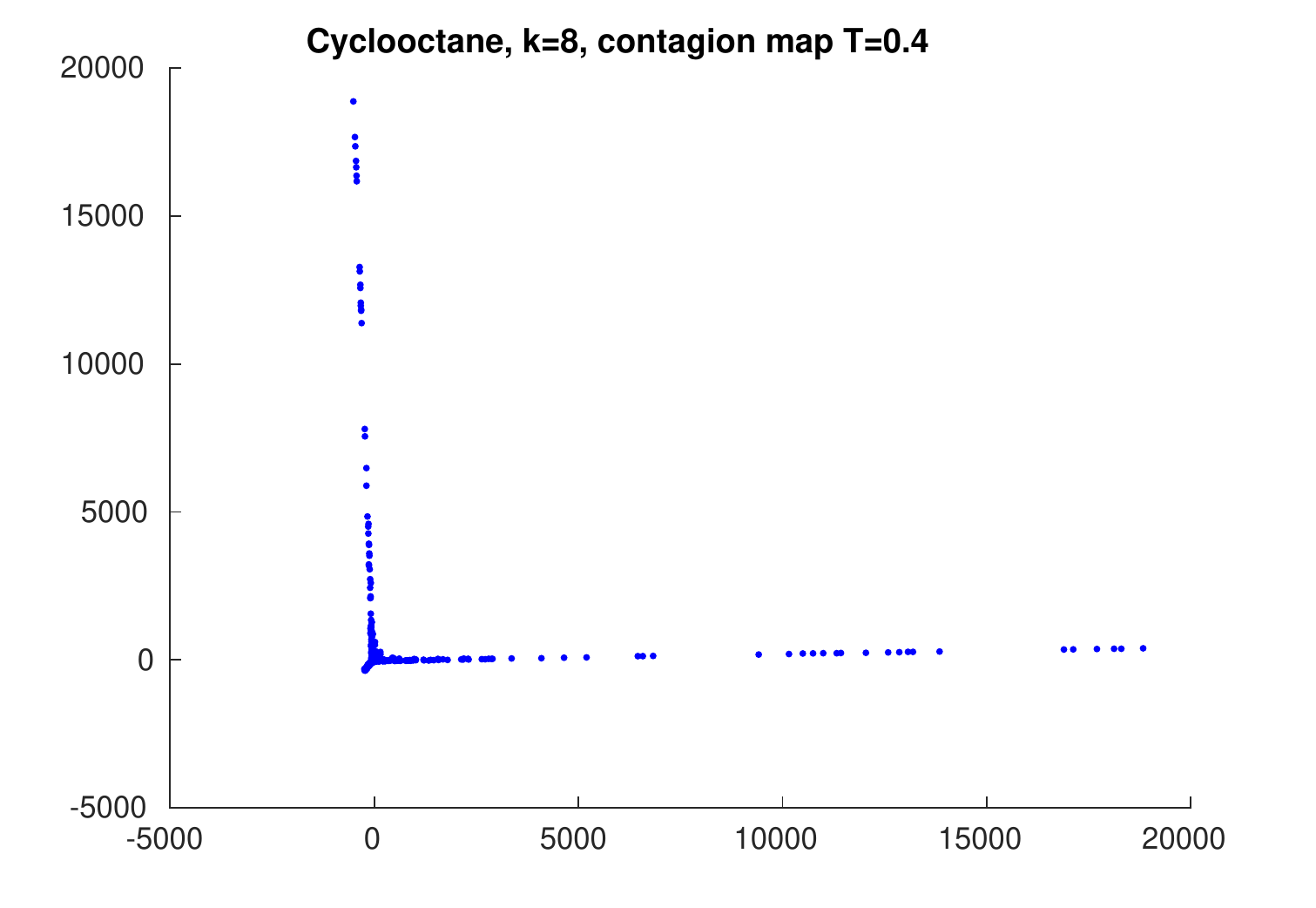}
\end{minipage}
\caption{Results for contagion maps for different values of $T$ on the data set of 6040 points on the conformation space of cyclo-octane (using an $8$-nearest-neighbour graph). (a--d) Residual variances for projections onto dimensions 1 to 10 in the contagion-map algorithm for (a) $T=0.1$, (b) $T=0.2$, (c) $T=0.3$, (b) $T=0.4$. (e--h) Visualizations of projections to 3D for (e) $T=0.1$, (f) $T=0.2$, (g) $T=0.3$, (h) $T=0.4$.}\label{cyclocontagionmap}
\end{figure}

Figure~\ref{cycloisomap} shows the residual variances for projections via MDS onto dimensions 1 to 10 based on shortest-path distances (i.e.~based on the entries in $D_{\rm iso}$), and the visualization of the projection to 3D. 
Figure~\ref{cyclocontagionmap} shows the residual variances for projections via MDS onto dimensions 1 to 10 based on activation times in contagions with thresholds $T=0.1,0.2,0.3,$ and $0.4$ (i.e.~based on the entries in $D_{\rm cont}$), and the visualizations of the projection to 3D. We see that Isomap and contagion maps with low thresholds ($T=0.1$ and $0.2$) detect the embedding dimension of the underlying space, suggesting the absence of noisy edges in the $8$-nearest-neighbour graph on this data set. Contagion maps with higher thresholds ($T=0.3$ and $0.4$) do not seem to reveal any meaningful structure. This is likely due to the contagion stabilizing before much of the graph has been activated, leading to many `infinite' activation times. 

\begin{figure}[H]
\centering
     \leftline{\hskip 5.00cm (a)}  
\includegraphics[scale=.4]{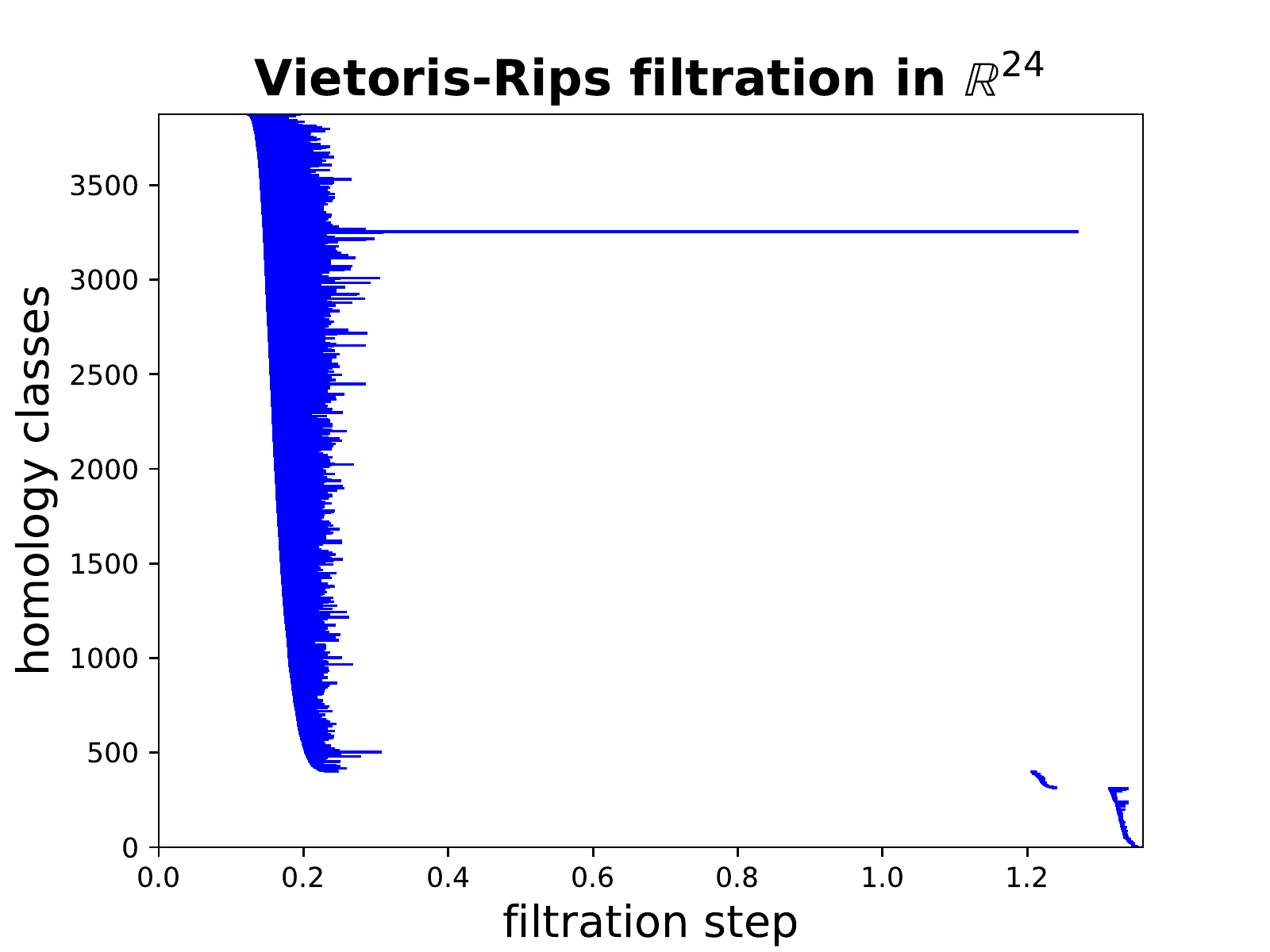}

\leftline{\hskip 0cm (b) \hskip 7cm (c)}
\begin{minipage}{0.45\textwidth}
\includegraphics[scale=.4]{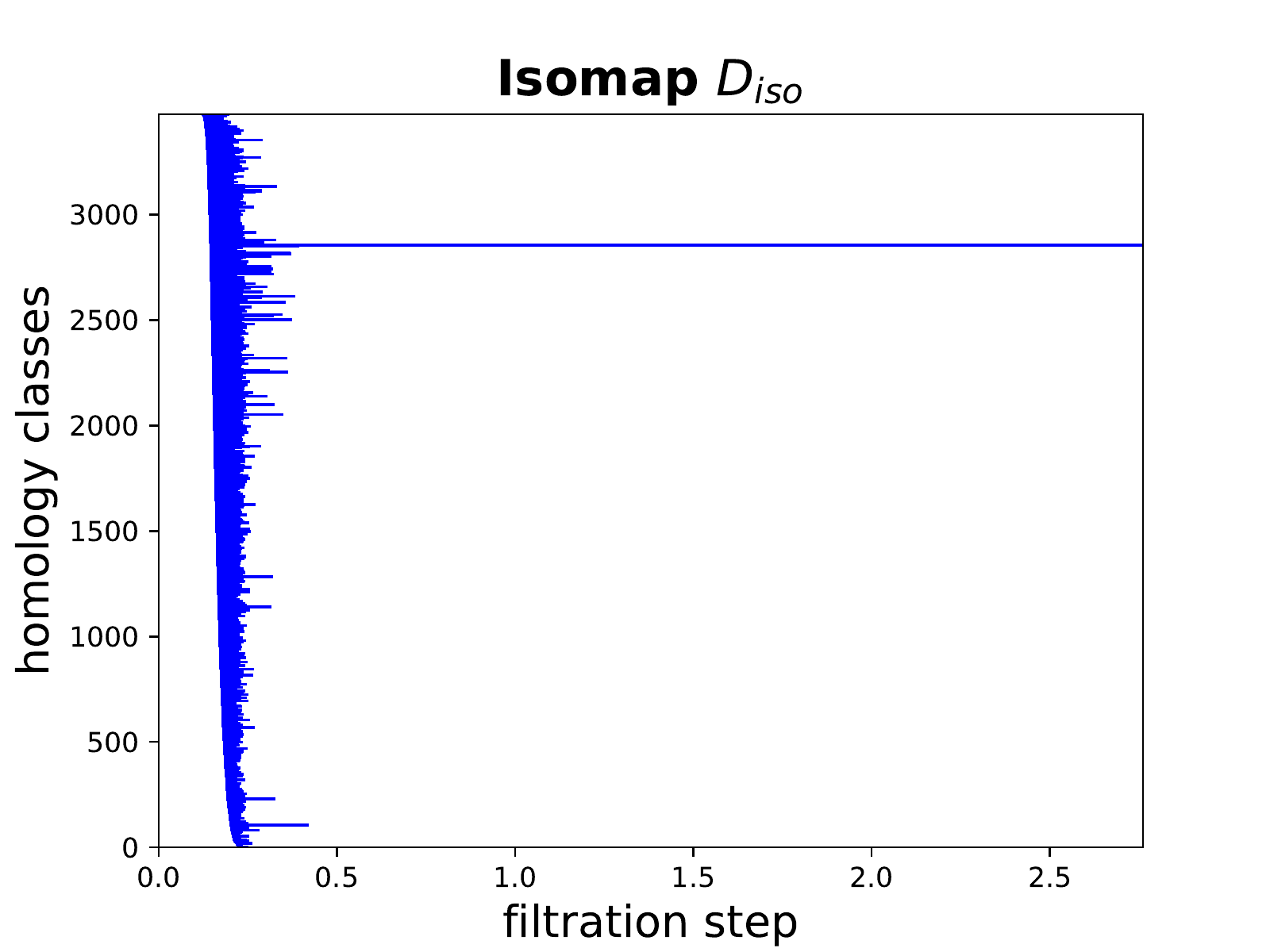}
\end{minipage}
\begin{minipage}{0.45\textwidth}
\includegraphics[scale=.4]{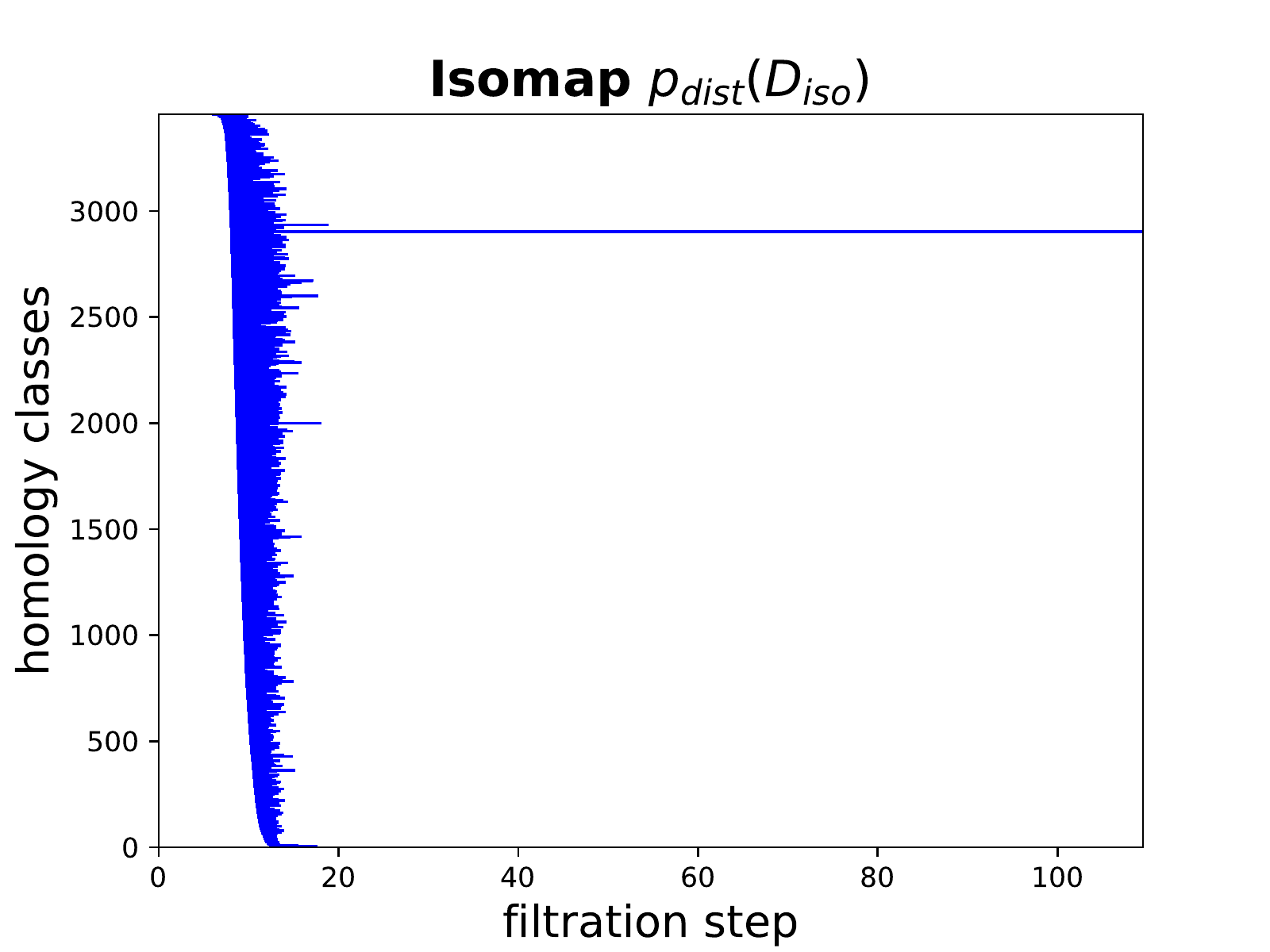}
\end{minipage}

\leftline{\hskip 0cm (d) \hskip 7cm (e)}
\begin{minipage}{0.45\textwidth}
\includegraphics[scale=.4]{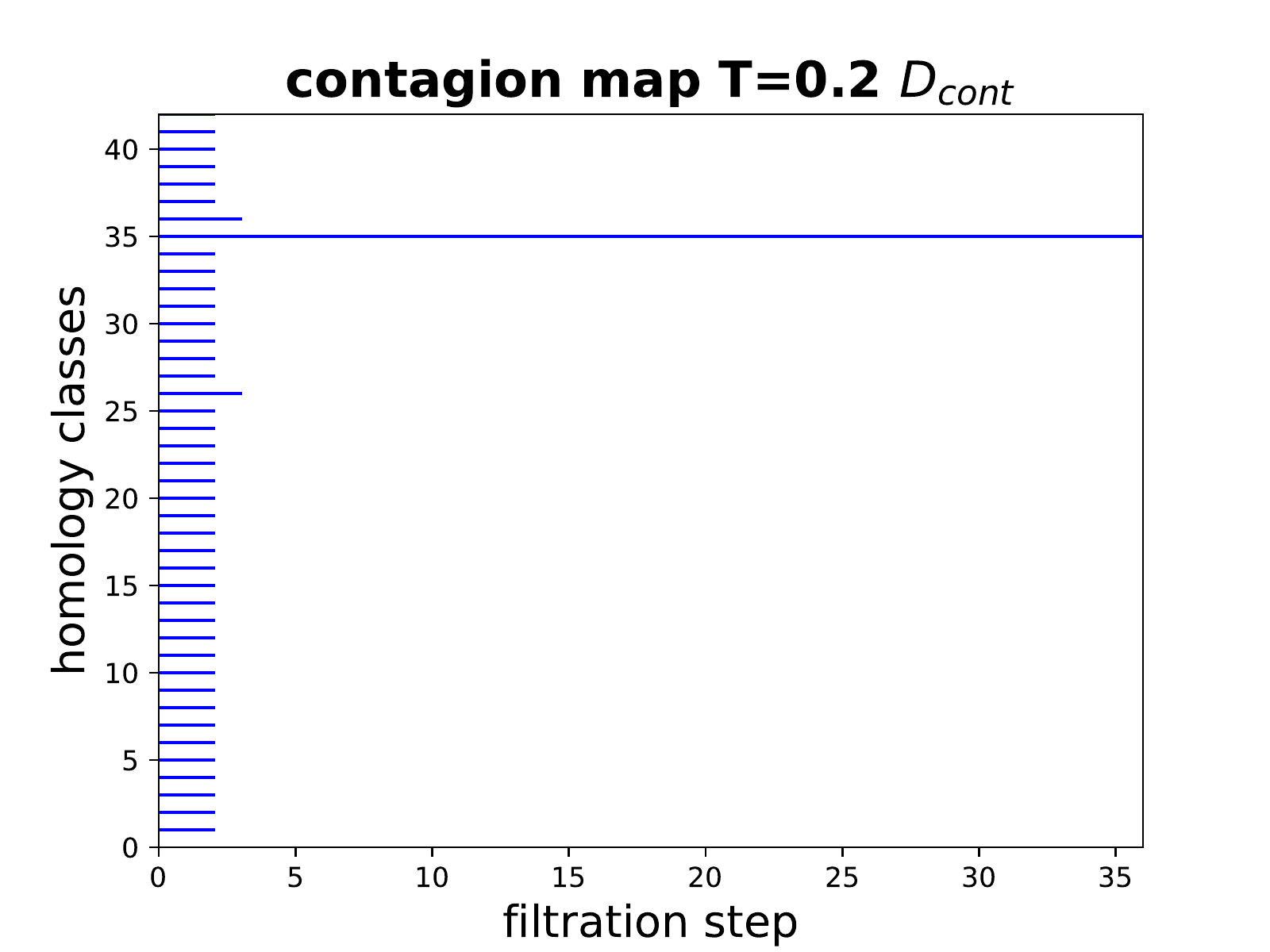}
\end{minipage}
\begin{minipage}{0.45\textwidth}
\includegraphics[scale=.4]{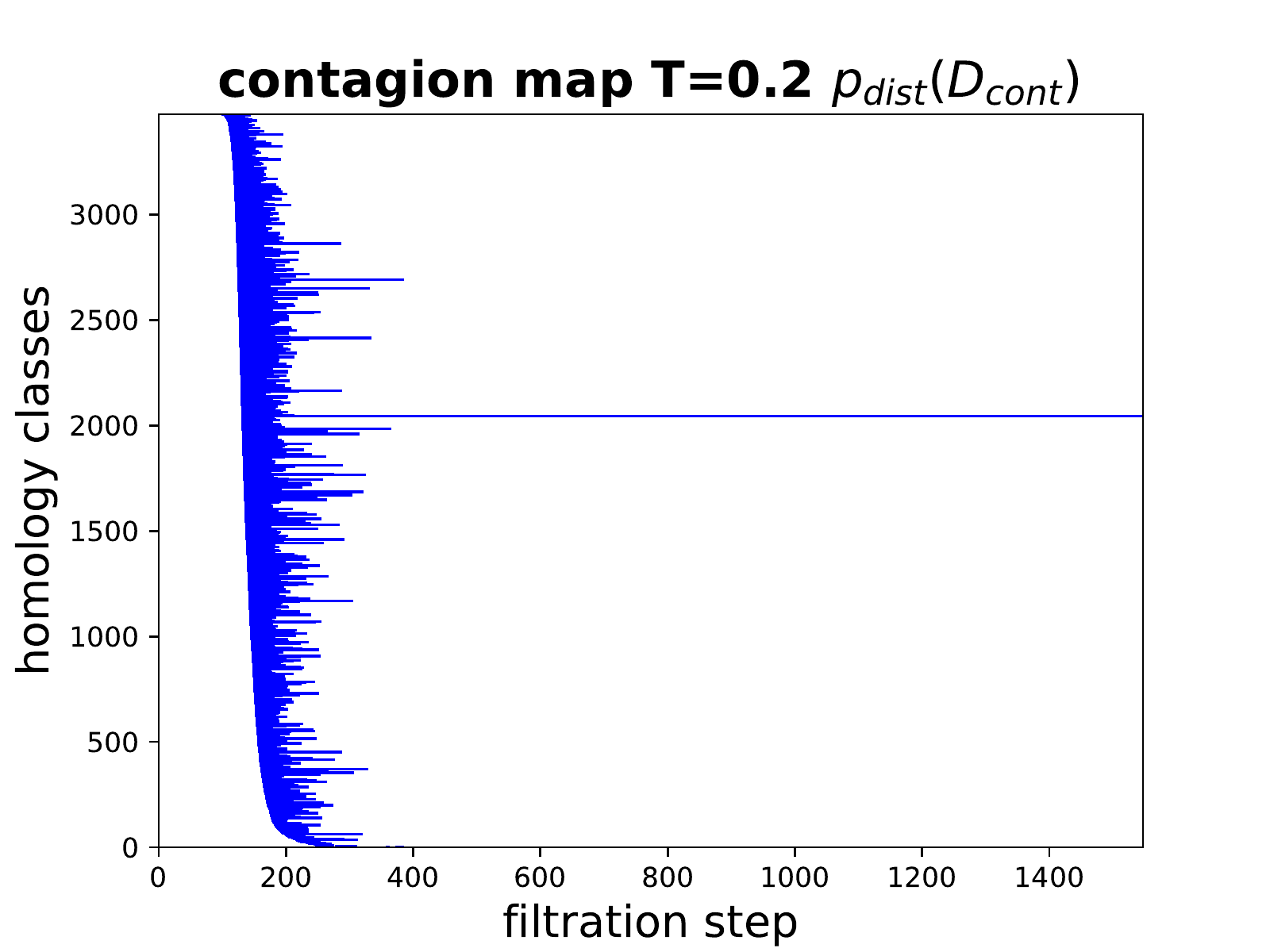}
\end{minipage}
\caption{Barcodes for the persistent homology in dimension $1$ of various Vietoris--Rips filtrations based on the cyclo-octane data set. (a) Vietoris--Rips filtration directly on the data points in $\mathbb{R}^{24}$. (b) Vietoris--Rips filtration based on the shortest-path distances in the $8$-nearest-neighbour graph (i.e.~based on the entries in $D_{\rm iso}$). (c) Vietoris--Rips filtration based on the points whose coordinate vectors are the rows of $D_{\rm iso}$ (corresponding to the $8$-nearest-neighbour graph). (d) Vietoris--Rips filtration based on the activation times of the contagion with threshold $T=0.2$ on the $8$-nearest-neighbour graph (i.e.~based on the entries in $D_{\rm cont}$). (e) Vietoris--Rips filtration based on the points whose coordinate vectors are the rows of $D_{\rm cont}$ (corresponding to the contagion with threshold $T=0.2$ on the $8$-nearest-neighbour graph).}\label{cyclo_barcodes}
\end{figure}

In Figure~\ref{cyclo_barcodes} we show barcodes corresponding to the persistent homology in dimension $1$ of various Vietoris--Rips filtrations based on the cyclo-octane data set of $6040$ points in $\mathbb{R}^{24}$. The barcode in panel (a) corresponds to the Vietoris--Rips filtration built directly on the data points in their ambient space $\mathbb{R}^{24}$. This barcode has one dominant bar, suggesting that the Klein bottle and the sphere whose union is the conformation space of cyclo-octane intersect in a way that makes the $1$-dimensional loop that is present in the homology of the Klein bottle over $\mathbb{Z}/2\mathbb{Z}$, but not over $\mathbb{Z}$, nullhomotopic. 
The other panels show barcodes corresponding to various Vietoris--Rips filtrations that, in a sense, mimic the Vietoris--Rips filtration built according to the intrinsic metric on the underlying manifold. Namely, they are Vietoris--Rips filtrations based on different versions of Isomap and contagion map, that is, based on estimates to the geodesic distance on the underlying manifold. Panel (b) shows the barcode corresponding to the shortest-path distances in the $8$-nearest-neighbour graph (i.e.~based on the entries in $D_{\rm iso}$). Panel (c) shows the Vietoris--Rips filtration based on the points whose coordinate vectors are the columns of $D_{\rm iso}$. Panel (d) shows the Vietoris--Rips filtration based on the activation times of the contagion with threshold $T=0.2$ on the $8$-nearest-neighbour graph (i.e.~based on the entries in $D_{\rm cont}$). Panel (e) shows the Vietoris--Rips filtration based on the points whose coordinate vectors are the columns of $D_{\rm cont}$. 
All of these barcodes have one dominant bar, consistent with the homology of the underlying manifold. The barcode in panel (d) has many bars with identical birth and death. This stems from the fact that activation times (in our contagion model) have integer values between $0$ and $2N$ (where $N$ is the number of node, or, equivalently, data points), and so a Vietoris--Rips filtration based on these values has only few filtration steps at which simplices are added. 

\section{Conclusion}\label{manifoldconclusions}
Isomap is a well established manifold-learning tool and is useful for many data sets. It can successfully handle curvature of data in many cases, and the freedom of choosing the parameter $k$ or $\epsilon$ when creating a neighbourhood graph allows it to handle noise to some extent. However, when faced with particularly sparse and noisy data, Isomap is prone to so-called `short-circuit errors', which in some cases cannot be avoided regardless of the choice of $k$ or $\epsilon$. For such data, contagion maps can yield better reconstructions. For a suitable choice of the threshold parameter $T$, single noisy edges that occur in a neighbourhood graph do not carry a contagion and thus do not distort the estimate of the geodesic distances via activation times significantly. In other words, with the right choice of $T$, contagion maps are able to `exploit social reinforcement to silence noise'. 

\section{Acknowledgements} 
We thank Vidit Nanda and Ulrike Tillmann for helpful discussions and useful comments on various versions of this paper. 


\bibliographystyle{plain}

\end{document}